\documentclass{article} 
\usepackage{iclr2027_conference,times}
\iclrfinalcopy

\usepackage{amsmath,amsfonts,bm}

\def\eqref#1{equation~\ref{#1}}

\def\1{\bm{1}}

\DeclareMathAlphabet{\mathsfit}{\encodingdefault}{\sfdefault}{m}{sl}
\SetMathAlphabet{\mathsfit}{bold}{\encodingdefault}{\sfdefault}{bx}{n}

\usepackage[utf8]{inputenc} 
\usepackage[T1]{fontenc}    
\usepackage[pagebackref=true]{hyperref} 
\usepackage{url}            
\usepackage{booktabs}       
\usepackage{amsfonts}       
\usepackage{nicefrac}       
\usepackage{microtype}      
\usepackage{xcolor}         
\usepackage{amsmath}
\usepackage{subcaption}
\usepackage{titletoc}

\usepackage{placeins}   

\title{Multi-Token Prediction and Alternative \\ Tokenization Improve Learning Latent \\ Multi-Hop Graph Search at Scale}

\author{Arvid Frydenlund \\
  University of Toronto, Vector Institute, \\
  Manulife Applied AI Research, \\ 
  \texttt{arvie@cs.toronto.edu}  \\\And
  FuTe Wong \\
  University of Toronto, Vector Institute, \\
  \texttt{fute@cs.toronto.edu} \\\And
  Katherine Lambert \\
  University of Toronto, \\
  \texttt{katherine.lambert@mail.utoronto.ca} \\\And 
  Frank Rudzicz \\
  Dalhousie University, Vector Institute, \\ 
  \texttt{frank@dal.ca} \\
  \\}
\usepackage{bm}
\usepackage{multicol}

\usepackage{amsmath}
\usepackage{amssymb}
\usepackage{subcaption}

\usepackage{algorithm}
\usepackage{algorithmic}

 \usepackage{graphicx} 
 \usepackage{rotating}
 \usepackage{float}

 \usepackage{tikz}

\newcommand{\tikzA}{%
  \begin{tikzpicture}[baseline=-0.65ex, scale=0.115, line width=0.15ex]
    \draw (0,0) circle (1);
    \draw (-0.68,-0.73) -- (0,0.85) -- (0.68,-0.73);
    \draw (-0.52,-0.15) -- (0.52,-0.15);
  \end{tikzpicture}%
}

\newcommand{\A}{\mbox{\tikzA}\,}

\DeclareRobustCommand{\tikzZoom}{%
  \begin{tikzpicture}[baseline=-0.65ex, scale=0.22, line width=0.16ex]
    \draw (0,0) circle (1);
    \draw (0.71,-0.71) -- (1.45,-1.45);
  \end{tikzpicture}%
}

\DeclareRobustCommand{\zoomfig}[1]{\hyperref[#1]{\mbox{\tikzZoom}}}

\DeclareRobustCommand{\tikzZoomOut}{%
  \begin{tikzpicture}[baseline=-0.65ex, scale=0.22, line width=0.16ex]
    \draw (0,0) circle (1);
    \draw (0.71,-0.71) -- (1.45,-1.45);
    \draw (-0.5,0) -- (0.5,0);
  \end{tikzpicture}%
}

\DeclareRobustCommand{\smallfig}[1]{\hyperref[#1]{\mbox{\tikzZoomOut}}}

\newsavebox{\figbox}

\begin{document}

\maketitle

\begin{abstract}
Given an unseen graph, a start node, $s$, and an end node, $t$, can a language model generate the shortest path from $s$ to $t$?
Prior work has found that models fail to learn this task as the graphs scale.
We show that this failure has two causes: the supervision signal and the tokenization.
First, we theoretically establish and empirically demonstrate the effectiveness of multi-token prediction (MTP), and show that the benefits of MTP are topologically agnostic.
Second, we introduce alternative tokenizations and show that the two causes are orthogonal.
Using these methods, we break through the scaling wall observed by prior work, consistently and stably learning to solve the task at scales far beyond previous limits.
\end{abstract}

\section{Introduction}\label{sec:intro}

Can a language model learn to search? A direct way to answer this 
is to test whether the model can recover a shortest path in an unseen graph from a symbolic description alone. 
Here, success would not measure world knowledge or memorization, but whether the model can carry out a multihop computation over structure it has not seen before.
Our purpose is not to benchmark models but to understand why the task is hard to learn despite being within the capacity of small transformers.
Prior work has studied learnable search in controlled settings as a model for general planning and multihop reasoning \citep{bachmann2024the, saparov2025transformers, cohen2025spectral}, outlined below.

A multihop reasoning task is a tuple $(\mathcal{W},\, q,\, a)$ where $\mathcal{W}$ is a world model (a set of facts, axioms, etc.), $q$ is a query, and $a$ is a target answer. The task is solved by a sequence of $k \geq 1$ inferential steps that connect $q$ to $a$, each licensed by some relation or rule derivable from $\mathcal{W}$.  Importantly, $\mathcal{W}$ induces a latent reasoning graph, $\mathcal{G} = (\mathcal{N},\, \mathcal{E})$, where nodes are reasoning states and edges are viable inference steps. The task solution is a path $\pi = qu_1 \ldots u_{k-1}a$. 
When $k > 1$, $a$ cannot be directly inferred from $q$ in a single step, but must be found by discovering a path $\pi$ in $\mathcal{G}$ which connects $q$ to $a$.  This is graph search; the difficulty of which is characterized by the topology of $\mathcal{G}$, with the branching factor and the path length $k$ determining the size of the search space. As the reasoning graph is latent, its topology is unknown, and we can not make topological assumptions. Crucially, search as a process requires no prior knowledge of $\mathcal{G}$'s topology: it discovers $\pi$ by traversal alone. This makes graph search a natural theoretical model for multihop reasoning as it captures the essential structural and compositional difficulty without presupposing topological properties (unstated in prior work; \A \ref{appx:t-model}).

Studying multihop reasoning using natural language data with LLMs introduces confounds that obscure whether a model is genuinely traversing a reasoning graph, as it may produce the `answer' via memorized semantic shortcuts rather than structural inference; shortcuts which come from co-occurrences in the inaccessible training data \citep{ju-etal-2024-investigating, balesni2024lessons, yang-etal-2025-large}. Indeed, LLMs often ignore relational structure necessary for multihop reasoning in favour of semantic shortcuts, even when it is explicitly given \citep{huang2024can, xu2025when, guan2025attention}. 

These confounds motivate the use of a highly controlled environment to study multihop reasoning via graph search, {\em which we inherit} \citep[][ \S \ref{sec:task}]{bachmann2024the, saparov2025transformers}.  Graph search in this environment is not easy for an LM to learn.  \citet{bachmann2024the} found the existence of a structural shortcut that can prevent learning the task on a specific graph topology with next-token prediction (NTP), which can be ameliorated via certain training conditions (\S \ref{sec::problem-NTP}). However, even with better training conditions, prior work has found the models fail to learn this task as graphs scale in size and path length, hitting a scaling wall at $\approx$50 nodes, where learning becomes increasingly unstable \citep[][especially their Fig. 6]{frydenlund-2025-language, saparov2025transformers}.


{\bf We investigate and break through this scaling wall}. As prior work has identified an issue with NTP, and suggested MTP as a potential solution, we 
begin by considering if NTP is the cause of poor scaling.  To do this, we extend the results of \citet{bachmann2024the} to general topologies (\S \ref{sec::SSE}), and then elucidate how MTP overcomes the issues of NTP in a (nearly) topological-agnostic manner, and the various benefits this has for learnability (\S \ref{sec:sse-mtp}).  
We then hypothesize two other suspects: the causal constraint and lack of permutation-invariance.  These motivate the development of alternative tokenization methods (\S \ref{sec:causal-constraint} and \ref{sec:STAN}). Together, these methods allow us to stably learn the task at scale.

Prior work has eliminated obvious alternative suspects: model size and scratchpads \citep{saparov2025transformers}.  
Larger models do not improve training as the graph scales. This aligns with our strong empirical results with small models (implying larger models are unnecessary),  
and theoretical results that transformers with $O(\log k)$ layers can express the task \citep{pmlr-v235-sanford24a, sanford2024understanding, frydenlund-2024-mystery, merrill2025a}.\footnote{This fact immediately rules out the possibility of true path-length generalization OOD performance under a statically sized transformer.  {\bf As prior work has found the task unlearnable at scale in-domain, we focus on solving the problem in-domain, as that is a prerequisite to OOD performance}. See limitations in \A \ref{app:limitations}.}
Test-time compute via scratchpads/CoT is not a viable solution.  We show that a BFS scratchpad, while learnable, introduces its own tokenization difficulties. Critically, it also reduces the latent reasoning graph to a linear chain, breaking the theoretical model (\A \ref{appx:sp-collapse}).



\subsection{The Shortest-Path Task}\label{sec:task}

Let $\mathcal{G} = (\mathcal{N},\, \mathcal{E})$ be an unweighted graph, directed or not, with $|\mathcal{N}|$ nodes and $|\mathcal{E}|$ edges.  Let $s, t \in \mathcal{N}$ be the start and end nodes of the shortest path, $\pi = u_0 u_1 \dots u_{k}$ of length $k+1$ nodes. 
Then, given some tokenization of $\mathcal{G}$ and a
query,  $Q =$ `$(\, s\, t\, )$' s.t.\ $s=u_0,\, t=u_k$, the task is to generate the shortest path from $s$ to $t$, node-by-node; 
thus modelling  $p(\pi_b\,|\, s_b,\, t_b,\, \mathcal{G}_b)$ for the $b^{\mathrm{th}}$ sample $(\mathcal{G}_b,\, \pi_b)$. 

Each graph is tokenized as a list of edges, $E = [\,\text{`}u_i\, u_j\, |\text{'} :\, \forall\, (u_i, \, u_j) \in \mathcal{E}\,]$ where `|' is the end-of-edge marker.  The list is shuffled before concatenation to avoid providing structural information \citep{bachmann2024the}, e.g.\ ordered by BFS from $s$ \citep{ge-etal-2025-graph}.  The {\em source-side prefix} is the concatenation $[Q, E]$ \citep{frydenlund-2024-mystery}.  
The {\em target-side} is $\pi$ tokenized with start- and end-of-task tokens.  As the source-side must be supplied, we restrict prediction to the target-side (\A \ref{appx:prefix}).  Thus, the task must be learned from the supervision provided by the single path, $\pi$. 

The graphs are synthetically generated; nodes are sampled from a global node vocabulary, $\mathcal{N}_b \subseteq \mathcal{V}$, and the edges are sampled via some topology-specific process.  Nodes are represented by their one-hot IDs and are thus {\em semanticless} in that there is no relationship between nodes, i.e.\ `atomic facts'.  This prevents homophily shortcuts \citep{huang2024can, mcpherson2001birds, xu2025when, guan2025attention}. {\bf Thus, the model must learn to solve the task via the structural information provided by the prefix alone.} We generate unique instances during training  \citep{saparov2025transformers, frydenlund-2025-language}.  This is critical to avoid spurious correlations  \citep{frydenlund-2024-mystery, wang2023can}. 

This is an {\em implicit latent} multihop reasoning task, excluding inference-time procedures such as beam search or explicit intermediate token generations via scratchpads or CoT  \citep[][\A \ref{appx:sp-collapse}]{pan2026opening, li2025implicit, nye2022show, wei2022chain}.  As we train on many different graphs and evaluate on unseen graphs, this is the  {\em in-context} setting of the task \citep[][\A \ref{appx:in-weights}]{noroozizadeh2025deep}. 
This contrasts with the {\em in-weights} 
setting, which trains on a single graph and evaluates on unseen paths.
{\bf Altogether, this setting allows us to isolate the defining challenge of multihop reasoning: the ability to compose and chain individual inference steps into a coherent solution path} (\A \ref{appx:t-model}).

\subsection{The Sequential Supervision Problem with Next-Token Prediction}\label{sec::problem-NTP}

\begin{figure}[h]
    \centering  
    \includegraphics[trim={1.0cm 0.71cm .27cm 3.66cm},clip, scale=0.38]{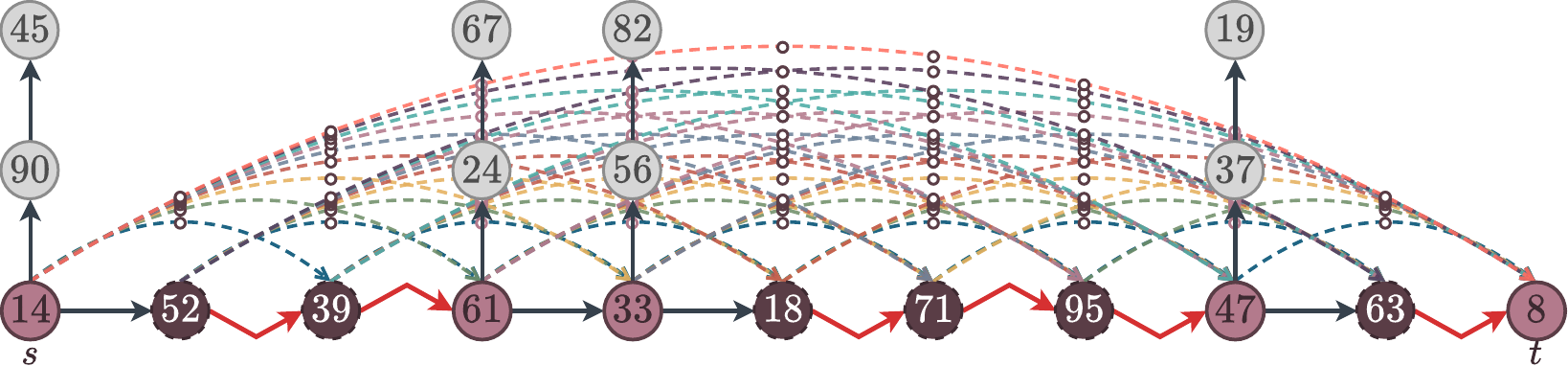}
    \caption{A path from 14 to 8, s.t. 14, 61, 33, and 47 are choice nodes.  The CHC absorbs supervision from non-choice nodes (zigzag red) under NTP.  Whereas MTP induces a dense quadratic latent multihop supervision supergraph (dashed rainbow).  The rainbow displays the hop-length mixture. This supergraph is agnostic to the underlying topology (See Fig. \ref{fig:topology-grid} in \protect\A \ref{appx:visualization} for the full example).} 
    \label{fig:placeholder}
\end{figure}

Given a single training instance, $(\mathcal{G},\, \pi)$, where $|\pi|=k+1$, NTP results in learning a joint distribution of
$p(\pi\,|\, s,\, t,\, \mathcal{E}) = p(s\,|\, s,\, t)\, p(u_1\,|\,s,\, t)\, \dots\, p(u_i\,|\,\pi_{<u_i},\, t)\, \dots\, p(t\,|\, \pi_{<t},\, t)$ with $k+1$ terms. 
The model learns positional shortcuts to trivially predict the first and last terms since $s$ and $t$ start and end $\pi$ \citep{frydenlund-2025-language}.  This leaves $k-1$ intra-sample terms, giving a {\em sequential sample efficiency} (SSE) of $k-1$ input-target supervision pairs or loss terms.  However, another topology-induced shortcut exists that can also absorb the supervision provided by these terms, further reducing the SSE.    

\citet{bachmann2024the} introduced the adversarially designed  {\em path-star task} to undermine the SSE under NTP with star-shaped graphs (Fig. \ref{fig:graph-ps}). Each graph has a centre node, $s$, with $d$ arms of length $k$.  The query is $(s,\, t)$, where $t$ ends an arm, making $\pi$ the arm containing $t$.  Let $d(u_i)$ be the degree of node $u_i$.  Then $d(s) = d$ and all other nodes have degree 2 or 1. This topology creates an unsympathetic interaction with NTP due to {\em teacher-forcing}, which conditions each prediction on the prior ground-truth sequence \citep{williams1989learning}.  Consider $p(u_i\,|\,\pi_{<u_i},\, t,\, \mathcal{E})$ when $1 < i \leq k-1$. By conditioning on the subpath, $\pi_{<u_i}$, the model can trivially predict that $u_i$ follows $u_{i-1}$ via the single edge $(u_{i-1},\, u_i) \in \mathcal{E}$ as $d(u_{i-1})=2$.  Such a prediction does not support learning multihop reasoning and, this shortcut, called the {\em Clever Hans Cheat} (CHC), will absorb all terms except $p(u_1\,|\,s,\, t)$, 
reducing the SSE from $k-1$ to 1 supervision pair on path-star graphs under NTP. 

This leads to very unfavourable learning conditions, where LMs using NTP fail to learn the task on graphs with only $9$ nodes ($d=2$ and $k=4$).  \citet{bachmann2024the} showed that a type of multi-token prediction (MTP) can ameliorate this complete failure, seemingly by restoring the SSE.    


However, the CHC not only reduces the amount of supervision, but also the type.  Consider the remaining term, $p(u_1\,|\,s,\, t)$. This requires $k$ hops to solve, as the model needs to reconstruct the subpath, $\pi(u_1,\, t) = u_1 \dots t$, to predict $u_1$ (done backwards). 
Now consider the terms the CHC absorbed.  Each term requires reconstruction of a subpath, each one hop shorter than the last, e.g.\, $p(u_2\,|\,su_1,\, t)$ requires  $|\pi(u_2,\, t)| = k - 1$, etc.  Thus, if not for the CHC, these terms would have provided a mixture of supervision at $k'=2,\,\dots,\,k-1$ hops, as each term is a {\bf decomposed subtask},
inducing a  {\bf curriculum-like} range of difficulties \citep{bengio2009curriculum}. As the original path-star task was trained on graphs of the same size, the model is only provided supervision at exactly $k$ hops.

The complete failure to learn on path-star graphs can be ameliorated by
training on a mixture of hop lengths, i.e. {\bf mixed multihop training} (MMHT; \A \ref{appx:MMHT-note}).  \citet{frydenlund-2025-language} showed that this can be achieved in multiple ways: by sampling graphs of $k' \leq k$ or sampling an arm-interior node to be the target, $t'$.  These methods explicitly mix values of $k' \leq k$ at the population level or 
inner-sample.
This can also be achieved implicitly intra-sample via MTP, which is equivalent to token masking and discrete diffusion (\A \ref{appx:masking-as-mtp}), or by training on different topologies (\S \ref{sec::SSE}).  Thus, both the density and the type of supervision matter. 
The importance of MMHT for efficient multihop learning has been established for other tasks \citep{pmlr-v235-sanford24a, pmlr-v291-wang25a, yao-etal-2025-language-models}.\footnote{\citet{pmlr-v235-sanford24a} introduced a simple in-context trace-back task called {\em Khops}.  They used an inner-sample mixture of $k$ but did not experimentally validate its importance.  We perform such an experiment in \A \ref{appx:khops-results}, which shows that the training requires an exponential number of steps as $k$ grows without MMHT.}

\citet{frydenlund-2025-language} considered MTP as it induces MMHT; while insufficient alone to break the scaling wall, this motivates a 
analysis of when the CHC induces topological shortcuts and prevents MMHT under NTP (\S \ref{sec::SSE}).\footnote{This is significant for a secondary reason; \citet{bachmann2024the} proved the existence of an adversarial topology for learning under NTP, and speculated that it will affect a broader set of tasks.  This has had a large impact, with many follow-ups proposing novel methods designed to overcome this problem (\A \ref{appx:PST-prior-work}).} 
{\bf We then show MTP overcomes this by simultaneously increasing supervision density and inducing implicit MMHT, with both benefits nearly topology-independent} (\S \ref{sec:sse-mtp}).

We then hypothesize two further suspects for the scaling wall: the causal constraint of causal LMs and permutation sensitivity. These motivate the development of alternative tokenization schemes which prove highly effective, breaking the scaling wall (\S \ref{sec:causal-constraint} and \S \ref{sec:STAN}). However, we find evidence that their success is not fully explained by their motivating hypotheses (\S \ref{sec:experiments}).

\section{Characterizing 
Sample Efficiency with General Topologies}\label{sec::SSE}

We now formalize when and how sequential supervision supports learning beyond star-graphs to
general topologies by characterizing the SSE under NTP.  We will then extend this to MTP in \S \ref{sec:sse-mtp}.

Let $A$ be the adjacency matrix of $\mathcal{G}$ s.t.\ $A(u,\, v) = 1$ if $(u,\, v) \in \mathcal{E}$, else $0$.
We say a node, $u_{i-1}$ at step $i-1$, is a {\em choice node} for a path, $\pi$, if there is more than one possible continuation from $u_{i-1}$, given the subpath $\pi_{\leq u_{i-1}} = \pi(s,\, u_{i-1})$.
We define $c(u_{i-1}, \pi_{\leq u_{i-1}})$ as the {\em choice set} of node $u_{i-1}$, containing the unseen neighbours of $u_{i-1}$ which may continue the prior path s.t.\ $c(u_{i-1},\, \pi_{\leq u_{i-1}}) = \{v \in \mathcal{N}\, :\, v \notin \pi_{\leq u_{i-1}}\, \mathrm{and}\, A(u_{i-1},\, v) = 1\}$.  The {\em choice degree} is $|c(u_{i-1},\, \pi_{\leq u_{i-1}})| = d(u_{i-1}) - \sum_{v \in \pi_{ \leq u_{i-1}}} A(u_{i-1}, v)$, thus $|c(s,\, \{s\})| = d(s)$ and $|c(u_{i-1},\, \pi_{\leq u_{i-1}})| < d(u_{i-1})$, otherwise. 
A node is a {\em choice node} when  $|c(u_{i-1},\, \pi_{\leq u_{i-1}})| > 1$, since if $|c(u_{i-1},\, \pi_{\leq u_{i-1}})| = 1$, there is only a single node that can continue the path and the CHC is sufficient for predicting it via the only viable edge $(u_{i-1},\, v \in c(u_{i-1},\, \pi_{\leq u_{i-1}}))$.
{\bf Only choice nodes support learning under NTP, as they require the model to perform a latent multihop traversal of the graph to predict the succeeding node from the choice set.  The CHC absorbs all other nodes.}  Thus, the SSE for general topologies depends on both $\mathcal{G}$ and $\pi$ and is just the number of choice nodes in $\pi$, $c(\pi)$.  

For undirected graphs, any interior node $u_{i-1}$ with $d(u_{i-1}) \geq 3$ is a choice node. Proof: $u_{i-1}$ is adjacent to only one observed node ($u_{i-2}$), as an edge to any earlier $u_j$ ($j < i-2$) would yield a shorter path to $u_{i-1}$, contradicting it being a shortest path; thus $|c(u_{i-1},\, \pi_{\leq u_{i-1}})| = d(u_{i-1}) - 1 \geq 2$.

Multiple choice nodes induce a subtask decomposition, implicitly providing MMHT. Proof sketch: assume $s$ and $u'=u_{i-1}$ are choice nodes in $\pi = s \dots u_{i-1} \dots t$.  Then multiple nodes could follow $u'$.  To determine which node in the choice set is correct, the model needs to reconstruct the subpath $\pi(u_{i-1},\, t)$, which requires $k' = k-i+1 < k$ hops. 
Hence, the subtask is simpler, with the position of the choice nodes controlling the sub-sample difficulty in terms of hop size. 
As a consequence, SSE and hop difficulty under NTP depend on both the graph topology and the sampled path.
We could maximize SSE and the mixture distribution by training on a topology that branches at each step in the path, i.e.\ $d$-ary trees.  This is not a scalable strategy when considering larger -- more difficult -- $k$, as it requires graphs of size $|\mathcal{N}| = d^k$.
{\bf However, MTP strongly reduces this dependency on topology.}  


\subsection{The Sequential Sample Efficiency of Multi-Token Prediction}\label{sec:sse-mtp}

MTP generalizes NTP to predict up to $n$ future tokens from a given position.  Consider conditioning on the prior-path, $\pi_{\leq u_{i-1}}$, i.e.\ up to position $i-1$.  Predicting one step into the future, $j=i$, recovers NTP.  Predicting two steps, $j=i+1$, results in a skip in the conditioning information since the prior node is unobserved, i.e.\ implicitly 
masked out.  Let $D$ be the distance matrix of $\mathcal{G}$, s.t.\ $D(u,\, v)$ is the distance between $u$ and $v$.  We extend the definition of a choice set to MTP as  $c(u_{j-1},\, \pi_{\leq u_{i-1}}) = \{v \in \mathcal{N}\, :\, v \notin \pi_{\leq u_{i-1}}\, \mathrm{and}\, D(u_{i-1},\, v) = j-i+1\}$ for $u_j$, when conditioning on $\pi_{\leq u_{i-1}}$.  Thus, the choice set for node $u_j$ contains all unseen nodes $j - i+1$ hops from $u_{i-1}$.    

Under NTP, when $j=i$, non-choice nodes with $|c(u_{i-1},\, \pi_{\leq u_{i-1}})| = 1$ do not contribute to supervision as they do not require multihop reasoning to predict the following node.  This is not true under MTP when $j-i > 0$, which introduces a skip of length $j-i$. This is because $u_j$ is $j-i+1$ edges from $u_{i-1}$, so any prediction must traverse $j-i+1$ edges and no single edge lookup suffices.
This multihop mechanism is different than when $|c(u_{i-1},\, \pi_{\leq u_{i-1}})| > 1$.  For either NTP or MTP, when we have a choice node,  multihop reasoning must be used to predict the following token from the choice set.  However, for NTP, construction of the choice set is trivial, whereas it is not for MTP with a $j-i > 0$ skip. 
Thus, MTP forces the model to perform an additional multihop computation, thereby increasing the SSE despite using the same sampled $\pi$ with the same $c(\pi)$.



To calculate the increase in SSE, we consider {\em Ranking-into-the-Future} (RITF) for our MTP method \citep{frydenlund-2025-language}. RITF allows for a dynamic future window $n$ s.t.\ $n=k$ maximally, which, as we will see, makes it optimal for this task.\footnote {Whereas static-$n$ methods only support $n \ll k$ hops, with $n\approx 4$ 
\citep{gloeckle2024better, cai2024medusa}.} 
It requires no model modification (allowing for a fair comparison with NTP), except for an optional additional prediction head.  Unlike \citet{frydenlund-2025-language}, who implemented RITF with a pairwise loss, we use the Plackett-Luce loss, which is probabilistic and is an ordered multi-label generalization of softmax-CE \citep{luce1959individual, plackett1975analysis, xia2008listwise, Frydenlund_Singh_Rudzicz_2022}. 
This defines the preference probability 

\begin{equation}\label{eq::PLD}
p(u_{i} \succ \dots \succ u_{k})
= \prod_{j=i}^{k} p(u_j \,|\,\pi_{<u_i},\, s,\, t,\, \mathcal{E})
= \prod_{j=i}^{k} \sigma\bigl(u_j^i;\, \mathcal{Z}_j\bigr),
\end{equation}
where $\succ$ denotes preference order, $u_j^i$ is the logit for node $u_j$ produced from the hidden state $h_i$, and $\sigma(\cdot\,;\mathcal{Z}_j)$ is softmax over the candidate set $\mathcal{Z}_j = \mathcal{N} \setminus \{u_\ell : i \leq \ell < j\}$.
Each succeeding softmax removes previously ranked nodes from $\mathcal{Z}_j$, so the first step is standard softmax-CE 
(\A \ref{appx:mtp} and \A \ref{appx:plackett-luce}).



We define NTP with positional shortcuts removed (Eq.\ \ref{eq:ntp}); MTP uses a superset of $k(k+1)/2 - 1$ terms (Eq.\ \ref{eq:mtp}).
Correcting for the $k - 1 - c(\pi)$ terms the CHC absorbs gives an SSE of $k(k+1)/2 - (k - c(\pi))$.
This residual topological dependency is linear, thus dominated by the quadratic MTP terms as $k$ grows, making MTP nearly topology-independent while NTP remains strongly topology-dependent.

\begin{minipage}{0.46\textwidth}
\noindent
  \begin{align}\label{eq:ntp}
      L_{\mathrm{NTP}} = -\sum_{i=1}^{k-1} \log p(u_i\,|\,\pi_{<u_i},\, s,\, t,\, \mathcal{E}).  
    \end{align}
\end{minipage}
\begin{minipage}{0.53\textwidth}
\noindent
    \begin{align}\label{eq:mtp}
      L_{\mathrm{MTP}} = -\sum_{i=1}^{k-1}\sum_{j=i}^{k} \log p(u_j\,|\,\pi_{<u_i},\, s,\, t,\, \mathcal{E}).
    \end{align}
\end{minipage}

Eq. \ref{eq:mtp} has an optimal SSE and hop mixture. 
{\em Proof sketch:} There are $k(k+1)/2$ non-trivial sub-paths in $\pi$, and each of these matches a term in the full version of Eq. \ref{eq:mtp}.  As no other sub-paths exist to support more terms, it must have optimal SSE  (same for an optimal hop mixture, proof in \A \ref{appx:proof-optimal}).


As we have shown, multiple choice nodes 
induce implicit MMHT, but this is topology-dependent for NTP.
{\bf Thus, MTP has multiple advantages over NTP for this task:  (1) It has optimal SSE with respect to the sampled path, (2) it induces implicit MMHT, and, importantly, (3) these advantages are nearly topologically independent.}  We empirically verify that MTP outperforms NTP on shortest path search and other symbolic multihop tasks (\S \ref{sec:experiments}, \A \ref{appx:khops}, \A \ref{appx:khops-gen}, and  \A \ref{appx:bfs-sp}).

We motivated the study of graph search as a theoretical model for multihop reasoning because the latent reasoning graph has unknown topology, thus making topological assumptions impermissible (\S \ref{sec:intro}). Given that our theoretical model was constructed to be agnostic to topology, and our finding that the theoretical advantages of MTP share this agnosticism, {\bf we {\em conjecture} that the advantages of MTP uncovered here apply to any multihop reasoning task.} This would make MTP a principled objective for any multihop reasoning task in general. 
Evaluating this 
is non-trivial, as our results rest on each target token corresponding to exactly one hop, which is untrue for natural language (\A \ref{app:limitations}).


A preliminary glance at Fig.\ \ref{fig:er-baseline} shows that MTP improves learnability at scale.  However, the results remain noisy with high variance across seeds.  This motivates investigating two further hypotheses.

\section{The Causal Constraint, Sparsity, and Fragmented Routing}\label{sec:causal-constraint}

Causal LMs derive expressive power from sequential computation, but are restricted to prior tokens only \citep{vaswani2017attention}. \citet{frydenlund-2024-mystery, saparov2025transformers} identified that the causal constraint hinders information routing. {\bf We show this problem is severely compounded with fragmented routing due to an interaction between the causal constraint, graph sparsity, and tokenization.} The model must route information across edges, which can only occur after an edge is observed. Since our graphs are sparse and $E$ is shuffled (to prevent structural shortcuts), the edge stream forms many disconnected components until most of $E$ is seen.\footnote{Non-sparse graphs only support tiny shortest paths, and thus do not support investigating scalability with $k$.} As a result, routing is deferred until most of $E$ has been processed, exhausting the model's sequential computation (Fig. \ref{fig:frag-routing} in \A \ref{appx:causal-constraint-vis}).


The causal constraint can be bypassed via repeating inputs \citep{xu-etal-2024-reading, springer2025repetition, kukic-etal-2026-sequence}.  Thus, if we tokenized the input as $[Q,\, E,\, E]$, then the second edge list can fully observe the first.  However, $E$ provides an inconsistent scaffold,  so we consider the node list, $N$, as $[Q,\, E,\, N]$ for a consistent means of accumulating edge information. $N$ is sorted by order of appearance of each node in the shuffled $E$, i.e.\ $E$ is shuffled, but $N$ is ordered relative to $E$.  

To test the hypothesis that the node list bypasses issues relating to routing, we consider both causal LMs and prefix LMs without causal masking on the source-side prefix \citep{raffel2020exploring, saparov2025transformers}.
Thus, if the {\em only} benefit to using the edge list is that it avoids routing issues, then we would expect to see performance gains with the causal LM over a baseline, but not the prefix LM.
We refer to experiments that include the node list as `+ $N$' compared to the baseline of only $[Q,\, E]$.  
 



\section{Positional Embeddings and Permutation Sensitivity}\label{sec:STAN}

 
Non-causal models like prefix LMs have been explored both to remove the attention restriction (as per above) and for their perceived permutation invariance \citep{frydenlund-2024-mystery, saparov2025transformers}. Permutation invariance is important; it reduces the space of possible tokenization orderings and spurious correlations.
Since permutation invariance of $\mathcal{E}$ is inherent to the task, this inductive bias is well-motivated. However, prior work failed to adequately address this, as {\bf a more clandestine issue breaks permutation invariance: positional embeddings} \citep{chen2023positional}. This is clandestine as positional embeddings are ubiquitous in LMs given their necessity for serial tokenization.  


Standard transformers take a $T$-length sequence of content token IDs $x$ alongside positional IDs $\phi$.
Tokenizing an edge as `$u$ $v$ |' means any node ID recurring at multiple positions is only distinguishable through $\phi$.
Positional embeddings are thus necessary {\em because tokens are not unique}, but they break the desired permutation invariance, so we make every `token' unique without $\phi$.

Note how the edge-marker creates {\em sequential annotations} of the edges. We follow \citet{saparov2025transformers} in concatenating and projecting the content and positional embeddings for the baseline models. 
Thus, the input to the model is really $z = [x, \phi]  \in \mathcal{Z}_{+}^{T \times 2}$.  We now do a simple trick, where we make each edge a unique {\em super-token} by treating it as a {\em structured} column input, $\hat{z} \in \mathcal{Z}_{+}^{T \times 3}$ s.t.\ $\hat{z}_{\hat{i}} = [x_i,\, x_{i+1},\, x_{i+2} = \text{`}|\text{'}]$, where the edge-marker annotation replaces $\phi_i$.  As no positional information leaks into the edge tokenization, this supports permutation invariance (Fig. \ref{fig:models} in \A \ref{appx:STAN-vis}).    


Thus, we have moved from sequential annotation (SEAN) to structural annotation (STAN). However, only edges contain 3 tokens of information. For other tokens, we replicate content tokens as padding. For node list tokens, where $x_i = u$, we use $\hat{z}_i = [x_i, x_i, \text{`}\texttt{/}\text{'}]$, where `$\texttt{/}$' is the node-marker, making $\mathcal{N}$ permutation-invariant. We retain sequential positional embeddings for the query and target tokens, necessary for output generation. 
As we were already concatenating content and positional embeddings, {\bf the primary change is how the input sequence is presented, not the model itself}.
We use `STAN' instead of `permutation invariance' to separate the method from the motivating hypothesis, since STAN can apply to causal LMs, which are incapable of permutation invariance.  STAN has the practical benefit of reducing the sequential tokenization of $E$ to $1/3$ the original length.



\section{Methodology}


We consider three topologies: Erd\H{o}s-R\'enyi graphs (ERG), Euclidean graphs (EG), and path-star graphs (PSG) \citep{erdds1959random, penrose2003random, bachmann2024the}. 
We use NTP during inference.  We consider the per-token accuracy (ACC) and the per-sequence accuracy (SACC), which is correct if all tokens are correct.  Both depend on a ground-truth sequence.  As the shortest path is not unique, the task accuracy is whether a valid shortest path was generated (TACC) (\A \ref{appx:label-smoothing}).  Each experiment is performed with 5 trial seeds and evaluated on $b=10$k $(\mathcal{G}_b,\, \pi_b)$ instances.

Unless stated, we consider small LMs with $L = 11$ layers and learned positional embeddings.
We sample $k' \in \{3,\, \dots,\ k\}$ and $|\mathcal{N}|' \in \{k+1,\, \dots,\ N\}$ for various max $k$ and $|\mathcal{N}|$ via a true curriculum.
Training runs to a max of 75k steps with a batch size of 1000 instances, stopping early at $>95\%$ TACC.
{\bf Each model for every experiment is trained from scratch} (for $\approx$ 2900 exps.), see \A \ref{appx:details}.

\section{Experiments, Results, and Discussion}\label{sec:experiments}

\begin{figure}[h]
    \centering
    \sbox{\figbox}{\includegraphics[trim={.35cm .35cm 3.cm 1.cm},clip, scale=.43]{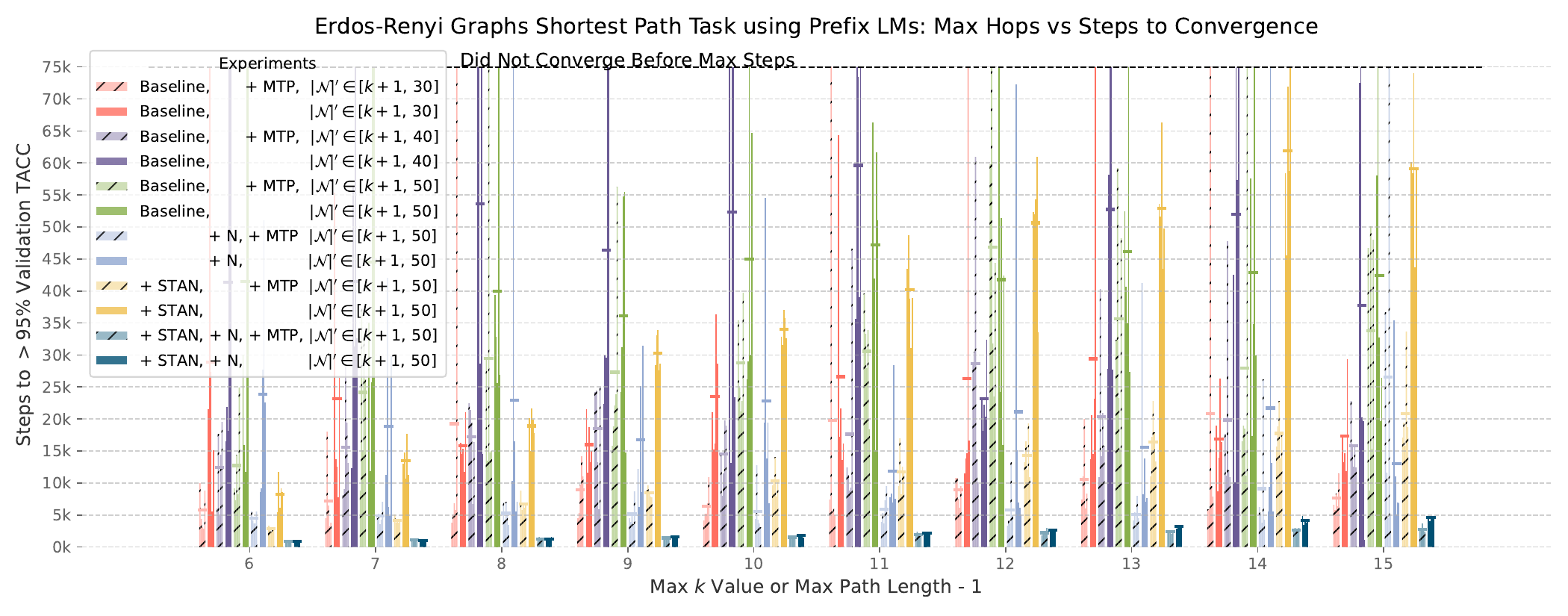}}%
    \usebox{\figbox}%
    \rlap{\hspace{0.6em}\raisebox{\dimexpr\ht\figbox-1.6ex\relax}[0pt][0pt]{\zoomfig{fig:er-baseline-full}}}%
    \caption{Baseline ($|\mathcal{N}| \in \{30,\, 40,\, 50\}$) and ablation ($|\mathcal{N}|=50$) results using prefix LMs on ERGs for the shortest-path task. Each experiment represents different seeds, illustrating the large variance.}
    \label{fig:er-baseline}
\end{figure}

\begin{figure}[h]
    \centering
    \sbox{\figbox}{\includegraphics[trim={.35cm .35cm 1.6cm .84cm},clip, scale=.42]{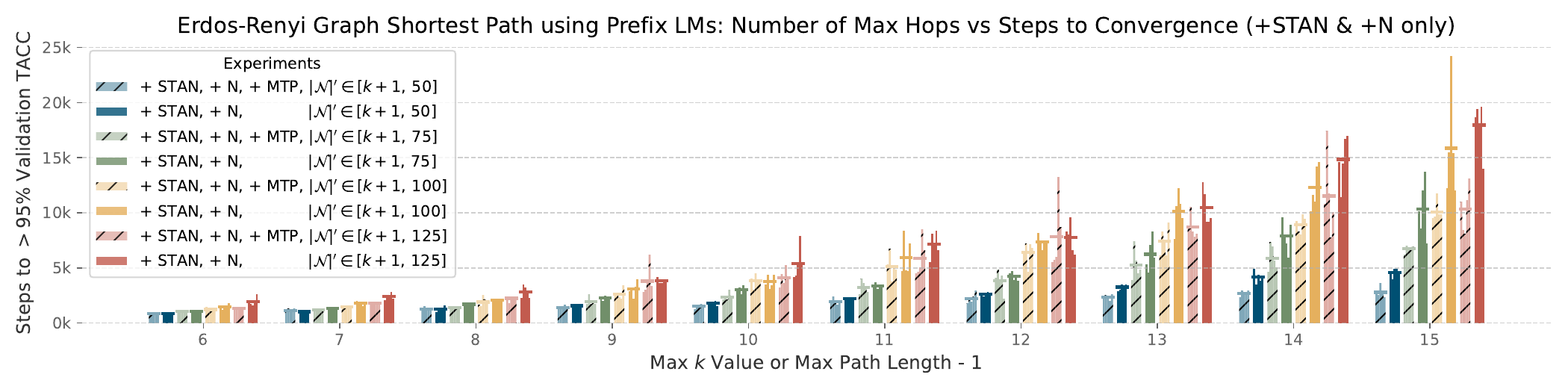}}%
    \usebox{\figbox}%
    \rlap{\hspace{0.6em}\raisebox{\dimexpr\ht\figbox-1.6ex\relax}[0pt][0pt]{\zoomfig{fig:er-new-full}}}%
    \caption{STAN and $N$ results using prefix LMs on ERGs.  The experiments at $|\mathcal{N}| = 50$ (blue) are also shown in Fig.\ \ref{fig:er-baseline}, demonstrating a dramatic improvement in learning and variance compared to the baselines.  This then allows us to scale to much larger graph sizes: $50$ through $75,\, 100$, and $125$.  Fig.\ \ref{fig:er-baseline} and Fig.\ \ref{fig:er-new-decoder-only} display a max of 75k steps vs. 25k here, showing a large difference between models.}
    \label{fig:er-new}
\end{figure}

\begin{figure}[h]
    \centering
    \sbox{\figbox}{\includegraphics[trim={.35cm .35cm 2.5cm .84cm},clip, scale=.42]{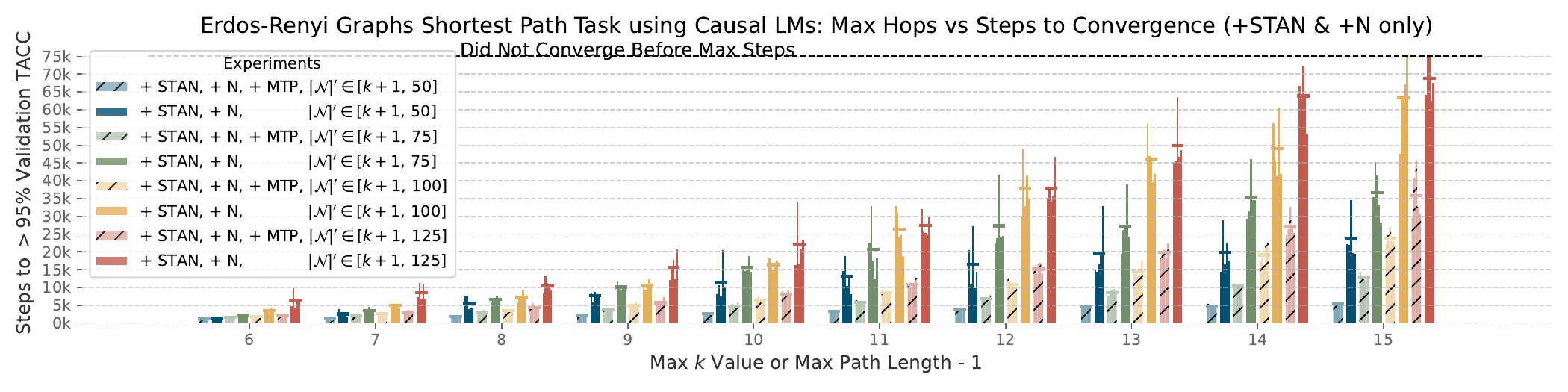}}%
    \usebox{\figbox}%
    \rlap{\hspace{0.6em}\raisebox{\dimexpr\ht\figbox-1.6ex\relax}[0pt][0pt]{\zoomfig{fig:er-new-decoder-only-full}}}%
    \caption{STAN and $N$ results using causal LMs on ERGs, for comparison to prefix LMs in Fig. \ref{fig:er-new}.  {\bf We see that MTP halves the cost of learning compared to NTP, thus demonstrating its importance to scaling, even if the tokenization methods account for the largest empirical gains.}}
    \label{fig:er-new-decoder-only}
\end{figure}

We consider a prefix LM baseline (SEAN and no node list) using ERGs at $|\mathcal{N}| \in \{30,\, 40,\, 50\}$ max nodes (Fig. \ref{fig:er-baseline}), then, we perform ablation at $|\mathcal{N}|=50$ on the STAN and $N$ methods (Fig. \ref{fig:er-baseline}). We then scale to $|\mathcal{N}| \in \{50,\, 75,\, 100,\, 125\}$ with STAN and $N$ with a prefix LM (Fig. \ref{fig:er-new}) and a causal LM (Fig. \ref{fig:er-new-decoder-only}).  Each NTP experiment is paired with a corresponding MTP experiment (in a lighter shade and hatched).  The results in Figs. \ref{fig:er-baseline}, \ref{fig:er-new}, and \ref{fig:er-new-decoder-only} are comparable, but be careful to note the scale of the y-axis and the graph sizes.  We sample paths and graphs in $k' \in [3,\, k]$ and $|\mathcal{N}|' \in [k+1,\, |\mathcal{N}|]$.

We see that the baselines are noisy (consistent with prior work on PSGs or variants; \citealt{saparov2025transformers, frydenlund-2025-language}), with a lower number of nodes being easier ($|\mathcal{N}|=30$ vs.\ $50$). However, scaling does not clearly depend on $k$, which is very unexpected, as fewer hops should be easier.


Among the ablations, including the node list ($+N$) has the most dramatic effect. Since $+N$ still improves performance with a prefix LM, its benefit extends beyond bypassing the causal constraint (this does not rule out that it also provides that benefit for causal LMs). Despite this, $+N$ alone leaves models noisy across trials. STAN resolves this variance and induces the expected behaviour where convergence scales with $k$, but does not consistently outperform the baseline. 
We also observe divergent trends between causal and prefix LM ablations, with causal LMs showing clearer trends, indicating a clear performance difference due to the different model parametrizations (Fig.\ \ref{fig:er-causal-lm-baseline}, \A \ref{appx:er-causal-lm-baseline}).




STAN and $N$ together significantly improve learnability for prefix LMs; consistently solving it in a fraction of the steps and leading to expected scaling behaviour (Fig. \ref{fig:er-new}). In solving the variance issue, we have greatly decreased the cost of future experimentation (\A \ref{appx:comp-cost}).  As expected, the prefix LMs outperform the causal LMs (Fig. \ref{fig:er-new}  vs. Fig. \ref{fig:er-new-decoder-only}). 
However, the fact that STAN also improves the causal LMs implies either that permutation invariance is not why it is effective or that causal LMs can, at least partially, get these benefits despite not being capable of true permutation invariance.

We demonstrate that MTP outperforms NTP on average across scales and methods, and that this difference increases as $k$ and $|\mathcal{N}|$ grow, thereby verifying its usefulness for scaling the task. {\em This also indicates that MTP and the alternative tokenization are orthogonal methods, and thus solve different underlying issues.}  While we see less difference between NTP and MTP on  EGs (Fig.\ \ref{fig:eg_k} in \A \ref{appx:eg}), we see a dramatic difference between them with PSGs in the following Section.   

\subsection{Path-Star Graphs}

For all PSG experiments, we use $L=8$, a max of 50k steps, and no true curriculum learning.  We consider models with a max $k \in [6,\, 15]$ and a max $d \in [2,\, 10]$.  By construction, the size of each graph is coupled to $k$ and $d$ s.t.\ $|\mathcal{N}| = dk + 1$.  Thus, our largest experiments consider $|\mathcal{N}| = 151$.  

\begin{figure}[h]
    \centering
    \sbox{\figbox}{\includegraphics[trim={.35cm .35cm 2.4cm .84cm},clip, scale=.43]{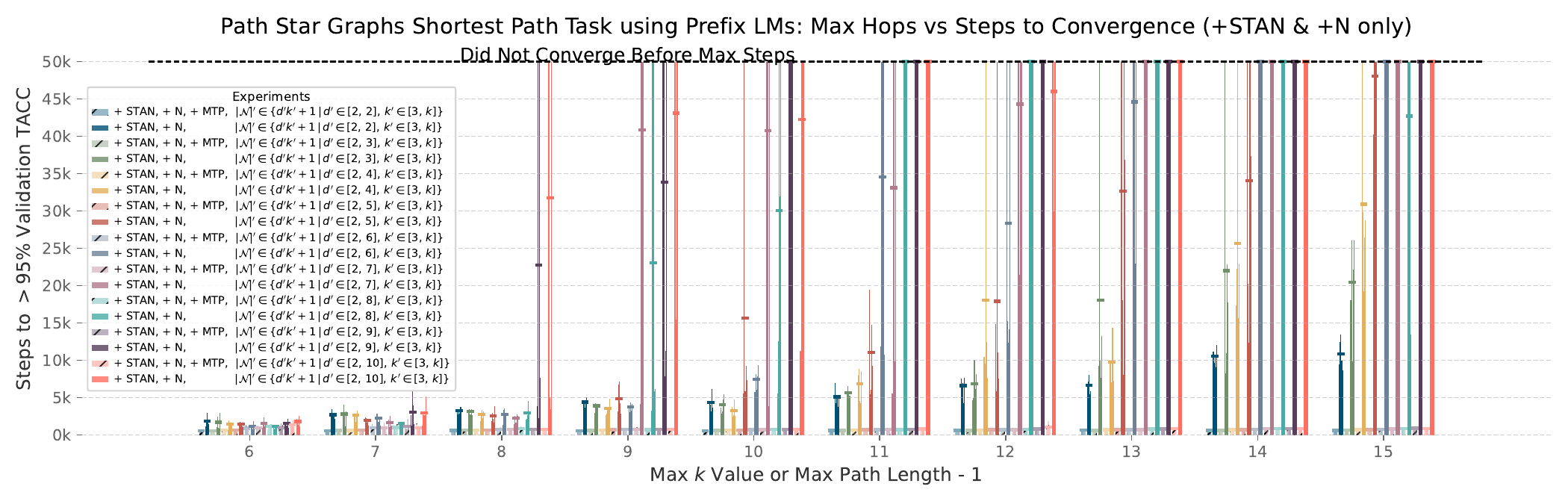}}%
    \usebox{\figbox}%
    \rlap{\hspace{0.6em}\raisebox{\dimexpr\ht\figbox-1.6ex\relax}[0pt][0pt]{\zoomfig{fig:psg_k-full}}}%
    \caption{STAN and $N$ results 
    on PSGs.  Each colour increments $d$. Notice that in the STAN and $N$ setting, MTP makes the task trivial to learn across all scales (please zoom in for the MTP results).}  
    \label{fig:psg_k}
\end{figure}

\begin{figure}[h]
    \centering
    \sbox{\figbox}{\includegraphics[trim={.35cm .35cm .35cm .86cm},clip, scale=.41]{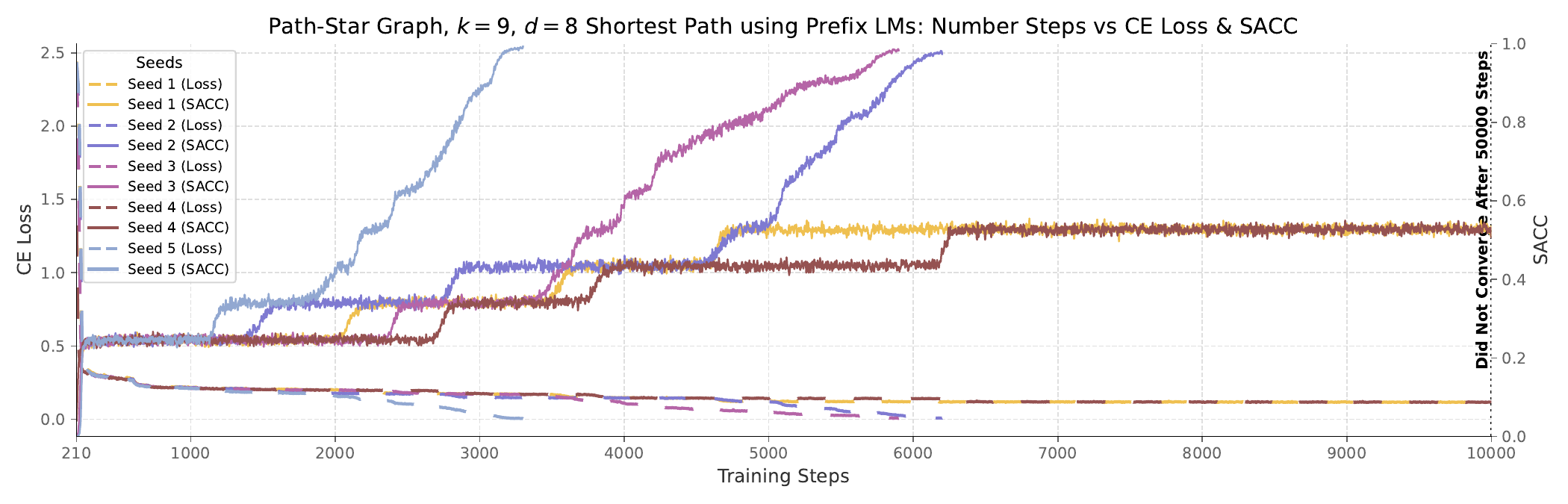}}%
    \usebox{\figbox}%
    \rlap{\hspace{0.6em}\raisebox{\dimexpr\ht\figbox-1.6ex\relax}[0pt][0pt]{\zoomfig{fig:psg_step-full}}}%
    \caption{The PSG experiment where $k=9$ and $d=8$ from Fig.\ \ref{fig:psg_k}.  Here, three seeds learn the task, and two seeds do not.  Jumps in SACC indicate learning the task for a specific $k'$ and $d'$ graph size.}  
    \label{fig:psg_step}
\end{figure}

Our results are shown in Fig.\ \ref{fig:psg_k}. To our knowledge, these are the first results to consistently solve PSGs using NTP beyond $k=4$ hops with only inter-sample MMHT 
and no other methods such as MTP, diffusion, RL, or novel architectures  (\A \ref{appx:PST-prior-work}).
As such, these results provide strong 
counterevidence to claims that NTP is insufficient to learn to plan \citep{bachmann2024the, wu2024can}. 
We can consistently scale NTP to $|\mathcal{N}|=65$ ($k=8$ and $d=8$) with STAN and +$N$, but these are insufficient alone to scale higher. {\bf Also applying MTP makes the task trivial across all scales.}  The fact that PSGs have the largest disparity between NTP and MTP conforms with theory, as PSGs are adversarial to NTP. 
An explanation for the size of this disparity is that PSGs allow solutions which do not generalize to arbitrary topology, e.g.\ the model can identify all final nodes by their degree, $d=1$, then walk back all arms to $s$ in parallel to identify the correct arm \citep{frydenlund-2024-mystery}. 



We observe the expected pattern where task difficulty grows with $k$ and $|\mathcal{N}|$; however, it is not gradual, but spikes sharply. To explain this, we consider the model's learning dynamics, and Fig.\ \ref{fig:psg_step} reveals why.  Here, rises in SACC between plateaus correspond to the model incrementally learning to solve the task at $d' < d$ and $k' < k$, from smallest to largest. Once solved at $d'$ or $k'$, those samples provide a near-zero loss and cease contributing to the learning signal. Thus, the models have {\em locked} themselves out of a hop-mixture including lower $k'$ terms.  Without the lower-$k'$ terms, models behave as if MMHT was never provided \citep{bachmann2024the, frydenlund-2025-language}, thus explaining the spikes. We confirm this by training on a restricted MMHT sampling only $k' \in [6,\, 9]$ rather than $k' \in [3,\, 9]$. These models fail to learn the task, just as if they had no MMHT (Fig.\ \ref{fig:restricte-psg} in \A \ref{appx:restricte-psg}).

{\em This confirms the importance of MMHT, while providing a new insight in that the actual mixture distribution is also critical.}  It also exposes two related issues: poor learning dynamics with NTP and failure to learn a generalized solution.  Importantly, the latter issue can occur only if the models fail to generalize their partial solutions at lower $k'$ to larger $k'$ (as Fig.\ \ref{fig:psg_step} demonstrates).

The CHC is a topological shortcut which causes the model to lose almost all supervision on PSGs.  This {\em lockout} phenomenon has a different root cause but leads to a similar loss of supervision and poor learning dynamics.   However, lockout may not be topology-dependent. 
We find that models can get stuck at high ACC without meeting the $> 95\%$ TACC threshold.  At such times, less than $1\%$ of the targets are contributing to the loss \citep{pezeshki2021gradient}.  Thus, such models have failed to generalize and may have locked themselves out of learning from the easier samples required for generalization.  


Sampling different graph sizes $k'$ induces an explicit MMHT, which we have shown is insufficient for learning with PSGs and NTP at scale.
MTP induces an implicit MMHT, and does not lead to lockout.
Thus, a final benefit of MTP is that the objective remains challenging -- so you don't lose your loss.

\section{Conclusion and Limitations}

We make several contributions showing that MTP and alternative tokenization enable learning latent multihop graph search at scale. We demonstrate the utility of shortest-path generation on unseen graphs as a controlled setting of latent multihop reasoning in transformer LMs. We extend \citet{bachmann2024the}'s adversarial path-star analysis to general topologies via a formal notion of choice nodes, characterizing precisely when and why supervision degrades under NTP. Our central finding, established both theoretically and empirically, is that MTP substantially improves supervision for this task through increased supervision density, mixed-hop supervision, and better learning dynamics, and, importantly, these advantages hold regardless of the underlying graph topology.

Multihop reasoning is intrinsically graph-structured, so our results on shortest-path search using MTP should extend to multihop reasoning more broadly. We empirically verify that similar benefits exist for MTP on other multihop reasoning tasks, including Khops \citep{pmlr-v235-sanford24a}, Khops-Gen (a novel variant), and a BFS scratchpad variant of shortest-path search (\A \ref{appx:khops}, \A \ref{appx:khops-gen}, and \A \ref{appx:bfs-sp}).
That MTP's advantages hold irrespective of topology was, itself, an unexpected finding.
Our theoretical model was constructed to be topology-agnostic and we did not expect MTP to inherit this property, which leads us to conjecture that MTP's benefits apply to any multihop reasoning task (\A \ref{app:limitations}).


A major trend in prior work is to use CoT or scratchpads to solve difficult reasoning tasks \citep{nye2022show, wei2022chain, cho2024arithmetic, alberghi2025on, zhang2026dagmath, xu2026no}, which changes how the task can be expressed, and thus its learnability \citep{wies2023subtask, dziri2023faith, feng2023towards, merrill2024the, abbe2024how, li2024chain, kim2025transformers}.
The BFS scratchpad we consider in \A \ref{appx:bfs-sp} is learnable, but its sequence length makes training impractical and it does not overcome the scaling wall.
More critically, it reduces the latent reasoning graph to a linear chain, breaking the theoretical model of \S \ref{sec:intro}, reducing it to the multihop backtrack of Khops-Gen.
Showing that the latent solution is the scalable solution bucks this trend.

While MTP yields strong theoretical and empirical gains, it alone does not break the scaling wall of prior work.
This motivated us to search for other suspects, and we identified two: the causal constraint of causal LMs, which results in a fragmented routing problem, and positional embeddings, which break the permutation invariance.
These hypotheses motivated two simple modifications: the node list ($+N$), for routing, and structural annotation (STAN), for permutation invariance, which we isolate with ablations over causal vs.\ prefix LMs, with and without $+N$ and STAN.

We find that both lead to strong empirical gains, allowing us to consistently learn the task at graph sizes well beyond the limits of prior work, and, together with MTP, to break the scaling wall.
Scaling to $2.5\times$ the size while training in $1/3$ as many steps (Fig. \ref{fig:er-baseline} vs. Fig. \ref{fig:er-new}) is a substantial gain given that the scaling trend is super-linear (Fig. \ref{fig:er-new}).
However, the ablations provide counter-evidence to these hypotheses: $+N$ improves prefix LMs, which should be unaffected by the causal constraint, and STAN improves causal LMs, which are incapable of true permutation invariance.
Both methods are highly effective, but their benefits must extend beyond the hypotheses that motivated them.
Why they work so well, especially in combination, remains an open question (\A \ref{app:limitations}).

Path-star graphs have received significant attention, spurring many follow-up works proposing novel methods to address their adversarial nature to NTP (\A \ref{appx:PST-prior-work}).
While PSGs are useful in exposing supervision pathologies,  they can be solved in ways that may not reflect arbitrary reasoning or planning. 
{\em We strongly caution against using PSGs as a benchmark.}  As we show, the task is trivial, given the proper conditions.  Thus, the main benefit of PSGs is as a tool for studying those conditions \citep{bachmann2024the, xu2026depth}.
 We also caution against viewing the results on PSGs with NTP as a generalizable limitation of NTP.  As we show, these limitations are limited to a specific adversarial topology, and NTP is actually sufficient for learning (at moderate scales and under the correct conditions). 
However, our results do support the claim that MTP is much more efficient than NTP at learning multihop graph search, and that this matters at scale.

\section*{AI use statement}


In this work, we used generative AI tools for code generation, proof generation and verification, and minimal writing and editing.  In particular, we used LLMs to assist in coding; however, this was done only for small sections of code at a time, allowing for human verification.  We engaged an LLM to verify the validity of our proofs (which led to various improvements).  We used AI to make the illustrations {\em but not the plots}, which were manually done to ensure integrity.  For the writing component, AI was mostly used to make arguments more succinct to meet space limitations in the main paper and help with the appendices.  This was monitored closely. We also used the AI tool {\em Grammarly} for proofreading. We have reviewed all AI-assisted work. We take responsibility for the final content of this work, including any artifacts produced with the aid of generative AI.

\section*{Reproducibility statement}


The main engineering challenge in this work is the online dataset and the PL-RITF cost function. We have already released these components, and the complete repository will follow upon acceptance.

\subsubsection*{Acknowledgments}


We thank Vaishnavh Nagarajan for his insightful feedback.  A significant chunk of my Ph.D. was dedicated to working on a problem he and Bachmann first proposed, so their work has had a big impact on me. We thank Abulhair Saparov for releasing his C++ graph generation code \citep{saparov2025transformers}. While we didn't use it directly, understanding it was critical for our own code. 

We also thank Shariyar Murtaza, Eugene Wen, and Manulife for providing a seamless transition of this project from my Ph.D. to working at Manulife.   

Resources used in preparing this research were provided, in part, by the Province of Ontario, the Government of Canada through CIFAR, companies sponsoring the Vector Institute, and the Digital Research Alliance of Canada.

\bibliography{acl_2025_post_submission, neurips_2026}

@article{vaswani2017attention,
  title={Attention is all you need},
  author={Vaswani, Ashish and Shazeer, Noam and Parmar, Niki and Uszkoreit, Jakob and Jones, Llion and Gomez, Aidan N and Kaiser, {\L}ukasz and Polosukhin, Illia},
  journal={Advances in neural information processing systems},
  volume={30},
  year={2017}
}

@inproceedings{bachmann2024the,
  title = 	 {The Pitfalls of Next-Token Prediction},
  author =       {Bachmann, Gregor and Nagarajan, Vaishnavh},
  booktitle = 	 {Proceedings of the 41st International Conference on Machine Learning},
  pages = 	 {2296--2318},
  year = 	 {2024},
  editor = 	 {Salakhutdinov, Ruslan and Kolter, Zico and Heller, Katherine and Weller, Adrian and Oliver, Nuria and Scarlett, Jonathan and Berkenkamp, Felix},
  volume = 	 {235},
  series = 	 {Proceedings of Machine Learning Research},
  month = 	 {21--27 Jul},
  publisher =    {PMLR},
  url = 	 {https://proceedings.mlr.press/v235/bachmann24a.html}
}

@article{monea2023pass,
  title={{PaSS}: Parallel speculative sampling},
  author={Monea, Giovanni and Joulin, Armand and Grave, Edouard},
  journal={arXiv preprint arXiv:2311.13581},
  year={2023}
}

@article{williams1989learning,
  title={A learning algorithm for continually running fully recurrent neural networks},
  author={Williams, Ronald J and Zipser, David},
  journal={Neural computation},
  volume={1},
  number={2},
  pages={270--280},
  year={1989},
  publisher={MIT Press One Rogers Street, Cambridge, MA 02142-1209, USA journals-info~…}
}

@inproceedings{NEURIPS2020_1457c0d6,
 author = {Brown, Tom and Mann, Benjamin and Ryder, Nick and Subbiah, Melanie and Kaplan, Jared D and Dhariwal, Prafulla and Neelakantan, Arvind and Shyam, Pranav and Sastry, Girish and Askell, Amanda and Agarwal, Sandhini and Herbert-Voss, Ariel and Krueger, Gretchen and Henighan, Tom and Child, Rewon and Ramesh, Aditya and Ziegler, Daniel and Wu, Jeffrey and Winter, Clemens and Hesse, Chris and Chen, Mark and Sigler, Eric and Litwin, Mateusz and Gray, Scott and Chess, Benjamin and Clark, Jack and Berner, Christopher and McCandlish, Sam and Radford, Alec and Sutskever, Ilya and Amodei, Dario},
 booktitle = {Advances in Neural Information Processing Systems},
 editor = {H. Larochelle and M. Ranzato and R. Hadsell and M.F. Balcan and H. Lin},
 pages = {1877--1901},
 publisher = {Curran Associates, Inc.},
 title = {Language Models are Few-Shot Learners},
 url = {https://proceedings.neurips.cc/paper_files/paper/2020/file/1457c0d6bfcb4967418bfb8ac142f64a-Paper.pdf},
 volume = {33},
 year = {2020}
}

@inproceedings{ott2019fairseq,
    title = "fairseq: A Fast, Extensible Toolkit for Sequence Modeling",
    author = "Ott, Myle  and
      Edunov, Sergey  and
      Baevski, Alexei  and
      Fan, Angela  and
      Gross, Sam  and
      Ng, Nathan  and
      Grangier, David  and
      Auli, Michael",
    editor = "Ammar, Waleed  and
      Louis, Annie  and
      Mostafazadeh, Nasrin",
    booktitle = "Proceedings of the 2019 Conference of the North {A}merican Chapter of the Association for Computational Linguistics (Demonstrations)",
    month = jun,
    year = "2019",
    address = "Minneapolis, Minnesota",
    publisher = "Association for Computational Linguistics",
    url = "https://aclanthology.org/N19-4009/",
    doi = "10.18653/v1/N19-4009",
    pages = "48--53"
}

@article{yang2019xlnet,
  title={Xlnet: Generalized autoregressive pretraining for language understanding},
  author={Yang, Zhilin and Dai, Zihang and Yang, Yiming and Carbonell, Jaime and Salakhutdinov, Russ R and Le, Quoc V},
  journal={Advances in neural information processing systems},
  volume={32},
  year={2019}
}

@inproceedings{liao-etal-2020-probabilistically,
    title = "Probabilistically Masked Language Model Capable of Autoregressive Generation in Arbitrary Word Order",
    author = "Liao, Yi  and
      Jiang, Xin  and
      Liu, Qun",
    booktitle = "Proceedings of the 58th Annual Meeting of the Association for Computational Linguistics",
    month = jul,
    year = "2020",
    address = "Online",
    publisher = "Association for Computational Linguistics",
    url = "https://aclanthology.org/2020.acl-main.24",
    doi = "10.18653/v1/2020.acl-main.24",
    pages = "263--274",
}

@inproceedings{
zhou2024what,
title={What Algorithms can Transformers Learn? A Study in Length Generalization},
author={Hattie Zhou and Arwen Bradley and Etai Littwin and Noam Razin and Omid Saremi and Joshua M. Susskind and Samy Bengio and Preetum Nakkiran},
booktitle={The Twelfth International Conference on Learning Representations},
year={2024},
url={https://openreview.net/forum?id=AssIuHnmHX}
}

@inproceedings{gu2018non,
  title={Non-Autoregressive Neural Machine Translation},
  author={Gu, Jiatao and Bradbury, James and Xiong, Caiming and Li, Victor OK and Socher, Richard},
  booktitle={International Conference on Learning Representations},
  year={2018}
}

@inproceedings{gu-kong-2021-fully,
    title = "Fully Non-autoregressive Neural Machine Translation: Tricks of the Trade",
    author = "Gu, Jiatao  and
      Kong, Xiang",
    booktitle = "Findings of the Association for Computational Linguistics: ACL-IJCNLP 2021",
    month = aug,
    year = "2021",
    address = "Online",
    publisher = "Association for Computational Linguistics",
    url = "https://aclanthology.org/2021.findings-acl.11",
    doi = "10.18653/v1/2021.findings-acl.11",
    pages = "120--133",
}

@inproceedings{lee-etal-2018-deterministic,
    title = "Deterministic Non-Autoregressive Neural Sequence Modeling by Iterative Refinement",
    author = "Lee, Jason  and
      Mansimov, Elman  and
      Cho, Kyunghyun",
    booktitle = "Proceedings of the 2018 Conference on Empirical Methods in Natural Language Processing",
    month = oct # "-" # nov,
    year = "2018",
    address = "Brussels, Belgium",
    publisher = "Association for Computational Linguistics",
    url = "https://aclanthology.org/D18-1149",
    doi = "10.18653/v1/D18-1149",
    pages = "1173--1182",
}

@inproceedings{ghazvininejad-etal-2019-mask,
    title = "Mask-Predict: Parallel Decoding of Conditional Masked Language Models",
    author = "Ghazvininejad, Marjan  and
      Levy, Omer  and
      Liu, Yinhan  and
      Zettlemoyer, Luke",
    booktitle = "Proceedings of the 2019 Conference on Empirical Methods in Natural Language Processing and the 9th International Joint Conference on Natural Language Processing (EMNLP-IJCNLP)",
    month = nov,
    year = "2019",
    address = "Hong Kong, China",
    publisher = "Association for Computational Linguistics",
    url = "https://aclanthology.org/D19-1633",
    doi = "10.18653/v1/D19-1633",
    pages = "6112--6121",
}

@InProceedings{pmlr-v235-sanford24a,
  title = 	 {Transformers, parallel computation, and logarithmic depth},
  author =       {Sanford, Clayton and Hsu, Daniel and Telgarsky, Matus},
  booktitle = 	 {Proceedings of the 41st International Conference on Machine Learning},
  pages = 	 {43276--43327},
  year = 	 {2024},
  editor = 	 {Salakhutdinov, Ruslan and Kolter, Zico and Heller, Katherine and Weller, Adrian and Oliver, Nuria and Scarlett, Jonathan and Berkenkamp, Felix},
  volume = 	 {235},
  series = 	 {Proceedings of Machine Learning Research},
  month = 	 {21--27 Jul},
  publisher =    {PMLR},
  url = 	 {https://proceedings.mlr.press/v235/sanford24a.html}
}

@inproceedings{
wies2023subtask,
title={Sub-Task Decomposition Enables Learning in Sequence to Sequence Tasks},
author={Noam Wies and Yoav Levine and Amnon Shashua},
booktitle={The Eleventh International Conference on Learning Representations },
year={2023},
url={https://openreview.net/forum?id=BrJATVZDWEH}
}

@inproceedings{frydenlund-2024-mystery,
    title = "The Mystery of the Pathological Path-star Task for Language Models",
    author = "Frydenlund, Arvid",
    editor = "Al-Onaizan, Yaser  and
      Bansal, Mohit  and
      Chen, Yun-Nung",
    booktitle = "Proceedings of the 2024 Conference on Empirical Methods in Natural Language Processing",
    month = nov,
    year = "2024",
    address = "Miami, Florida, USA",
    publisher = "Association for Computational Linguistics",
    url = "https://aclanthology.org/2024.emnlp-main.695",
    doi = "10.18653/v1/2024.emnlp-main.695",
    pages = "12493--12516",
}

@article{Frydenlund_Singh_Rudzicz_2022, title={Language Modelling via Learning to Rank}, volume={36}, url={https://ojs.aaai.org/index.php/AAAI/article/view/21308}, DOI={10.1609/aaai.v36i10.21308}, abstractNote={We consider language modelling (LM) as a multi-label structured prediction task by re-framing training from solely predicting a single ground-truth word to ranking a set of words which could continue a given context. To avoid annotating top-k ranks, we generate them using pre-trained LMs: GPT-2, BERT, and Born-Again models. This leads to a rank-based form of knowledge distillation (KD). We also develop a method using N-grams to create a non-probabilistic teacher which generates the ranks without the need of a pre-trained LM. We confirm the hypotheses: that we can treat LMing as a ranking task and that we can do so without the use of a pre-trained LM. We show that rank-based KD generally gives a modest improvement to perplexity (PPL) -- though often with statistical significance -- when compared to Kullback–Leibler-based KD. Surprisingly, given the naivety of the method, the N-grams act as competitive teachers and achieve similar performance as using either BERT or a Born-Again model teachers. Unsurprisingly, GPT-2 always acts as the best teacher. Using it and a Transformer-XL student on Wiki-02, rank-based KD reduces a cross-entropy baseline from 65.27 to 55.94 and against a KL-based KD of 56.70.}, number={10}, journal={Proceedings of the AAAI Conference on Artificial Intelligence}, author={Frydenlund, Arvid and Singh, Gagandeep and Rudzicz, Frank}, year={2022}, month={Jun.}, pages={10636-10644} }

@inproceedings{yin-etal-2024-semformer,
    title = "Semformer: Transformer Language Models with Semantic Planning",
    author = "Yin, Yongjing  and
      Ding, Junran  and
      Song, Kai  and
      Zhang, Yue",
    editor = "Al-Onaizan, Yaser  and
      Bansal, Mohit  and
      Chen, Yun-Nung",
    booktitle = "Proceedings of the 2024 Conference on Empirical Methods in Natural Language Processing",
    month = nov,
    year = "2024",
    address = "Miami, Florida, USA",
    publisher = "Association for Computational Linguistics",
    url = "https://aclanthology.org/2024.emnlp-main.1039",
    doi = "10.18653/v1/2024.emnlp-main.1039",
    pages = "18669--18680",
}

@inproceedings{
hu2025learning,
title={The Belief State Transformer},
author={Edward S. Hu and Kwangjun Ahn and Qinghua Liu and Haoran Xu and Manan Tomar and Ada Langford and Dinesh Jayaraman and Alex Lamb and John Langford},
booktitle={The Thirteenth International Conference on Learning Representations},
year={2025},
url={https://openreview.net/forum?id=ThRMTCgpvo}
}

@inproceedings{chen-etal-2024-masked,
    title = "Masked Thought: Simply Masking Partial Reasoning Steps Can Improve Mathematical Reasoning Learning of Language Models",
    author = "Chen, Changyu  and
      Wang, Xiting  and
      Lin, Ting-En  and
      Lv, Ang  and
      Wu, Yuchuan  and
      Gao, Xin  and
      Wen, Ji-Rong  and
      Yan, Rui  and
      Li, Yongbin",
    editor = "Ku, Lun-Wei  and
      Martins, Andre  and
      Srikumar, Vivek",
    booktitle = "Proceedings of the 62nd Annual Meeting of the Association for Computational Linguistics (Volume 1: Long Papers)",
    month = aug,
    year = "2024",
    address = "Bangkok, Thailand",
    publisher = "Association for Computational Linguistics",
    url = "https://aclanthology.org/2024.acl-long.320",
    doi = "10.18653/v1/2024.acl-long.320",
    pages = "5872--5900",
}

@article{
huang2024can,
title={Can {LLM}s Effectively Leverage Graph Structural Information through Prompts, and Why?},
author={Jin Huang and Xingjian Zhang and Qiaozhu Mei and Jiaqi Ma},
journal={Transactions on Machine Learning Research},
issn={2835-8856},
year={2024},
url={https://openreview.net/forum?id=L2jRavXRxs},
note={}
}

@article{zhang2022unveiling,
  title={Unveiling transformers with {LEGO}: a synthetic reasoning task},
  author={Zhang, Yi and Backurs, Arturs and Bubeck, S{\'e}bastien and Eldan, Ronen and Gunasekar, Suriya and Wagner, Tal},
  journal={arXiv preprint arXiv:2206.04301},
  year={2022}
}

@inproceedings{
yao2023tree,
title={Tree of Thoughts: Deliberate Problem Solving with Large Language Models},
author={Shunyu Yao and Dian Yu and Jeffrey Zhao and Izhak Shafran and Thomas L. Griffiths and Yuan Cao and Karthik R Narasimhan},
booktitle={Thirty-seventh Conference on Neural Information Processing Systems},
year={2023},
url={https://openreview.net/forum?id=5Xc1ecxO1h}
}

@inproceedings{10.1145/3626772.3657775,
author = {Tang, Jiabin and Yang, Yuhao and Wei, Wei and Shi, Lei and Su, Lixin and Cheng, Suqi and Yin, Dawei and Huang, Chao},
title = {GraphGPT: Graph Instruction Tuning for Large Language Models},
year = {2024},
isbn = {9798400704314},
publisher = {Association for Computing Machinery},
address = {New York, NY, USA},
url = {https://doi.org/10.1145/3626772.3657775},
doi = {10.1145/3626772.3657775},
booktitle = {Proceedings of the 47th International ACM SIGIR Conference on Research and Development in Information Retrieval},
pages = {491–500},
numpages = {10},
location = {Washington DC, USA},
series = {SIGIR '24}
}

@article{erdds1959random,
  title={On random graphs I},
  author={Erdős, P and R{\'e}nyi, A},
  journal={Publ. math. debrecen},
  volume={6},
  number={1},
  pages={290--297},
  year={1959}
}

@inproceedings{ren2024survey,
  title={A survey of large language models for graphs},
  author={Ren, Xubin and Tang, Jiabin and Yin, Dawei and Chawla, Nitesh and Huang, Chao},
  booktitle={Proceedings of the 30th ACM SIGKDD Conference on Knowledge Discovery and Data Mining},
  pages={6616--6626},
  year={2024}
}

@inproceedings{
wang2023can,
title={Can Language Models Solve Graph Problems in Natural Language?},
author={Heng Wang and Shangbin Feng and Tianxing He and Zhaoxuan Tan and Xiaochuang Han and Yulia Tsvetkov},
booktitle={Thirty-seventh Conference on Neural Information Processing Systems},
year={2023},
url={https://openreview.net/forum?id=UDqHhbqYJV}
}

@article{luo2024graphinstruct,
  title={Graphinstruct: Empowering large language models with graph understanding and reasoning capability},
  author={Luo, Zihan and Song, Xiran and Huang, Hong and Lian, Jianxun and Zhang, Chenhao and Jiang, Jinqi and Xie, Xing and Jin, Hai},
  journal={arXiv preprint arXiv:2403.04483},
  year={2024}
}

@article{guo2023gpt4graph,
  title={{GPT4Graph}: Can large language models understand graph structured data? an empirical evaluation and benchmarking},
  author={Guo, Jiayan and Du, Lun and Liu, Hengyu and Zhou, Mengyu and He, Xinyi and Han, Shi},
  journal={arXiv preprint arXiv:2305.15066},
  year={2023}
}

@inproceedings{yuan2024gracore,
    title = "{G}ra{C}o{R}e: Benchmarking Graph Comprehension and Complex Reasoning in Large Language Models",
    author = "Yuan, Zike  and
      Liu, Ming  and
      Wang, Hui  and
      Qin, Bing",
    editor = "Rambow, Owen  and
      Wanner, Leo  and
      Apidianaki, Marianna  and
      Al-Khalifa, Hend  and
      Eugenio, Barbara Di  and
      Schockaert, Steven",
    booktitle = "Proceedings of the 31st International Conference on Computational Linguistics",
    month = jan,
    year = "2025",
    address = "Abu Dhabi, UAE",
    publisher = "Association for Computational Linguistics",
    url = "https://aclanthology.org/2025.coling-main.531/",
    pages = "7925--7948"
}

@inproceedings{
tang2025grapharena,
title={GraphArena: Evaluating and Exploring Large Language Models on Graph Computation},
author={Jianheng Tang and Qifan Zhang and Yuhan Li and Nuo Chen and Jia Li},
booktitle={The Thirteenth International Conference on Learning Representations},
year={2025},
url={https://openreview.net/forum?id=Y1r9yCMzeA}
}

@article{liu2023evaluating,
  title={Evaluating large language models on graphs: Performance insights and comparative analysis},
  author={Liu, Chang and Wu, Bo},
  journal={arXiv preprint arXiv:2308.11224},
  year={2023}
}

@article{perozzi2024let,
  title={Let your graph do the talking: Encoding structured data for {LLM}s},
  author={Perozzi, Bryan and Fatemi, Bahare and Zelle, Dustin and Tsitsulin, Anton and Kazemi, Mehran and Al-Rfou, Rami and Halcrow, Jonathan},
  journal={arXiv preprint arXiv:2402.05862},
  year={2024}
}

@inproceedings{
saparov2025transformers,
title={Transformers Struggle to Learn to Search},
author={Abulhair Saparov and Srushti Ajay Pawar and Shreyas Pimpalgaonkar and Nitish Joshi and Richard Yuanzhe Pang and Vishakh Padmakumar and Mehran Kazemi and Najoung Kim and He He},
booktitle={The Thirteenth International Conference on Learning Representations},
year={2025},
url={https://openreview.net/forum?id=9cQB1Hwrtw}
}

@inproceedings{
khona2024toward,
title={Towards an Understanding of Stepwise Inference in Transformers: A Synthetic Graph Navigation Model},
author={Mikail Khona and Maya Okawa and Jan Hula and Rahul Ramesh and Kento Nishi and Robert P. Dick and Ekdeep Singh Lubana and Hidenori Tanaka},
booktitle={Forty-first International Conference on Machine Learning},
year={2024},
url={https://openreview.net/forum?id=8VEGkphQaK}
}

@inproceedings{
fatemi2024talk,
title={Talk like a Graph: Encoding Graphs for Large Language Models},
author={Bahare Fatemi and Jonathan Halcrow and Bryan Perozzi},
booktitle={The Twelfth International Conference on Learning Representations},
year={2024},
url={https://openreview.net/forum?id=IuXR1CCrSi}
}

@article{dai2024revisiting,
  title={Revisiting the Graph Reasoning Ability of Large Language Models: Case Studies in Translation, Connectivity and Shortest Path},
  author={Dai, Xinnan and Wen, Qihao and Shen, Yifei and Wen, Hongzhi and Li, Dongsheng and Tang, Jiliang and Shan, Caihua},
  journal={arXiv preprint arXiv:2408.09529},
  year={2024}
}

@inproceedings{wu2024can,
  title={Can Graph Learning Improve Planning in LLM-based Agents?},
  author={Wu, Xixi and Shen, Yifei and Shan, Caihua and Song, Kaitao and Wang, Siwei and Zhang, Bohang and Feng, Jiarui and Cheng, Hong and Chen, Wei and Xiong, Yun and others},
  booktitle={The Thirty-eighth Annual Conference on Neural Information Processing Systems},
  year={2024}
}

@inproceedings{
sanford2024understanding,
title={Understanding Transformer Reasoning Capabilities via Graph Algorithms},
author={Clayton Sanford and Bahare Fatemi and Ethan Hall and Anton Tsitsulin and Mehran Kazemi and Jonathan Halcrow and Bryan Perozzi and Vahab Mirrokni},
booktitle={The Thirty-eighth Annual Conference on Neural Information Processing Systems},
year={2024},
url={https://openreview.net/forum?id=AfzbDw6DSp}
}

@inproceedings{
nye2022show,
title={Show Your Work: Scratchpads for Intermediate Computation with Language Models},
author={Maxwell Nye and Anders Johan Andreassen and Guy Gur-Ari and Henryk Michalewski and Jacob Austin and David Bieber and David Dohan and Aitor Lewkowycz and Maarten Bosma and David Luan and Charles Sutton and Augustus Odena},
booktitle={Deep Learning for Code Workshop},
year={2022},
url={https://openreview.net/forum?id=HBlx2idbkbq}
}

@article{wei2022chain,
  title={Chain-of-thought prompting elicits reasoning in large language models},
  author={Wei, Jason and Wang, Xuezhi and Schuurmans, Dale and Bosma, Maarten and Ichter, Brian and Xia, Fei and Chi, Ed and Le, Quoc V and Zhou, Denny},
  journal={Advances in neural information processing systems},
  volume={35},
  pages={24824--24837},
  year={2022}
}

@inproceedings{
kim2025transformers,
title={Transformers Provably Solve Parity Efficiently with Chain of Thought},
author={Juno Kim and Taiji Suzuki},
booktitle={The Thirteenth International Conference on Learning Representations},
year={2025},
url={https://openreview.net/forum?id=n2NidsYDop}
}

@inproceedings{
abbe2024how,
title={How Far Can Transformers Reason? The Globality Barrier and Inductive Scratchpad},
author={Emmanuel Abbe and Samy Bengio and Aryo Lotfi and Colin Sandon and Omid Saremi},
booktitle={The Thirty-eighth Annual Conference on Neural Information Processing Systems},
year={2024},
url={https://openreview.net/forum?id=FoGwiFXzuN}
}

@inproceedings{
dziri2023faith,
title={Faith and Fate: Limits of Transformers on Compositionality},
author={Nouha Dziri and Ximing Lu and Melanie Sclar and Xiang Lorraine Li and Liwei Jiang and Bill Yuchen Lin and Sean Welleck and Peter West and Chandra Bhagavatula and Ronan Le Bras and Jena D. Hwang and Soumya Sanyal and Xiang Ren and Allyson Ettinger and Zaid Harchaoui and Yejin Choi},
booktitle={Thirty-seventh Conference on Neural Information Processing Systems},
year={2023},
url={https://openreview.net/forum?id=Fkckkr3ya8}
}

@inproceedings{
feng2023towards,
title={Towards Revealing the Mystery behind Chain of Thought: A Theoretical Perspective},
author={Guhao Feng and Bohang Zhang and Yuntian Gu and Haotian Ye and Di He and Liwei Wang},
booktitle={Thirty-seventh Conference on Neural Information Processing Systems},
year={2023},
url={https://openreview.net/forum?id=qHrADgAdYu}
}

@inproceedings{
merrill2024the,
title={The Expressive Power of Transformers with Chain of Thought},
author={William Merrill and Ashish Sabharwal},
booktitle={The Twelfth International Conference on Learning Representations},
year={2024},
url={https://openreview.net/forum?id=NjNGlPh8Wh}
}

@inproceedings{
li2024chain,
title={Chain of Thought Empowers Transformers to Solve Inherently Serial Problems},
author={Zhiyuan Li and Hong Liu and Denny Zhou and Tengyu Ma},
booktitle={The Twelfth International Conference on Learning Representations},
year={2024},
url={https://openreview.net/forum?id=3EWTEy9MTM}
}

@inproceedings{
liu2023transformers,
title={Transformers Learn Shortcuts to Automata},
author={Bingbin Liu and Jordan T. Ash and Surbhi Goel and Akshay Krishnamurthy and Cyril Zhang},
booktitle={The Eleventh International Conference on Learning Representations },
year={2023},
url={https://openreview.net/forum?id=De4FYqjFueZ}
}

@inproceedings{ruoss-etal-2023-randomized,
    title = "Randomized Positional Encodings Boost Length Generalization of Transformers",
    author = "Ruoss, Anian  and
      Del{\'e}tang, Gr{\'e}goire  and
      Genewein, Tim  and
      Grau-Moya, Jordi  and
      Csord{\'a}s, R{\'o}bert  and
      Bennani, Mehdi  and
      Legg, Shane  and
      Veness, Joel",
    editor = "Rogers, Anna  and
      Boyd-Graber, Jordan  and
      Okazaki, Naoaki",
    booktitle = "Proceedings of the 61st Annual Meeting of the Association for Computational Linguistics (Volume 2: Short Papers)",
    month = jul,
    year = "2023",
    address = "Toronto, Canada",
    publisher = "Association for Computational Linguistics",
    url = "https://aclanthology.org/2023.acl-short.161",
    doi = "10.18653/v1/2023.acl-short.161",
    pages = "1889--1903",
}

@inproceedings{
li2024functional,
title={Functional Interpolation for Relative Positions improves Long Context Transformers},
author={Shanda Li and Chong You and Guru Guruganesh and Joshua Ainslie and Santiago Ontanon and Manzil Zaheer and Sumit Sanghai and Yiming Yang and Sanjiv Kumar and Srinadh Bhojanapalli},
booktitle={The Twelfth International Conference on Learning Representations},
year={2024},
url={https://openreview.net/forum?id=rR03qFesqk}
}

@inproceedings{
mcleish2024transformers,
title={Transformers Can Do Arithmetic with the Right Embeddings},
author={Sean Michael McLeish and Arpit Bansal and Alex Stein and Neel Jain and John Kirchenbauer and Brian R. Bartoldson and Bhavya Kailkhura and Abhinav Bhatele and Jonas Geiping and Avi Schwarzschild and Tom Goldstein},
booktitle={The 4th Workshop on Mathematical Reasoning and AI at NeurIPS'24},
year={2024},
url={https://openreview.net/forum?id=cBFsFt1nDW}
}

@inproceedings{
zhou2024transformers,
title={Transformers Can Achieve Length Generalization But Not Robustly},
author={Yongchao Zhou and Uri Alon and Xinyun Chen and Xuezhi Wang and Rishabh Agarwal and Denny Zhou},
booktitle={ICLR 2024 Workshop on Mathematical and Empirical Understanding of Foundation Models},
year={2024},
url={https://openreview.net/forum?id=DWkWIh3vFJ}
}

@inproceedings{
cho2024position,
title={Position Coupling: Improving Length Generalization of Arithmetic Transformers Using Task Structure},
author={Hanseul Cho and Jaeyoung Cha and Pranjal Awasthi and Srinadh Bhojanapalli and Anupam Gupta and Chulhee Yun},
booktitle={The Thirty-eighth Annual Conference on Neural Information Processing Systems},
year={2024},
url={https://openreview.net/forum?id=5cIRdGM1uG}
}

@inproceedings{
cho2024arithmetic,
title={Arithmetic Transformers Can Length-Generalize in Both Operand Length and Count},
author={Hanseul Cho and Jaeyoung Cha and Srinadh Bhojanapalli and Chulhee Yun},
booktitle={The Thirteenth International Conference on Learning Representations},
year={2025},
url={https://openreview.net/forum?id=eIgGesYKLG}
}

@inproceedings{goodman-etal-2020-teaforn,
    title = "{T}ea{F}or{N}: Teacher-Forcing with N-grams",
    author = "Goodman, Sebastian  and
      Ding, Nan  and
      Soricut, Radu",
    editor = "Webber, Bonnie  and
      Cohn, Trevor  and
      He, Yulan  and
      Liu, Yang",
    booktitle = "Proceedings of the 2020 Conference on Empirical Methods in Natural Language Processing (EMNLP)",
    month = nov,
    year = "2020",
    address = "Online",
    publisher = "Association for Computational Linguistics",
    url = "https://aclanthology.org/2020.emnlp-main.702",
    doi = "10.18653/v1/2020.emnlp-main.702",
    pages = "8704--8717",
}

@inproceedings{qi-etal-2020-prophetnet,
    title = "{P}rophet{N}et: Predicting Future N-gram for Sequence-to-{S}equence{P}re-training",
    author = "Qi, Weizhen  and
      Yan, Yu  and
      Gong, Yeyun  and
      Liu, Dayiheng  and
      Duan, Nan  and
      Chen, Jiusheng  and
      Zhang, Ruofei  and
      Zhou, Ming",
    editor = "Cohn, Trevor  and
      He, Yulan  and
      Liu, Yang",
    booktitle = "Findings of the Association for Computational Linguistics: EMNLP 2020",
    month = nov,
    year = "2020",
    address = "Online",
    publisher = "Association for Computational Linguistics",
    url = "https://aclanthology.org/2020.findings-emnlp.217",
    doi = "10.18653/v1/2020.findings-emnlp.217",
    pages = "2401--2410",
}

@inproceedings{pal-etal-2023-future,
    title = "Future Lens: Anticipating Subsequent Tokens from a Single Hidden State",
    author = "Pal, Koyena  and
      Sun, Jiuding  and
      Yuan, Andrew  and
      Wallace, Byron  and
      Bau, David",
    editor = "Jiang, Jing  and
      Reitter, David  and
      Deng, Shumin",
    booktitle = "Proceedings of the 27th Conference on Computational Natural Language Learning (CoNLL)",
    month = dec,
    year = "2023",
    address = "Singapore",
    publisher = "Association for Computational Linguistics",
    url = "https://aclanthology.org/2023.conll-1.37",
    doi = "10.18653/v1/2023.conll-1.37",
    pages = "548--560",
}

@inproceedings{
cai2024medusa,
title={Medusa: Simple {LLM} Inference Acceleration Framework with Multiple Decoding Heads},
author={Tianle Cai and Yuhong Li and Zhengyang Geng and Hongwu Peng and Jason D. Lee and Deming Chen and Tri Dao},
booktitle={Forty-first International Conference on Machine Learning},
year={2024},
url={https://openreview.net/forum?id=PEpbUobfJv}
}

@inproceedings{
gloeckle2024better,
title={Better \& Faster Large Language Models via Multi-token Prediction},
author={Fabian Gloeckle and Badr Youbi Idrissi and Baptiste Roziere and David Lopez-Paz and Gabriel Synnaeve},
booktitle={Forty-first International Conference on Machine Learning},
year={2024},
url={https://openreview.net/forum?id=pEWAcejiU2}
}

@inproceedings{
wu2024do,
title={Do Language Models Plan Ahead for Future Tokens?},
author={Wilson Wu and John Xavier Morris and Lionel Levine},
booktitle={First Conference on Language Modeling},
year={2024},
url={https://openreview.net/forum?id=BaOAvPUyBO}
}

@inproceedings{men-etal-2024-unlocking,
    title = "Unlocking the Future: Exploring Look-Ahead Planning Mechanistic Interpretability in Large Language Models",
    author = "Men, Tianyi  and
      Cao, Pengfei  and
      Jin, Zhuoran  and
      Chen, Yubo  and
      Liu, Kang  and
      Zhao, Jun",
    editor = "Al-Onaizan, Yaser  and
      Bansal, Mohit  and
      Chen, Yun-Nung",
    booktitle = "Proceedings of the 2024 Conference on Empirical Methods in Natural Language Processing",
    month = nov,
    year = "2024",
    address = "Miami, Florida, USA",
    publisher = "Association for Computational Linguistics",
    url = "https://aclanthology.org/2024.emnlp-main.440",
    doi = "10.18653/v1/2024.emnlp-main.440",
    pages = "7713--7724",
}

@inproceedings{xia2022speculative,
    title = "Speculative Decoding: Exploiting Speculative Execution for Accelerating Seq2seq Generation",
    author = "Xia, Heming  and
      Ge, Tao  and
      Wang, Peiyi  and
      Chen, Si-Qing  and
      Wei, Furu  and
      Sui, Zhifang",
    editor = "Bouamor, Houda  and
      Pino, Juan  and
      Bali, Kalika",
    booktitle = "Findings of the Association for Computational Linguistics: EMNLP 2023",
    month = dec,
    year = "2023",
    address = "Singapore",
    publisher = "Association for Computational Linguistics",
    url = "https://aclanthology.org/2023.findings-emnlp.257/",
    doi = "10.18653/v1/2023.findings-emnlp.257",
    pages = "3909--3925"
}

@article{chen2023accelerating,
  title={Accelerating large language model decoding with speculative sampling},
  author={Chen, Charlie and Borgeaud, Sebastian and Irving, Geoffrey and Lespiau, Jean-Baptiste and Sifre, Laurent and Jumper, John},
  journal={arXiv preprint arXiv:2302.01318},
  year={2023}
}

@inproceedings{leviathan2023fast,
  title={Fast inference from transformers via speculative decoding},
  author={Leviathan, Yaniv and Kalman, Matan and Matias, Yossi},
  booktitle={Proceedings of the 40th International Conference on Machine Learning},
  volume={202},
  pages={19274--19286},
  year={2023},
  organization={PMLR}
}

@article{zheng2026temporal,
  title={Temporal Guidance for Large Language Models},
  author={Zheng, Hong-Kai and Li, Piji},
  journal={arXiv preprint arXiv:2601.21744},
  year={2026}
}

@article{mikolov2013efficient,
  title={Efficient estimation of word representations in vector space},
  author={Mikolov, Tomas and Chen, Kai and Corrado, Greg and Dean, Jeffrey},
  journal={arXiv preprint arXiv:1301.3781},
  year={2013}
}

@inproceedings{bengio2009curriculum,
  title={Curriculum learning},
  author={Bengio, Yoshua and Louradour, J{\'e}r{\^o}me and Collobert, Ronan and Weston, Jason},
  booktitle={Proceedings of the 26th annual international conference on machine learning},
  pages={41--48},
  year={2009}
}

@inproceedings{
ye2025beyond,
title={Beyond Autoregression: Discrete Diffusion for Complex Reasoning and Planning},
author={Jiacheng Ye and Jiahui Gao and Shansan Gong and Lin Zheng and Xin Jiang and Zhenguo Li and Lingpeng Kong},
booktitle={The Thirteenth International Conference on Learning Representations},
year={2025},
url={https://openreview.net/forum?id=NRYgUzSPZz}
}

@inproceedings{
ye2025implicit,
title={Implicit Search via Discrete Diffusion: A Study on Chess},
author={Jiacheng Ye and Zhenyu Wu and Jiahui Gao and Zhiyong Wu and Xin Jiang and Zhenguo Li and Lingpeng Kong},
booktitle={The Thirteenth International Conference on Learning Representations},
year={2025},
url={https://openreview.net/forum?id=A9y3LFX4ds}
}

@article{cohen2025spectral,
  title={Spectral Journey: How Transformers Predict the Shortest Path},
  author={Cohen, Andrew and Gromov, Andrey and Yang, Kaiyu and Tian, Yuandong},
  journal={arXiv preprint arXiv:2502.08794},
  year={2025}
}

@article{chen2023positional,
  title={Positional Information Matters for Invariant In-Context Learning: A Case Study of Simple Function Classes},
  author={Chen, Yongqiang and Xie, Binghui and Zhou, Kaiwen and Han, Bo and Bian, Yatao and Cheng, James},
  journal={arXiv preprint arXiv:2311.18194},
  year={2023}
}

@inproceedings{
austin2021structured,
title={Structured Denoising Diffusion Models in Discrete State-Spaces},
author={Jacob Austin and Daniel D. Johnson and Jonathan Ho and Daniel Tarlow and Rianne van den Berg},
booktitle={Advances in Neural Information Processing Systems},
editor={A. Beygelzimer and Y. Dauphin and P. Liang and J. Wortman Vaughan},
year={2021},
url={https://openreview.net/forum?id=h7-XixPCAL}
}

@article{ahn2025efficient,
  title={Efficient Joint Prediction of Multiple Future Tokens},
  author={Ahn, Kwangjun and Lamb, Alex and Langford, John},
  journal={arXiv preprint arXiv:2503.21801},
  year={2025}
}

@article{liu2024deepseek,
  title={{DeepSeek}-V3 technical report},
  author={Liu, Aixin and Feng, Bei and Xue, Bing and Wang, Bingxuan and Wu, Bochao and Lu, Chengda and Zhao, Chenggang and Deng, Chengqi and Zhang, Chenyu and Ruan, Chong and others},
  journal={arXiv preprint arXiv:2412.19437},
  year={2024}
}

@inproceedings{
gerontopoulos2025multi,
title={Multi-Token Prediction Needs Registers},
author={Anastasios Gerontopoulos and Spyros Gidaris and Nikos Komodakis},
booktitle={The Thirty-ninth Annual Conference on Neural Information Processing Systems},
year={2025},
url={https://openreview.net/forum?id=WDdBhcwzGe}
}

@article{stern2018blockwise,
  title={Blockwise parallel decoding for deep autoregressive models},
  author={Stern, Mitchell and Shazeer, Noam and Uszkoreit, Jakob},
  journal={Advances in Neural Information Processing Systems},
  volume={31},
  year={2018}
}

@inproceedings{frydenlund-2025-language,
    title = "Language Models, Graph Searching, and Supervision Adulteration: When More Supervision is Less and How to Make More More",
    author = "Frydenlund, Arvid",
    editor = "Che, Wanxiang  and
      Nabende, Joyce  and
      Shutova, Ekaterina  and
      Pilehvar, Mohammad Taher",
    booktitle = "Proceedings of the 63rd Annual Meeting of the Association for Computational Linguistics (Volume 1: Long Papers)",
    month = jul,
    year = "2025",
    address = "Vienna, Austria",
    publisher = "Association for Computational Linguistics",
    url = "https://aclanthology.org/2025.acl-long.1409/",
    doi = "10.18653/v1/2025.acl-long.1409",
    pages = "29011--29059",
    ISBN = "979-8-89176-251-0"
}

@inproceedings{ye-etal-2024-language,
    title = "Language is All a Graph Needs",
    author = "Ye, Ruosong  and
      Zhang, Caiqi  and
      Wang, Runhui  and
      Xu, Shuyuan  and
      Zhang, Yongfeng",
    editor = "Graham, Yvette  and
      Purver, Matthew",
    booktitle = "Findings of the Association for Computational Linguistics: EACL 2024",
    month = mar,
    year = "2024",
    address = "St. Julian{'}s, Malta",
    publisher = "Association for Computational Linguistics",
    url = "https://aclanthology.org/2024.findings-eacl.132/",
    pages = "1955--1973"
}

@article{besta2024graph, title={Graph of Thoughts: Solving Elaborate Problems with Large Language Models}, volume={38}, url={https://ojs.aaai.org/index.php/AAAI/article/view/29720}, DOI={10.1609/aaai.v38i16.29720}, abstractNote={We introduce Graph of Thoughts (GoT): a framework that
advances prompting capabilities in large language models
(LLMs) beyond those offered by paradigms such as Chain-of-Thought or Tree of Thoughts (ToT). The key idea and primary advantage of GoT is the ability to model the information generated by an LLM as an arbitrary graph, where units of information (&quot;LLM thoughts&quot;) are vertices, and edges correspond
to dependencies between these vertices. This approach enables combining arbitrary LLM thoughts into synergistic outcomes, distilling the essence of whole networks of thoughts,
or enhancing thoughts using feedback loops. We illustrate
that GoT offers advantages over state of the art on different
tasks, for example increasing the quality of sorting by 62%
over ToT, while simultaneously reducing costs by &gt;31%.
We ensure that GoT is extensible with new thought transformations and thus can be used to spearhead new prompting
schemes. This work brings the LLM reasoning closer to human thinking or brain mechanisms such as recurrence, both
of which form complex networks}, number={16}, journal={Proceedings of the AAAI Conference on Artificial Intelligence}, author={Besta, Maciej and Blach, Nils and Kubicek, Ales and Gerstenberger, Robert and Podstawski, Michal and Gianinazzi, Lukas and Gajda, Joanna and Lehmann, Tomasz and Niewiadomski, Hubert and Nyczyk, Piotr and Hoefler, Torsten}, year={2024}, month={Mar.}, pages={17682-17690} }

@InProceedings{pmlr-v291-wang25a,
  title = 	 {Learning Compositional Functions with Transformers from Easy-to-Hard Data},
  author =       {Wang, Zixuan and Nichani, Eshaan and Bietti, Alberto and Damian, Alex and Hsu, Daniel and Lee, Jason D and Wu, Denny},
  booktitle = 	 {Proceedings of Thirty Eighth Conference on Learning Theory},
  pages = 	 {5632--5711},
  year = 	 {2025},
  editor = 	 {Haghtalab, Nika and Moitra, Ankur},
  volume = 	 {291},
  series = 	 {Proceedings of Machine Learning Research},
  month = 	 {30 Jun--04 Jul},
  publisher =    {PMLR},
  url = 	 {https://proceedings.mlr.press/v291/wang25a.html}
}

@inproceedings{
ramesh2024compositional,
title={Compositional Capabilities of Autoregressive Transformers: A Study on Synthetic, Interpretable Tasks},
author={Rahul Ramesh and Ekdeep Singh Lubana and Mikail Khona and Robert P. Dick and Hidenori Tanaka},
booktitle={Forty-first International Conference on Machine Learning},
year={2024},
url={https://openreview.net/forum?id=L1eJ3NKPCd}
}

@inproceedings{
alberghi2025on,
title={On the Bias of Next-Token Predictors Toward Systematically Inefficient Reasoning: A Shortest-Path Case Study},
author={Riccardo Alberghi and Elizaveta Demyanenko and Luca Biggio and Luca Saglietti},
booktitle={The Thirty-ninth Annual Conference on Neural Information Processing Systems},
year={2025},
url={https://openreview.net/forum?id=l8razJItEy}
}

@inproceedings{
yang2024looped,
title={Looped Transformers are Better at Learning Learning Algorithms},
author={Liu Yang and Kangwook Lee and Robert D Nowak and Dimitris Papailiopoulos},
booktitle={The Twelfth International Conference on Learning Representations},
year={2024},
url={https://openreview.net/forum?id=HHbRxoDTxE}
}

@inproceedings{
saunshi2025reasoning,
title={Reasoning with Latent Thoughts: On the Power of Looped Transformers},
author={Nikunj Saunshi and Nishanth Dikkala and Zhiyuan Li and Sanjiv Kumar and Sashank J. Reddi},
booktitle={The Thirteenth International Conference on Learning Representations},
year={2025},
url={https://openreview.net/forum?id=din0lGfZFd}
}

@inproceedings{
fan2025looped,
title={Looped Transformers for Length Generalization},
author={Ying Fan and Yilun Du and Kannan Ramchandran and Kangwook Lee},
booktitle={The Thirteenth International Conference on Learning Representations},
year={2025},
url={https://openreview.net/forum?id=2edigk8yoU}
}

@InProceedings{giannou2023looped,
  title = 	 {Looped Transformers as Programmable Computers},
  author =       {Giannou, Angeliki and Rajput, Shashank and Sohn, Jy-Yong and Lee, Kangwook and Lee, Jason D. and Papailiopoulos, Dimitris},
  booktitle = 	 {Proceedings of the 40th International Conference on Machine Learning},
  pages = 	 {11398--11442},
  year = 	 {2023},
  editor = 	 {Krause, Andreas and Brunskill, Emma and Cho, Kyunghyun and Engelhardt, Barbara and Sabato, Sivan and Scarlett, Jonathan},
  volume = 	 {202},
  series = 	 {Proceedings of Machine Learning Research},
  month = 	 {23--29 Jul},
  publisher =    {PMLR},
  url = 	 {https://proceedings.mlr.press/v202/giannou23a.html}
}

@article{mcpherson2001birds,
  title={Birds of a feather: Homophily in social networks},
  author={McPherson, Miller and Smith-Lovin, Lynn and Cook, James M},
  journal={Annual review of sociology},
  volume={27},
  number={1},
  pages={415--444},
  year={2001},
  publisher={Annual Reviews 4139 El Camino Way, PO Box 10139, Palo Alto, CA 94303-0139, USA}
}

@inproceedings{
luo2024reasoning,
title={Reasoning on Graphs: Faithful and Interpretable Large Language Model Reasoning},
author={Linhao Luo and Yuan-Fang Li and Gholamreza Haffari and Shirui Pan},
booktitle={The Twelfth International Conference on Learning Representations},
year={2024},
url={https://openreview.net/forum?id=ZGNWW7xZ6Q}
}

@inproceedings{
xu2025when,
title={When Structure Doesn{\textquoteright}t Help: {LLM}s Do Not Read Text-Attributed Graphs as Effectively as We Expected},
author={Haotian Xu and Yuning You and Tengfei Ma},
booktitle={The Fourth Learning on Graphs Conference},
year={2025},
url={https://openreview.net/forum?id=ijQTdlHJze}
}

@inproceedings{
guan2025attention,
title={Attention Mechanisms Perspective: Exploring {LLM} Processing of Graph-Structured Data},
author={Zhong Guan and Likang Wu and Hongke Zhao and Ming He and Jianping Fan},
booktitle={Forty-second International Conference on Machine Learning},
year={2025},
url={https://openreview.net/forum?id=16oBNUlWyB}
}

@inproceedings{wu-etal-2025-graph,
    title = "Graph-R1: Incentivizing the Zero-Shot Graph Learning Capability in {LLM}s via Explicit Reasoning",
    author = "Wu, Yicong  and
      Lu, Guangyue  and
      Zuo, Yuan  and
      Zhang, Huarong  and
      Wu, Junjie",
    editor = "Christodoulopoulos, Christos  and
      Chakraborty, Tanmoy  and
      Rose, Carolyn  and
      Peng, Violet",
    booktitle = "Proceedings of the 2025 Conference on Empirical Methods in Natural Language Processing",
    month = nov,
    year = "2025",
    address = "Suzhou, China",
    publisher = "Association for Computational Linguistics",
    url = "https://aclanthology.org/2025.emnlp-main.1220/",
    doi = "10.18653/v1/2025.emnlp-main.1220",
    pages = "23909--23927",
    ISBN = "979-8-89176-332-6"
}

@inproceedings{
wang2024llms,
title={{LLM}s as Zero-shot Graph Learners: Alignment of {GNN} Representations with {LLM} Token Embeddings},
author={Duo Wang and Yuan Zuo and Fengzhi Li and Junjie Wu},
booktitle={The Thirty-eighth Annual Conference on Neural Information Processing Systems},
year={2024},
url={https://openreview.net/forum?id=32g9BWTndc}
}

@inproceedings{chen2024llaga,
author = {Chen, Runjin and Zhao, Tong and Jaiswal, Ajay and Shah, Neil and Wang, Zhangyang},
title = {LLaGA: large language and graph assistant},
year = {2024},
publisher = {JMLR.org},
booktitle = {Proceedings of the 41st International Conference on Machine Learning},
articleno = {306},
numpages = {15},
location = {Vienna, Austria},
series = {ICML'24}
}

@inproceedings{
kong2025gofa,
title={{GOFA}: A Generative One-For-All Model for Joint Graph Language Modeling},
author={Lecheng Kong and Jiarui Feng and Hao Liu and Chengsong Huang and Jiaxin Huang and Yixin Chen and Muhan Zhang},
booktitle={The Thirteenth International Conference on Learning Representations},
year={2025},
url={https://openreview.net/forum?id=mIjblC9hfm}
}

@inproceedings{ijcai2024p898,
  title     = {A Survey of Graph Meets Large Language Model: Progress and Future Directions},
  author    = {Li, Yuhan and Li, Zhixun and Wang, Peisong and Li, Jia and Sun, Xiangguo and Cheng, Hong and Yu, Jeffrey Xu},
  booktitle = {Proceedings of the Thirty-Third International Joint Conference on Artificial Intelligence, {IJCAI-24}},
  publisher = {International Joint Conferences on Artificial Intelligence Organization},
  editor    = {Kate Larson},
  pages     = {8123--8131},
  year      = {2024},
  month     = {8},
  note      = {Survey Track},
  doi       = {10.24963/ijcai.2024/898},
  url       = {https://doi.org/10.24963/ijcai.2024/898},
}

@inproceedings{kukic-etal-2026-sequence,
    title = "Sequence Repetition Enhances Token Embeddings and Improves Sequence Labeling with Decoder-only Language Models",
    author = "Kuki{\'c}, Matija Luka  and
      {\v{C}}uljak, Marko  and
      Duki{\'c}, David  and
      Tutek, Martin  and
      {\v{S}}najder, Jan",
    editor = "Demberg, Vera  and
      Inui, Kentaro  and
      Marquez, Llu{\'i}s",
    booktitle = "Findings of the {A}ssociation for {C}omputational {L}inguistics: {EACL} 2026",
    month = mar,
    year = "2026",
    address = "Rabat, Morocco",
    publisher = "Association for Computational Linguistics",
    url = "https://aclanthology.org/2026.findings-eacl.339/",
    doi = "10.18653/v1/2026.findings-eacl.339",
    pages = "6442--6456",
    ISBN = "979-8-89176-386-9"
}

@inproceedings{xu-etal-2024-reading,
    title = "Re-Reading Improves Reasoning in Large Language Models",
    author = "Xu, Xiaohan  and
      Tao, Chongyang  and
      Shen, Tao  and
      Xu, Can  and
      Xu, Hongbo  and
      Long, Guodong  and
      Lou, Jian-Guang  and
      Ma, Shuai",
    editor = "Al-Onaizan, Yaser  and
      Bansal, Mohit  and
      Chen, Yun-Nung",
    booktitle = "Proceedings of the 2024 Conference on Empirical Methods in Natural Language Processing",
    month = nov,
    year = "2024",
    address = "Miami, Florida, USA",
    publisher = "Association for Computational Linguistics",
    url = "https://aclanthology.org/2024.emnlp-main.871/",
    doi = "10.18653/v1/2024.emnlp-main.871",
    pages = "15549--15575"
}

@inproceedings{
springer2025repetition,
title={Repetition Improves Language Model Embeddings},
author={Jacob Mitchell Springer and Suhas Kotha and Daniel Fried and Graham Neubig and Aditi Raghunathan},
booktitle={The Thirteenth International Conference on Learning Representations},
year={2025},
url={https://openreview.net/forum?id=Ahlrf2HGJR}
}

@inproceedings{
li2026scaling,
title={Scaling Reasoning Hop Exposes Weaknesses: Demystifying and Improving Hop Generalization in Large Language Models},
author={Zhaoyi Li and Jiatong Li and Gangwei Jiang and Linqi Song and Defu Lian and Ying Wei},
booktitle={The Fourteenth International Conference on Learning Representations},
year={2026},
url={https://openreview.net/forum?id=qK4JKOu0Gx}
}

@article{pan2026opening,
  title={Opening the Black Box: A Survey on the Mechanisms of Multi-Step Reasoning in Large Language Models},
  author={Pan, Liangming and Liang, Jason and Ye, Jiaran and Yang, Minglai and Lu, Xinyuan and Zhu, Fengbin},
  journal={arXiv preprint arXiv:2601.14270},
  year={2026}
}

@inproceedings{yao-etal-2025-language-models,
    title = "Language models can learn implicit multi-hop reasoning, but only if they have lots of training data",
    author = "Yao, Yuekun  and
      Du, Yupei  and
      Zhu, Dawei  and
      Hahn, Michael  and
      Koller, Alexander",
    editor = "Christodoulopoulos, Christos  and
      Chakraborty, Tanmoy  and
      Rose, Carolyn  and
      Peng, Violet",
    booktitle = "Proceedings of the 2025 Conference on Empirical Methods in Natural Language Processing",
    month = nov,
    year = "2025",
    address = "Suzhou, China",
    publisher = "Association for Computational Linguistics",
    url = "https://aclanthology.org/2025.emnlp-main.490/",
    doi = "10.18653/v1/2025.emnlp-main.490",
    pages = "9684--9702",
    ISBN = "979-8-89176-332-6"
}

@inproceedings{
noroozizadeh2025deep,
title={Deep sequence models tend to memorize geometrically; it is unclear why},
author={Shahriar Noroozizadeh and Vaishnavh Nagarajan and Elan Rosenfeld and Sanjiv Kumar},
booktitle={Forty-third International Conference on Machine Learning},
year={2026},
url={https://openreview.net/forum?id=CT2tSmahVQ}
}

@article{trainin2026discrete,
  title={Discrete Diffusion Models Exploit Asymmetry to Solve Lookahead Planning Tasks},
  author={Trainin, Itamar and Ravfogel, Shauli and Abend, Omri and Feder, Amir},
  journal={arXiv preprint arXiv:2602.19980},
  year={2026}
}

@inproceedings{
liu2025lmtp,
title={L-{MTP}: Leap Multi-Token Prediction Beyond Adjacent Context for Large Language Models},
author={Xiaohao Liu and Xiaobo Xia and Weixiang Zhao and Manyi Zhang and Xianzhi Yu and Xiu Su and Shuo Yang and See-Kiong Ng and Tat-Seng Chua},
booktitle={The Thirty-ninth Annual Conference on Neural Information Processing Systems},
year={2025},
url={https://openreview.net/forum?id=0VDmWjW456}
}

@book{luce1959individual,
  title={Individual choice behavior},
  author={Luce, R Duncan and others},
  volume={4},
  year={1959},
  publisher={Wiley New York}
}

@article{plackett1975analysis,
  title={The analysis of permutations},
  author={Plackett, Robin L},
  journal={Journal of the Royal Statistical Society: Series C (Applied Statistics)},
  volume={24},
  number={2},
  pages={193--202},
  year={1975},
  publisher={Wiley Online Library}
}

@inproceedings{cao2007learning,
  title={Learning to rank: from pairwise approach to listwise approach},
  author={Cao, Zhe and Qin, Tao and Liu, Tie-Yan and Tsai, Ming-Feng and Li, Hang},
  booktitle={Proceedings of the 24th international conference on Machine learning},
  pages={129--136},
  year={2007}
}

@ARTICLE{6471714,
  author={Zhang, Min-Ling and Zhou, Zhi-Hua},
  journal={IEEE Transactions on Knowledge and Data Engineering}, 
  title={A Review on Multi-Label Learning Algorithms}, 
  year={2014},
  volume={26},
  number={8},
  pages={1819-1837},
  doi={10.1109/TKDE.2013.39}}

@book{bakir2007predicting,
  title={Predicting Structured Data},
  editor={Bak{\i}r, G{\"o}khan and Hofmann, Thomas and Sch{\"o}lkopf, Bernhard and Smola, Alexander J. and Taskar, Ben and Vishwanathan, S. V. N.},
  year={2007},
  publisher={MIT Press}
}

@article{elhage2021mathematical,
  title={A mathematical framework for transformer circuits},
  author={Elhage, Nelson and Nanda, Neel and Olsson, Catherine and Henighan, Tom and Joseph, Nicholas and Mann, Ben and Askell, Amanda and Bai, Yuntao and Chen, Anna and Conerly, Tom and others},
  journal={Transformer Circuits Thread},
  volume={1},
  number={1},
  pages={12},
  year={2021}
}

@misc{pybind11,
  author = {Wenzel Jakob and Jason Rhinelander and Dean Moldovan},
  year = {2016},
  note = {https://github.com/pybind/pybind11},
  title = {pybind11 --- Seamless operability between C++11 and Python}
}

@book{siek2001boost,
  title={The Boost Graph Library: User Guide and Reference Manual},
  author={Siek, Jeremy G and Lee, Lie-Quan and Lumsdaine, Andrew},
  year={2002},
  publisher={Addison-Wesley}
}

@book{penrose2003random,
  title={Random geometric graphs},
  author={Penrose, Mathew},
  volume={5},
  year={2003},
  publisher={OUP Oxford}
}

@InProceedings{pmlr-v235-back-de-luca24a,
  title = 	 {Simulation of Graph Algorithms with Looped Transformers},
  author =       {Back De Luca, Artur and Fountoulakis, Kimon},
  booktitle = 	 {Proceedings of the 41st International Conference on Machine Learning},
  pages = 	 {2319--2363},
  year = 	 {2024},
  editor = 	 {Salakhutdinov, Ruslan and Kolter, Zico and Heller, Katherine and Weller, Adrian and Oliver, Nuria and Scarlett, Jonathan and Berkenkamp, Felix},
  volume = 	 {235},
  series = 	 {Proceedings of Machine Learning Research},
  month = 	 {21--27 Jul},
  publisher =    {PMLR},
  url = 	 {https://proceedings.mlr.press/v235/back-de-luca24a.html}
}

@article{bengio2003neural,
  title={A neural probabilistic language model},
  author={Bengio, Yoshua and Ducharme, R{\'e}jean and Vincent, Pascal and Jauvin, Christian},
  journal={Journal of machine learning research},
  volume={3},
  number={Feb},
  pages={1137--1155},
  year={2003}
}

@inproceedings{NIPS2015_7137debd,
 author = {Dai, Andrew M and Le, Quoc V},
 booktitle = {Advances in Neural Information Processing Systems},
 editor = {C. Cortes and N. Lawrence and D. Lee and M. Sugiyama and R. Garnett},
 pages = {},
 publisher = {Curran Associates, Inc.},
 title = {Semi-supervised Sequence Learning},
 url = {https://proceedings.neurips.cc/paper_files/paper/2015/file/7137debd45ae4d0ab9aa953017286b20-Paper.pdf},
 volume = {28},
 year = {2015}
}

@article{li2025implicit,
  title={Implicit reasoning in large language models: A comprehensive survey},
  author={Li, Jindong and Fu, Yali and Fan, Li and Liu, Jiahong and Shu, Yao and Qin, Chengwei and Yang, Menglin and King, Irwin and Ying, Rex},
  journal={arXiv preprint arXiv:2509.02350},
  year={2025}
}

@article{zhu2025scaling,
  title={Scaling latent reasoning via looped language models},
  author={Zhu, Rui-Jie and Wang, Zixuan and Hua, Kai and Zhang, Tianyu and Li, Ziniu and Que, Haoran and Wei, Boyi and Wen, Zixin and Yin, Fan and Xing, He and others},
  journal={arXiv preprint arXiv:2510.25741},
  year={2025}
}

@article{raffel2020exploring,
  title={Exploring the limits of transfer learning with a unified text-to-text transformer},
  author={Raffel, Colin and Shazeer, Noam and Roberts, Adam and Lee, Katherine and Narang, Sharan and Matena, Michael and Zhou, Yanqi and Li, Wei and Liu, Peter J},
  journal={Journal of machine learning research},
  volume={21},
  number={140},
  pages={1--67},
  year={2020}
}

@inproceedings{
zhang2026dagmath,
title={{DAG}-Math: Graph-of-Thought Guided Mathematical Reasoning in {LLM}s},
author={Yuanhe Zhang and Ilja Kuzborskij and Jason D. Lee and Chenlei Leng and Fanghui Liu},
booktitle={The Fourteenth International Conference on Learning Representations},
year={2026},
url={https://openreview.net/forum?id=ylr6WArKQN}
}

@inproceedings{jin-etal-2024-graph,
    title = "Graph Chain-of-Thought: Augmenting Large Language Models by Reasoning on Graphs",
    author = "Jin, Bowen  and
      Xie, Chulin  and
      Zhang, Jiawei  and
      Roy, Kashob Kumar  and
      Zhang, Yu  and
      Li, Zheng  and
      Li, Ruirui  and
      Tang, Xianfeng  and
      Wang, Suhang  and
      Meng, Yu  and
      Han, Jiawei",
    editor = "Ku, Lun-Wei  and
      Martins, Andre  and
      Srikumar, Vivek",
    booktitle = "Findings of the Association for Computational Linguistics: ACL 2024",
    month = aug,
    year = "2024",
    address = "Bangkok, Thailand",
    publisher = "Association for Computational Linguistics",
    url = "https://aclanthology.org/2024.findings-acl.11/",
    doi = "10.18653/v1/2024.findings-acl.11",
    pages = "163--184"
}

@inproceedings{ge-etal-2025-graph,
    title = "Can Graph Descriptive Order Affect Solving Graph Problems with {LLM}s?",
    author = "Ge, Yuyao  and
      Liu, Shenghua  and
      Bi, Baolong  and
      Wang, Yiwei  and
      Mei, Lingrui  and
      Feng, Wenjie  and
      Chen, Lizhe  and
      Cheng, Xueqi",
    editor = "Che, Wanxiang  and
      Nabende, Joyce  and
      Shutova, Ekaterina  and
      Pilehvar, Mohammad Taher",
    booktitle = "Proceedings of the 63rd Annual Meeting of the Association for Computational Linguistics (Volume 1: Long Papers)",
    month = jul,
    year = "2025",
    address = "Vienna, Austria",
    publisher = "Association for Computational Linguistics",
    url = "https://aclanthology.org/2025.acl-long.321/",
    doi = "10.18653/v1/2025.acl-long.321",
    pages = "6404--6420",
    ISBN = "979-8-89176-251-0"
}

@inproceedings{
liu2024sophia,
title={Sophia: A Scalable Stochastic Second-order Optimizer for Language Model Pre-training},
author={Hong Liu and Zhiyuan Li and David Leo Wright Hall and Percy Liang and Tengyu Ma},
booktitle={The Twelfth International Conference on Learning Representations},
year={2024},
url={https://openreview.net/forum?id=3xHDeA8Noi}
}

@misc{Kenneth2013RandomIP,
  title={Random integer partitions with restricted numbers of parts},
  author={Locey, Kenneth},
  year={2013},
  url={https://api.semanticscholar.org/CorpusID:124354951},
  howpublished={Semantic Scholar CorpusID:124354951}
}

@inproceedings{yuan-etal-2026-barriers,
    title = "Barriers to Discrete Reasoning with Transformers: A Survey Across Depth, Exactness, and Bandwidth",
    author = "Yuan, Michelle  and
      Sun, Weiyi  and
      Rezaeian, Amir H.  and
      Singh, Jyotika  and
      Ghoshal, Sandip  and
      Wang, Yao-Ting  and
      Ballesteros, Miguel  and
      Benajiba, Yassine",
    editor = "Demberg, Vera  and
      Inui, Kentaro  and
      Marquez, Llu{\'i}s",
    booktitle = "Proceedings of the 19th Conference of the {E}uropean Chapter of the {A}ssociation for {C}omputational {L}inguistics (Volume 1: Long Papers)",
    month = mar,
    year = "2026",
    address = "Rabat, Morocco",
    publisher = "Association for Computational Linguistics",
    url = "https://aclanthology.org/2026.eacl-long.87/",
    doi = "10.18653/v1/2026.eacl-long.87",
    pages = "1966--1978",
    ISBN = "979-8-89176-380-7"
}

@inproceedings{
lee2025selfimproving,
title={Self-Improving Transformers Overcome Easy-to-Hard and Length Generalization Challenges},
author={Nayoung Lee and Ziyang Cai and Avi Schwarzschild and Kangwook Lee and Dimitris Papailiopoulos},
booktitle={Forty-second International Conference on Machine Learning},
year={2025},
url={https://openreview.net/forum?id=ZtX0MBT6mf}
}

@inproceedings{
lioubashevski2025looking,
title={Looking Beyond the Top-1: Transformers Determine Top Tokens in Order},
author={Daria Lioubashevski and Tomer M. Schlank and Gabriel Stanovsky and Ariel Goldstein},
booktitle={Forty-second International Conference on Machine Learning},
year={2025},
url={https://openreview.net/forum?id=2B11W1Z6ID}
}

@inproceedings{
qharabagh2026learning,
title={Learning to Execute Graph Algorithms Exactly with Graph Neural Networks},
author={Muhammad Fetrat Qharabagh and Artur Back de Luca and George Giapitzakis and Kimon Fountoulakis},
booktitle={Forty-third International Conference on Machine Learning},
year={2026},
url={https://openreview.net/forum?id=YKEmoqwkE9}
}

@inproceedings{
pezeshki2021gradient,
title={Gradient Starvation: A Learning Proclivity in Neural Networks},
author={Mohammad Pezeshki and S{\'e}kou-Oumar Kaba and Yoshua Bengio and Aaron Courville and Doina Precup and Guillaume Lajoie},
booktitle={Advances in Neural Information Processing Systems},
editor={A. Beygelzimer and Y. Dauphin and P. Liang and J. Wortman Vaughan},
year={2021},
url={https://openreview.net/forum?id=aExAsh1UHZo}
}

@article{ye2023diffusion,
  title={Diffusion language models can perform many tasks with scaling and instruction-finetuning},
  author={Ye, Jiasheng and Zheng, Zaixiang and Bao, Yu and Qian, Lihua and Gu, Quanquan},
  journal={arXiv preprint arXiv:2308.12219},
  year={2023}
}

@inproceedings{
setlur2024rl,
title={{RL} on Incorrect Synthetic Data Scales the Efficiency of {LLM} Math Reasoning by Eight-Fold},
author={Amrith Setlur and Saurabh Garg and Xinyang Geng and Naman Garg and Virginia Smith and Aviral Kumar},
booktitle={The Thirty-eighth Annual Conference on Neural Information Processing Systems},
year={2024},
url={https://openreview.net/forum?id=9m87e9Keq1}
}

@inproceedings{
zhou2026qsharp,
title={\$Q{\textbackslash}sharp\$: Provably Optimal Distributional {RL} for {LLM} Post-Training},
author={Jin Peng Zhou and Kaiwen Wang and Jonathan Daniel Chang and Zhaolin Gao and Nathan Kallus and Kilian Q Weinberger and Kiant{\'e} Brantley and Wen Sun},
booktitle={The Thirty-ninth Annual Conference on Neural Information Processing Systems},
year={2025},
url={https://openreview.net/forum?id=MrRdupucYb}
}

@article{thankaraj2025looking,
  title={Looking beyond the next token},
  author={Thankaraj, Abitha and Jiang, Yiding and Kolter, J Zico and Bisk, Yonatan},
  journal={arXiv preprint arXiv:2504.11336},
  year={2025}
}

@article{bobbili2025pita,
  title={PITA: Preference-Guided Inference-Time Alignment for LLM Post-Training},
  author={Bobbili, Sarat Chandra and Dinesha, Ujwal and Narasimha, Dheeraj and Shakkottai, Srinivas},
  journal={arXiv preprint arXiv:2507.20067},
  year={2025}
}

@inproceedings{
zuhri2025predicting,
title={Predicting the Order of Upcoming Tokens Improves Language Modeling},
author={Zayd Muhammad Kawakibi Zuhri and Erland Hilman Fuadi and Alham Fikri Aji},
booktitle={Forty-third International Conference on Machine Learning},
year={2026},
url={https://openreview.net/forum?id=wmROqgop27}
}

@article{prakash2025can,
  title={What Can You Do When You Have Zero Rewards During RL?},
  author={Prakash, Jatin and Buvanesh, Anirudh},
  journal={arXiv preprint arXiv:2510.03971},
  year={2025}
}

@article{monea2025breadcrumbs,
  title={Breadcrumbs Reasoning: Memory-Efficient Reasoning with Compression Beacons},
  author={Monea, Giovanni and Feldman, Yair and Padmanabhan, Shankar and Brantley, Kiant{\'e} and Artzi, Yoav},
  journal={arXiv preprint arXiv:2510.13797},
  year={2025}
}

@inproceedings{
mahajan2026beyond,
title={Beyond Multi-Token Prediction: Pretraining {LLM}s with Future Summaries},
author={Divyat Mahajan and Sachin Goyal and Badr Youbi Idrissi and Mohammad Pezeshki and Ioannis Mitliagkas and David Lopez-Paz and Kartik Ahuja},
booktitle={The Fourteenth International Conference on Learning Representations},
year={2026},
url={https://openreview.net/forum?id=aeYIFVn4vb}
}

@inproceedings{
teoh2026nextlatent,
title={Next-Latent Prediction Transformers Learn Compact World Models},
author={Jayden Teoh and Manan Tomar and Kwangjun Ahn and Edward S. Hu and Pratyusha Sharma and Riashat Islam and Alex Lamb and John Langford},
booktitle={Bridging Planning and Reasoning in Natural Language with Foundational Models},
year={2026},
url={https://openreview.net/forum?id=Lh4ayjJIAW}
}

@article{pipis2025wait,
  title={Wait, Wait, Wait... Why Do Reasoning Models Loop?},
  author={Pipis, Charilaos and Garg, Shivam and Kontonis, Vasilis and Shrivastava, Vaishnavi and Krishnamurthy, Akshay and Papailiopoulos, Dimitris},
  journal={arXiv preprint arXiv:2512.12895},
  year={2025}
}

@inproceedings{patel-etal-2026-xlm,
    title = "x{LM}: A Python Package for Non-Autoregressive Language Models",
    author = "Patel, Dhruvesh  and
      Maram, Durga Prasad  and
      Chintha, Sai Sreenivas  and
      Rozonoyer, Benjamin  and
      McCallum, Andrew",
    editor = "Croce, Danilo  and
      Leidner, Jochen  and
      Moosavi, Nafise Sadat",
    booktitle = "Proceedings of the 19th Conference of the {E}uropean Chapter of the {A}ssociation for {C}omputational {L}inguistics (Volume 3: System Demonstrations)",
    month = mar,
    year = "2026",
    address = "Rabat, Marocco",
    publisher = "Association for Computational Linguistics",
    url = "https://aclanthology.org/2026.eacl-demo.31/",
    doi = "10.18653/v1/2026.eacl-demo.31",
    pages = "445--456",
    ISBN = "979-8-89176-382-1"
}

@inproceedings{
patel2025insertion,
title={Insertion Language Models: Sequence Generation with Arbitrary-Position Insertions},
author={Dhruvesh Patel and Aishwarya Sahoo and Avinash Amballa and Tahira Naseem and Tim G. J. Rudner and Andrew McCallum},
booktitle={NeurIPS 2025 Workshop on Structured Probabilistic Inference {\&} Generative Modeling},
year={2025},
url={https://openreview.net/forum?id=IbVGvSDFQV}
}

@inproceedings{
bohnet2026analysis,
title={Analysis of Optimality of Large Language Models on Planning Problems},
author={Bernd Bohnet and Michael Curtis Mozer and Aaron T Parisi and Kevin Swersky and William A Cunningham and Kathleen Kenealy and Noah Fiedel},
booktitle={Third Conference on Language Modeling},
year={2026},
url={https://openreview.net/forum?id=YzMhzSuaIn}
}

@inproceedings{
xu2026depth,
title={The Depth Ceiling: On the Limits of Large Language Models in Discovering Latent Planning},
author={Yi Xu and Philipp Jettkant and Laura Ruis},
booktitle={Third Conference on Language Modeling},
year={2026},
url={https://openreview.net/forum?id=K8FSifPQ3s}
}

@inproceedings{
huang2026transformers,
title={How Transformers Learn to Plan via Multi-Token Prediction},
author={Jianhao Huang and Zhanpeng Zhou and Renqiu Xia and Baharan Mirzasoleiman and Weijie J Su and Wei Huang},
booktitle={Third Conference on Language Modeling},
year={2026},
url={https://openreview.net/forum?id=1uj8eRv3aJ}
}

@article{wang2026trainability,
  title={On the Trainability of Masked Diffusion Language Models via Blockwise Locality},
  author={Wang, Yuxiang and Xiang, Yu and Zhou, Baojian and Zhao, Qifang and Jiang, Keyue and Xiao, Yanghua and Xu, Xiaoxiao},
  journal={arXiv preprint arXiv:2604.24832},
  year={2026}
}

@article{peng2026efficient,
  title={Efficient Pre-Training with Token Superposition},
  author={Peng, Bowen and Gigant, Th{\'e}o and Quesnelle, Jeffrey},
  journal={arXiv preprint arXiv:2605.06546},
  year={2026}
}

@InProceedings{pmlr-v130-reddi21a,
  title = 	 { RankDistil: Knowledge Distillation for Ranking },
  author =       {Reddi, Sashank and Kumar Pasumarthi, Rama and Menon, Aditya and Singh Rawat, Ankit and Yu, Felix and Kim, Seungyeon and Veit, Andreas and Kumar, Sanjiv},
  booktitle = 	 {Proceedings of The 24th International Conference on Artificial Intelligence and Statistics},
  pages = 	 {2368--2376},
  year = 	 {2021},
  editor = 	 {Banerjee, Arindam and Fukumizu, Kenji},
  volume = 	 {130},
  series = 	 {Proceedings of Machine Learning Research},
  month = 	 {13--15 Apr},
  publisher =    {PMLR},
  url = 	 {https://proceedings.mlr.press/v130/reddi21a.html}
}

@inproceedings{lee-etal-2024-methods,
    title = "Methods, Applications, and Directions of Learning-to-Rank in {NLP} Research",
    author = "Lee, Justin  and
      Bernier-Colborne, Gabriel  and
      Maharaj, Tegan  and
      Vajjala, Sowmya",
    editor = "Duh, Kevin  and
      Gomez, Helena  and
      Bethard, Steven",
    booktitle = "Findings of the Association for Computational Linguistics: NAACL 2024",
    month = jun,
    year = "2024",
    address = "Mexico City, Mexico",
    publisher = "Association for Computational Linguistics",
    url = "https://aclanthology.org/2024.findings-naacl.123/",
    doi = "10.18653/v1/2024.findings-naacl.123",
    pages = "1900--1917"
}

@article{xiao2026mimo,
  title={{MiMo-V2-Flash} Technical Report},
  author={Xiao, Bangjun and Xia, Bingquan and Yang, Bo and Gao, Bofei and Shen, Bowen and Zhang, Chen and He, Chenhong and Lou, Chiheng and Luo, Fuli and Wang, Gang and others},
  journal={arXiv preprint arXiv:2601.02780},
  year={2026}
}

@article{team2026qwen3,
  title={Qwen3.5-Omni Technical Report},
  author={Team, Qwen},
  journal={arXiv preprint arXiv:2604.15804},
  year={2026}
}

@inproceedings{
zhang2026soft,
title={Soft Thinking: Unlocking the Reasoning Potential of {LLM}s in Continuous Concept Space},
author={Zhen Zhang and Xuehai He and Weixiang Yan and Ao Shen and Chenyang Zhao and Xin Eric Wang},
booktitle={The Thirty-ninth Annual Conference on Neural Information Processing Systems},
year={2026},
url={https://openreview.net/forum?id=ByQdHPGKgU}
}

@inproceedings{
tan2026think,
title={Think Silently, Think Fast: Dynamic Latent Compression of {LLM} Reasoning Chains},
author={Wenhui Tan and Jiaze Li and Jianzhong Ju and Zhenbo Luo and Ruihua Song and Jian Luan},
booktitle={The Thirty-ninth Annual Conference on Neural Information Processing Systems},
year={2025},
url={https://openreview.net/forum?id=AQsko3PPUe}
}

@article{tang2026multiplex,
  title={Multiplex Thinking: Reasoning via Token-wise Branch-and-Merge},
  author={Tang, Yao and Dong, Li and Hao, Yaru and Dong, Qingxiu and Wei, Furu and Gu, Jiatao},
  journal={arXiv preprint arXiv:2601.08808},
  year={2026}
}

@inproceedings{
chen2025pearl,
title={{PEARL}: Towards Permutation-Resilient {LLM}s},
author={Liang Chen and Li Shen and Yang Deng and Xiaoyan Zhao and Bin Liang and Kam-Fai Wong},
booktitle={The Thirteenth International Conference on Learning Representations},
year={2025},
url={https://openreview.net/forum?id=txoJvjfI9w}
}

@inproceedings{NIPS2017_f22e4747,
 author = {Zaheer, Manzil and Kottur, Satwik and Ravanbakhsh, Siamak and Poczos, Barnabas and Salakhutdinov, Russ R and Smola, Alexander},
 booktitle = {Advances in Neural Information Processing Systems},
 editor = {I. Guyon and U. Von Luxburg and S. Bengio and H. Wallach and R. Fergus and S. Vishwanathan and R. Garnett},
 pages = {},
 publisher = {Curran Associates, Inc.},
 title = {Deep Sets},
 url = {https://proceedings.neurips.cc/paper_files/paper/2017/file/f22e4747da1aa27e363d86d40ff442fe-Paper.pdf},
 volume = {30},
 year = {2017}
}

@inproceedings{
nie2026large,
title={Large Language Diffusion Models},
author={Shen Nie and Fengqi Zhu and Zebin You and Xiaolu Zhang and Jingyang Ou and Jun Hu and Jun Zhou and Yankai Lin and Ji-Rong Wen and Chongxuan Li},
booktitle={The Thirty-ninth Annual Conference on Neural Information Processing Systems},
year={2026},
url={https://openreview.net/forum?id=KnqiC0znVF}
}

@inproceedings{aynetdinov-akbik-2025-babies,
    title = "Babies Learn to Look Ahead: Multi-Token Prediction in Small {LM}s",
    author = "Aynetdinov, Ansar  and
      Akbik, Alan",
    editor = "Charpentier, Lucas  and
      Choshen, Leshem  and
      Cotterell, Ryan  and
      Gul, Mustafa Omer  and
      Hu, Michael Y.  and
      Liu, Jing  and
      Jumelet, Jaap  and
      Linzen, Tal  and
      Mueller, Aaron  and
      Ross, Candace  and
      Shah, Raj Sanjay  and
      Warstadt, Alex  and
      Wilcox, Ethan Gotlieb  and
      Williams, Adina",
    booktitle = "Proceedings of the First BabyLM Workshop",
    month = nov,
    year = "2025",
    address = "Suzhou, China",
    publisher = "Association for Computational Linguistics",
    url = "https://aclanthology.org/2025.babylm-main.41/",
    pages = "566--577"
}

@inproceedings{
jenner2024evidence,
title={Evidence of Learned Look-Ahead in a Chess-Playing Neural Network},
author={Erik Jenner and Shreyas Kapur and Vasil Georgiev and Cameron Allen and Scott Emmons and Stuart Russell},
booktitle={The Thirty-eighth Annual Conference on Neural Information Processing Systems},
year={2024},
url={https://openreview.net/forum?id=8zg9sO4ttV}
}

@inproceedings{
lee2024teaching,
title={Teaching Arithmetic to Small Transformers},
author={Nayoung Lee and Kartik Sreenivasan and Jason D. Lee and Kangwook Lee and Dimitris Papailiopoulos},
booktitle={The Twelfth International Conference on Learning Representations},
year={2024},
url={https://openreview.net/forum?id=dsUB4bst9S}
}

@article{shen2023positional,
  title={Positional description matters for transformers arithmetic},
  author={Shen, Ruoqi and Bubeck, S{\'e}bastien and Eldan, Ronen and Lee, Yin Tat and Li, Yuanzhi and Zhang, Yi},
  journal={arXiv preprint arXiv:2311.14737},
  year={2023}
}

@inproceedings{
xu2026no,
title={How Far Ahead Do {LLM}s Plan? Uncovering the Latent Horizon in Chain-of-Thought Reasoning},
author={Liyan Xu and Mo Yu and Fandong Meng and Jie Zhou},
booktitle={Forty-third International Conference on Machine Learning},
year={2026},
url={https://openreview.net/forum?id=pZqloVjqHd}
}

@article{zhao2026shattered,
  title={Shattered Compositionality: Counterintuitive Learning Dynamics of Transformers for Arithmetic},
  author={Zhao, Xingyu and Sharma, Darsh and Uppaal, Rheeya and Zhong, Yiqiao},
  journal={arXiv preprint arXiv:2601.22510},
  year={2026}
}

@article{tan2026pair,
  title={Pair-In, Pair-Out: Latent Multi-Token Prediction for Efficient LLMs},
  author={Tan, Wenhui and Li, Minghao and Ma, Xiaoqian and Fan, Siqi and Huang, Xiusheng and Zhang, Liujie and Song, Ruihua and Chen, Weihang},
  journal={arXiv preprint arXiv:2605.27255},
  year={2026}
}

@inproceedings{
tong2026generalization,
title={Generalization in {LLM} Problem Solving: The Case of the Shortest Path},
author={Yao Tong and Jiayuan Ye and Anastasia Borovykh and Reza Shokri},
booktitle={The Fourteenth International Conference on Learning Representations},
year={2026},
url={https://openreview.net/forum?id=RnRHNEeqvI}
}

@article{pochinkov2025parascopes,
  title={ParaScopes: What do Language Models Activations Encode About Future Text?},
  author={Pochinkov, Nicky and Volkova, Yulia and Vasileva, Anna and Chereddy, Sai VR},
  journal={arXiv preprint arXiv:2511.00180},
  year={2025}
}

@article{shirai2025less,
  title={Less is More: Learning Graph Tasks with Just LLMs},
  author={Shirai, Sola and Srinivas, Kavitha and Dolby, Julian and Katz, Michael and Samulowitz, Horst and Sohrabi, Shirin},
  journal={arXiv preprint arXiv:2508.10115},
  year={2025}
}

@article{mordig2026rethinking,
  title={Rethinking Easy-to-Hard: Limits of Curriculum Learning in Post-Training for Deductive Reasoning},
  author={Mordig, Maximilian and Opedal, Andreas and Liu, Weiyang and Sch{\"o}lkopf, Bernhard},
  journal={arXiv preprint arXiv:2603.27226},
  year={2026}
}

@inproceedings{
behrouz2025best,
title={Best of Both Worlds: Advantages of Hybrid Graph Sequence Models},
author={Ali Behrouz and Ali Parviz and Mahdi Karami and Clayton Sanford and Bryan Perozzi and Vahab Mirrokni},
booktitle={Forty-second International Conference on Machine Learning},
year={2025},
url={https://openreview.net/forum?id=GJKe8WYHxq}
}

@inproceedings{zhong2026toward,
    title = "Toward Consistent World Models with Multi-Token Prediction and Latent Semantic Enhancement",
    author = "Zhong, Qimin  and
      Liao, Hao  and
      Qin, Haiming  and
      Zhou, Mingyang  and
      Mao, Rui  and
      Chen, Wei  and
      Chao, Naipeng",
    editor = "Liakata, Maria  and
      Moreira, Viviane P.  and
      Zhang, Jiajun  and
      Jurgens, David",
    booktitle = "Proceedings of the 64th Annual Meeting of the {A}ssociation for {C}omputational {L}inguistics (Volume 1: Long Papers)",
    month = jul,
    year = "2026",
    address = "San Diego, California, United States",
    publisher = "Association for Computational Linguistics",
    url = "https://aclanthology.org/2026.acl-long.618/",
    doi = "10.18653/v1/2026.acl-long.618",
    pages = "13582--13602",
    ISBN = "979-8-89176-390-6"
}

@article{zhong2025understanding,
  title={Understanding and Enhancing the Planning Capability of Language Models via Multi-Token Prediction},
  author={Zhong, Qimin and Liao, Hao and Wang, Siwei and Zhou, Mingyang and Wu, Xiaoqun and Mao, Rui and Chen, Wei},
  journal={arXiv preprint arXiv:2509.23186},
  year={2025}
}

@inproceedings{
svete2026on,
title={On the Reasoning Abilities of Masked Diffusion Language Models},
author={Anej Svete and Ashish Sabharwal},
booktitle={The Fourteenth International Conference on Learning Representations},
year={2026},
url={https://openreview.net/forum?id=BVnIsh4Nz1}
}

@inproceedings{
wang2026do,
title={Finding the Minimal Parameter Budget for Implicit Reasoning: A Data Complexity Driven Scaling Law for Language Models},
author={Xinyi Wang and Shawn Tan and Shenbo Xu and Mingyu Jin and William Yang Wang and Rameswar Panda and Yikang Shen},
booktitle={Forty-third International Conference on Machine Learning},
year={2026},
url={https://openreview.net/forum?id=iuIPAhZpxz}
}

@inproceedings{
ustaomeroglu2026internal,
title={Internal Planning in Language Models: Characterizing Horizon and Branch Awareness},
author={Muhammed Ustaomeroglu and Baris Askin and Gauri Joshi and Carlee Joe-Wong and Guannan Qu},
booktitle={The Fourteenth International Conference on Learning Representations},
year={2026},
url={https://openreview.net/forum?id=dqGWQdFdTC}
}

@inproceedings{
dong2025emergent,
title={Emergent Response Planning in {LLM}s},
author={Zhichen Dong and Zhanhui Zhou and Zhixuan Liu and Chao Yang and Chaochao Lu},
booktitle={Forty-second International Conference on Machine Learning},
year={2025},
url={https://openreview.net/forum?id=Ce79P8ULPY}
}

@article{lee2025distinct,
  title={Distinct computations emerge from compositional curricula in in-context learning},
  author={Lee, Jin Hwa and Lampinen, Andrew K and Singh, Aaditya K and Saxe, Andrew M},
  journal={arXiv preprint arXiv:2506.13253},
  year={2025}
}

@article{liu2026language,
  title={What do language models learn and when? The implicit curriculum hypothesis},
  author={Liu, Emmy and Sun, Kaiser and Li, Millicent and Lee, Isabelle and Tjuatja, Lindia and Huang, Jen-tse and Neubig, Graham},
  journal={arXiv preprint arXiv:2604.08510},
  year={2026}
}

@inproceedings{luo-etal-2025-lets,
    title = "Let{'}s Be Self-generated via Step by Step: A Curriculum Learning Approach to Automated Reasoning with Large Language Models",
    author = "Luo, Kangyang  and
      Ding, Zichen  and
      Weng, Zhenmin  and
      Qiao, Lingfeng  and
      Zhao, Meng  and
      Li, Xiang  and
      Yin, Di  and
      Shu, Jinlong",
    editor = "Che, Wanxiang  and
      Nabende, Joyce  and
      Shutova, Ekaterina  and
      Pilehvar, Mohammad Taher",
    booktitle = "Findings of the Association for Computational Linguistics: ACL 2025",
    month = jul,
    year = "2025",
    address = "Vienna, Austria",
    publisher = "Association for Computational Linguistics",
    url = "https://aclanthology.org/2025.findings-acl.795/",
    doi = "10.18653/v1/2025.findings-acl.795",
    pages = "15343--15420",
    ISBN = "979-8-89176-256-5"
}

@inproceedings{ma-etal-2025-problem,
    title = "Problem-Solving Logic Guided Curriculum In-Context Learning for {LLM}s Complex Reasoning",
    author = "Ma, Xuetao  and
      Jiang, Wenbin  and
      Huang, Hua",
    editor = "Che, Wanxiang  and
      Nabende, Joyce  and
      Shutova, Ekaterina  and
      Pilehvar, Mohammad Taher",
    booktitle = "Findings of the Association for Computational Linguistics: ACL 2025",
    month = jul,
    year = "2025",
    address = "Vienna, Austria",
    publisher = "Association for Computational Linguistics",
    url = "https://aclanthology.org/2025.findings-acl.440/",
    doi = "10.18653/v1/2025.findings-acl.440",
    pages = "8394--8412",
    ISBN = "979-8-89176-256-5"
}

@inproceedings{
abramov2025grokking,
title={Grokking in the Wild: Data Augmentation for Real-World Multi-Hop Reasoning with Transformers},
author={Roman Abramov and Felix Steinbauer and Gjergji Kasneci},
booktitle={Forty-second International Conference on Machine Learning},
year={2025},
url={https://openreview.net/forum?id=lyUJH51URt}
}

@inproceedings{
hao2025training,
title={Training Large Language Models to Reason in a Continuous Latent Space},
author={Shibo Hao and Sainbayar Sukhbaatar and DiJia Su and Xian Li and Zhiting Hu and Jason E Weston and Yuandong Tian},
booktitle={Second Conference on Language Modeling},
year={2025},
url={https://openreview.net/forum?id=Itxz7S4Ip3}
}

@inproceedings{pham-etal-2025-clozemath,
    title = "{C}loze{M}ath: Improving Mathematical Reasoning in Language Models by Learning to Fill Equations",
    author = "Pham, Quang Hieu  and
      Nguyen, Thuy Duong  and
      Pham, Tung  and
      Luu, Anh Tuan  and
      Nguyen, Dat Quoc",
    editor = "Che, Wanxiang  and
      Nabende, Joyce  and
      Shutova, Ekaterina  and
      Pilehvar, Mohammad Taher",
    booktitle = "Findings of the Association for Computational Linguistics: ACL 2025",
    month = jul,
    year = "2025",
    address = "Vienna, Austria",
    publisher = "Association for Computational Linguistics",
    url = "https://aclanthology.org/2025.findings-acl.738/",
    doi = "10.18653/v1/2025.findings-acl.738",
    pages = "14322--14329",
    ISBN = "979-8-89176-256-5"
}

@inproceedings{
zhu2025reasoning,
title={Reasoning by Superposition: A Theoretical Perspective on Chain of Continuous Thought},
author={Hanlin Zhu and Shibo Hao and Zhiting Hu and Jiantao Jiao and Stuart Russell and Yuandong Tian},
booktitle={The Thirty-ninth Annual Conference on Neural Information Processing Systems},
year={2025},
url={https://openreview.net/forum?id=UdOEZgWJLc}
}

@inproceedings{
park2025iclr,
title={{ICLR}: In-Context Learning of Representations},
author={Core Francisco Park and Andrew Lee and Ekdeep Singh Lubana and Yongyi Yang and Maya Okawa and Kento Nishi and Martin Wattenberg and Hidenori Tanaka},
booktitle={The Thirteenth International Conference on Learning Representations},
year={2025},
url={https://openreview.net/forum?id=pXlmOmlHJZ}
}

@article{lepori2026language,
  title={Language Models Struggle to Use Representations Learned In-Context},
  author={Lepori, Michael A and Linzen, Tal and Yuan, Ann and Filippova, Katja},
  journal={arXiv preprint arXiv:2602.04212},
  year={2026}
}

@inproceedings{iyer-etal-2026-beyond,
    title = "Beyond Tokens: Concept-Level Training Objectives for {LLM}s",
    author = "Iyer, Laya  and
      Somani, Pranav  and
      Guo, Alice  and
      Jurafsky, Dan  and
      Shani, Chen",
    editor = "Demberg, Vera  and
      Inui, Kentaro  and
      Marquez, Llu{\'i}s",
    booktitle = "Proceedings of the 19th Conference of the {E}uropean Chapter of the {A}ssociation for {C}omputational {L}inguistics (Volume 2: Short Papers)",
    month = mar,
    year = "2026",
    address = "Rabat, Morocco",
    publisher = "Association for Computational Linguistics",
    url = "https://aclanthology.org/2026.eacl-short.34/",
    doi = "10.18653/v1/2026.eacl-short.34",
    pages = "457--474",
    ISBN = "979-8-89176-381-4"
}

@article{zhang2026concept,
  title={Learning Concepts, Not Tokens: Self-Supervised Semantic Alignment for Language Models},
  author={Zhang, Christine and Jurafsky, Dan and Shani, Chen},
  journal={arXiv preprint arXiv:2603.29123v2},
  year={2026}
}

@inproceedings{zhou-etal-2025-graph,
    title = "Each graph is a new language: Graph Learning with {LLM}s",
    author = "Zhou, Huachi  and
      Du, Jiahe  and
      Zhou, Chuang  and
      Yang, Chang  and
      Xiao, Yilin  and
      Xie, Yuxuan  and
      Huang, Xiao",
    editor = "Che, Wanxiang  and
      Nabende, Joyce  and
      Shutova, Ekaterina  and
      Pilehvar, Mohammad Taher",
    booktitle = "Findings of the Association for Computational Linguistics: ACL 2025",
    month = jul,
    year = "2025",
    address = "Vienna, Austria",
    publisher = "Association for Computational Linguistics",
    url = "https://aclanthology.org/2025.findings-acl.902/",
    doi = "10.18653/v1/2025.findings-acl.902",
    pages = "17548--17559",
    ISBN = "979-8-89176-256-5"
}

@article{liu2009learning,
  title={Learning to rank for information retrieval},
  author={Liu, Tie-Yan},
  journal={Foundations and Trends{\textregistered} in Information Retrieval},
  volume={3},
  number={3},
  pages={225--331},
  year={2009},
  publisher={Emerald Publishing Limited}
}

@inproceedings{xia2008listwise,
  title={Listwise approach to learning to rank: theory and algorithm},
  author={Xia, Fen and Liu, Tie-Yan and Wang, Jue and Zhang, Wensheng and Li, Hang},
  booktitle={Proceedings of the 25th international conference on Machine learning},
  pages={1192--1199},
  year={2008}
}

@inproceedings{zhu2020listwise,
  title={Listwise learning to rank by exploring unique ratings},
  author={Zhu, Xiaofeng and Klabjan, Diego},
  booktitle={Proceedings of the 13th international conference on web search and data mining},
  pages={798--806},
  year={2020}
}

@inproceedings{ma2021learning,
  title={Learning-to-rank with partitioned preference: Fast estimation for the Plackett-Luce model},
  author={Ma, Jiaqi and Yi, Xinyang and Tang, Weijing and Zhao, Zhe and Hong, Lichan and Chi, Ed and Mei, Qiaozhu},
  booktitle={International Conference on Artificial Intelligence and Statistics},
  pages={928--936},
  year={2021},
  organization={PMLR}
}

@inproceedings{
ye2025transformers,
title={Transformers Provably Learn Algorithmic Solutions for Graph Connectivity, But Only with the Right Data},
author={Qilin Ye and Deqing Fu and Robin Jia and Vatsal Sharan},
booktitle={Forty-third International Conference on Machine Learning},
year={2026},
url={https://openreview.net/forum?id=AhRuQzYHzY}
}

@inproceedings{
merrill2025a,
title={A Little Depth Goes a Long Way: The Expressive Power of Log-Depth Transformers},
author={William Merrill and Ashish Sabharwal},
booktitle={The Thirty-ninth Annual Conference on Neural Information Processing Systems},
year={2025},
url={https://openreview.net/forum?id=5pHfYe10iX}
}

@article{cui2026latent,
  title={How Do Latent Reasoning Methods Perform Under Weak and Strong Supervision?},
  author={Cui, Yingqian and Dai, Zhenwei and He, Bing and Shi, Zhan and Liu, Hui and Sun, Rui and Liu, Zhiji and Xing, Yue and Tang, Jiliang and Dumoulin, Benoit},
  journal={arXiv preprint arXiv:2602.22441},
  year={2026}
}

@inproceedings{
rizvi2026illusion,
title={The Illusion of Superposition? A Principled Analysis of Latent Thinking in Language Models},
author={Michael Rizvi-Martel and Guillaume Rabusseau and Marius Mosbach},
booktitle={Third Conference on Language Modeling},
year={2026},
url={https://openreview.net/forum?id=49kLxSRDzL}
}

@article{tang2026k,
  title={K-Forcing: Joint Next-K-Token Decoding via Push-Forward Language Modeling},
  author={Tang, Zhiwei and He, Yuanyu and Han, Yizheng and Zhao, Wangbo and Tang, Jiasheng and Wang, Fan and Zhuang, Bohan},
  journal={arXiv preprint arXiv:2606.10820},
  year={2026}
}

@article{xiong2026superthoughts,
  title={SuperThoughts: Reasoning Tokens in Superposition},
  author={Xiong, Zheyang and Garg, Shivam and Yu, Max and Shrivastava, Vaishnavi and Zhao, Haoyu and Kyrillidis, Anastasios and Papailiopoulos, Dimitris},
  journal={arXiv preprint arXiv:2606.13862},
  year={2026}
}

@inproceedings{
ankner2024hydra,
title={Hydra: Sequentially-Dependent Draft Heads for Medusa Decoding},
author={Zachary Ankner and Rishab Parthasarathy and Aniruddha Nrusimha and Christopher Rinard and Jonathan Ragan-Kelley and William Brandon},
booktitle={First Conference on Language Modeling},
year={2024},
url={https://openreview.net/forum?id=FbhjirzvJG}
}

@inproceedings{
wei2026provable,
title={Provable Benefits of {RLVR} over {SFT} for Reasoning Models: Learning to Backtrack Efficiently},
author={Stanley Wei and Juno Kim},
booktitle={ICLR 2026 2nd Workshop on Deep Generative Model in Machine Learning: Theory, Principle and Efficacy},
year={2026},
url={https://openreview.net/forum?id=Xawq2LhmeN}
}

@inproceedings{
zhang2026nitp,
title={{NITP}: Next Implicit Token Prediction for {LLM} Pre-training},
author={Xiangdong Zhang and Debing Zhang and Shaofeng Zhang and Xiaohan Qin and Yu Cheng and Junchi Yan},
booktitle={Forty-third International Conference on Machine Learning},
year={2026},
url={https://openreview.net/forum?id=oPQmCKS1tV}
}

@article{samragh2025your,
  title={Your {LLM} knows the future: Uncovering its multi-token prediction potential},
  author={Samragh, Mohammad and Kundu, Arnav and Harrison, David and Nishu, Kumari and Naik, Devang and Cho, Minsik and Farajtabar, Mehrdad},
  journal={arXiv preprint arXiv:2507.11851},
  year={2025}
}

@article{noci2026thinking,
  title={Thinking into the Future: Latent Lookahead Training for Transformers},
  author={Noci, Lorenzo and Bachmann, Gregor and Moosavi-Dezfooli, Seyed-Mohsen and Nabi, Moin},
  journal={arXiv preprint arXiv:2603.20219},
  year={2026}
}

@article{bechler2026lost,
  title={Lost in Tokenization: Fundamental Trade-offs in Graph Tokenization for Transformers},
  author={Bechler-Speicher, Maya and Yehudai, Gilad and Harari, Gil and Sanford, Clayton and Globerson, Amir and Bruna, Joan},
  journal={arXiv preprint arXiv:2605.22471},
  year={2026}
}

@inproceedings{
yehudai2026depth,
title={Depth-Width Tradeoffs for Transformers on Graph Tasks},
author={Gilad Yehudai and Clayton Sanford and Maya Bechler-Speicher and Orr Fischer and Ran Gilad-Bachrach and Amir Globerson},
booktitle={The Thirty-ninth Annual Conference on Neural Information Processing Systems},
year={2025},
url={https://openreview.net/forum?id=A2pmNL7L1E}
}

@inproceedings{wen2026frontier,
  title={Frontier Language Models Struggle to Copy: Text Can Be Better Viewed in 2D},
  author={Wen, Haodong and Zhang, Yiran and Chen, Yingfa and Lyu, Kaifeng},
  booktitle={ICML 2026 Workshop on Foundations of Deep Generative Models: Understanding Memorization, Generalization, and Reasoning},
  year={2026},
}

@inproceedings{
wang2024grokking,
title={Grokking of Implicit Reasoning in Transformers: A Mechanistic Journey to the Edge of Generalization},
author={Boshi Wang and Xiang Yue and Yu Su and Huan Sun},
booktitle={The Thirty-eighth Annual Conference on Neural Information Processing Systems},
year={2024},
url={https://openreview.net/forum?id=D4QgSWxiOb}
}

@inproceedings{
london2025pause,
title={Pause Tokens Strictly Increase the Expressivity of Constant-Depth Transformers},
author={Charles London and Varun Kanade},
booktitle={The Thirty-ninth Annual Conference on Neural Information Processing Systems},
year={2025},
url={https://openreview.net/forum?id=eG5oh8l1WZ}
}

@inproceedings{
suleymanzade2026mux,
title={{MUX}: Continuous Reasoning via Multiplexed Tokens},
author={Ayhan Suleymanzade and Halil Alperen Gozeten and Ismail Ilkan Ceylan and Jinwoo Kim},
booktitle={ICLR 2026 Workshop on Logical Reasoning of Large Language Models},
year={2026},
url={https://openreview.net/forum?id=Ee78Uqsq0a}
}

@inproceedings{yang-etal-2018-hotpotqa,
    title = "{H}otpot{QA}: A Dataset for Diverse, Explainable Multi-hop Question Answering",
    author = "Yang, Zhilin  and
      Qi, Peng  and
      Zhang, Saizheng  and
      Bengio, Yoshua  and
      Cohen, William  and
      Salakhutdinov, Ruslan  and
      Manning, Christopher D.",
    editor = "Riloff, Ellen  and
      Chiang, David  and
      Hockenmaier, Julia  and
      Tsujii, Jun{'}ichi",
    booktitle = "Proceedings of the 2018 Conference on Empirical Methods in Natural Language Processing",
    month = oct # "-" # nov,
    year = "2018",
    address = "Brussels, Belgium",
    publisher = "Association for Computational Linguistics",
    url = "https://aclanthology.org/D18-1259/",
    doi = "10.18653/v1/D18-1259",
    pages = "2369--2380"
}

@inproceedings{de-cao-etal-2019-question,
    title = "Question Answering by Reasoning Across Documents with Graph Convolutional Networks",
    author = "De Cao, Nicola  and
      Aziz, Wilker  and
      Titov, Ivan",
    editor = "Burstein, Jill  and
      Doran, Christy  and
      Solorio, Thamar",
    booktitle = "Proceedings of the 2019 Conference of the North {A}merican Chapter of the Association for Computational Linguistics: Human Language Technologies, Volume 1 (Long and Short Papers)",
    month = jun,
    year = "2019",
    address = "Minneapolis, Minnesota",
    publisher = "Association for Computational Linguistics",
    url = "https://aclanthology.org/N19-1240/",
    doi = "10.18653/v1/N19-1240",
    pages = "2306--2317"
}

@inproceedings{ding-etal-2019-cognitive,
    title = "Cognitive Graph for Multi-Hop Reading Comprehension at Scale",
    author = "Ding, Ming  and
      Zhou, Chang  and
      Chen, Qibin  and
      Yang, Hongxia  and
      Tang, Jie",
    editor = "Korhonen, Anna  and
      Traum, David  and
      M{\`a}rquez, Llu{\'i}s",
    booktitle = "Proceedings of the 57th Annual Meeting of the Association for Computational Linguistics",
    month = jul,
    year = "2019",
    address = "Florence, Italy",
    publisher = "Association for Computational Linguistics",
    url = "https://aclanthology.org/P19-1259/",
    doi = "10.18653/v1/P19-1259",
    pages = "2694--2703"
}

@inproceedings{sun-etal-2018-open,
    title = "Open Domain Question Answering Using Early Fusion of Knowledge Bases and Text",
    author = "Sun, Haitian  and
      Dhingra, Bhuwan  and
      Zaheer, Manzil  and
      Mazaitis, Kathryn  and
      Salakhutdinov, Ruslan  and
      Cohen, William",
    editor = "Riloff, Ellen  and
      Chiang, David  and
      Hockenmaier, Julia  and
      Tsujii, Jun{'}ichi",
    booktitle = "Proceedings of the 2018 Conference on Empirical Methods in Natural Language Processing",
    month = oct # "-" # nov,
    year = "2018",
    address = "Brussels, Belgium",
    publisher = "Association for Computational Linguistics",
    url = "https://aclanthology.org/D18-1455/",
    doi = "10.18653/v1/D18-1455",
    pages = "4231--4242"
}

@inproceedings{sun-etal-2019-pullnet,
    title = "{P}ull{N}et: Open Domain Question Answering with Iterative Retrieval on Knowledge Bases and Text",
    author = "Sun, Haitian  and
      Bedrax-Weiss, Tania  and
      Cohen, William",
    editor = "Inui, Kentaro  and
      Jiang, Jing  and
      Ng, Vincent  and
      Wan, Xiaojun",
    booktitle = "Proceedings of the 2019 Conference on Empirical Methods in Natural Language Processing and the 9th International Joint Conference on Natural Language Processing (EMNLP-IJCNLP)",
    month = nov,
    year = "2019",
    address = "Hong Kong, China",
    publisher = "Association for Computational Linguistics",
    url = "https://aclanthology.org/D19-1242/",
    doi = "10.18653/v1/D19-1242",
    pages = "2380--2390"
}

@inproceedings{saxena-etal-2020-improving,
    title = "Improving Multi-hop Question Answering over Knowledge Graphs using Knowledge Base Embeddings",
    author = "Saxena, Apoorv  and
      Tripathi, Aditay  and
      Talukdar, Partha",
    editor = "Jurafsky, Dan  and
      Chai, Joyce  and
      Schluter, Natalie  and
      Tetreault, Joel",
    booktitle = "Proceedings of the 58th Annual Meeting of the Association for Computational Linguistics",
    month = jul,
    year = "2020",
    address = "Online",
    publisher = "Association for Computational Linguistics",
    url = "https://aclanthology.org/2020.acl-main.412/",
    doi = "10.18653/v1/2020.acl-main.412",
    pages = "4498--4507"
}

@inproceedings{lin-etal-2018-multi,
    title = "Multi-Hop Knowledge Graph Reasoning with Reward Shaping",
    author = "Lin, Xi Victoria  and
      Socher, Richard  and
      Xiong, Caiming",
    editor = "Riloff, Ellen  and
      Chiang, David  and
      Hockenmaier, Julia  and
      Tsujii, Jun{'}ichi",
    booktitle = "Proceedings of the 2018 Conference on Empirical Methods in Natural Language Processing",
    month = oct # "-" # nov,
    year = "2018",
    address = "Brussels, Belgium",
    publisher = "Association for Computational Linguistics",
    url = "https://aclanthology.org/D18-1362/",
    doi = "10.18653/v1/D18-1362",
    pages = "3243--3253"
}

@inproceedings{han-etal-2025-reasoning,
    title = "Reasoning with Graphs: Structuring Implicit Knowledge to Enhance {LLM}s Reasoning",
    author = "Han, Haoyu  and
      Xie, Yaochen  and
      Liu, Hui  and
      Tang, Xianfeng  and
      Nag, Sreyashi  and
      Headden, William  and
      Li, Yang  and
      Luo, Chen  and
      Ji, Shuiwang  and
      He, Qi  and
      Tang, Jiliang",
    editor = "Che, Wanxiang  and
      Nabende, Joyce  and
      Shutova, Ekaterina  and
      Pilehvar, Mohammad Taher",
    booktitle = "Findings of the Association for Computational Linguistics: ACL 2025",
    month = jul,
    year = "2025",
    address = "Vienna, Austria",
    publisher = "Association for Computational Linguistics",
    url = "https://aclanthology.org/2025.findings-acl.1319/",
    doi = "10.18653/v1/2025.findings-acl.1319",
    pages = "25698--25714",
    ISBN = "979-8-89176-256-5"
}

@inproceedings{
das2018go,
title={Go for a Walk and Arrive at the Answer: Reasoning Over Paths in Knowledge Bases using Reinforcement Learning},
author={Rajarshi Das and Shehzaad Dhuliawala and Manzil Zaheer and Luke Vilnis and Ishan Durugkar and Akshay Krishnamurthy and Alex Smola and Andrew McCallum},
booktitle={International Conference on Learning Representations},
year={2018},
url={https://openreview.net/forum?id=Syg-YfWCW},
}

@inproceedings{wen-etal-2024-mindmap,
    title = "{M}ind{M}ap: Knowledge Graph Prompting Sparks Graph of Thoughts in Large Language Models",
    author = "Wen, Yilin  and
      Wang, Zifeng  and
      Sun, Jimeng",
    editor = "Ku, Lun-Wei  and
      Martins, Andre  and
      Srikumar, Vivek",
    booktitle = "Proceedings of the 62nd Annual Meeting of the Association for Computational Linguistics (Volume 1: Long Papers)",
    month = aug,
    year = "2024",
    address = "Bangkok, Thailand",
    publisher = "Association for Computational Linguistics",
    url = "https://aclanthology.org/2024.acl-long.558/",
    doi = "10.18653/v1/2024.acl-long.558",
    pages = "10370--10388"
}

@inproceedings{cheng-etal-2024-structure,
    title = "Structure Guided Prompt: Instructing Large Language Model in Multi-Step Reasoning by Exploring Graph Structure of the Text",
    author = "Cheng, Kewei  and
      Ahmed, Nesreen K.  and
      Willke, Theodore L.  and
      Sun, Yizhou",
    editor = "Al-Onaizan, Yaser  and
      Bansal, Mohit  and
      Chen, Yun-Nung",
    booktitle = "Proceedings of the 2024 Conference on Empirical Methods in Natural Language Processing",
    month = nov,
    year = "2024",
    address = "Miami, Florida, USA",
    publisher = "Association for Computational Linguistics",
    url = "https://aclanthology.org/2024.emnlp-main.528/",
    doi = "10.18653/v1/2024.emnlp-main.528",
    pages = "9407--9430"
}

@inproceedings{
saparov2023language,
title={Language Models Are Greedy Reasoners: A Systematic Formal Analysis of Chain-of-Thought},
author={Abulhair Saparov and He He},
booktitle={The Eleventh International Conference on Learning Representations },
year={2023},
url={https://openreview.net/forum?id=qFVVBzXxR2V}
}

@inproceedings{lin-etal-2025-implicit,
    title = "Implicit Reasoning in Transformers is Reasoning through Shortcuts",
    author = "Lin, Tianhe  and
      Xie, Jian  and
      Yuan, Siyu  and
      Yang, Deqing",
    editor = "Che, Wanxiang  and
      Nabende, Joyce  and
      Shutova, Ekaterina  and
      Pilehvar, Mohammad Taher",
    booktitle = "Findings of the Association for Computational Linguistics: ACL 2025",
    month = jul,
    year = "2025",
    address = "Vienna, Austria",
    publisher = "Association for Computational Linguistics",
    url = "https://aclanthology.org/2025.findings-acl.493/",
    doi = "10.18653/v1/2025.findings-acl.493",
    pages = "9470--9487",
    ISBN = "979-8-89176-256-5"
}

@inproceedings{
yang2025number,
title={Number Cookbook: Number Understanding of Language Models and How to Improve It},
author={Haotong Yang and Yi Hu and Shijia Kang and Zhouchen Lin and Muhan Zhang},
booktitle={The Thirteenth International Conference on Learning Representations},
year={2025},
url={https://openreview.net/forum?id=BWS5gVjgeY}
}

@inproceedings{yuchi-etal-2026-llms,
    title = "{LLM}s Know More About Numbers than They Can Say",
    author = "Yuchi, Fengting  and
      Du, Li  and
      Eisner, Jason",
    editor = "Demberg, Vera  and
      Inui, Kentaro  and
      Marquez, Llu{\'i}s",
    booktitle = "Proceedings of the 19th Conference of the {E}uropean Chapter of the {A}ssociation for {C}omputational {L}inguistics (Volume 2: Short Papers)",
    month = mar,
    year = "2026",
    address = "Rabat, Morocco",
    publisher = "Association for Computational Linguistics",
    url = "https://aclanthology.org/2026.eacl-short.47/",
    doi = "10.18653/v1/2026.eacl-short.47",
    pages = "659--673",
    ISBN = "979-8-89176-381-4"
}

@inproceedings{yang-etal-2025-large,
    title = "Do Large Language Models Perform Latent Multi-Hop Reasoning without Exploiting Shortcuts?",
    author = "Yang, Sohee  and
      Kassner, Nora  and
      Gribovskaya, Elena  and
      Riedel, Sebastian  and
      Geva, Mor",
    editor = "Che, Wanxiang  and
      Nabende, Joyce  and
      Shutova, Ekaterina  and
      Pilehvar, Mohammad Taher",
    booktitle = "Findings of the Association for Computational Linguistics: ACL 2025",
    month = jul,
    year = "2025",
    address = "Vienna, Austria",
    publisher = "Association for Computational Linguistics",
    url = "https://aclanthology.org/2025.findings-acl.205/",
    doi = "10.18653/v1/2025.findings-acl.205",
    pages = "3971--3992",
    ISBN = "979-8-89176-256-5"
}

@inproceedings{ju-etal-2024-investigating,
    title = "Investigating Multi-Hop Factual Shortcuts in Knowledge Editing of Large Language Models",
    author = "Ju, Tianjie  and
      Chen, Yijin  and
      Yuan, Xinwei  and
      Zhang, Zhuosheng  and
      Du, Wei  and
      Zheng, Yubin  and
      Liu, Gongshen",
    editor = "Ku, Lun-Wei  and
      Martins, Andre  and
      Srikumar, Vivek",
    booktitle = "Proceedings of the 62nd Annual Meeting of the Association for Computational Linguistics (Volume 1: Long Papers)",
    month = aug,
    year = "2024",
    address = "Bangkok, Thailand",
    publisher = "Association for Computational Linguistics",
    url = "https://aclanthology.org/2024.acl-long.486/",
    doi = "10.18653/v1/2024.acl-long.486",
    pages = "8987--9001"
}

@article{balesni2024lessons,
  title={Lessons from studying two-hop latent reasoning},
  author={Balesni, Mikita and Korbak, Tomek and Evans, Owain},
  journal={arXiv preprint arXiv:2411.16353},
  year={2024}
}

@inproceedings{sun2024think,
  title={Think-on-graph: Deep and responsible reasoning of large language model on knowledge graph},
  author={Sun, Jiashuo and Xu, Chengjin and Tang, Lumingyuan and Wang, Saizhuo and Lin, Chen and Gong, Yeyun and Ni, Lionel and Shum, Heung-Yeung and Guo, Jian},
  booktitle={International Conference on Learning Representations},
  volume={2024},
  pages={3868--3898},
  year={2024}
}

@article{soviany2022curriculum,
  title={Curriculum learning: A survey},
  author={Soviany, Petru and Ionescu, Radu Tudor and Rota, Paolo and Sebe, Nicu},
  journal={International Journal of Computer Vision},
  volume={130},
  number={6},
  pages={1526--1565},
  year={2022},
  publisher={Springer}
}

@article{he2026reasoning,
  title={Reasoning with latent tokens in diffusion language models},
  author={He, Andre and Welleck, Sean and Fried, Daniel},
  journal={arXiv preprint arXiv:2602.03769},
  year={2026}
}

@inproceedings{
wang2026the,
title={The Power of Power Law: Asymmetry Enables Compositional Reasoning},
author={Zixuan Wang and Xingyu Dang and Jason D. Lee and Kaifeng Lyu},
booktitle={Forty-third International Conference on Machine Learning},
year={2026},
url={https://openreview.net/forum?id=K83orsg2X9}
}

@article{newell2007computer,
  title={Computer science as empirical inquiry: symbols and search},
  author={Newell, Allen and Simon, Herbert A.},
  journal={Communications of the ACM},
  volume={19},
  number={3},
  pages={113--126},
  year={1976},
  publisher={ACM}
}

@book{newell1972human,
  title={Human problem solving},
  author={Newell, Allen and Simon, Herbert Alexander},
  volume={104},
  number={9},
  year={1972},
  publisher={Prentice-hall Englewood Cliffs, NJ}
}

@inproceedings{
minegishi2026emergent,
title={Emergent Analogical Reasoning in Transformers},
author={Gouki Minegishi and Jingyuan Feng and Hiroki Furuta and Takeshi Kojima and Yusuke Iwasawa and Yutaka Matsuo},
booktitle={Forty-third International Conference on Machine Learning},
year={2026},
url={https://openreview.net/forum?id=GxTkgMBiz8}
}
\bibliographystyle{iclr2027_conference}


\newpage

\appendix
\let\oldsection\section
\renewcommand{\section}{\FloatBarrier\oldsection}

\startcontents[appendixtoc]
\section*{Appendix Table of Contents}
\printcontents[appendixtoc]{}{1}{} 
 
\section{Notation and Abbreviations}

Notation:
\begin{itemize}
    \item  $\mathcal{G} = (\mathcal{N},\, \mathcal{E})$: the graph with a set of nodes and edges.  We use the style $\mathcal{G}$ to indicate sets.  Note that we are considering the in-context version of the task, so we sample many different graphs $\mathcal{G}_b$, where $b \leq 75$M. We drop the specific index since we are almost always talking about a particular graph instance.
    \item $G$, $Q$, $N$, and $E$: the sequentially tokenized graph, query, node list and edge list.
    \item $[Q, E]$: the concatenation of $Q$ and $E$.
    \item $|\mathcal{N}|$:  the number of nodes.  Also, the maximum number of nodes in a given experiment.
    \item $\mathcal{V}$: the set of vocabulary tokens.  In practice, $|\mathcal{V}| = |\mathcal{N}| + 23$ special tokens (not all special tokens are used for all tasks; however, they are included in each model for simplicity).  
    \item  $\pi =u_0 u_1 u_2 \dots u_{k-1} u_{k} = s u_1 u_2 \dots u_{k-1} t$  : the shortest path of length $k+1$, starting at $s$ and ending at $t$.  We refer to both of these as nodes and tokens, depending on the context.  
    \item $\pi_{<u_i}$: a sub-path of $\pi$
    \item $k$: the number of hops, generally equivalent to the path length minus one.  Also, the maximum path length in a given experiment.  
    \item $k'$: a sampled or intermediate value of $\leq k$
    \item $|\mathcal{N}|'$: a sampled or  intermediate number of nodes $\leq|\mathcal{N}|$
    \item $d$: the degree of a node or the degree parameter of path-star graphs, depending on context.   
    \item $A$: the adjacency matrix.
    \item $D$: the distance matrix. 
    \item $c(u, \pi_{\leq u})$: the {\em choice set} of node $u$.
    \item $c(\pi)$: the number of choice nodes in $\pi$.
    \item $x$, $\phi$, $z$: the input, positional embedding, or concatenated ids.  
    \item $L$: the number of layers in the model.
\end{itemize}

Abbreviations:
\begin{itemize}
    \item NTP and MTP: next- and multi-token prediction.
    \item CHC: the Clever Hans Cheat.  A topologically-dependent learning shortcut \citep{bachmann2024the}.
    \item MMHT and MMLT: Mixed Multihop Training and Mixed Multi-length Training (\A \ref{appx:MMLT}).  
    \item SSE: Sequential Sample Efficiency.  The number of input-target supervision pairs that contribute to learning in a single instance.
    \item ERG, EG, and PSG: Erd\H{o}s-Renyi, Euclidean, and path-star graph topologies.  
    \item SEAN and STAN: one-dimensional sequential annotation and multidimensional structural annotation.  
    \item ACC, SACC, and TACC: per-token accuracy, sequential accuracy, and task accuracy (which depends on the task, and so is different for shortest path, Khops, Khops-Gen, and the scratchpad tasks). 
\end{itemize}

\subsection{A Note on MMHT and Curriculum Terminology}\label{appx:MMHT-note}

MMHT has been called `composition' or `subtask decomposition' \citep{frydenlund-2025-language, wies2023subtask}.  Others have called it a `curriculum' in a broad sense, or `implicit curriculum via data mixture' \citep{pmlr-v291-wang25a, yao-etal-2025-language-models}.  Subtask decomposition addresses how a task can be recursively decomposed and how this can be expressed implicitly and intra-sample via sequential supervision, which is a property of the task.  Curriculum learning is generally explicit and inner-sample, and more often concerned with the order the samples are presented in, i.e.\ easy to hard \citep{bengio2009curriculum}.  
Thus, we use `mixed multihop training' as a generic name to cover both inner- and intra-sample data mixing, which is not concerned with order or explicitness, or with whether the specific task is compositional (the task can be fuzzily compositional \citep{frydenlund-2025-language}). 

We employ a true curriculum data sampling algorithm for sampling path lengths. Because tasks can also have implicit MMHT due to being truly compositional or via MTP this distinction is important.

In general, `curriculum' is an overloaded term, and no longer applies to a specific pre-training supervision method \citep{bengio2009curriculum}.  It can be applied to in-context learning \citep{luo-etal-2025-lets, ma-etal-2025-problem, lee2025distinct}.  \citet{hao2025training} used a masking curriculum where parts of reasoning traces are systematically removed to promote the model relying on continuous CoT inputs instead.  It can also be applied across tasks in a multitask setting \citep{liu2026language}.  Note that they use  `curriculum' to describe a phenomenon where models naturally learn tasks in order of difficulty, instead of the curriculum being a form of training supervision.

\section{Further Discussion of Contributions, Limitations, and Scope}\label{app:limitations}

We summarize our contributions for quick reference.

\begin{enumerate}

\item {\bf Theoretical Model Framing.} 
We spell out the theoretical model underlying the study of multihop reasoning via shortest-path search, which prior work assumes but leaves unstated. 
We inherit this controlled setting \citep[][\A \ref{appx:t-model}]{bachmann2024the, saparov2025transformers}.

\item {\bf Topological shortcuts under NTP.}
We generalize the Clever Hans Cheat from path-star graphs to arbitrary topologies via choice nodes, giving an SSE of $c(\pi)$ which is thus strongly topology-dependent under NTP (\S \ref{sec::SSE}).
\citet{bachmann2024the}'s speculation that the problem is broad has spurred much follow-up work (\A \ref{appx:PST-prior-work}); by characterizing exactly when the CHC arises, and how it affects learning, we {\em delimit} its extent.

\item {\bf MTP is optimal, NTP is not.}
We prove that MTP attains an optimal SSE and hop mixture over the sampled path, giving $k(k+1)/2 - (k - c(\pi))$ terms after correcting for those the CHC absorbs (\S \ref{sec:sse-mtp} and \A \ref{appx:proof-optimal}).
This residual topological dependency is linear, and thus dominated by the quadratic MTP terms as $k$ grows, making MTP nearly topology-independent while NTP remains strongly topology-dependent (\S \ref{sec:sse-mtp}).
We also show that masked discrete diffusion yields a strict subset of MTP's terms (\A \ref{appx:discrete-proof}).

\item {\bf A novel MTP method.}
We introduce a probabilistic listwise Plackett-Luce formulation of Ranking-into-the-Future, in contrast to the non-probabilistic pairwise and listwise losses of \citet{frydenlund-2025-language} and \citet{zuhri2025predicting} respectively (\A \ref{appx:plackett-luce}).

\item {\bf Fragmented routing and the node list.}
We show that the causal constraint is severely compounded by graph sparsity and edge-list shuffling, deferring routing until most of $E$ has been observed and exhausting the model's sequential computation (\S \ref{sec:causal-constraint}).
This motivates including the node list ($+N$)  to bypass the causal constraint and accumulate edge information.

\item {\bf Permutation sensitivity and structural annotation.}
Positional embeddings break the permutation invariance inherent to the task (\S \ref{sec:STAN}).
This motivates structural annotation (STAN), which makes each edge a unique {\em super-token}, removing positional leakage.

\item {\bf Breaking the scaling wall.}
{\em This is the focus of this work.}
Prior work could not stably learn the task beyond $50$ nodes.
STAN and $+N$ account for the largest empirical gains, and together with MTP they break this wall.
We show that (a) we learn stably up to $|\mathcal{N}| = 125$ in about $1/3$ the steps (\S \ref{sec:experiments}).  {\bf Note that the scaling trend is super-linear (Fig.\ \ref{fig:er-new}), so scaling to $2.5\times$ the graph size is a substantial gain}; (b) the variance issue between seeds reported by prior work is solved, greatly reducing the cost of running future experiments (\A \ref{appx:comp-cost}); (c) MTP beats NTP across scale with this gap growing with $k$ and $|\mathcal{N}|$; (d) {\bf the performance of MTP is orthogonal to the tokenization.  Thus, they solve different underlying scaling problems}; and (e) we solve path-star graphs with NTP well beyond prior work using only inter-sample MMHT and MTP makes them trivial at all scales.

\item {\bf Auxiliary tasks (\A \ref{appx:aux-tasks}).}
We consider three further latent multihop tasks: Khops (\A \ref{appx:khops}) \citep{pmlr-v235-sanford24a}, Khops-Gen (a generative variant we introduce, \A \ref{appx:khops-gen}), and a BFS scratchpad (BFS-SP, \A \ref{appx:bfs-sp}).
These serve two purposes.
First, they show that MTP's benefits are not limited to shortest-path search, but hold across symbolic multihop tasks.
Second, they let us show that while BFS-SP is learnable, it suffers its own tokenization issues and its sequence length grows too quickly to scale. More importantly, we show it collapses the latent reasoning graph into a linear chain, which breaks the established theoretical model. (\A \ref{appx:sp-collapse}).  Thus, it is not a viable `solution'.  
We show this by reducing it to Khops-Gen.
Additionally, we find that hop-length is not the only dimension of difficulty that matters, as mixed multi-length training (MMLT, \A \ref{appx:MMLT}) is necessary for learning Khops-Gen.

\end{enumerate}

This work considers a scaling issue encountered in prior work.
We overcame that scaling wall via both alternative tokenization and MTP.
{\bf The fact that MTP's advantages are nearly topology-independent is an unexpected finding.}
We analyzed MTP to explain why it ameliorates a known, topology-{\em specific} pathology, and the topology-independence fell out of that analysis.
However, this leads to a natural conjecture.
We motivated graph search as a model for multihop reasoning precisely because the latent reasoning graph has unknown topology, so no topological assumptions are permissible.
If MTP's advantages do not depend on topology, then they should hold for any multihop reasoning task, making MTP a principled training objective for multihop reasoning in general.
Had this been our starting hypothesis, this would be a different paper, with methodology built around testing that conjecture rather than around scaling.

Evaluating it is non-trivial.
Our results rest on the fact that each target token corresponds to exactly one hop, so a window spanning $k$ tokens also spans $k$ hops.
This correspondence breaks for natural language.
A potential solution is non-contiguous windows \citep{liu2025lmtp}, but designing hop-aligned targets for arbitrary tasks would itself be a non-trivial research contribution.
{\em So it remains a conjecture and not an established claim.}

Regardless, this paper would still be necessary.
It is by investigating the scaling issue that we establish MTP's benefits, both theoretically and empirically.
But MTP alone does not break the scaling wall.
The tokenization methods account for the majority of the empirical gains, and they are orthogonal to MTP, so any work concerned with scaling must consider both.
This orthogonality makes our methodology disjointed,
but this is a reflection of the underlying causes being disjointed.  
This also means MTP's benefits hold independently of our tokenization changes.
{\em A paper on MTP alone would be cleaner, but, unfortunately, the most interesting finding is not necessarily the most important one in terms of performance gain.}

{\bf The most significant limitation is that our motivating hypotheses do not fully explain why STAN and $+N$ work so well.}  Both methods are well motivated.
Fragmented routing and the loss of permutation invariance are real properties of the setting, and we show that both hold (\S \ref{sec:causal-constraint}, \S \ref{sec:STAN}).
The question is not whether these issues exist, but the extent to which they contribute to the scaling wall.

We ablate using two model parameterizations in an attempt to answer this, since each hypothesis predicts where its method should have no effect.
The causal constraint does not apply to prefix LMs, and causal LMs cannot be permutation-invariant.
The design is thus falsifiable, but it falsified our own explanations.
$+N$ still improves prefix LMs, and STAN still improves causal LMs.
Importantly, the counter-evidence shows that each benefit extends beyond its motivating hypothesis, not that the hypothesis is wrong.
{\bf Indeed, prefix LMs outperform causal LMs on ERGs (Fig.\ \ref{fig:er-new} vs.\ Fig.\ \ref{fig:er-new-decoder-only}), which shows that the causal constraint matters.}
Note that both issues ultimately stem from the causal constraint; 
Fragmented routing arises from the interaction of causal masking with graph sparsity and tokenization, and permutation invariance requires both the removal of causal masking and positional embeddings.
Thus, while this design can falsify a hypothesis, it cannot isolate the two, since changing the parameterization changes the conditions for both at once.

STAN in particular is hard to ablate further as it does three modifications that are intrinsically coupled: it moves positional information into the annotation channel, it makes $E$ permutation-invariant, and it reduces $E$ to $1/3$ its length.  Length is an interesting factor as it affects expressiveness \citep{sanford2024understanding, merrill2024the, london2025pause}.  \citet{yehudai2026depth} likewise changed the tokenization and the token count together and attributed the gain to the tokenization.
Note also that $+N$ increases the sequence length while STAN decreases it, and both improve performance; thus, raw length changes alone do not explain their performance.
Also, STAN removes the variance across trials and induces the expected behaviour, where convergence scales with $k$ (\S \ref{sec:experiments}).
The baseline does not behave this way.\footnote{This {\em may} be an artifact of the max-step cut-off.  However, as the spiked PSGs results show, training longer may not necessarily show a gradual pattern.}
A task whose difficulty grows with the number of hops should look like the former, not the latter.
The deficiency therefore lies in the original tokenization rather than in the task, even if we cannot say precisely what that deficiency is.

We thus have strong positive empirical results but a negative result for our hypotheses.
We leave closing this gap to future work.
We note four things.
First, the ablations still establish the tokenization as a cause of the scaling wall.
The same model, task, and data learn differently under different tokenizations, so the failure lies in the tokenization and what remains open is which of its properties is responsible.
Second, fragmented routing and the loss of permutation invariance are real issues, so these experiments were necessary regardless of which hypothesis ultimately explains the gains.
Third, neither method breaks the theoretical model (\A \ref{appx:bfs-sp}).
Finally, we emphasize that the MTP results are theoretically grounded and orthogonal to the tokenization, and thus hold with or without STAN and $+N$.

{\bf We observe the lockout phenomenon only on PSGs}.
This is a novel finding, only made visible by scaling NTP on PSGs beyond prior work (as prior models fail to the CHC before lockout can manifest).
Unlike the CHCH, its mechanism, where targets of different lengths cease contributing once solved, is not obviously tied to the PSG topology.
However, since PSGs are an adversarial topology to learning under NTP, we may only see this phenomenon in this adversarial setting.
We suspect looped models would solve this, as tying parameters across hops forces the model to reuse the same computation across hops \citep{giannou2023looped, pmlr-v235-back-de-luca24a, yang2024looped, fan2025looped, saunshi2025reasoning, zhu2025scaling}.  We do not test this as it would require training an entirely new set of models.
Note that this concerns in-domain learning dynamics and is separate from the case for looped models in OOD settings (below).

{\bf We cannot fully formalize learnability.}
We demonstrate scalability trends empirically.
On the theoretical side, we formalize some characteristics of learnability, showing that sequential sample efficiency is strongly topology-dependent under NTP, but becomes nearly topology-independent and increases substantially under MTP.
However, other characteristics that affect learnability remain unformalized, such as the number of paths from a choice node \citep{xu2026depth}.
We thus cannot formalize learnability, and so scalability, as one can for simpler tasks \citep{pmlr-v291-wang25a}.
Given the theoretical model, such a formalization would be a holy grail for learnability theory. We make partial progress towards this; 
however, the larger point, and a key takeaway from our work, is
that precisely specifying learnability for this task is very difficult.

We address other limitations, which we consider as questions of scope.  

{\bf Why did you not consider out-of-distribution generalization?}
{\em Our focus is the scaling wall, which is an in-domain failure.}
Prior work could not learn the task at scale even when training and testing on the same distribution.
Solving the scaling issue in-domain is a prerequisite to asking anything about OOD.
While this makes OOD evaluation an obvious next step, it also places it beyond the scope of the motivating phenomenon of this work.
Moreover, OOD evaluation is not trivial.
There are at least three OOD dimensions, and each is difficult to evaluate for its own reasons.

The first is path length and graph size.
This requires positional embeddings capable of length generalization \citep{ruoss-etal-2023-randomized, li2024functional}.
We use learnt embeddings because they carry no inductive bias, which lets us rule them out as a cause of the scaling wall.
Using OOD-capable embeddings only makes sense now that the scaling wall is solved, and would have otherwise been a confound for the question we were asking.
Beyond this, we know a static model cannot scale to arbitrary $k$, so looped models would be required \citep{giannou2023looped, saunshi2025reasoning, fan2025looped}.
But looped models tie parameters across latent hops, which changes in-domain learnability in its own right.
Any OOD result would thus be entangled with an in-domain change, which is a further confound.
In both cases, we would have to train an entirely new set of models, as this cannot be evaluated with the ones we have.
{\bf Note this is the only OOD question that actually concerns scale.}

The second is topology.
\citet{saparov2025transformers} consider exactly this, training on one topology to test generalization to another, and it did not solve the scaling wall.
It also raises the question of which topologies support generalization to others.
Our analysis of choice nodes suggests trees may be the best candidate, but this is beyond the in-domain scaling issue.

The third is task.
Does learning to search help on other graph tasks?
Search is a fundamental building block, so it is reasonable to expect that it does, but this too is beyond the scaling question.

{\bf Why did you not consider other topologies?}
Adding topologies multiplies the experiments required, leading to experimental scope creep.
We consider three topologies, whereas prior work prioritize one \citep{bachmann2024the, saparov2025transformers}.
This is because we wish to answer questions about how topology affects scale.
The CHC \citep{bachmann2024the} and our generalization to arbitrary topology make path-star graphs the most adversarial topology to NTP, which is why they get so much attention in the paper.
We then show MTP is nearly topology-independent.
Path length is what matters for scaling, so the relevant topologies are those supporting long paths.
This rules out dense and small-world graphs, where most paths are short.
It also makes deep trees computationally prohibitive, since long paths require exponentially more nodes. 
We would consider any topology with a property expected to contradict the theory, but testing topologies without such a reason has diminishing returns at a large computational cost in our experimental setup(\A \ref{appx:comp-cost}).

{\bf Why did you not consider other tasks?}
Connectivity, connected components, and reachability all reduce to shortest path search, so a model that learns that should learn these.
However, determining the correct tokenization for these tasks is non-trivial, as we see both with direct shortest path and the BFS scratchpad. 
Weighted shortest paths in the style of Dijkstra's algorithm further require tokenizing edge weights, which introduces a representational choice we would then have to control for.
{\em Our focus is not on benchmarking language models on various graph tasks.}

{\bf Why not report FLOPs?} We report steps rather than throughput because our claim concerns supervision; thus, FLOPs are not the appropriate metric. This is not a benchmarking paper.

{\bf Why not consider other MTP methods?} We introduce a novel MTP without comparison to other methods.
Ours is designed to support our theory.  All we need to demonstrate is that such a method exists and performs well.
This is not a benchmarking paper of MTP methods on this task.
Readers wanting a rank-based MTP evaluated on natural language data should see \citet{zuhri2025predicting}.

{\bf Why did you not consider discrete diffusion?}
We do consider it, theoretically.
Masked discrete diffusion models (MDM) induce MMHT in the same way MTP does (\A \ref{appx:masking-as-mtp}), and produce a subset of the MTP cost terms for this task (\A \ref{appx:discrete-proof}).
However, MDM changes both the loss and the model parameterization at once, combining a masking-based loss with a non-causal architecture.
This is not a controlled comparison, and inducing the same cost terms does not imply the same expressibility or learnability \citep{svete2026on}.
More importantly, MDM can potentially generate in any order.
This admits the reverse `solution', where the path is produced from $t$ back to $s$ for PSGs \citep{bachmann2024the}.  This requires no multihop reasoning, and thus potentially bypasses the very thing we care about.  Thus, they are a much harder model to study even in controlled settings.

{\bf Why did you not use LLMs or larger models?}
Prior work has already shown that LLMs cannot solve the task on path-star graphs \citep{bachmann2024the, xu2026depth}, and that model size is not why the task is hard to learn at scale \citep{saparov2025transformers}.
There is no benefit in repeating established negative results.
Small models can express the problem, and we show they can learn it at scale.
A separate question is whether our methods transfer to larger or pretrained models.
MTP has been part of foundation models for some time \citep{liu2024deepseek, xiao2026mimo, team2026qwen3}, and the node list is easily applicable to LLMs.
STAN induces a new input format, and there is no reason to expect this to harm a larger model trained from scratch.
Applying STAN to a pretrained LLM may be an engineering problem, but that sits outside the question of learnability, and is not a claim we make.
The purpose of this work is not to benchmark LLMs on the task, but to understand why the task is hard to learn, which requires a controlled setting LLMs are ill-suited for (\A \ref{appx:t-model}).

{\bf Why did you not probe the model or use explainability methods?}
{\em Because the experimental design makes them unnecessary.}
Probing is required when a correct output does not entail that the intended computation occurred, as memorization, world knowledge, or surface heuristics can produce it.
This is why we spend so long on the theoretical model (\A \ref{appx:t-model}).
Assumptions 1--5 remove these (\A \ref{appx:t-model-assumptions}).  As test graphs are unseen, the edge set exists only in the current input, and a correct path entails a multihop search over it.
{\bf Task performance is therefore direct evidence of the behaviour being studied, rather than a proxy for it.}
Fig.\ \ref{fig:psg_step} is an example of this: SACC decomposed by $k'$ and $d'$ reads out which sub-problems the model has solved and when it acquires them, and the restricted MMHT experiment (\A \ref{appx:restricte-psg}) confirms the reading by intervention.  We claim that latent multihop computation occurs, not how the model computes it, and nothing in this work depends on the latter.

{\bf Is it not already known that MTP helps on this task?}
\textit{That MTP helps is known; why it helps is not fully known, and prior results were tied to specific topologies, PSGs, at small scales.}
\citet{bachmann2024the} proposed `teacherless' training for path-star graphs and attributed its benefit to avoiding the CHC, but did not give a full account.
The CHC results from a bad interaction with teacher-forcing, yet removing teacher-forcing entirely is not the answer.
Teacherless training induces only the $k$ terms where $i=1$ in Eq.\ \ref{eq:mtp}, which are the hardest, as they condition on no target-side tokens, whereas the remaining terms require $1 < i \leq j$, and thus use partial teacher-forcing.
\citet{frydenlund-2025-language} also used MTP without giving a complete explanation.
\citet{gloeckle2024better} speculated that MTP upweights informally defined `choice points', which are not the choice nodes we define (\A \ref{appx:mtp}).
\citet{huang2026transformers} proved that MTP and NTP induce different learning dynamics for $d=2$, $k=2$ PSGs (concurrent to this work).
The scaling wall only appears as the graphs grow, and \citet{xu2026depth} learned PSGs up to $k=5$ with NTP alone.
The small-scale MTP results should be compared against this NTP result.
Results at these scales are therefore not results about the phenomenon we address.
We give the mechanism for general topologies, show that the benefit does not depend on the topology, and demonstrate this at scales far beyond these works.
Several other works have applied MTP to PSGs (\A \ref{appx:PST-prior-work}), but did so to benchmark a method rather than to explain why MTP helps.

\begin{figure}[h]
    \centering
    \includegraphics[width=0.4\linewidth, trim={5.0cm 16.5cm 6.cm 5.cm},clip]{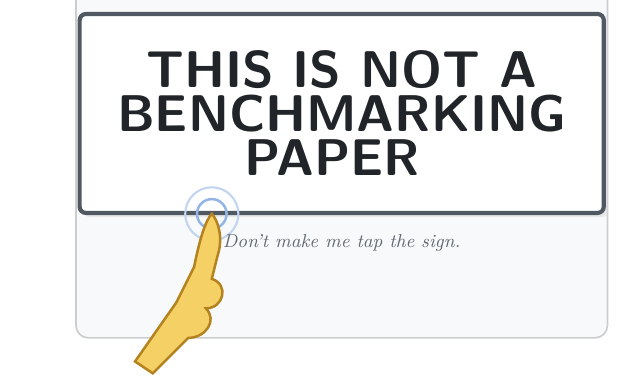}
    \caption{An illustration of the work's scope}
    \label{fig:mem}
\end{figure}

See Fig.\ \ref{fig:mem} for further clarification of the work's scope.  

\section{Graph Search as a Theoretical Model of Multihop Reasoning}\label{appx:t-model}

This section sets out the theoretical model we adopt.
Modelling multihop reasoning as graph search is common, but doing so commits one to a set of claims which are rarely stated. We first define a multihop reasoning task and the graph it induces, and show that the definition admits two settings depending on how the learning problem is defined (\A \ref{appx:t-model-def}).
We then state the five claims the model commits us to, and justify them by establishing their prior use (\A \ref{appx:t-model-claims}).
We then expand on assumptions our own setting adds (\A \ref{appx:t-model-assumptions}).
Finally, we describe the second setting, in which the graph is held in the weights, and which supports answering different questions from our setting (\A \ref{appx:in-weights}).

This section also acts as a tutorial to our experimental setup, which is inherited from prior work \citep{bachmann2024the, saparov2025transformers, frydenlund-2025-language}.

\subsection{Defining a Multihop Reasoning Task}\label{appx:t-model-def}

In \S \ref{sec:intro}, we defined a multihop reasoning task as a tuple $(\mathcal{W},\, q,\, a)$.
We now expand on this.

$\mathcal{W}$ is any information which licenses inference, i.e.\ a set of facts, axioms, rules, or relations.
The task is solved by $k \geq 1$ inferential steps connecting $q$ to $a$, where each step is licensed by $\mathcal{W}$.
Thus, $\mathcal{W}$ induces the reasoning graph $\mathcal{G} = (\mathcal{N},\, \mathcal{E})$, and the solution is a path $\pi = q\, u_1 \ldots u_{k-1}\, a$ in $\mathcal{G}$.\footnote{This connection is so ingrained that even the term `multihop' invokes imagery of a graph structure.  Hopping on what?  Compare with the nearly synonymous but more neutral term `multistep' reasoning.  To our knowledge, there is no definitional difference between the two.}

$\mathcal{W}$ need not surface $\mathcal{G}$, which may instead have to be inferred.
This matters because inferring $\mathcal{G}$ and traversing it are separate problems, and only the latter is unique to multihop reasoning.
Either way, $\pi$ is latent, and finding it is what makes the task multihop.
Importantly, the topology is not known in advance.
$\mathcal{G}$ is determined by $\mathcal{W}$, which is arbitrary, so no topology can be built into a solution method, even when $\mathcal{G}$ is given explicitly and its topology is readable at inference.

Multihop question answering (QA) is an important special case.
Here, $\mathcal{W}$ is a knowledge base $\mathfrak{K} = (\mathfrak{E},\, \mathfrak{R})$ of entities $\mathfrak{E}$ and relations $\mathfrak{R}$, $q$ is a question, and $a$ is its answer.
$\mathcal{G}$ is then a knowledge graph, with entities as nodes and relations as edges.\footnote{Note that $\mathcal{G}$ may be inferred or explicitly given in multihop QA as well.  Many methods construct an explicit graph from text before reasoning over it \citep{de-cao-etal-2019-question, ding-etal-2019-cognitive, cheng-etal-2024-structure}, while others search over a given knowledge graph directly \citep{sun2024think}.  Thus, whether $\mathcal{W}$ surfaces $\mathcal{G}$ is independent of whether the task is posed in natural language.}
The task is solved by finding a path $\pi = (n_0,\, n_1,\, \ldots,\, n_k)$ in $\mathcal{G}$ s.t.\ $n_0$ corresponds to $q$ and $n_k$ corresponds to $a$.

However, in multihop QA, the graph is often a single static dataset.  
This leads to two distinct theoretical models.  
$\mathcal{W}$ can either be supplied as many independent instances, which supports inference over an unobserved $\mathcal{W}$, or as a single fixed (partially) observed dataset to be learnt during training.\footnote{Again, this is regardless of whether $\mathcal{G}$ is explicitly given or inferred.}
Call these the {\em in-context} and {\em in-weights} settings \citep{noroozizadeh2025deep}.\footnote{Both settings require memory and differ only in where it resides.  In-context, $\mathcal{W}$ is held in the hidden states, and is thus transient and rebuilt for each instance.  In-weights, $\mathcal{W}$ is held in the parameters, i.e.\ {\em parametric memory}, and is thus persistent and shared across instances.  Note that this is not LLM `in-context learning'.  We do not apply an unseen task to a pretrained LLM; rather, we train small LMs from scratch.  
They are related, as both must produce in-context representations to solve the task.
}  In-context, $\mathcal{W}$ varies across instances, so memorization is not viable and the model must compute over the provided context.
In-weights, $\mathcal{W}$ is the same for every instance, so memorization is viable and, indeed, is necessary.
Note that the task definition is agnostic to this choice, as the split arises from how the learning problem is set up, i.e.\ what varies across the data and what is fixed.
Thus, the same definition yields two distinct theoretical models, each of which supports answering different questions (\A \ref{appx:in-weights}).

\subsection{The Underlying Claims}\label{appx:t-model-claims}

Modelling multihop reasoning as graph search reduces to the following five claims.
\begin{enumerate}
    \item[Claim 1.] All multihop reasoning tasks induce a reasoning graph.

    \item[Claim 2.] All multihop reasoning tasks require traversal of the reasoning graph.\footnote{Claim 2 concerns what the task requires, not how the model achieves it.
A model may produce $a$ without composing at all, e.g.\ by exploiting a statistical correlation between $q$ and $a$.
This is a shortcut, and is the failure mode we study.
Note that this differs from composing during training and storing the result, which we return to in \A \ref{appx:in-weights}.}

    \item[Claim 3.] Search is a specific type of traversal often used in multihop reasoning.

    \item[Claim 4.] A theoretical model of multihop reasoning must apply to an arbitrary reasoning graph.

    \item[Claim 5.] Graph search is such a model.
\end{enumerate}

We use the term `claim' to unify these five ideas, as they are not of the same kind.
Claims 1 and 2 hold by construction from the definition (\A \ref{appx:t-model-def}).
Claim 3 is definitional, and that search is often used is established by the literature.
Claim 4 is a modelling assumption; one could instead want a restricted model, whereas we require generality because $\mathcal{W}$ is arbitrary.
Claim 5 is the resulting theoretical choice, and follows from search assuming no structure in $\mathcal{G}$ while isolating the behaviour we study.

To our knowledge, no prior work explicitly states all five claims.
However, there is a long history of prior work which assumes one or more of them, and we justify the claims by establishing their prior use.
Because multihop reasoning and graph search are very broad areas, these claims are often not the focus of -- or even stated in -- prior work, and thus must be inferred from their methods.\footnote{This is not a comprehensive survey, and even wider claims, such that all of general problem solving can be modelled as graph search, go back to the dawn of AI \citep{newell1972human, newell2007computer}.}

\subsubsection{Claims 1, 2, and 3}

Claims 1, 2, and 3 are highly coupled, and prior work generally supports them together.
As noted, the quantifiers in Claims 1 and 2 are by construction, so prior work serves to establish coverage, i.e.\ specific cases where multihop reasoning induces a reasoning graph or requires traversal.
Claim 3 follows from elementary graph search algorithms, and prior work establishes that this also holds in the multihop context.

For Claim 1, the reasoning graph is often implicit.
However, it can also be explicit, and appears under various names such as a `knowledge graph', `knowledge base', or `entity graph'.
Prior work which constructs an explicit graph from implicit information provides evidence for the existence of an underlying reasoning graph.
Works which support Claims 1 through 3 fall into three groups: those which construct an explicit graph from unstructured text, those which surface reasoning structure at inference time, and those which search over a given graph directly.

{\bf Constructing an explicit reasoning graph.}
\citet{yang-etal-2018-hotpotqa} introduced one of the earliest multihop question-answering datasets for LLMs.
They explicitly constructed their dataset as a graph in which edges represent Wikipedia hyperlinks and nodes are the first paragraph of each article (Claim 1).
They studied question answering, which they framed as a search on the reasoning graph (Claim 2).
They also demonstrated that different kinds of questions induce different types of multihop reasoning tasks; however, traversal or search is a necessary subroutine in each of these (Claims 2 and 3).

\citet{de-cao-etal-2019-question} framed multihop question answering as `an inference problem over a graph', i.e.\ search over a graph (Claim 3).
To support this, they converted text to an explicit graph, with nodes representing named entities and edges as relations between them (Claim 1).
Similarly, \citet{ding-etal-2019-cognitive} constructs and searches over a `cognitive' graph (Claims 1 and 2).

\citet{sun-etal-2018-open} used text to augment missing gaps in a knowledge base, demonstrating that the graph can mix both explicit (knowledge base) and implicit (inferred from text) graph structures (Claim 1).
They also plainly stated the relation that multihop question answering has to graph traversal: `many questions require multi-hop reasoning, which follows a path from a seed node mentioned in the question to the target answer node' (Claim 2).
Note that Claim 3 can also be inferred from this, if finding the path is non-trivial.

Further evidence for Claims 1, 2, and 3 similar to the above works can be found in \citep{das2018go, lin-etal-2018-multi, sun-etal-2019-pullnet, saxena-etal-2020-improving, luo2024reasoning, han-etal-2025-reasoning, yang-etal-2025-large}.
Also consider \citet{cheng-etal-2024-structure}, who developed a method which `explicitly converts unstructured text into a graph via LLMs and instructs them to navigate this graph using task-specific strategies to formulate responses' specifically for multihop tasks.

{\bf Surfacing the reasoning graph at inference time.}
Not only have these claims been observed in the literature since the dawn of the LLM age, but Claims 1 and 2 are implied by the most ubiquitous method employed for solving multihop reasoning tasks with LLMs: Chain-of-Thought (CoT) \citep{wei2022chain}.
CoT combines prompting with test-time compute.
The CoT chain generates a reasoning path.
Thus, CoT explicitly surfaces an underlying latent path in the latent graph, which assumes such a graph exists (Claim 1) and that traversal is required (Claim 2).
Follow-up work extended this by surfacing more complex structures such as trees and full graphs \citep[][Claims 1 and 2]{yao2023tree, besta2024graph, jin-etal-2024-graph, wen-etal-2024-mindmap}.
These structures require search to traverse (Claim 3), leading to the direct incorporation of actual search algorithms such as breadth- or depth-first search (BFS or DFS) into the inference procedure \citep{yao2023tree}.
\citet{han-etal-2025-reasoning} considered prompting an LLM to explicitly generate a fine-grained graph over which reasoning is then conducted, and compared this to CoT, ToT, and GoT for various multihop reasoning tasks, including search (Claims 1, 2, and 3).

{\bf Searching over a given graph.}
\citet{sun2024think} introduced `think-on-graph', which treats an LLM as an agent that searches over an explicit knowledge graph for multihop reasoning.
They motivated this by identifying that other methods which use a knowledge graph do not search over the graph directly but rather retrieve information from the graph, which is then incorporated into the prompt context.
This converts the graph to a different format which is easier to incorporate into an LLM.
However, by doing this, the LLM is then forced to perform search over a latent sub-graph, thus inducing a `loose coupling' with the original explicit graph.
To solve this, their method forces the model to perform beam search over the explicit knowledge graph using test-time compute.
This work thus addresses how the reasoning graph can be treated as implicit or explicit, and the relation of search to reasoning (Claims 1, 2, and 3).

\subsubsection{Claims 4 and 5}

Not all multihop reasoning tasks are framed as search tasks.
For example, arithmetic, aggregation, and analogical reasoning all require multihop reasoning but not `search'.
However, solving each of these requires traversing the reasoning graph (this is why Claims 2 and 3 are separate).
An arithmetic expression induces a chain or tree of operations, given by the expression itself, and aggregation induces a set of nodes sharing a relation.
This structure is known in advance, and can thus be built into a solution method as an inductive bias.
Search assumes no such structure.

Claim 4 largely follows from the definition, as $\mathcal{G}$ is determined by $\mathcal{W}$ and $\mathcal{W}$ is arbitrary (\A \ref{appx:t-model-def}).
It is a modelling assumption rather than a corollary because a restricted model is a legitimate thing to want, e.g.\ one covering only the structures induced by a particular class of questions.
We require generality, as we are modelling multihop reasoning rather than a subclass of it.
\citet{cheng-etal-2024-structure} make the diversity explicit, acknowledging `the diversity of queries and their corresponding graph structures', and categorizing reasoning tasks such that each `is aligned with a unique graph structure', including graph sorting, graph querying, and logical inference.

Claim 4 is uncontroversial, but can be violated in practice.
The path-star literature illustrates this danger, as the topology admits solutions which exploit it and do not generalize.  The reverse `solution', which requires no multihop reasoning is the prime example \citep{bachmann2024the}, but other non-generalizable solutions exist as well \citep{frydenlund-2024-mystery}.
Thus, results obtained under one topology need not hold under another.
A theoretical model which assumed a topology would then describe that topology rather than multihop reasoning in general.

Claim 4 gives rise to Claim 5.
Search assumes no topology, and is thus defined for an arbitrary reasoning graph.
Importantly, search enables this while also allowing us to isolate the behaviour we want to study.
This is because search depends only on the structure of $\mathcal{G}$; to find $\pi$, we need to know which nodes connect to which, and nothing about what any node means.
We can thus define the task as a multihop traversal over a structural description of $\mathcal{G}$, without retrieval, entity resolution, or language generation.

{\bf This provides a clean window into how models learn structural reasoning, i.e.\ whether the model learns to traverse $\mathcal{G}$.
Consider a search task defined with and without isolating structural traversal from these other aspects.
If a model does better with them, the gain cannot have come from traversal, as traversal is what we isolated.
Thus, it came from bypassing traversal, i.e.\ a shortcut.
If a model does worse, then the added aspects are the cause, and they are worth studying on their own.
Thus either case justifies isolating traversal. Note that this justifies isolation, but this doesn't imply that 
traversal does not interact with these other aspects.  However, such interactions should be studied under a separate theoretical model, which we will discuss in \A \ref{appx:in-weights}.}

The works above used graphs as a practical model in order to develop methods.
Graph search has also been used as a theoretical model.

\citet{saparov2025transformers} stated this most directly.
They noted that search `is the backbone of many reasoning problems as it is equivalent to proof search in a simplified logic which is a subset of almost any logic'.
Thus, `the model must solve this task if there is any chance to generalize to more complex search and reasoning tasks'.

\citet{bachmann2024the} designed the path-star task as `a minimal setting that captures the core essence of what it means to solve problems with lookahead, without irrelevant confounding factors'.
They expected their failure result to extend to more general tasks such as story telling.

\citet{khona2024toward} framed their setup in order `to design and study an abstraction which enables a precise understanding of stepwise inference in Transformers'.
They argued that `tasks which see maximum benefit from stepwise inference can be cast as a graph navigation problem', and gave examples such as the Tower of Hanoi.

Note that prior work often considered graph search under the name planning \citep{bachmann2024the, khona2024toward}.
This speaks to the task being achieved latently across layers.
Planning requires committing to a path before acting on it, so if, instead, a model is allowed to act and revise that action, it is no longer a planning problem. 

\citet{cohen2025spectral} motivated the controlled setting, and chose graph search to instantiate it.
They noted there was no `quantitative, verifiable definition of planning or reasoning', and that benchmarks `cannot distinguish ``true reasoning/planning'' from pattern matching or retrieval due to web-scale training and possible data contamination'.
Their solution was to `study language model internal dynamics in settings with carefully controlled, structured data', using graph connectivity.
Similarly, \citet{sanford2024understanding} asked `how can we analyze reasoning capabilities of neural networks?' and answered that `one approach is to study algorithms executable with their internal representation'.

Finally, Claim 5 concerns the theoretical model, and thus holds for both settings.
\citet{noroozizadeh2025deep} adopted the same model in-weights, describing path-finding on path-star graphs as `a (symbolic) implicit reasoning task'.

{\bf Importantly, while these works used graph search as a theoretical model, they did not formally specify it, leaving the underlying claims to be inferred.
This is why we state them here.}
Making the commitments explicit allows them to be discussed or rejected rather than left implicit.
Thus, future work can state which claims it adopts, and which it does not.

\subsection{Our Instantiation}\label{appx:t-model-assumptions}

The claims above establish graph search as the theoretical model, i.e.\ they justify that this is the thing to study.
We now consider the assumptions we make when instantiating that model in practice.
That is, we still need an experimental setup which forces the model to solve the task by traversal alone.
{\bf Importantly, each assumption is made because it prevents alternative ways to produce a correct answer without traversing.}
These assumptions are separate from the claims, as one can adopt the theoretical model but reject the instantiation.  Importantly, these assumptions are not made to make the task easier to learn.  Indeed, most restrict how the model must learn to solve the task by avoiding `solutions' which do not require multihop reasoning or search.

{\bf Assumption 1: traversal is latent.}
We exclude chain-of-thought, scratchpads, and test-time search, so that the traversal occurs within the forward pass in a single step.
\citet{bachmann2024the} give the argument for why search on path-star graphs must be latent.
The first emitted token commits to an arm, so the target must already have been located before generation begins, and the search cannot be spread across the generated sequence.\footnote{This, and the fact that there are alternative potential paths, makes it a planning task.}

By enforcing that the model performs a latent computation of the path, we prevent a solution via enumeration. 
When test-time compute is permitted, the model easily learns the one-hop adjacency and then walks from the start node until it reaches the target. 
The walk is not quite Markovian, as the model learns not to repeat paths it has already produced, and we never train it to do this.
This finds a path without performing multihop reasoning.  \citet{khona2024toward} observed the same ordering in a setting which permitted test-time compute, where their models first learned to avoid generating edges that do not exist in the graph, and only later learned to reach the goal node.

Test-time compute admits a second kind of solution.
Here the model does perform multihop reasoning, but it does so over a known structure.  An example of this is using a BFS scratchpad reduces the search state to a linear chain which the model then follows via a multihop traversal.
As such, it cannot model multihop reasoning in general (Claim 4).
We demonstrate that a BFS scratchpad does this, thus breaking the theoretical model (and that it does not scale, making it an impractical `solution' to boot; \A \ref{appx:bfs-sp}, \A \ref{appx:sp-collapse}).

Both solutions need test-time compute.
Thus, we disallow it and thus require a latent solution.

{\bf Assumption 2: the target is a shortest path.}
We could instead ask for reachability, connectivity, or any path from $s$ to $t$.
The problem is that these do not constrain the length of the output.
The model can then backtrack and wander, and the generated path itself provides test-time compute.
This thus allows for the enumeration solution above, where the walk is the generation.
Requiring a shortest path fixes the output length, thus preventing such a solution.
The model must instead compare the alternatives at each choice node, and it is this which makes the task a search.

{\bf Assumption 3: the edge list is shuffled.}
The edges are presented in a random order (\S \ref{sec:task}).
If the order is allowed to carry information, then the edges can be listed in breadth-first order from $s$ \citep{ge-etal-2025-graph}.
This gives the same structural reduction as the scratchpad above, except that it is supplied in the input rather than generated.
Randomizing the order prevents any such ordering.
\citet{ge-etal-2025-graph} compared BFS, DFS, and PageRank edge orderings against a shuffled list, and found BFS ordering improved shortest-path accuracy by 70.1\%.
\citet{yuan2024gracore} similarly found that sequentially ordering node names outperformed random ordering, with the largest gains on path reasoning.

{\bf Assumption 4: nodes are semanticless.}
Node identifiers carry no relationship to one another, which prevents semantic and homophily shortcuts \citep{huang2024can, mcpherson2001birds, xu2025when, guan2025attention}.
One could retain semantics and audit each shortcut as it arises, but the set of possible shortcuts is not known in advance, whereas removing semantics guarantees their absence.
Without doing this, then producing a correct answer would not be direct evidence that a $k$-hop traversal was performed.
Semantic relations between nodes allow the answer to be produced by other means, so accuracy no longer measures the behaviour we wish to study.
Removing semantics restores the correspondence, as the only information available is the structure given in the prefix.
Semanticless nodes require relabelling, since no natural corpus is semanticless.
One could relabel a natural graph and preserve its topology, but Assumption 5 requires a fresh graph for every instance, which a fixed corpus does not supply.
Synthetic data is therefore not an independent choice but a consequence of these two assumptions (\A \ref{appx:controlled-settings}).

{\bf Assumption 5: training instances are unique.}
Uniqueness addresses two separate problems, which arise at two different scales.
The first is an alternative solution, in-weights storage, where a fixed graph, or a small set of them, lets the model store the structure in its weights.
Either the answer is recalled outright, or the $k$-hop composition is amortized over training, and the forward pass reduces to a lookup of its result.
Amortization requires a fixed graph, since the cost is paid once and reused across instances (\A \ref{appx:in-weights}).
Thus, this leaves the forward pass as the only place the traversal can occur, which is the object of our study.
The second is a learning problem.
A large but finite set of graphs is past the in-weights regime, but repeated subgraphs give spurious correlations  \citep{frydenlund-2024-mystery, wang2023can}.
Generating every instance uniquely removes both, leaving the forward pass as the only place the traversal can occur.

The last two assumptions are of a different kind from the first three.
Assumptions 1 to 3 would be adopted by anyone asking our question, whereas Assumptions 4 and 5 are specific to it.
Semantics and memorization are not confounds in general, and there are questions for which they are the object of study rather than something to be removed.
Dropping Assumption 5 gives rise to the in-weights setting, which we describe next (\A \ref{appx:in-weights}).
Assumption 4 is an independent choice within that setting, and multihop question answering drops both.

\subsection{The In-Weights Model}\label{appx:in-weights}

Assumptions 4 and 5 are the ones most likely to give pause.
Recall that multihop question answering is the special case in which $\mathcal{W}$ is a knowledge base, $\mathfrak{K}=(\mathfrak{E},\, \mathfrak{R})$, so that the nodes of $\mathcal{G}$ are entities and its edges are relations (\A \ref{appx:t-model-def}).
Both assumptions are violated there.
Entities carry meaning and stand in relation to one another, and $\mathfrak{K}$ is fixed and seen many times over training rather than generated anew for each question.
Thus, these two assumptions rule out the most ubiquitous multihop reasoning task in NLP.

Claim 5 warned that the aspects we remove may interact with traversal, and studying that interaction requires a setting in which they are present.
The definition already provides one (\A \ref{appx:t-model-def});
rather than supplying $\mathcal{W}$ anew for each instance, fix it and learn it during training.
$\mathcal{G}$ then resides in the weights rather than in the context, providing the {\em in-weights setting}.

The two settings support different questions.
We ask whether a model can learn to perform multihop traversal without memorization.
The in-weights setting asks how a model learns when memorization is allowed.
Importantly, the two settings have very different learning dynamics.

\citet{noroozizadeh2025deep} demonstrated how much this change matters.
They studied the path-star task with the graph held in the weights and learnt over training, rather than supplied in the context.
Importantly, theirs is a controlled study, and it is controlled in the same way ours is, so the only substantive difference is whether $\mathcal{G}$ is supplied in the context or held in the weights.
The paths had a fixed length, so there is no MMHT. The train and test paths are disjoint, so they evaluate on held-out paths (though every edge was seen during training).
Whereas the original in-context task is unlearnable with NTP and without MMHT on small graphs, they easily scaled to massive graphs of up to $5 \times 10^4$ nodes with $k$ of 6 to 10.
Thus, the task is the same and the controls are the same, so only the setting can account for this difference in results, which implies different learning dynamics.  

They attribute this to how the graph is stored.
The model organizes the node embeddings so that their geometry reflects the multihop structure of the graph.
This geometry arises from training on the individual edges alone, with no path supervision whatsoever.
Importantly, the multihop aspect of the task then collapses to a lookup at inference.
The composition still happens, but it is amortized over training rather than performed at inference.
This matters because multihop reasoning is generally taken to be a traversal performed at inference, and not the learning of compositional facts.
Thus, dropping Assumption 5 does not relax our setting, rather, it replaces it.

The in-weights setting does not merely make traversal easier, it can remove the need for it.
This is counterintuitive given how the setting is generally studied.
Work on multihop QA holds the knowledge base in the weights, yet asks whether the model traverses it at inference, such as by locating a bridge entity within the forward pass \citep{yang-etal-2025-large, balesni2024lessons}.
Traversal is assumed to be the thing which is learnt, and a failure is read as an inability to compose.
However, the setting itself admits a solution in which no traversal occurs at inference, which is the remarkable finding of \citet{noroozizadeh2025deep}.
A failure may then indicate that the structure was not stored, rather than that the model cannot traverse it.
Thus, an in-weights result may be about how structure is stored rather than about multihop reasoning as it is usually meant.

The two settings are therefore not variants of a single model, they are different models.
Which of the two a result is about determines what that result can be taken to show.
We strongly support \citet{noroozizadeh2025deep}'s emphasis on this point, where they wrote that `a takeaway from [their] work is to be deliberate not to conflate insights from the in-context setting with that of the in-weights setting (a distinction that is rarely made explicit in literature)'.
They give a fuller account of the distinction, along with an organized list of prior works under each setting, and we defer to it rather than duplicating it here.

The two settings are best read as limits.
In the in-weights limit, $\mathcal{G}$ is fixed and can be committed to memory in full, so traversal can in principle be eliminated.
In the in-context limit, $\mathcal{G}$ is unseen and nothing about it can have been stored, so all of the structure must be computed at inference.
Most tasks of interest sit between the two.
Multihop QA is the clearest case, as the entities and relations are held in the weights, whereas the question, and often retrieved evidence, are supplied at inference.
Frequently occurring compositions may well be stored, whereas rare ones must be composed.
Note that we offer this as a conjecture about where real tasks sit rather than as a result.

Semantics is an independent choice within the in-weights setting.
\citet{noroozizadeh2025deep} kept their nodes semanticless, and \citet{wang2026do} compared semantic and semanticless representations of the same knowledge graph, so the setting does not require semantics.
Indeed, they found the natural language forms gave noisy scaling trends whereas random identifiers gave a clear and reproducible one, which they attribute to the information carried by the lexical form of the entity and relation names.
Multihop QA drops both.
However, retaining semantics makes auditing unavoidable, since the set of possible shortcuts is not known in advance.
This auditing is what limits the scale, and \citet{yang-etal-2025-large} and \citet{balesni2024lessons} are both restricted to two hops as a result (\A \ref{appx:controlled-settings}).

\citet{minegishi2026emergent} trained a single-layer transformer on two-hop compositions over a fixed relational graph, plus an analogical mapping between two copies of it, proving a in-weights setting that supports analogical reasoning.
They also formalized analogy as a functor between categories, and showed that it is acquired later than composition.

\citet{wang2026do} studied scale in the in-weights setting along a different axis, namely the parameter budget.
They trained on the individual triples of a fixed knowledge graph and tested on triples which follow from the latent rules the graph was generated under.
They found that the minimal parameter budget required scales linearly with what they call the graph search entropy, which is the entropy rate of a maximal entropy random walk over the graph, and that the largest model was not the best one.

\citet{zhong2025understanding} studied multi-token prediction in the in-weights setting.
They found that multi-token prediction improved path prediction over next-token prediction, and that the gain was largest on the paths which required the most inference beyond what was observed in the training data.
They attribute this to a transfer layer which learns the powers of the adjacency matrix, which is the same object as the distance matrix our skips require the model to compute (\S \ref{sec:sse-mtp}).
The difference is where it resides, as the powers are stored in the weights of a single graph there, whereas here they must be computed within the forward pass for a graph which has never been seen.
Multi-token prediction thus helped in both settings, and by supplying supervision on the same structure, but the learning dynamics of the two settings differ.

{\bf Given the different learning dynamics in the two settings, we strongly recommend that future researchers be very clear about which setting they are evaluating, as this detail is often a single line hidden in the method section, or is not stated at all and must be inferred from the experimental details.}
\citet{tong2026generalization} illustrates the difficulty.
They considered generalization to unseen graphs and to longer path lengths for in-weights shortest path search.
Their use of multiple graphs makes the setup look like the in-context task, and the test graphs even have disjoint node sets, but the number of graphs is minimal, and the models were pretrained on random walks over all of them, so the graphs are only unseen during finetuning.
They found that models generalize across graphs, which is expected given that pretraining, but that even in the in-weights setting they fail to generalize to out-of-distribution lengths.

\section{Extended Relevant Work}\label{appx:related}

\subsection{Graphs and LLMs}

There is a strong interest in applying LLMs to non-text-based tasks, despite LLMs being designed for and trained on language \citep{NEURIPS2020_1457c0d6}.
However, applying LLMs to graph-based tasks is challenging due to a mismatch between data and model design.
LLMs were designed to process sequential data.
Thus, graphs must be encoded linearly for an LM; however, there is no canonical linearization of a graph, and the chosen order has a large effect on performance \citep{fatemi2024talk, ge-etal-2025-graph, yuan2024gracore}.

LLMs often perform surprisingly poorly on graph-based tasks \citep{wang2023can, liu2023evaluating, fatemi2024talk, guo2023gpt4graph, dai2024revisiting, perozzi2024let, ge-etal-2025-graph, tang2025grapharena}.

Importantly, prior work has shown that LLMs fail to understand and exploit graph topology when it is given as part of the prompt \citep{huang2024can, xu2025when, guan2025attention}.
Instead, LLMs largely ignore explicit structure in favour of inferring structure from node semantics \citep{huang2024can}.
This is possible for some tasks due to local homophily, a phenomenon in which semantically similar nodes tend to connect more closely in natural data \citep{mcpherson2001birds}.
\citet{xu2025when} corroborated these findings under different graph encodings, stating that `most structural encoding strategies offer marginal or even negative gains' and that this was consistent across LLM sizes and even for a `reasoning' model.
This led them to suggest that using structure is unnecessary for LLMs.
Also, the attention patterns of LLMs do not correspond to graph structures, implying a lack of traversal across the graphs to solve tasks \citep{guan2025attention}.

Prior work has considered how to use natural language to describe graph structure and semantics \citep{ye-etal-2024-language, chen2024llaga, zhou-etal-2025-graph}.
\citet{dai2024revisiting} showed the limits of using text descriptions of graphs for graph tasks.

\citet{park2025iclr} showed how in-context representations of graph structure emerge for graph navigation on grid topologies.
Note that here `in-context' is in-context learning using LLMs.
The provided graph tokenization is a series of random walks over the graph.
Interestingly, this will act as a replication of the adjacency list, which will bypass the causal constraint as in \S \ref{sec:causal-constraint}.
This was followed up by \citet{lepori2026language}, who showed that even though models can use in-context learning to represent graph structure, models often then fail to take advantage of this structure in downstream tasks.

Even though knowledge graphs are a rich source of structured information, incorporating this into LLMs is an open problem \citep[][among many others]{ijcai2024p898, 10.1145/3626772.3657775, chen2024llaga, wang2024llms, luo2024reasoning, kong2025gofa, wu-etal-2025-graph}.



\subsection{Controlled Settings and Synthetic Data}\label{appx:controlled-settings}

Studying graph search in natural language introduces confounds that obscure whether a model is genuinely traversing the reasoning graph.
The latent graph $\mathcal{G}$ is unknown and must be inferred from text; world knowledge and memorization can produce correct answers without actual path discovery; and surface-level correlations can yield shortcuts, particularly when $k$ is small.
There are also retrieval and reading-comprehension limitations which are unrelated to the multihop search process itself.
These confounds make it difficult to isolate the core difficulty of multihop reasoning.

We control for them by using language models trained from scratch on synthetic data, where there is no prior knowledge to exploit, the graph structure is explicit and varied, and the only path to a correct answer is structural traversal.
{\bf This makes learnable graph search a controlled environment for studying multihop reasoning, where success is evidence of latent structural computation.}

Controlled settings with synthetic data are widely used when studying reasoning in LMs.
\citet{saparov2023language} generated synthetic ontologies with fictional concept names in order to avoid confounds from knowledge acquired during pretraining, and analyzed the generated chain-of-thought rather than the predicted label.
More broadly, many used synthetic symbolic tasks to ensure a controlled environment when studying multihop reasoning \citep{zhang2022unveiling, ramesh2024compositional, khona2024toward, wang2024grokking}.

\citet{yang-etal-2025-large} provides a nice comparison with our work under different motivations, as they studied the extent to which pretrained LLMs can perform latent multihop reasoning without shortcuts.
Considering LLMs introduces multiple practical issues, which we avoid.
First, they need to identify when and how LLMs employ shortcuts.
This leads them to focus their inquiry on a single type of shortcut based on co-occurrence information \citep{ju-etal-2024-investigating}.
This doesn't rule out other types of shortcuts.
Second, because LLM data is generally inaccessible, they must use a `proxy corpus' to create a test set such that head and answer entities never occurred in the proxy corpus.
This does not rule out such co-occurrences happening in the true training data.
This co-occurrence-based filtering led to the creation of a small dataset, which only considers 1- and 2-hop questions.
Finally, as most LLMs default to using CoT, they need to suppress test-time compute behaviour by framing the question and prompting the model to `fill in the blank' with an immediate answer.
They find that the specific semantics of the bridge entities is the largest determiner of performance.
For example, the best models perform latent multihop reasoning when bridging on a county but fail when bridging on a year.
This does not rule out that these models just have poor representations of years, which is a known issue \citep{yang2025number, yuchi-etal-2026-llms}.

Similarly, \citet{balesni2024lessons} used synthetic data to try to bypass these practical issues with LLMs.
Their argument is similar to the use of a proxy corpus, in that synthetic data will have a low chance of being in the training data, where they stated that `by using synthetic facts, we rule out memorization and reasoning shortcuts as explanations for two-hop performance'.
Again, this data issue limits their study to 2-hop questions.
They found that fine-tuned LLMs fail to reason when composing two synthetic facts, and require that one fact be non-synthetic, learned from pretraining.
This result implies that memorization or neural language representation remains a confound in this setup.
\citet{li2026scaling} `resynthesized' the data by randomly substituting entity names and numerical values.
This was done to ameliorate the related -- and broader -- issue of data leakage from LLM pretraining.

{\bf We are not framing the limitations of these works negatively; rather, they are examples that demonstrate how difficult it is to control for various confounds when using LLMs with real-world data; while these issues can be alleviated, they can not be entirely ruled out, and they affect the conclusions of these works.
These examples thus justify the use of a strictly controlled setting using purely synthetic data and language models trained from scratch to avoid these confounds.}

\citet{lin-etal-2025-implicit} found a structural shortcut that can be learnt for simple multihop addition and subtraction statements.
This shortcut arises due to an asymmetry between addition and subtraction, where the former is commutative and the latter is not.
This allows the model to learn addition and subtraction when the variable is left-side via chaining terms, which is not a viable method for subtraction when the variable is on the right-side.
This shortcut can be analyzed in relation to our work, where addition and left-side subtraction induce a latent linear graph (hence the `chain' solution), where right-side subtraction induces a latent tree.\footnote{We disagree with their conclusion that `language models rely on shortcuts for implicit reasoning' which is also implied in the title `Implicit Reasoning in Transformers is Reasoning through Shortcuts'.  This is a very broad conclusion. While their work supports the claim that language models can learn shortcuts in lieu of implicit reasoning, and they identify a very interesting shortcut, the causes for this shortcut will not apply to most reasoning tasks.}

\subsection{Expressibility and Learnability of Graph and other Compositional Tasks}\label{appx:expressibility}

There are theoretical motivations for applying small LMs to graph-based tasks. %
Small LMs provide a controlled environment, allowing one to probe the limits of expressibility and learnability of LMs and the underlying model, such as transformers \citep{pmlr-v235-back-de-luca24a, sanford2024understanding, khona2024toward, cohen2025spectral, qharabagh2026learning}. %

{\em Expressibility} concerns the question: can model $M$ compute task $T$, which relies on a precise definition of both $M$ and $T$.  This includes the computational budget of $M$.  For transformers, this concerns the number of layers, the embedding dimension, and sequential computation (which is often tied to the task).  Expressibility results for graph tasks are stated in terms of the number of layers and the embedding dimension, as a function of the graph size. %
This means one can select a model which is large enough for the task by construction, and attribute any remaining failure to learnability. %
This is how we select our model size (\A \ref{appx:details}). %

{\em Learnability} asks the question: can model $M$ actually learn task $T$ under learning conditions $C$.  A task being expressible implies that there exists a model parameterization that can solve the task, but not that the task is easily learnable, as finding that parameterization may be very difficult.  The path-star task is a prime example of this, where it is easy to express \citep{frydenlund-2024-mystery}, but very difficult to learn (because it is designed to be adversarial to learn) \citep[][\A \ref{appx:PST-prior-work}]{bachmann2024the}. %

{\bf Expressibility.}
\citet{sanford2024understanding} developed four computational classes for transformer expressibility on graph tasks, stated over the depth $L$ and the embedding dimension $m$ with $N = O(|\mathcal{N}| + |E|)$.
These range from a single wide-enough layer (Depth1), to $L = O(\log N)$ with $m = O(N^{\epsilon})$ (LogDepth), to either adding a polynomial number of pause tokens (LogDepthPause) or instead widening to $m = O(N^{1/2+\epsilon})$ (LogDepthWide).
They then grouped graph tasks into three families: retrieval tasks such as edge existence are solved in Depth1, parallelizable tasks such as connectivity are solved in LogDepth, and search tasks are the hardest family.
Shortest path is a search task, and is solved in LogDepthWide, so logarithmic depth and polynomial width suffice without any pause tokens.

A transformer can express various multihop reasoning tasks with $k$ hops using $\log(k)$ layers. %
This includes Khops \citep{pmlr-v235-sanford24a}, shortest path on path-star graphs \citep{frydenlund-2024-mystery}, and graph connectivity \citep{sanford2024understanding, merrill2025a, ye2025transformers}. %
These are instances of a general phenomenon identified by \citet{liu2023transformers}, who showed that a shallow non-recurrent transformer can exactly simulate any semiautomaton on a length-$T$ input with $O(\log T)$ layers, and with $O(1)$ layers when the associated group is solvable.
They called these {\em shortcut solutions}.\footnote{Their use of `shortcut' refers to the depth of the model being smaller than the number of sequential reasoning steps, and such a solution is exact.  This differs from our use of the term, which refers to a solution that exploits a spurious feature of the data and is not exact.}

Depth can be traded off against width.
\citet{yehudai2026depth} showed that one can trade off model depth for dimensional width when solving various graph tasks. %
They show that this is valuable, since width is parallelizable on GPUs, leading to empirical speedups. %
Their lower bounds are stated for a specific tokenization, where each token is a row of the adjacency matrix, and thus, as discussed below with \citet{bechler2026lost}, expressibility results are tied to the tokenization used. %
This illustrates that one must be very careful when translating different results across graph tasks and experimental setups. %
We discuss their comparison of graph tokenizations in \A \ref{appx:alt-token}. %
\citet{yao-etal-2025-language-models} proved a similar trade-off for their $k$-hop task, where the product of the depth, width, number of heads, and bits of precision must grow linearly in $k$. %
Their bound holds when the attention pattern does not depend on the query entity, and they noted that shortcuts using fewer layers exist for some relation sets. %

\citet{london2025pause} proved that including `pause tokens' strictly increases the expressibility of transformers. %
\citet{sanford2024understanding} also found that these lead to a different expressibility family. %
Our inclusion of the node list will act in a similar way to pause tokens, and so this change in expressibility may explain the observed change in learnability. %
Note that shortest path is expressible without pause tokens given sufficient width, so this is not a statement that the extra positions are necessary.
These are asymptotic regimes, and satisfying their form at a fixed graph size does not certify a particular finite model, so we cannot rule out that $+N$ changes what our model can express.
What is interesting is that STAN decreases the overall token count for the edges to 1/3. %
Thus, this improvement can not be explained solely by raw token count. %

\citet{bechler2026lost} considered the expressibility of three different graph tokenization schemes for graphs: adjacency lists (per node), random walks, and spectral encodings. %
They show that while the latter two encodings are derived from adjacency lists, transformer models can not convert between encodings under a fixed depth, and that the different representations have different expressibility for certain graph tasks. %
Note that the latter two `tokenizations' are more of a feature preprocessing that encodes graph-global information into each node encoding. %
This is orthogonal to our work, which only encodes singular edge or node information into each token (via raw edge- and node-lists), and encoding graph-global information would run contrary to our core scientific question, which is how language models can learn graph-global information from local encodings efficiently at scale.  

{\bf Our model design follows from these results.}
These bounds imply that our tasks are expressible by small transformers, as the required depth and width grow slowly in the size of the graph and in the number of hops. %
We therefore use models of $L=11$ layers and $m=256$ dimensions or smaller (\A \ref{appx:details}), which satisfy the depth and width required at our graph sizes. %
Any remaining failure is then attributable to learnability rather than to expressibility. %

{\bf Learnability.}
A model with enough depth and width to express a task will not necessarily learn the intended solution
\citet{liu2023transformers} found that these `shortcut' solutions are learned by standard training, but that they are brittle.
Their models failed to generalize to unseen sequence lengths and degraded under limited supervision, while a recurrent LSTM was unaffected.
They showed that a recency-biased scratchpad could guide the transformer towards the recurrent solution instead, so what is learned depends on the supervision rather than on the depth.
\citet{ye2025transformers}, who consider in-weights graph connectivity, noted that training models within their expressibility limits can lead to learning the underlying algorithmic solution; however, exceeding this will result in learning shortcut solutions. %
\citet{khona2024toward} found no qualitative change in the learning dynamics of their in-weights shortest-path task as they varied the number of layers. %
They instead attributed failure to a simplicity bias, where the model prefers a shallower solution than the task requires. %
They then found a critical embedding dimension, around $20$ for a graph of $200$ nodes, above which their model generalized to all held-out node pairs, so width and not depth was the binding constraint in their setting.
\citet{yao-etal-2025-language-models} likewise found that their model performed the task by looking up one hop per layer, which is what their bound predicts when the width is held fixed.
\citet{sanford2024understanding} expected shortest path to require a very large model, and only their pre-trained $60$M-parameter model approached optimal performance, whereas we solve the harder in-context version with $\approx6$M parameters trained from scratch.

\citet{li2026scaling} found that LLMs do not generalize to longer hop lengths, even when provided with test-time compute. %
\citet{xu2026depth} studied how many latent planning steps a model can learn to perform. %
They found that transformers trained from scratch discovered at most three latent steps, while finetuned frontier models reached five and a frontier model reached seven with few-shot prompting. %
They also found that a strategy discovered at five steps generalized to eight steps at test time, separating the difficulty of discovering a strategy from the difficulty of executing it, and that this separation was not resolved by scale. %
\citet{wang2026do} similarly found that the largest model was not the best, and that the optimal parameter budget scaled with the entropy of the graph search task rather than with model capacity. %

\citet{lee2025selfimproving} showed how a `self-improving' process can act as a form of extended curriculum. %
They show OOD length generalization by training the model on its own OOD predictions, thereby extending its in-domain distribution without extra human supervision or annotation. %
This technique is interesting in that it blurs the lines of what OOD means. %
They consider mazes, which are topologically equivalent to random trees with $|\mathcal{N}|=30$. %


How task decomposition affects learnability has been studied \citep{wies2023subtask, dziri2023faith, abbe2024how, pmlr-v291-wang25a}. %
This often concerns length or other OOD generalization \citep{zhou2024what, zhou2024transformers}. %
\citet{zhao2026shattered} considered the learning dynamics of arithmetic and sorting tasks.  They find that models learn multiple sub-skills in parallel.  This is similar to our finding that models do not learn to generalize across $k'$ on path-star graphs, where each $k'$ is a sub-skill.  They concluded that training data correlation and distributional shifts were the cause of this.  However, that can not be the reason for failure in our case, as we only consider in-domain distributions and go to great lengths to avoid spurious correlations with the training data. %

As mentioned in Sec \ref{sec::problem-NTP}, \citet{yao-etal-2025-language-models} considered a synthetic natural language $k\leq4$-hop task.  They empirically demonstrated that the number of unique training instances needed to learn the task grows exponentially with $k$.  MMHT improves this to be polynomial with $k$.  They called this an increase in `training budget', but the growth is in the number of instances, not training steps (as is our main concern).  We suggest readers consider this an `annotation budget', where, despite using a synthetic dataset, they asked the question, `if one has a $k$-hop task, how many training instances do we need to annotate?' and answered that it grows exponentially with $k$ without the use of MMHT. An increase in data, even when it is `non-factual', can help accelerate grokking \citep{abramov2025grokking}.  \citet{wang2026the} showed that for some tasks, uniform data sampling is sub-optimal, and sampling from a power distribution leads to a significant improvement in learnability. %

Compositional tasks also relate to curriculum learning due to the impact of MMHT \citep{frydenlund-2025-language, yao-etal-2025-language-models}.  \citet{mordig2026rethinking} found no advantages to using a difficulty-based curriculum on arithmetic and logic tasks.  The benefits of a true explicit inter-sample curriculum may be masked or already provided by implicit intra-sample MMHT, for many compositional or multihop reasoning tasks.  They considered that difficulty can be characterized in various ways.  We believe this to be an important point that is often neglected.  We also show this as the number-of-hops and task-length are different sources of difficulty (leading to both MMHT and MMLT in \A \ref{appx:MMLT}). %

We recommend \citet{yuan-etal-2026-barriers} as a primer on theoretical issues concerning discrete reasoning with language models. %

\subsection{The Path-Star Task and Path-Star Graphs}\label{appx:PST-prior-work}

Here we list works that consider either the path-star task proper, a modified version, or path-star graphs.\footnote{We went through each work listed on Semantic Scholar which cited \citet{bachmann2024the} and filtered these for ones which build on their work.  This was done in May 2026.  Citations are listed in temporal order; however, publication dates are used where possible.  This means that some published papers actually predate preprint-only papers when considering preprint dates.}  The task originally set out by \citet{bachmann2024the} did not use MMHT.  This is important as introducing MMHT, which can be done in various ways -- and ways which may not be obvious -- will significantly change the difficulty of the task \citep{frydenlund-2025-language}.  For example, various works introduce MMHT: \citep{saparov2025transformers, frydenlund-2025-language, prakash2025can, patel2025insertion, patel-etal-2026-xlm, patel2025insertion, bohnet2026analysis}.  Others modify or replace the graph topology \citep{saparov2025transformers, frydenlund-2025-language, lee2025selfimproving}, or introduce other changes \citep{thankaraj2025looking, pipis2025wait}. {\bf This is often done without consideration that this itself changes the learnability of the task.}  See \citet{xu2026depth} for a work which explicitly considers the original task without using MMHT. They find that small LMs can learn the task for $k=4$ for $d \in \{2,\, 3,\, 4,\, 5\}$ and that certain LLMs can be finetuned to learn it consistently up to $k=5$.  Interestingly, when this happens, they show that such models also get OOD generation to up to $k' = k+4$.   

We group the works by method (some using multiple methods): {\bf next-token prediction:} \citep{saparov2025transformers, lee2025selfimproving, xu2026depth}, {\bf multi-token prediction:} \citep{yin-etal-2024-semformer, frydenlund-2025-language,  gerontopoulos2025multi, zuhri2025predicting, mahajan2026beyond, teoh2026nextlatent, huang2026transformers}, {\bf discrete diffusion:}  \citep{ye2023diffusion, frydenlund-2024-mystery, ye2025beyond, patel2025insertion, patel-etal-2026-xlm, trainin2026discrete, wang2026trainability}, {\bf reinforcement learning:} \citep{setlur2024rl, zhou2026qsharp, bobbili2025pita, prakash2025can, monea2025breadcrumbs, wei2026provable}\footnote{\citet{wei2026provable} is slightly different but we included it as we believe it will be of interest to others doing work with RL.}, {\bf novel architectures} (sometimes to support MTP): \citep{yin-etal-2024-semformer, hu2025learning, ahn2025efficient, mahajan2026beyond, teoh2026nextlatent}, and {\bf scratchpads or intermediate sequential reasoning:} \citep{frydenlund-2025-language, thankaraj2025looking, alberghi2025on, xu2026depth}. 

{\bf We reiterate that we do not believe the path-star task or path-star graphs should be used to benchmark novel methods.  Solving the task is not the point of the task; understanding why it is difficult is the point.}

Concurrent to this work, \citet{huang2026transformers} studied the learning dynamics of MTP on the path-star task and proved that MTP and NTP induce different dynamics.
Their theoretical analysis was limited to $d=2$, $k=2$ graphs and a two-layer disentangled transformer.
Their experiments used $d \in \{2, 5\}$ with $k=4$.
These results should be compared against the NTP results of \citet{xu2026depth}, who learned graphs of this size without MTP, and against our MTP results.  We encourage the reader to consult their work directly, as the methodology and scale differ substantially from ours.

\citet{bohnet2026analysis} showed an equivalence between Blocksworld and the topology of path-star graphs, where each stack in Blocksworld corresponds to an arm in a path-star graph.  To identify a stack containing a target block, the model must traverse the stack to find a target block, which requires solving the same task as the path-star task.  Thus, a subproblem of Blocksword is expressible as the path-star task.  However, they use stacks of different sizes, thus inducing MMHT \citep{frydenlund-2025-language}.  Also, they only consider using reasoning-enhanced LLMs and evaluating the task via one-shot in-context learning with (non-latent) CoT.  Hence, it was evaluated under very different conditions, and any insights on learnability or performance are very different. 

\subsection{Alternative Inputs or Tokenizations}\label{appx:alt-token}

Our alternative input scheme, STAN, relates to multiple lines of work.  First, work on permutation invariance \citep{NIPS2017_f22e4747, chen2023positional}.  This relates to alternative model parameterizations which forego the causal constraint, including non- and iterative-autoregressive models, discrete diffusion models, and other architectures \citep{gu2018non, ghazvininejad-etal-2019-mask, monea2023pass, austin2021structured, nie2026large, yang2019xlnet, liao-etal-2020-probabilistically, chen2025pearl}. 

Second, work on positional embeddings to structure task information. Much like the path-star task, multi-digit arithmetic is a seemingly simple latent multihop reasoning task that is non-trivial for LMs to learn \citep{lee2024teaching, zhou2024what}.\footnote{Seemingly simple from a human perspective, but one should be careful not to assume that this implies that it is easy to learn for a model. Thank you to Graeme Hirst for pointing that out.
} This often has to do with the positional embeddings used \citep{shen2023positional}.  One solution which makes the task more learnable is to structure the input by reusing or coupling positional embeddings for both operands \citep{mcleish2024transformers, cho2024position, cho2024arithmetic}.  For STAN, we essentially reuse the same `positional' token `|' for all edges (and likewise we reuse `/' for all the nodes).

\citet{wen2026frontier} recently showed that LLMs fail to learn simple copy tasks.  To address this, they proposed a 2D variant of RoPe that couples positions along the new dimension. They showed (a) empirically allows the models to solve the task and (b) expands the expressivity of 1-layer models to solve this task.  Note that this couples positional embeddings; however, the underlying inductive bias is that similar positions have similar or related content.  As such, an alternative strategy is to couple the content tokens under a single positional embedding, which is what we do.\footnote{They also showed that 2D RoPe applies to general models on natural language text, with line breaks being used to determine the second axis. Whereas our strategy is based on a data-specific inductive bias: that an edge is a single entity.}  

Third, work on compressing or reducing the size of the input stream, which often also relates to the use of continuous representations as inputs \citep{zhang2026soft, hao2025training, tan2026think, tang2026multiplex}.  This has been used in conjunction with MTP to improve both training and inference efficiencies \citep{peng2026efficient, tan2026pair, xiong2026superthoughts}.  

As an aside, this relates to continuous chain-of-thought, which autoregressively inputs the model's prior hidden-state into the model instead of decoding this to a discrete token  \citep{hao2025training}.  This leads to the concept that this hidden- or latent-state is a `super-position' of multiple {\em potential} discrete tokens, and that these can encode multiple reasoning paths in parallel, thus acting as a form of implicit search across multiple paths (see \citet{cui2026latent} and \citet{rizvi2026illusion} for demarcations of this behaviour). \citet{zhu2025reasoning} then proved that this continuous CoT allows a two-layer LM to express graph reachability more efficiently than discrete CoT.\footnote{Where graph reachability is equivalent to finding any path, not just the shortest path.}  On the surface, such work looks highly related to ours; however, it is different despite having similar terminology. First, when they describe a latent process, they are not performing implicit latent multihop reasoning across layers as in our setup.  Rather, reasoning is performed {\em explicitly} by sequentially generating autoregressive `latent' tokens, which are `latent' by the fact that they are not articulated or surfaced as discrete tokens.  Thus, expressibility and learnability are very different.  Second, they hypothesize that their compressed superposition inputs are encoding multiple search paths.  This differs from us and the work above, which explicitly compresses multiple sequential tokens from a single path.  Note that in our latent multihop reasoning, such multi-path `super-position' hidden-states may also exist, just across the model layers  \citep{saparov2025transformers, cohen2025spectral}.  

Finally, work on alternative encodings specifically designed for graph data (often using a GNN as an encoder or `prefix')  \citep[][among many others]{ren2024survey,   perozzi2024let, shirai2025less, behrouz2025best, bechler2026lost}. \citet{yehudai2026depth} compared three graph tokenizations: edge-lists, rows of the adjacency matrix, and Laplacian eigenvectors.
They found that adjacency rows outperformed edge-lists on three molecular property prediction tasks.
This reduced the number of tokens from $O(|\mathcal{E}|)$ to $O(|\mathcal{N}|)$, which was their stated motivation.
Like STAN, this changes how local edge information is packed into tokens while also reducing the token count.
Unlike our setup, they did not separate the effect of the tokenization from that of the shorter sequence.

\subsection{Multi-Token Prediction}\label{appx:mtp}

Multi-token prediction is a generalization of the NTP learning paradigm \citep{bengio2003neural, NIPS2015_7137debd}. NTP uses a sequential self-supervision rule that algorithmically creates a single target for each input.
MTP generalizes this sequential self-supervision rule to create multiple targets per input, where the labels are in some future window of size $n$ \citep{stern2018blockwise, goodman-etal-2020-teaforn, qi-etal-2020-prophetnet, monea2023pass, gloeckle2024better, cai2024medusa}.  Thus, MTP is actually a form of multi-label or structured prediction \citep{6471714, bakir2007predicting, liu2009learning}.  

Various LLMs have incorporated MTP into their pre-training \citep{liu2024deepseek, xiao2026mimo, team2026qwen3}.  In addition to improving inference and regular pre-training, MTP can also improve RL post-training \citep{xiao2026mimo}.  MTP also benefits small LMs \citep{aynetdinov-akbik-2025-babies}.

MTP provides benefits during training, where it can lead to better models and/or faster convergence \citep{gloeckle2024better, peng2026efficient}. 
\citet{gloeckle2024better} speculated that MTP upweights `choice points', citing \citet{bachmann2024the}, via a counting argument over a single fixed sequence.
Their choice points are not formally defined and are not the choice nodes we define, which follow from the graph topology.
Their tasks were induction over natural language and polynomial arithmetic, neither of which is a search over a graph, so our theoretical model does not hold for them.
They speculated as to why MTP helps and did not prove it.

It can also improve inference-time throughput when paired with speculative decoding  \citep{stern2018blockwise, cai2024medusa, xia2022speculative, chen2023accelerating, leviathan2023fast}.  However, as our work concerns learnability and scalability, we do not consider the benefits of MTP for inference.\footnote{Even though RITF can achieve faster inference with speculative decoding \citep{zuhri2025predicting}.} 

The main practical challenge of MTP is inducing multiple predictions at a single time step.  This can be achieved in various ways; however, the most common way is to modify the architecture to parameterize $n$ predictions for each step \citep{gloeckle2024better, cai2024medusa}.  This can be made dynamic by using a sequential model \citep{ankner2024hydra, liu2024deepseek}.  

Alternatively, \citet{gerontopoulos2025multi} introduced the idea that new hidden-states can be induced via inputting mock `register' tokens.  These are a kind of masked token with positional information telling the model which target token to predict from a given register input.  
A major limitation of this is the increased sequence length.  As such, they need to subsample tokens in the future window, resulting in non-contiguous windows \citep{liu2025lmtp}. 
\citet{zheng2026temporal} used a method which combines both projective MTP and mock registers.\footnote{
Interestingly, \citet{zheng2026temporal} framed future predictions as being `deprived of [] critical immediate context', which leads to `these heads naturally producing a high-entropy, context-deficient distribution that perfectly fits the definition of an amateur' \citep{zheng2026temporal}.  {\bf This runs contrary to the main argument of our work and \citet{frydenlund-2025-language}, where this context-deficiency and high-entropy are precisely why MTP is beneficial for multihop reasoning tasks.}  The lack of immediate context is what creates a mixture of hop sizes, leading to an increase in the number of loss terms and thus SSE.  The high entropy of some of these terms is an indication of the curriculum-like mix of difficulties among these terms.  Thus, what they are describing as a deficiency can also be a benefit depending on the tasks for which MTP is applied. } The relationship between MTP and sequence length has been explored in other recent works \citep{peng2026efficient, tan2026pair, xiong2026superthoughts}.

 {\bf Another alternative for MTP with a dynamic future window can be achieved without modifying the model or the inputs.  This is done by treating the MTP as a multi-label or structured prediction task, where future tokens are predicted using a structured loss function that encodes the future predictions into a given pre-existing hidden-state} \citep{yin-etal-2024-semformer, frydenlund-2025-language}.    

The most intuitive method for this is the BoW MTP method introduced by \citet{yin-etal-2024-semformer}.\footnote{They introduced this as a strong baseline to test their novel `Semformer' model against on the path-star task, where the BoW baseline nearly works as well (explained by the induction of MMHT \citep{frydenlund-2025-language}).  Kudos to them for developing such a strong baseline when the goal of the paper was to establish the value of their Semformer model.  To the best of our knowledge, their BoW model is the first method to treat MTP as a structured prediction task. {\bf  Given the number of recent works that also treat MTP as a structured prediction task, we believe the scientific value of this `baseline' was understated in their work.}}
This models the assumption that each future token is equally valid at position $i$.  Thus, BoW is equivalent to uniform label smoothing over the future window \citep{frydenlund-2025-language}.  {\bf This lack of order is a major limitation for this method}.  We know that the above assumption is not true, and it ignores temporal or sequential information for which we have explicit supervision.  This issue can be solved by using a structured loss that preserves the order, using the {\em learning-to-rank} paradigm \citep{luce1959individual, plackett1975analysis, cao2007learning, xia2008listwise, liu2009learning, pmlr-v130-reddi21a, lee-etal-2024-methods}.   Examples of these are a pairwise ranking loss \citep{frydenlund-2025-language} or the Plackett-Luce loss defined in Eq.\ \ref{eq::PLD}. See \A \ref{appx:plackett-luce} for more technical details of Plackett-Luce. 

{\bf The major limitation of this approach is that all tokens in the dynamic future window must be unique in order to support ranking, so the size of that window is data-dependent.}  As such, the benefits of this MTP approach will be minimal when there are many repeated targets in the sequence.  Practically, this issue can be addressed by just masking out repeats  \citep[][\A \ref{appx:khops}]{zuhri2025predicting}.    

Other recent works considered a structured BoW loss \citep{mahajan2026beyond, peng2026efficient}. \citet{zuhri2025predicting} introduced a near-identical rank-based MTP method as \citet{frydenlund-2025-language}, except they used a non-probabilistic listwise loss instead of a non-probabilistic pairwise loss (and we use the probabilistic listwise Plackett-Luce loss in this work).  Both works applied it to the path-star task; however, \citet{zuhri2025predicting} also considered other tasks in natural language modelling as well, and their work was otherwise far more focused on evaluating the method itself, whereas \citet{frydenlund-2025-language} was focused on the effect of MMHT on the path-star task.  

Various works showing that hidden-states naturally encode future information support the structured prediction view of MTP  \citep{pal-etal-2023-future, wu2024do, pochinkov2025parascopes}.   
\citet{men-etal-2024-unlocking}, \citet{jenner2024evidence}, and \citet{dong2025emergent} considered how this behaviour leads to planning.   Models trained using NTP also encode their most probable tokens in early layers, with their next probable token being encoded next, etc. \citep{lioubashevski2025looking}.  We suspect that MTP will lead to a similar saturation order, and the layer one uses to perform MTP may affect such behaviour. \citet{zhong2026toward} showed how projection-based MTP promotes compressed belief-states, and that it can lead to incorrectly prioritizing far-future tokens.  They applied it to in-weights path search, both on synthetic ER graphs and urban street graphs. \citet{zhong2025understanding} also showed the effectiveness of MTP on in-weights path search.  \citet{ustaomeroglu2026internal} considers the interpretability of models trained for the in-context shortest path task for both NTP and MTP; however, they use a max size of $N=28$ and $k=\{4,\, 6\}$.  

\citet{tang2026k} used noise to induce a joint distribution over the $n$ predictions, which relates to both MTP, masked discrete diffusion, and other noise-based generative models.

\citet{samragh2025your} showed how to convert an NTP-trained LLM to generate multiple tokens via masking, similar to  \citet{gerontopoulos2025multi}.  

\citet{noci2026thinking} showed how to self-supervise continuous chain-of-thought via MTP where the continuous chain-of-thought is used to create hidden-states to support MTP.  

As an aside, \citet{suleymanzade2026mux} considered how to train continuous chain-of-thought via direct supervision on the latent thought tokens.  They use distillation from a teacher model to produce reasoning traces.  They then create a weighted token embedding to represent an entire $n$-length sub-span of the full trace.     
This has an insightful relationship with MTP and multi-label prediction.  Because no ground-truth reasoning traces exist, they can not use the sequential self-supervision rule employed by MTP to algorithmically establish ground-truth multi-tokens and thus must resort to a pre-existing LLM, leading to a distillation process.  However, once individual targets were found, they then used a weighted BoW loss to compress these into a single latent supertoken.  They considered three different weightings, which are handcrafted or data-independent.  This is because the teacher model can not form a distribution that is restricted to just these sequential labels.  Thus, distillation provides labels but not weights.  This means that the core problem is the same as the multi-label version of MTP, with the complication that the ground truths can not be algorithmically determined, and they solved the weighted BoW problem using handcrafted weighting functions.  \citet{Frydenlund_Singh_Rudzicz_2022} showed how this can be converted to a ranking task.  This work also showed that no single weighting outperformed all the others in all situations.  However, these different weightings created different orderings between the tokens.  This suggests that the inductive bias we assume with RITF in that the tokens are ordered from near to far may not be universal.  They also choose the weighting functions specifically for their ability to reconstruct the original labels, suggesting these methods may be useful for MTP.    

\subsection{Masking, Non-Autoregressive Models, and Discrete Diffusion Models}\label{appx:masking-as-mtp}

Token masking, non-autoregressive models (NAR), masked discrete diffusion models (MDM) are closely related to MTP when applied to multihop reasoning tasks due to the fact that all three methods induce MMHT in a similar way \citep{frydenlund-2025-language}.  Whereas MTP results in a latent skip of conditional information, this is done explicitly with masking, which is a necessary aspect to both training and inference of NAR and MDM models \citep{lee-etal-2018-deterministic, ghazvininejad-etal-2019-mask, austin2021structured, nie2026large}.  This may also explain the success of MDM on other symbolic multihop reasoning tasks \citep{ye2025beyond, ye2025implicit, he2026reasoning}.
This connection is easy to see in light of the method of \citet{gerontopoulos2025multi}, which explicitly uses masks or registers to support MTP. 
One outcome of this is that discrete diffusion induces a subset of the MTP cost terms for graph tasks (See \A \ref{appx:discrete-proof}).  Note that inducing the same cost terms does not imply that expressibility or learnability is the same between LMs and discrete diffusion models \cite{svete2026on}. Additionally, NAR and MDM combine a masking-based loss with a non-causal model architecture.  This is related to our use of prefix LMs.

\citet{bachmann2024the} used a `teacher-less' model, which masks out all input tokens \citep{monea2023pass}.  This is a special case where it is equivalent to a 'fully' non-autoregressive model in terms of its induced cost terms \citep{gu2018non, gu-kong-2021-fully}.\footnote{The `teacherless' model is slightly different in terms of parameterization, which masks all target-side predictions to be independent.} It is also an MTP model, as this `teacher-less' method leads to learning the terms in Eq.\ \ref{eq:mtp} when $i=1$.  This also helps explain why this `teacherless' method is less preformative than other MTP methods, as it only induces $k$ terms, and these are the most difficult terms as they fully mask or skip all target-side conditional information.  


\citet{hao2025training} used a masking curriculum to remove parts of reasoning traces.  This promotes the model to instead use its own auto-regressively produced latent states to conduct multihop reasoning.  This masking out steps in the reasoning trace may induce MMHT.  Masking has been shown to be effective in other graph and reasoning tasks, which frames it as a form of `dependency regularization' \citep{luo2024graphinstruct, chen-etal-2024-masked, pham-etal-2025-clozemath}.

\section{A Visualization of the Sequential Supervision Problem under NTP and the Benefits of MTP}\label{appx:visualization}

\begin{figure}[h]
\centering
\includegraphics[trim={0.cm 0.cm 0.cm 0.0cm},clip, scale=0.7]{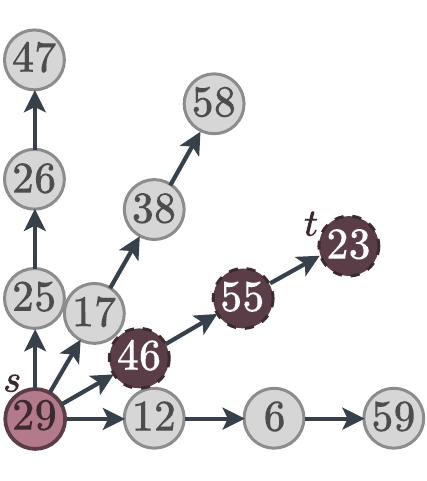}
\caption{A path-star graph with query, $Q=(29,\, 23)$.
Node 29 is the sole choice node, with 55 and 23 predictable by shortcuts.}
\label{fig:graph-ps}
\end{figure}

\begin{figure}[h]
\centering
\includegraphics[trim={0.cm 0.0cm 0.cm 0.cm},clip, scale=0.7]{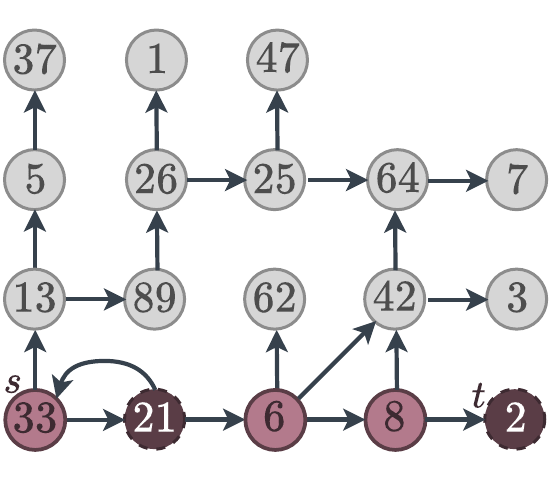}
\caption{A directed graph with $Q=(33,\, 2)$.
33, 6, and 8 are choice nodes.
{\bf 21 is not, despite having out-degree 2; as 33 is in the path, it is removed from consideration.}  The backward edge from 21 to 33 is why the choice set must be defined over unseen neighbours rather than by out-degree alone.}
\label{fig:graph-choice1}
\end{figure}

\newlength{\topow}
\setlength{\topow}{0.87\linewidth}   

\begin{figure}[]
    \centering
    \begin{subfigure}[b]{1.0\textwidth}
        \centering
        \includegraphics[trim={.65cm 0.7cm .25cm 2.8cm},clip, width=\topow]{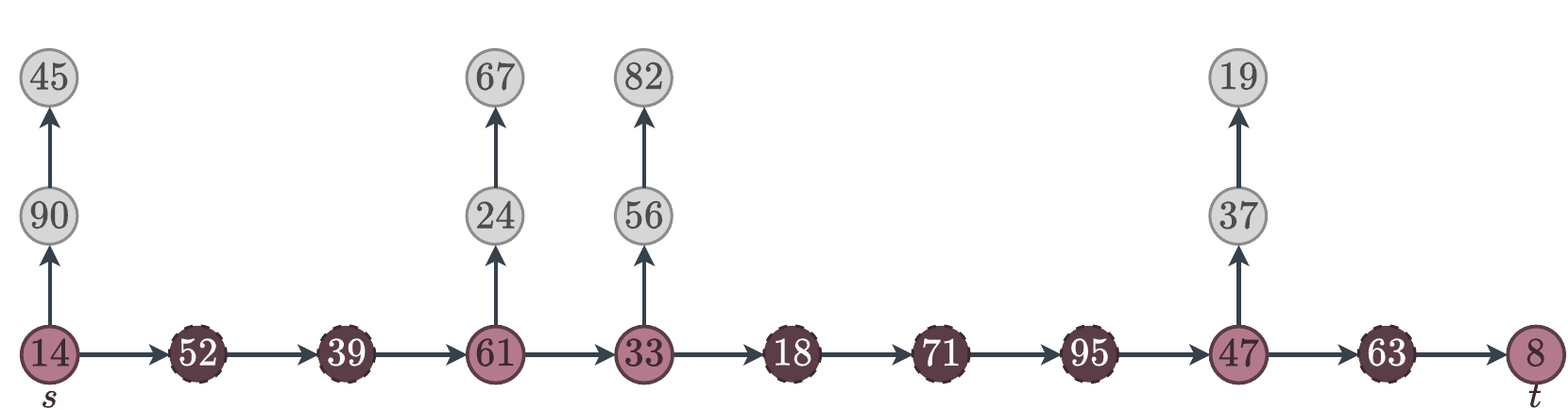}
        \caption{A graph with a sampled path $s$ to $t$.  Choice nodes (light maroon) support learning under NTP.}
        \label{fig:topology-grid-a}
    \end{subfigure}\\[-2pt]
    \begin{subfigure}[b]{1.0\textwidth}
        \centering
        \includegraphics[trim={.65cm 0.7cm .25cm 2.8cm},clip,width=\topow]{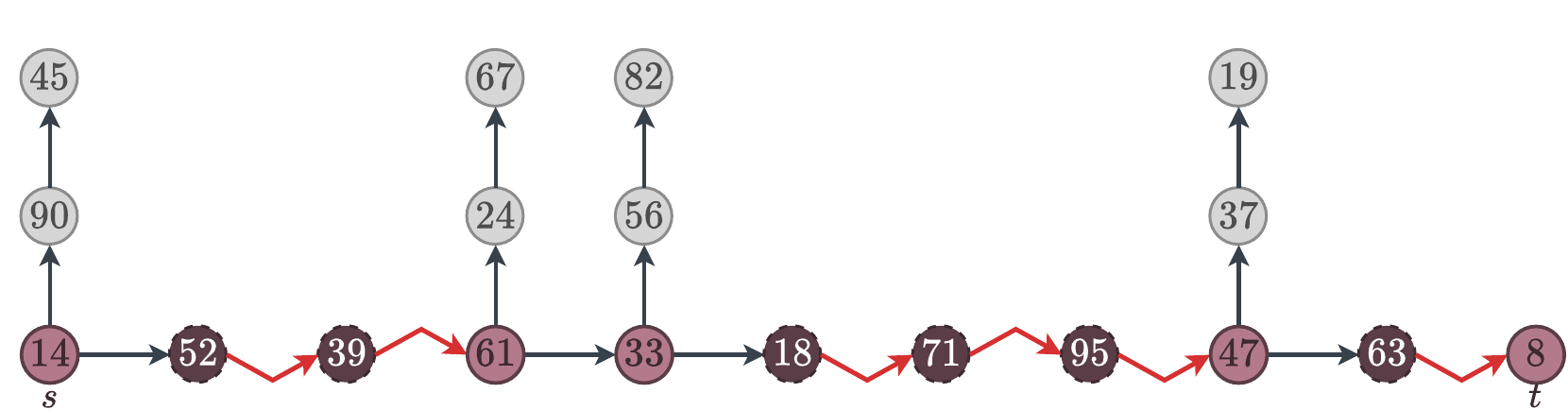}
        \caption{Non-choice nodes (dark maroon) support trivial predictions via the CHC (red zigzag edges), absorbing sequential supervision \citep{bachmann2024the}.  This shows where it happens on an arbitrary topology.}
        \label{fig:topology-grid-b}
    \end{subfigure}\\[-2pt]
    \begin{subfigure}[b]{1.0\textwidth}
        \centering
        \includegraphics[trim={.65cm 0.7cm .25cm 2.8cm},clip,width=\topow]{figs/panels/reg-panel-c.pdf}
        \caption{MTP induces dense quadratic (to $k$) supervision.  MTP restores non-choice nodes by inducing supervision connections.  Unlike regular edges, these are latent connections (dashed), and thus are not in the source-side prefix and do not support the CHC.  They also require multihop learning (tiny nodes), with the edge colour indicating the number of hops.}
        \label{fig:topology-grid-c}
    \end{subfigure}\\[-2pt]
    \begin{subfigure}[b]{1.0\textwidth}
        \centering
        \includegraphics[trim={.65cm 0.7cm .25cm 2.8cm},clip,width=\topow]{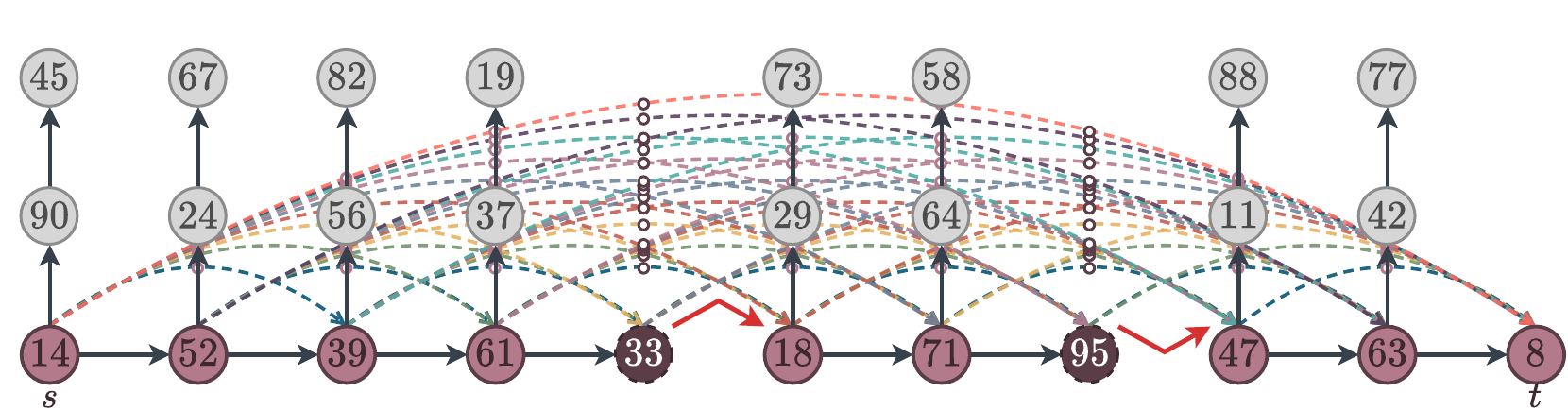}
        \caption{A {\em good} graph topology leading to a path with more choice nodes.  Notice how the MTP supergraph remains the same, showing that MTP is topologically independent (outside of the NTP connections).}
        \label{fig:topology-grid-good}
    \end{subfigure}\\[-2pt]
    \begin{subfigure}[b]{1.0\textwidth}
        \centering
        \includegraphics[trim={.65cm 0.7cm .25cm 2.8cm},clip,width=\topow]{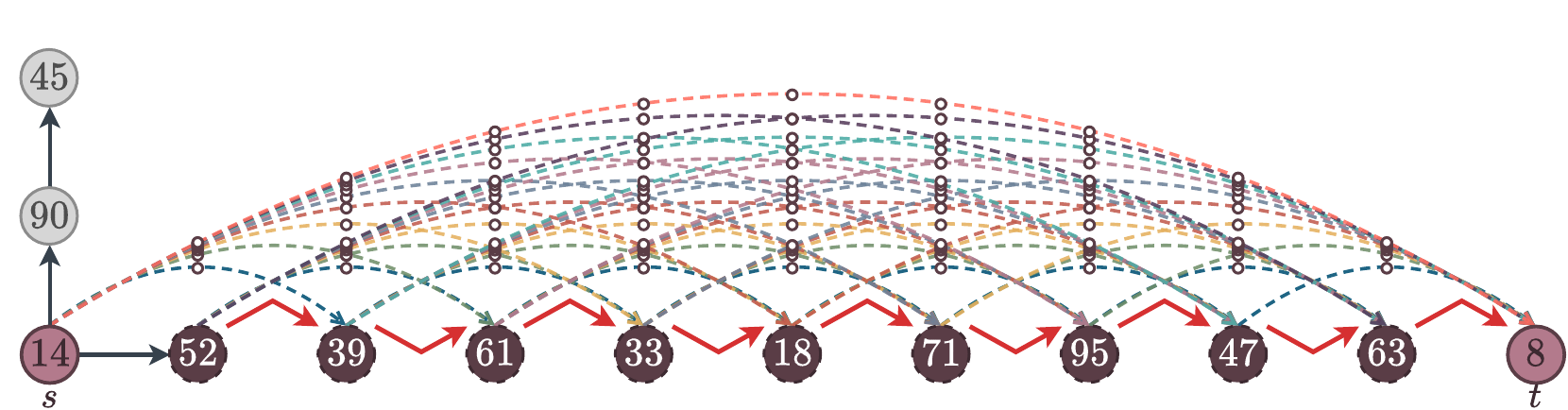}
        \caption{A {\em bad} graph topology, such as with an adversarial path-star graph leading to a path with a single choice node.  However, the MTP supergraph remains the same.}
        \label{fig:topology-grid-bad}
    \end{subfigure}
    \caption{{\bf The Benefits of MTP.} MTP supervision (c) is dense, unlike NTP (b), and topology-invariant as shown by (d)-(e).}
    \label{fig:topology-grid}
\end{figure}

\section{A Visualization of the Causal Constraint, Sparsity, and Fragmented Routing}\label{appx:causal-constraint-vis}

\newlength{\fragw}
\setlength{\fragw}{0.86\linewidth}   

\begin{figure}[H]
    \centering
    \begin{subfigure}{1.0\textwidth}
        \centering
        \includegraphics[trim={0cm 0.0cm .0cm 1.8cm},clip,width=\fragw]{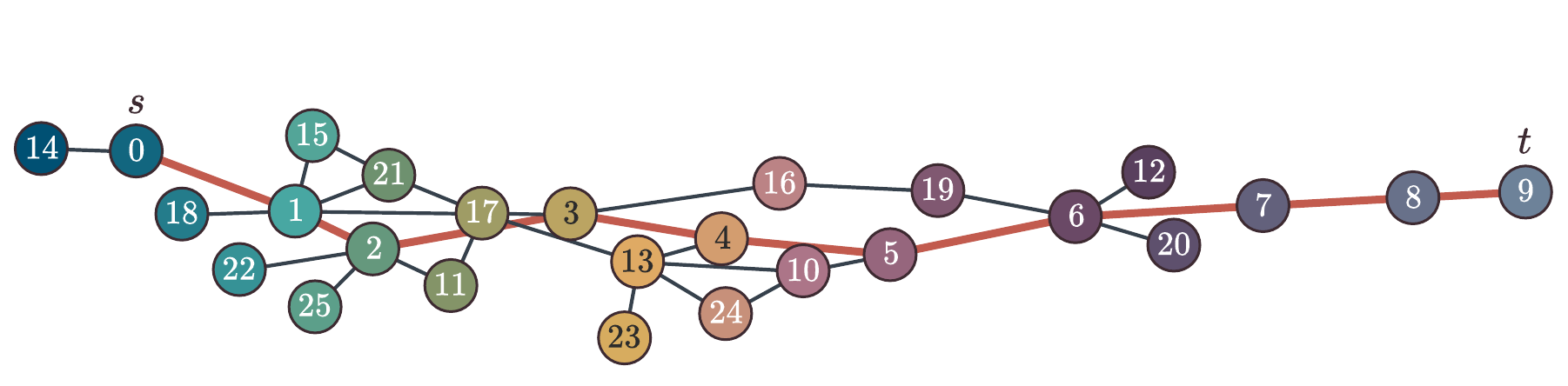}
        \caption{A sparse graph with a path from $s$ to $t$ highlighted.  Node colours help identify edges.}
        \label{fig:frag-a}
    \end{subfigure}%
    \\1
    \begin{subfigure}{1.0\textwidth}
        \centering
        \includegraphics[width=\fragw]{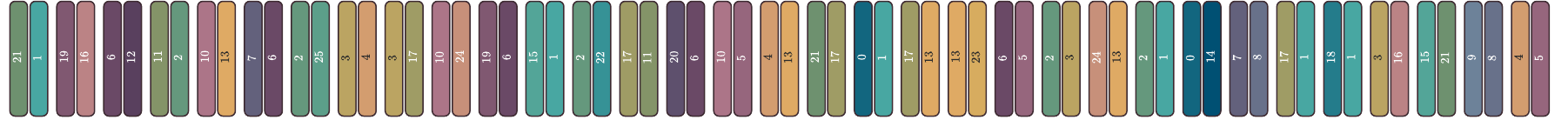}
        \caption{The shuffled edge list $E$ in the source-side prefix, providing a {\em stream} of edges under causality.}
        \label{fig:frag-b}
    \end{subfigure}%
    \\
    \begin{subfigure}{1.0\textwidth}
        \centering
        \includegraphics[width=\fragw]{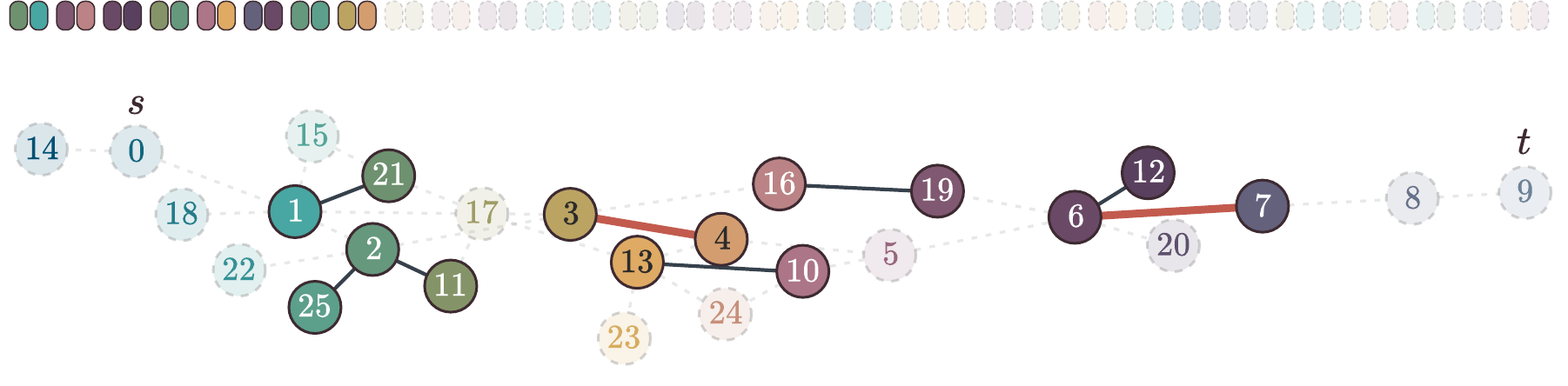}
        \caption{After streaming $8$ of $33$ edges, the observable graph is $18$ isolated components.}
        \label{fig:frag-c}
    \end{subfigure}%
    \\
    \begin{subfigure}{1.0\textwidth}
        \centering
        \includegraphics[width=\fragw]{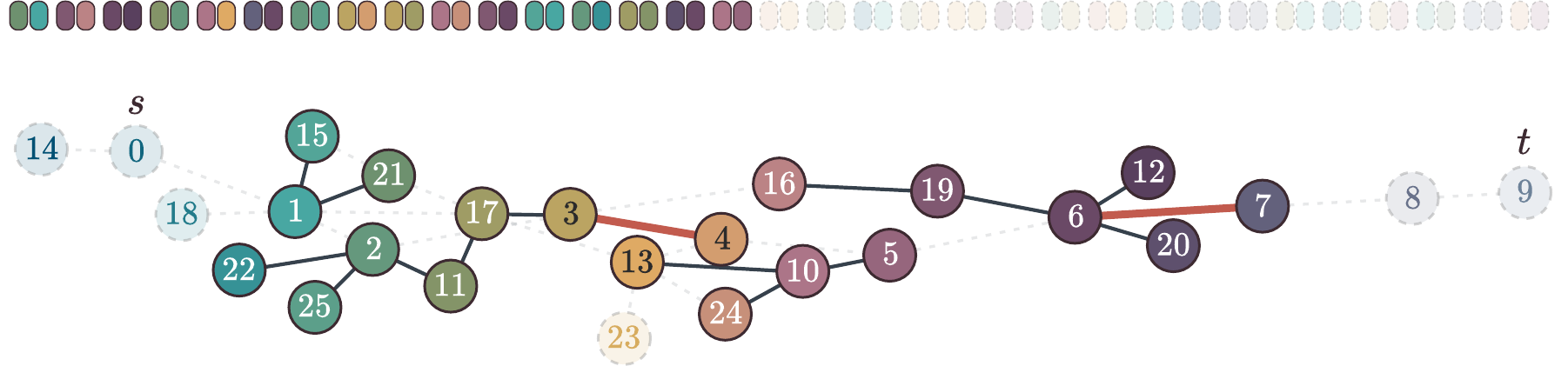}
        \caption{At the halfway point, $16$ of $33$ edges, with $10$ components.}
        \label{fig:frag-d}
    \end{subfigure}%
    \\
    \begin{subfigure}{1.0\textwidth}
        \centering
        \includegraphics[width=\fragw]{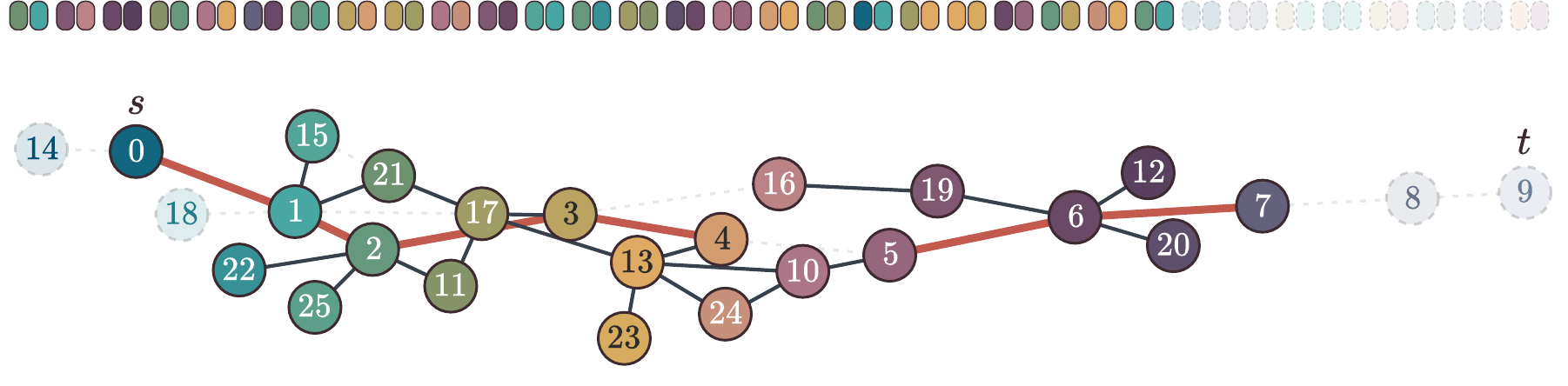}
        \caption{At $25$ of $33$ edges, there are $5$ components.  However, $s$ and $t$ are still unconnected.}
        \label{fig:frag-e}
    \end{subfigure}%
    \\
    \begin{subfigure}{1.0\textwidth}
        \centering
        \includegraphics[width=\fragw]{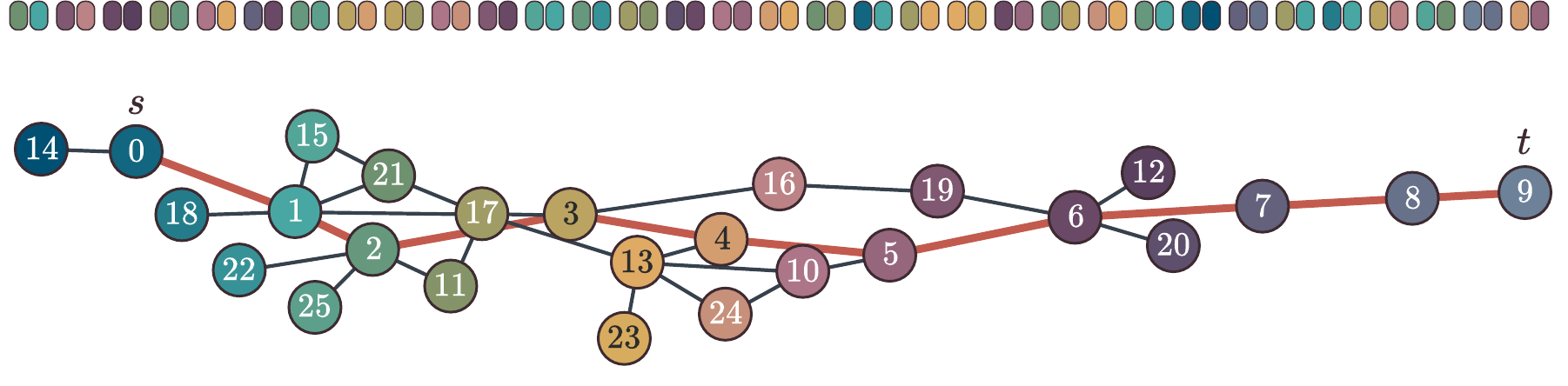}
        \caption{Only once all edges are observed does the shortest path from $s$ to $t$ exist.}
        \label{fig:frag-f}
    \end{subfigure}
    \caption{{\bf Fragmented routing.} A sparse graph with $|\mathcal{V}| = 26$,
    $|\mathcal{E}| = 33$ and $k = 9$. Panels (c)--(f) show the graph as a stream of edges under a causal LM.  As information can only
    be routed across an edge once that edge has been observed, routing will be fragmented (with dotted edges and faded nodes not yet observed).
    The strip above each of (c)--(f) reproduces the stream of $E$ from (b) at a given time in the stream.}
    \label{fig:frag-routing}
\end{figure}

\section{A Visualization of Tokenization and Causal Parameterization}\label{appx:STAN-vis}

This section provides visualizations of the two tokenizations: sequential annotation (SEAN) vs. structural annotation (STAN), and the two model parameterizations: causal LM vs. prefix-LM.  See Fig. \ref{fig:models}. 

\newcommand{\mdwidth}{1.\linewidth}
\begin{figure}[H]
  \centering
  \begin{subfigure}{\linewidth}
    \centering
    \includegraphics[width=\mdwidth]{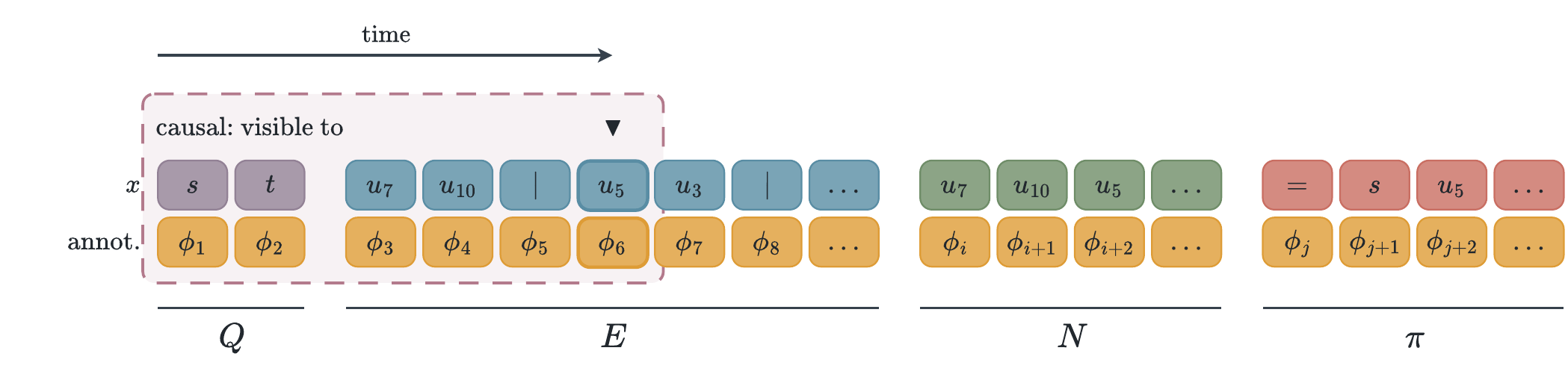}
    \caption{Causal attention with SEAN. Every token attends only to its left
      context, so the highlighted query token cannot see the remainder of the
      edge list $E$ that defines the graph it must search over.}
    \label{fig:model-causal}
  \end{subfigure}\\[1.2ex]
  \begin{subfigure}{\linewidth}
    \centering
    \includegraphics[width=\mdwidth]{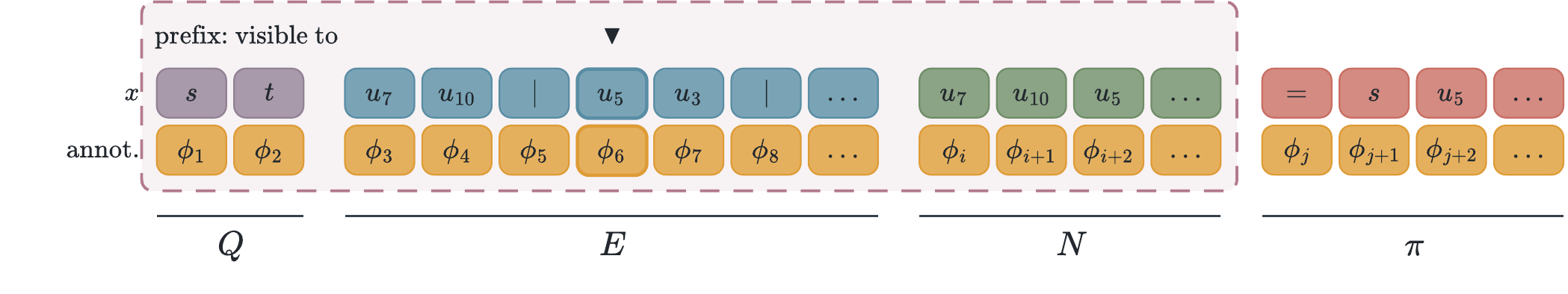}
    \caption{Prefix attention with SEAN. The query $Q$, edge list $E$ and node
      list $N$ form a bidirectional prefix, so the same query token attends to
      the whole graph; only the path $\pi$ remains causally masked.}
    \label{fig:model-prefix}
  \end{subfigure}\\[1.2ex]
  \begin{subfigure}{\linewidth}
    \centering
    \includegraphics[width=\mdwidth]{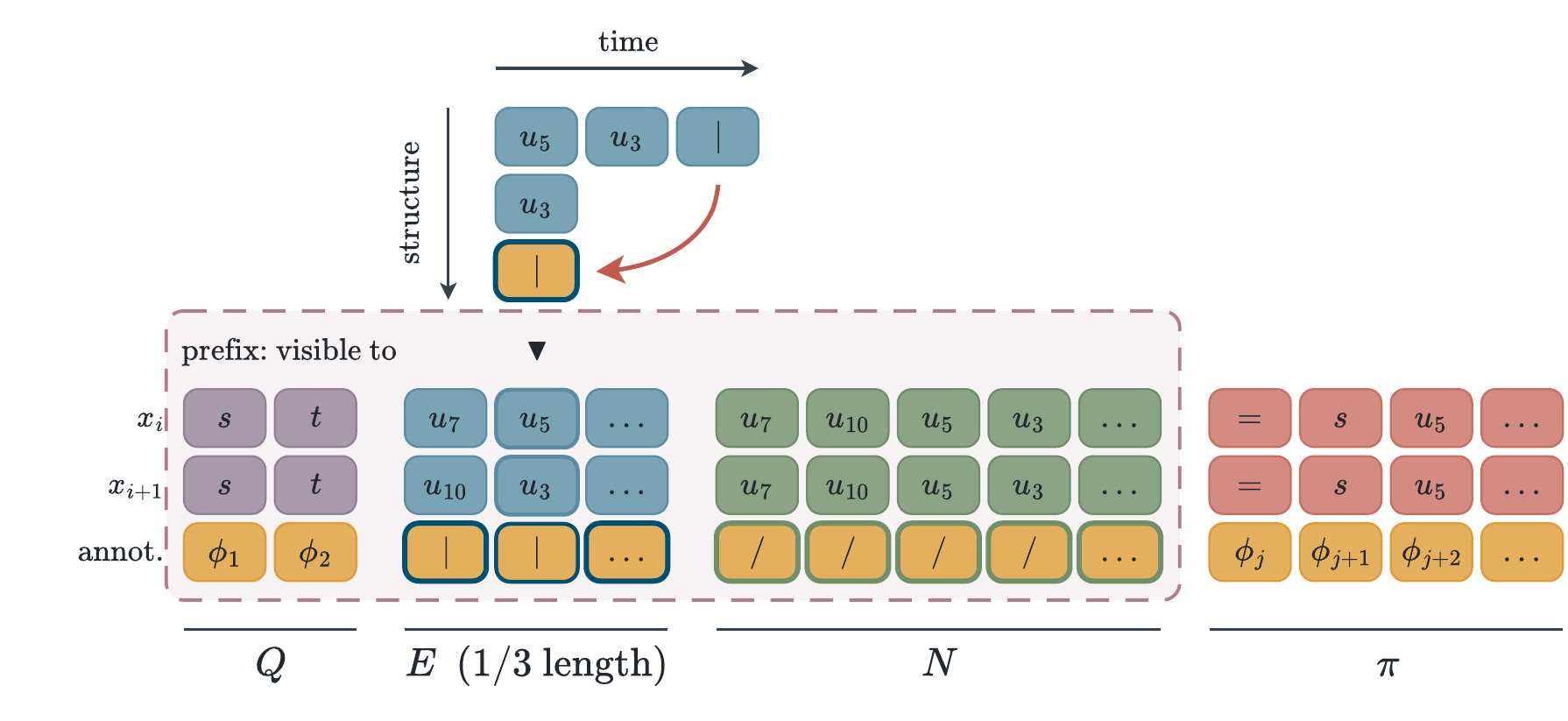}
    \caption{Prefix attention with STAN. The structure that SEAN spends separator tokens on is moved into the annotation channel, so an edge $u\,v$ occupies a single position.  One can view this as a transposition from the time dimension to the structural dimension.  As a consequence, $E$ becomes $1/3$ the length.  Combining STAN with the prefix-LM allows for true permutation invariance.}
    \label{fig:model-stan}
  \end{subfigure}
  \caption{{\bf Attention pattern (causal vs.\ prefix) and tokenization
    (SEAN vs.\ STAN).}
    The bottom row of every panel is the \emph{annotation channel}: SEAN fills
    it with positional annotations $\phi$, whereas STAN overwrites it with
    structural markers ($\mid$ for edge boundaries, $/$ for node-list entries).
    The shaded region marks what the highlighted token ($\blacktriangledown$)
    can attend to.}
  \label{fig:models}
\end{figure}

\section{Ranking-into-the-Future with Plackett-Luce}\label{appx:plackett-luce}

This section gives a brief tutorial on learning-to-rank and Plackett-Luce, and details how we apply it to MTP.

Consider a supervised task where, for each input $x_i$, we are given labels $y_i = y_i^1, \dots, y_i^n$ and a preferred ordering $r_i$ among them, and we want a model whose scores $l_i = l_i^1, \dots, l_i^n = h_i W_o$ reflect this ordering. Learning-to-rank methods differ based on what kind of structural information $r_i$ provides. A \emph{point-wise} approach treats $r_i$ as a set of interval scores and regresses $l_i$ against them directly. A \emph{pairwise} approach instead assumes $r_i$ only gives partial or relative information, e.g.\ $y_i^5 \succ y_i^3$, and converts this into pairwise comparisons, optimizing $l_i^5 > l_i^3$, etc.; this is the approach used for RITF by \citet{frydenlund-2025-language}. A \emph{listwise} approach assumes $r_i$ gives a full (or near-full) ordering over $y_i$ and models this directly \citep{cao2007learning, xia2008listwise, liu2009learning}.

MTP is naturally listwise. We construct $r_i$ using the MTP self-supervision rule, letting $y_i$ be the next $n$ tokens after position $i$, and assuming tokens closer in time are preferred to those farther away. This gives a full ordering over $y_i$, but only a partial ordering with respect to the vocabulary, since $n \lll |\mathcal{V}|$ in general.\footnote{We also assume the $n$ future tokens are unique, which holds for shortest paths by construction. We further assume no ties exist; this is not quite true here, since multiple valid shortest paths can create ties.  We ignore this for MTP but not NTP (see \A \ref{appx:label-smoothing}).  As a side note, modelling unknown ties exactly requires considering order permutations, which is computationally expensive \citep{ma2021learning}, though \citet{Frydenlund_Singh_Rudzicz_2022} showed known ties can be modelled efficiently.}

There are multiple ways to implement a listwise loss \citep{cao2007learning, liu2009learning, zhu2020listwise}. \citet{zuhri2025predicting} use a non-probabilistic loss based on ListNet \citep{cao2007learning}. We instead use the probabilistic Plackett-Luce loss, which models the maximum likelihood of the observed ordering, i.e.\ $\log p(r_i^1 \succ \dots \succ r_i^n)$ \citep{luce1959individual, plackett1975analysis, xia2008listwise}.

For MTP, the labels $y_i^1, \dots, y_i^n$ are just the path nodes $u_i, \dots, u_{i+n-1}$, and $r_i$ realizes the temporal order $u_i \succ u_{i+1} \succ \dots \succ u_{i+n-1}$ used in the main text (Eq.~\ref{eq::PLD}). We therefore drop the generic $y_i^j/r_i^j$ notation from here on and write everything in terms of the nodes directly, matching Eq.~\ref{eq::PLD}.

Recall from the main text that $u_j^i$ denotes the logit for node $u_j$ produced from the hidden state $h_i$, and that the softmax in Eq.~\ref{eq::PLD} is taken over $\mathcal{Z}_j = \mathcal{N} \setminus \{u_i, \dots, u_{j-1}\}$, the set of nodes not yet ranked at step $j$. We restate this here without the softmax shorthand $\sigma$, and correct $\mathcal{N}$ to $\mathcal{V}$: the normalization is technically over the full token vocabulary, not just the node vocabulary, since the logit vector $l_i = h_i W_o \in \mathbb{R}^{|\mathcal{V}|}$ assigns mass to every vocabulary item, node or otherwise.\footnote{This distinction is immaterial to learning, since non-node tokens are never part of $\pi$ and so never enter $\mathcal{Z}_j$'s complement, but it is necessary to get the partition function right as actually implemented.} Thus,
\begin{equation}\label{eq::PLD2}
\begin{split}
p(u_i \succ \dots \succ u_{i+n-1}) &= \prod_{j=i}^{i+n-1} p(u_j \,|\, u_i, \dots, u_{j-1}) \\
&= \prod_{j=i}^{i+n-1} \frac{e^{u_j^i}}{\displaystyle\sum_{v \,\in\, \mathcal{Z}_j} e^{v^i}}, \quad \text{where } \mathcal{Z}_j = \mathcal{V} \setminus \{u_i,\, \dots,\, u_{j-1}\}.
\end{split}
\end{equation}
For implementation, it is convenient to reindex by rank-step rather than by absolute future offset: let $l_i \in \mathbb{R}^{|\mathcal{V}|}$ be the logit vector at step $i$, permuted so that $l_i^j = u_{i+j-1}^i$ for $j = 1, \dots, n$ -- i.e.\ position $j$ holds the logit of the token $j$ steps into the future, so rank order and temporal order coincide for MTP. Since $\mathcal{Z}_j$ excludes exactly the $j-1$ already-ranked nodes, and these occupy the first $j-1$ positions of $l_i$ after reindexing, the set-sum over $\mathcal{Z}_j$ collapses to a contiguous range, $\sum_{v \,\in\, \mathcal{Z}_j} e^{v^i} = \sum_{v=j}^{|\mathcal{V}|} e^{l_i^v}$, giving
\begin{equation}
\begin{split}
p(l_i^{1} \succ \dots \succ l_i^{n}) & = \prod_{j=1}^{n} p(l_i^{j} \,|\, l_i^{1}, \dots, l_i^{j-1} ) = \prod_{j = 1}^{n} \frac{e^{l_i^{j}}}{\sum_{v = j}^{|\mathcal{V}|} e^{l_i^{v}}},
\end{split}
\end{equation}
and the loss for a single time-step at $i$ is simply the negative log-likelihood of this,
\begin{equation}
L_{\mathrm{RITF}} = -\log p(l_i^1 \succ  \dots \succ l_i^{n})= \sum_{j = 1}^{n} \log \sum_{v=j}^{|\mathcal{V}|} e^{l_i^{v}} - {l_i^{j}} = \sum_{j = 1}^{n} \log\Big(Z_i - \sum_{v < j} e^{l_i^{v}}\Big) - {l_i^{j}},
\end{equation}
where $Z_i = \sum_{v=1}^{|\mathcal{V}|} e^{l_i^{v}}$ is the full partition function or normalizing term i.e.\ the scalar realization of summing over $\mathcal{Z}_1 = \mathcal{V}$. This distribution represents a generative process of sampling a softmax distribution {\em without replacement} and works by re-normalizing the distribution at each rank-step $j$ by excluding previous samples. This is described by the iterative implementation in Alg. \ref{alg:pl}.

\begin{algorithm}
\caption{The Plackett-Luce Loss Given a Ground-Truth Rank Ordering via an Iterative Renormalization Process.}
\label{alg:pl}
\begin{algorithmic}[1]
\REQUIRE Vector of ordered logits $l_i = [l_i^1, l_i^2, \dots, l_i^{|\mathcal{V}|}]$ at time-step $i$
\REQUIRE The number of rank labels $n$

\STATE \textbf{Initialize:} 
\STATE $Z_i \leftarrow \sum_{u=1}^{|\mathcal{V}|} e^{l_i^u}$ \COMMENT{Compute initial full partition function}
\STATE $\mathcal{L} \leftarrow 0$ \COMMENT{Initialize log-likelihood accumulator}

\FOR{rank-step $j = 1$ \TO $n$}
    \STATE $Z_i^j \leftarrow Z_i$ \COMMENT{When $j=1$ this is just regular softmax-CE}
    \IF{$j > 1$}
        \STATE $Z_i^j \leftarrow Z_i - \sum_{v=1}^{j-1} e^{l_i^v}$ \COMMENT{Exclude accumulated past sample masses from partition function}
    \ENDIF
    
    \STATE $p_j \leftarrow \frac{e^{l_i^j}}{Z_i^j}$ \COMMENT{Compute step-dependent renormalized probability}
    \STATE $\mathcal{L} \leftarrow \mathcal{L} + \log(p_j)$ \COMMENT{Accumulate rank-step likelihood}
\ENDFOR

\RETURN $-\mathcal{L}$
\end{algorithmic}
\end{algorithm}

\begin{figure}[]
  \centering
  \includegraphics[width=0.95\linewidth]{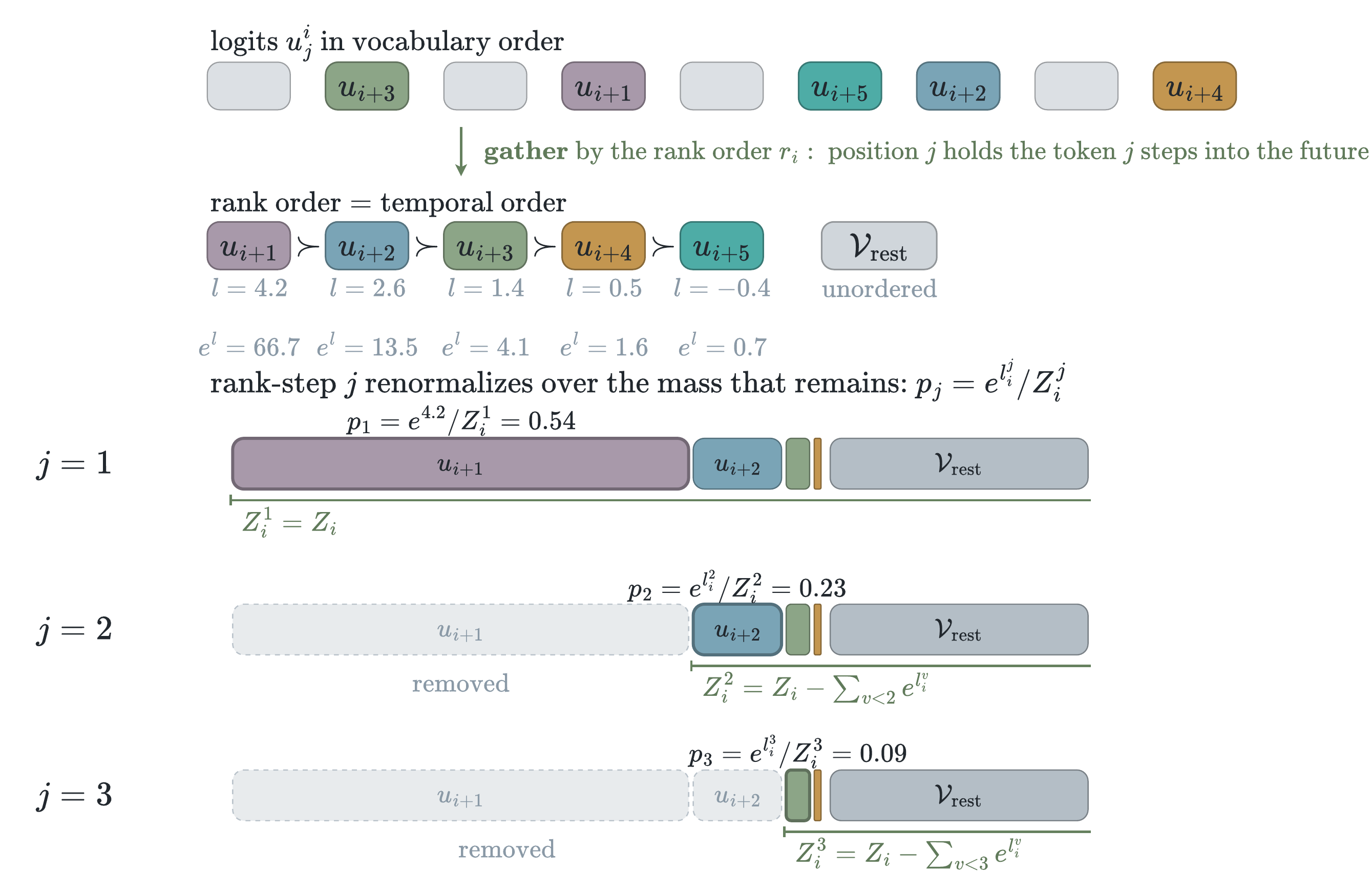}
  \caption{The Plackett-Luce loss of the MTP head (Alg.~\ref{alg:pl}), with
    concrete example logits. A gather puts the logits into the rank order
    $r_i$, which for MTP \emph{is} the temporal order: position $j$ holds
    the token $j$ steps ahead. Bar widths are the softmax of the logits
    shown, on one absolute scale across all three rows, so mass leaving the
    partition function is visible in place. At rank-step $j$ the
    already-ranked tokens are struck out and $Z_i^j$ spans only the
    remainder: each ranked token is removed from the partition function, so
    the next step renormalizes and competes only over what is left, i.e.\
    \emph{sampling without replacement} in the context of it modelling a generative process.}
  \label{fig:model-pl-mtp-b}
\end{figure}

\citet{Frydenlund_Singh_Rudzicz_2022} showed how this iteration over the rank-steps in Alg. \ref{alg:pl} can be vectorized for efficient parallel computation on a GPU.  This is done by treating $ \sum_{v < j} e^{l_i^{v}}$ as a padded cumulative summation.  This is illustrated in Fig.~\ref{fig:model-pl-mtp-b}.  Thus, the main cost compared to NTP using regular softmax-CE is a gathering call to order the logits by the ground-truth rank order to support the cumulative summation.  Note that Alg.\ \ref{alg:pl} is written in probability space for exposition.  Our implementation works in log-space, obtaining each $\log Z_i^j$ by a cumulative log-sum-exp over the candidate set rather than by subtraction.  This is stable whenever $n \lll |\mathcal{V}|$, which holds in all our experiments.  The instability that remains arises only as $n$ approaches $|\mathcal{V}|$, where $\mathcal{Z}_j$ reduces to a single element and the two quantities being cancelled become equal.

\begin{figure}[]
  \centering
  \includegraphics[width=0.95\linewidth]{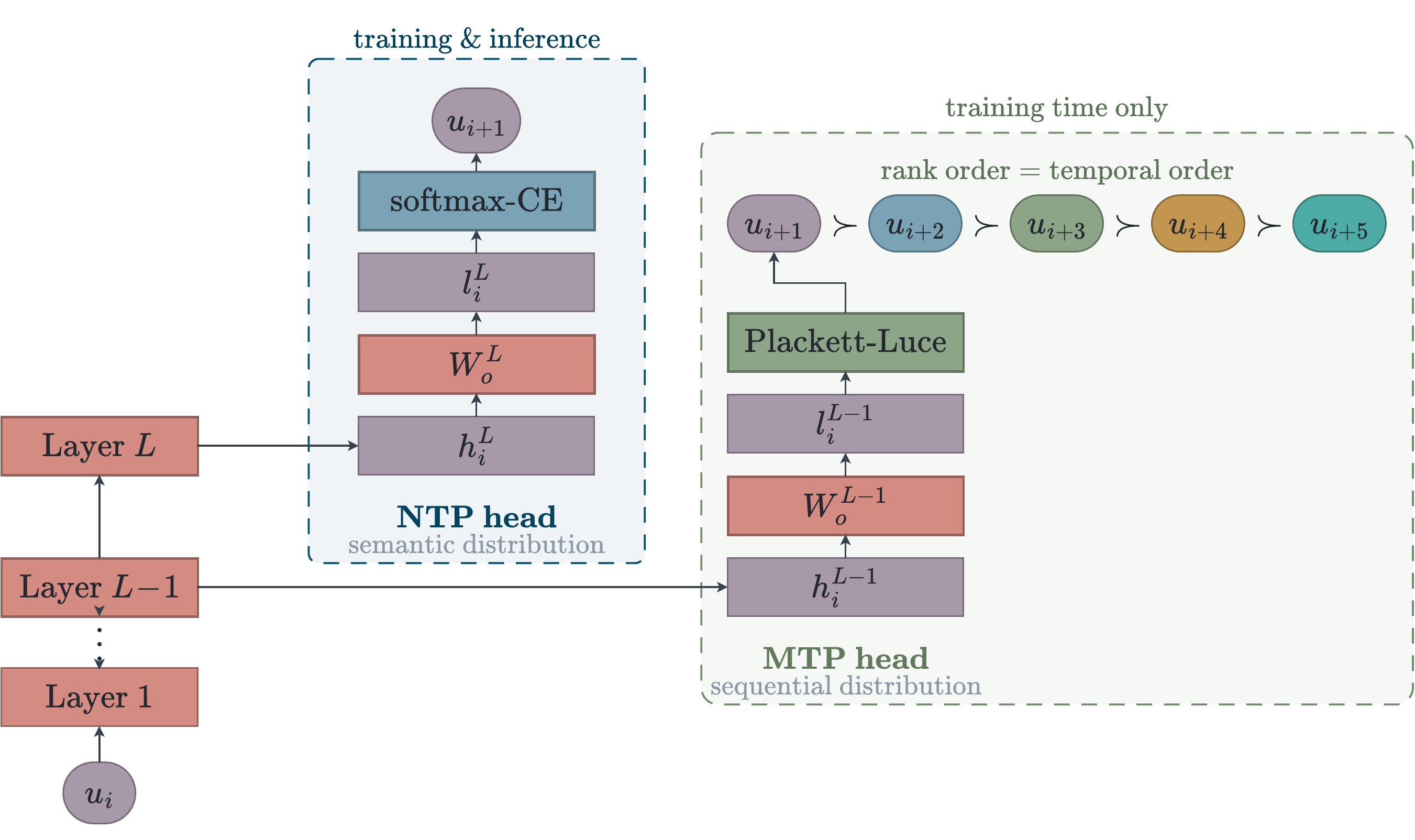}
  \caption{Architecture. The \textbf{NTP head} uses the last layer $L$,
    feeding the usual unembedding $W_o^L$ and a softmax-CE loss over the
    next token: the \emph{semantic} distribution. It is active during both
    training and inference. The \textbf{MTP head} instead taps the
    second-to-last layer $L\!-\!1$, feeding a second projection
    $W_o^{L-1}$ -- the only new parameters (or $W_o^L$ can be reused) --
    into a Plackett-Luce loss over the ordering of the next $n$ tokens: the
    \emph{sequential} distribution. The MTP head exists at training time
    only, so inference is unchanged. Squared boxes are parameters and
    operations, while rounded boxes are data.}
  \label{fig:model-pl-mtp-a}
\end{figure}

We do not actually want to use the hidden-state from the final layer to create our logits, i.e. $l_i = h_iW_o$, since this hidden-state parameterizes the model's \emph{semantic} next-token distribution where high-scoring tokens there are semantically related \citep{mikolov2013efficient}. Whereas the ranking loss instead needs a \emph{sequential} distribution, where high-scoring tokens are temporally related. Reusing the same hidden-state would conflate these two distributions. Instead, we use the hidden-state produced by the second-to-last layer, as illustrated in Fig.~\ref{fig:model-pl-mtp-a}.

\section{Proof of Optimal MTP Sequential Sample Efficiency}\label{appx:proof-optimal}

We prove that MTP covers every possible non-trivial sub-path of $\pi$, establishing optimality of SSE and hop mixture.
Term $(i,\, j)$ conditions on $\pi_{<u_i}$, so its last observed node is $u_{i-1}$ and it corresponds to the sub-path $(u_{i-1},\, u_j)$, putting the full set of $k(k+1)/2$ terms in bijection with the non-trivial sub-paths of $\pi$.
Note that $s$ is given in the query and is never a target, so it is not a term in either loss.
The one positional shortcut is the $(k,\, k)$ term, which predicts $t$ at the final position, and it is excluded by the sum bounds of both Eqs.~\ref{eq:ntp} and \ref{eq:mtp}.  Predicting $t$ is only trivial at the final position, where it is given by the query and the sequence has ended.  Predicting $t$ at an earlier position is not trivial, as the model must also determine that the path terminates there, which requires knowing the remaining distance.
Eq.~\ref{eq:mtp} is therefore the full set less that term, giving $k(k+1)/2 - 1$, and Eq.~\ref{eq:ntp} is its diagonal less that term, giving $k-1$.
Relative to predicting all $k+1$ nodes, NTP loses two terms, for $s$ and for $t$, while MTP loses the $i=0$ row and one further term.
The $i=0$ row goes because it contains the prediction of $s$ and because its remaining terms condition on $\pi_{<u_0} = \emptyset$, which is identical to $\pi_{<u_1} = \{s\}$ and so duplicates the $i=1$ row.

A non-trivial path contains at least 1 edge. 
There are $C = k(k+1)/2$ non-trivial sub-paths in a $k$ length path. Let $\pi=u_0,\,\dots,\,u_k$ be a shortest path with $k+1$ nodes and $k$ edges.   
Let $e_1,\, \dots,\, e_k$ be the edges in the path with $e_i = (u_{i-1},\, u_{i})$.
Then any two nodes form a non-trivial sub-path s.t.\ 
\begin{equation}
    P = \binom{k+1}{2} = \frac{(k+1)!}{2!( (k+1)-2 )!} = \frac{(k+1)k}{2}.
\end{equation}

Then, MTP has optimal sequential sample efficiency as every term in the loss corresponds to a non-trivial sub-path.  Proof: The full version of Eq.\ \ref{eq:mtp} is,

\begin{align} 
L^{\mathrm{full}}_{\mathrm{MTP}} &= \sum_{i=1}^{k} \sum_{j=i}^{k} \log p(u_j \mid \pi_{<u_i}, s, t, \mathcal{E}) \notag \\[10pt]
&= \begin{aligned}[t]
    &\log p(u_1 \mid u_0, \dots) + \log p(u_2 \mid u_0, \dots) \\ 
    &+ \dots + \log p(u_k \mid u_0, \dots) 
    \end{aligned} \tag{$i=1$} \\[10pt]
&+ \begin{aligned}[t]
    &\log p(u_2 \mid u_0, u_1, \dots) + \log p(u_3 \mid u_0, u_1, \dots) \\ 
    &+ \dots + \log p(u_k \mid u_0, u_1, \dots) 
    \end{aligned} \tag{$i=2$} \\[10pt]
&\vdots \notag \\
&+ \log p(u_k \mid u_0, \dots, u_{k-1}, \dots)  \tag{$i=k$} 
\end{align}

s.t.\ $i$ is the start of a sub-path and $j$ the end of that subpath.
Thus, each pair of indices $(i, j)$ in the above Eq.\ corresponds to a unique non-trivial subpath. 
The total number of terms in the expansion of $L_{\mathrm{MTP}}$ is
\begin{equation}
P = \sum_{i=1}^{k} (k - i + 1) = k + (k-1) + \dots + 1 = \frac{(k+1)k}{2} = C.
\end{equation}

{\bf Since $P = C$, every possible non-trivial sub-path maps to exactly one unique term in the $L^{\mathrm{full}}_{\mathrm{MTP}}$ loss. Then, as no other sub-path exists, there can be no other terms in the loss, thus achieving optimal sequential sample efficiency.  Likewise, as no other sub-path exists, there can be no other paths of different lengths, tthus having an optimal mixture of lengths.}  Eq.\ \ref{eq:mtp} is $L^{\mathrm{full}}_{\mathrm{MTP}}$ less the $(k,\, k)$ term, so it retains one term for every non-trivial sub-path that is not trivially predicted, and the same optimality holds.  Note that this assumes an autoregressive model, i.e.\ models the target sequence causally, as with both the prefix LMs and causal LMs.  We consider non-autoregressive discrete diffusion below (\A \ref{appx:discrete-proof}). This also assumes that the tokenization for the task is such that each token provides a single hop (\S \ref{sec:sse-mtp}). 

An additional property is that, under MTP, the proportion of terms with hop length $h$ follows a linearly decreasing distribution. 

  Proof: The full MTP loss has $k(k+1)/2$ terms total. Term $(i,j)$ predicts $u_j$ conditioned on $\pi_{<u_i}$, so the last observed node is $u_{i-1}$ and the hop length is:
  $$h = j - (i-1) = j - i + 1$$

  The number of terms with hop length $h$ is:
  $$N_h = |\{(i,j) : 1 \leq i \leq j \leq k,\ j - i + 1 = h\}| = |\{(i,\ i+h-1) : 1 \leq i \leq k-h+1\}| = k - h + 1$$

  So the proportion of terms at hop length $h$ is:
  $$P(H = h) = \frac{k - h + 1}{k(k+1)/2} = \frac{2(k-h+1)}{k(k+1)}, \quad h = 1, \ldots, k$$

  This sums to 1: $\sum_{h=1}^k \frac{2(k-h+1)}{k(k+1)} = \frac{2}{k(k+1)} \cdot \frac{k(k+1)}{2} = 1$.

  Thus, the distribution is non-uniform and linearly decreasing: $P(H=1) = \frac{2}{k+1}$ vs $P(H=k) = \frac{2}{k(k+1)}$, a ratio of $k$.

  {\bf We visualize both properties: the quadratic increase in supervision and the distribution of hop lengths as the MTP rainbow in Fig. \ref{fig:topology-grid}.  This applies to any task with MTP (Fig. \ref{fig:khops-gen-task} and Fig. \ref{fig:bfs-sp-fig}).}

  The distribution of lengths is interesting, as \citet{wang2026the} recently showed that there exist latent compositional multihop reasoning tasks where the learnability of the task increases under a data sample distribution that follows a power-law instead of a uniform distribution.\footnote{`data samples' here concerns the vocabulary samples, so by analogy to our task, this would be sampling node-ids non-uniformly.} Intuitively, this restricts the sample space during learning, which allows for learning the underlying algorithm without being distracted by data-specific values, and once this is done, the model can then learn to generalize to unseen/unsampled values.\footnote{See the mixed multi {\em length} training for Khops-Gen (\A \ref{appx:MMLT}).  This restricts how many node-ids appear per sample.
The marginal distribution over node-ids remains uniform.}    
  We suspect that a similar phenomenon may exist for the sample distribution across hop sizes, as once the model learns to compose for a small number of hops, it should be easier to generalize that to a larger number of hops; thus, one may expect that we should sample $k'$ non-uniformly and prioritize small values over larger ones. What we show is that MTP implicitly provides such a non-uniform distribution over $k$, weighted towards the easier terms.
  NTP induces one term per position, so each hop length has multiplicity one, and the best case is a uniform mixture (only on choice-node dense path).
MTP gives hop length $h$ a multiplicity of $k-h+1$, which is what produces the skew.
This is thus another potential benefit of MTP.  However, we do not know how to isolate this specific property to empirically validate this hypothesis (though it may be possible to demonstrate this with NTP (under a carefully chosen topology)).

\subsection{Masked Discrete Diffusion Models Supervise A Subset of the Sub-Paths Supervised by MTP}\label{appx:discrete-proof}
 
Here we compare masked discrete diffusion models (MDM) to MTP in terms of the induced loss terms.
Several works have proposed MDM for the path-star task (\A \ref{appx:PST-prior-work}).
Masking induces supervision over sub-paths of varying hop length, which creates similar density and mixed-hop supervision as MTP (\S \ref{sec::SSE}).
This offers insight into why MDM can learn the task, as masking provides a form of the supervision we attribute to MTP.\footnote{
It does not fully explain the behaviour, as MDM can generate in any order, which admits the reverse `solution' for PSGs. However, this is topology-specific as no reverse `solution' exists for all graphs. Also, \citet{frydenlund-2024-mystery} trained iterative autoregressive models on PSGs, which assume no order of generation and were therefore free to learn the reverse order without reverse supervision.
They did not, and instead learnt the task proper.
This suggests that the supervision is what matters, rather than the freedom to generate in any order.}
The analysis also bounds what masking provides, as MDM supervises a subset of the sub-paths supervised by MTP and induces at most $k-1$ terms per sample.

This analysis holds under the theoretical model of the shortest-path task (\A \ref{appx:bfs-sp}), where nodes are semanticless and the target is a single path, so a term's supervision is determined by which sub-path it reconstructs.
It is an analysis of which sub-paths each loss supervises rather than of learnability directly, and we do not claim it transfers to tasks where the relevant information is distributed across many tokens.
 
 Let $\pi = u_0, u_1, \ldots, u_k$ be a shortest path and let $M \subseteq \{1, \ldots, k-1\}$ be a sampled mask that designates which path tokens are unobserved, with $\bar{M} = \{0,\, \ldots,\, k\} \setminus M$ being the complementary observed positions.  Note $u_0=s$ and $u_k=t$ are always observed.   The MDM loss for a given mask $M$ is 
 \begin{equation}
 L_{\mathrm{MDM}}(M) = \sum_{j \in M} \log p(u_j \mid \pi_{\bar{M}},\, \mathcal{E}),    
 \end{equation}
 where $\pi_{\bar{M}} = \{u_i : i \in \bar{M}\}$ are the observed tokens.  
 
 The masked positions $M$ partition naturally into maximal contiguous blocks, each of which defines a subpath. 
 Let such a subpath be $\pi(u_a, u_b)$ with $u_a$ and $u_b$ being observed nodes forming the \emph{left} and \emph{right observed boundaries}, respectively, i.e.\ $a, b \in \bar{M}$, of length $k' = b - a$, s.t.\ the interior nodes $u_{a+1},\, \ldots,\, u_{b-1}$ are all masked, i.e.\ $\{a+1,\, \dots,\, b-1\} \subseteq M$.
 
 In the case where $\pi(u_0,\, u_k)$ is the fully masked path, then:
 
 1. Full masking induces $k-1$ loss terms.
 $M = \{1,\ldots,k-1\}$ yields one block with $b - a - 1 = k - 0 - 1 = k-1$ interior nodes, hence $k-1$ terms:
 $$L_{\mathrm{MDM}}(\{1,\,\ldots,\,k-1\}) = \sum_{j=1}^{k-1} \log p(u_j \mid u_0=s,\, u_k=t,\, \mathcal{E}).$$
Let $S(L)$ denote the set of sub-paths supervised by a loss $L$.  {\bf Thus, every MDM term supervises the same sub-path as a unique MTP term, so $S(L_{\mathrm{MDM}}(M)) \subset S(L_{\mathrm{MTP}})$ (Eq.~\ref{eq:mtp}), although the corresponding terms differ in their conditioning, as MDM also observes the right boundary.  Since $L_{\mathrm{MDM}}(M)$ induces at most $k-1$ terms, this means that MDM has a strictly worse SSE than MTP.}
 
 2. Each MDM term for interior node $u_j \in \pi(u_a,\, u_b)$ predicts $u_j$ given left boundary $u_a$ as its nearest observed left context, which corresponds to MTP term $(a+1,\, j)$ in Eq.~\ref{eq:mtp}, and conditions on $\pi_{<u_{a+1}}$ with $u_a$ as the last observed node.  The MDM term additionally conditions on the right boundary $u_b$, but the same sub-path $(a+1,\, j)$ is supervised.  {\bf Thus, every MDM term corresponds to a unique MTP term, and $L_{\mathrm{MDM}}(M)$ induces a subset of the terms in $L_{\mathrm{MTP}}$ (Eq.~\ref{eq:mtp}).  Since $L_{\mathrm{MDM}}(M)$ induces at most $k-1$ terms, this means that MDM has a strictly worse SSE than MTP.}
The conditioning difference is discussed below.

 3. Full masking induces a mixture of $k-1$ hop lengths.
 Each term predicts $u_j$ from the left boundary $u_0$, requiring a $j$-hop computation. The terms $j = 1, \ldots, k-1$ cover hop lengths $1, 2, \ldots, k-1$, resulting in a mixture of all $k-1$ hop lengths.  {\bf Thus, MDM induces a uniform mixture of hop lengths between $1$ and $k-1$, whereas MTP induces a non-uniform mixture weighted toward shorter subpaths, as hop length $h$ appears in $k - h + 1$ terms of $L_{\mathrm{MTP}}$.}

 4. Full masking induces a maximum number of loss terms with the longest hops.
 Since $|M| \leq k-1$ for any mask, any MDM sample induces at most $k-1$ terms. Full masking achieves this bound. It also achieves the maximum hop length $k-1$; any partial mask produces subpaths of length $k' < k$.  This must necessarily break up the longest contiguous subpath, thus reducing the maximum hop to $k'-1 < k-1$.

 Now consider the case where we have partial masking.  
 
 5. Observing interior nodes splits the path into shorter subpaths with fewer total terms.
 Observing $r$ additional interior nodes partitions $M$ into $r+1$ blocks. Each block $\pi(u_a,\, u_b)$ contributes $k' - 1 = b - a - 1$ terms. The total across all blocks is $k - 1 - r$ terms, i.e. one fewer per additional observed node.
 
 6. Shorter subpaths induce shorter hop mixtures.
 Subpath $\pi(u_a, u_b)$ of length $k' = b - a$ induces hops $1, \ldots, k'-1$ from its left boundary $u_a$. Since $k' \leq k$, the maximum hop per subpath decreases. The overall hop mixture is the union over all blocks, which covers only a subset of $\{1,\, \ldots,\, k-1\}$.  
 
 7.  For a general subpath $\pi(u_a, u_b)$ of length $k' = b - a$, the MDM terms predict $u_j$ for $j = a+1, \ldots, b-1$ with $u_a$ as the nearest observed left context.  Each such term corresponds to MTP term $(a+1, j)$ in Eq.~\ref{eq:mtp}, i.e. a contiguous prefix of row $i = a+1$, covering hops $1, \ldots, k'-1$ from $u_a$.  Thus, for any mask $M$, every MDM term across every block supervises the same sub-path as a unique MTP term.  {\bf Thus $S(L_{\mathrm{MDM}}(M)) \subset S(L_{\mathrm{MTP}})$ holds for all masks $M$, not just full masking, with the same difference in conditioning noted above.}

  In this analysis, we have made a Markovian assumption that only the left and right boundaries of the subpath matter. This is not strictly true for the shortest path. Recall that a choice node in $\pi$ is one where multiple nodes could follow that node. A choice node does not depend only on the current observed node, but also on the prior observed path, i.e.\ not just the left boundary, since the model can use the observed partial path to exclude nodes from the choice set. However, this is a property that affects \emph{learnability}, not the analysis of \emph{which loss terms are induced}, which is our focus here.
  
  Note, however, that this non-Markovian property is precisely why the reduction in terms and the change in MMHT distribution need not imply worse learnability for MDM. This is because the MDM term also conditions on the right boundary $u_b$, and on observed nodes elsewhere in the sequence, which may provide additional or different structural information beyond what the corresponding MTP term observes, and which may compensate for the fewer terms (we expect this to be more true of natural language tasks, where information is spread across multiple tokens).

  For completeness with the prior section on MTP, full-masking induces a uniform distribution over $k'$, and partial masking (using random Bernoulli masking) induces a geometric distribution.  

  {\bf The main takeaway here is that, while MTP and MDM are different, a majority of the benefits of MTP when applied to multihop reasoning are also enjoyed by MDM due to the masking procedure in the loss.  Thus, we would expect that MDM also outperforms NTP on such tasks.}  This may also explain the success of MDM on other symbolic multihop reasoning tasks \citep{ye2025beyond, ye2025implicit, he2026reasoning}.

\section{Extended Experimental Details}\label{appx:details}

{\bf Model.}
Unless stated, we consider LMs with $L = 11$ layers, $256$ dim., and 4 attention heads. %
Embeddings are initialized as one-hot but are learnable. %
Positional embeddings are learned. %
These sizes were chosen so that the task is expressible by the model, making any failure more plausibly attributable to learnability than to expressibility (\A \ref{appx:expressibility}).

{\bf Training.}
For optimization, we use Sophia \citep{liu2024sophia} and a linear-then-decreasing scheduler. %
We used a learning rate of $0.0005$, a warm-up of $500$ steps, and no L1 or L2 regularization. %
We also do not use dropout or attention dropout. %
We find that this is critical for these kinds of discrete algorithm tasks. %
The effective batch size is 1000 instances, leading to a max of $75$M instances ($>12$B tokens). %
Training is evaluated every 100 steps and stopped after achieving $>95\%$ TACC (averaged across $k' \in \{k-5, \dots, k\}$) on consecutive evals, or to a max of 75k steps. %
{\bf Each model for every experiment is trained from scratch} (for $\approx$ 2900 exps. outlined in \A \ref{appx:comp-cost}). %

{\bf Data.}
We consider three topologies: Erd\H{o}s-Renyi graphs (ERG) with an edge probability of $1/|\mathcal{N}|'$, the giant threshold for a sampled size $|\mathcal{N}|'$, Euclidean graphs (EG) with a connectivity radius of $1/\sqrt{|\mathcal{N}|'}$, and path-star graphs (PSG) \citep{erdds1959random, penrose2003random, bachmann2024the}. 
The ERG and EG random processes produce graphs with many unconnected components.
We merge all components randomly for ERG, or by distance for EG, using a single edge per component to support paths with large $k$.
When sampling paths at some $k'$, we use an \textit{ad hoc} method that also tries to choose a path in which the start node is a choice node (which is already done by construction for PSG).
This tries to ensure that a sampled path of $k'+1$ actually requires $k'$ hops to solve (\A \ref{appx:klookahead}).

We sample $k' \in \{3,\, \dots,\ k\}$ and  $|\mathcal{N}|' \in \{k+1,\, \dots,\ N\}$ for various max $k$ and $|\mathcal{N}|$ via a true curriculum (sampling inverse to the training SACC at $k'$). %
We smooth these sampling probabilities by $1/2k$; otherwise, the model would engage in catastrophic forgetting of solutions at various $k'$ it had perfected prior. %
We do not use a curriculum for PSG. %
The curriculum removes the plateaus in Fig.\ \ref{fig:psg_step}, which are the phenomenon we are measuring. %

{\bf Implementation.}
We use the vectorized GPU-friendly version of Plackett-Luce (\A \ref{appx:plackett-luce}). 
We use Fairseq to implement our models \citep{ott2019fairseq}. %
We needed to write our online datastream in C++ in order to generate enough graph instances. %
{\bf In general, because our models are so small, the datastream became the computational bottleneck and limited scaling to larger graphs.} %
For this we made use of PyBind \citep{pybind11} and Boost Graph Library \citep{siek2001boost}. %
We used the C++ code provided by \citet{saparov2025transformers} to understand technical issues such as learning PyBind and how to seed different threads. %

\subsection{Computation} \label{appx:comp-cost}

All experiments were run using a NVIDIA L40S 48GB GPU, 16-32 CPUs, and 64GB of RAM.  However, the shortest path task and Khop models can easily be trained on a 12GB TitanX.  The BFS scratchpad experiments require a larger GPU and are more compute-intensive due to the significant increase in token count.  Thus, the main computational cost of this work is in the number of experiments run.  Also, experiments like the baselines and the PSGs with NTP that fail to converge are the most expensive in terms of compute, as we must run them to the maximum number of steps.  Thus, while the max values we choose are arbitrary in terms of experimental design, they were practically chosen so that each experiment took a maximum of 24 hours to run.    

Given this, we provide a {\em rough} estimate of the computational cost required to conduct the main experiments. Table \ref{tab:cost} lists the number of experiments in Figs.\ \ref{fig:er-baseline}, \ref{fig:er-new}, \ref{fig:er-causal-lm-baseline}, \ref{fig:er-new-decoder-only}, \ref{fig:eg_k}, and \ref{fig:psg_k}.  This lists a total of 2900 experiments conducted.  Our cluster did not allow for ease tracking of hours (due to preemption), so we assume that on average each experiment took half a day or 12 GPU hours.  Thus, there are 34800 total GPU hours.  If one did not have access to a cluster such as we did, a price estimate for this using Runpod either using an L4 GPU (24GB vRAM, 50GB RAM, 12 CPU) at 0.39/hour and an RTX3090  (24GB vRAM, 125GB RAM, 16 CPU) at 0.49/hour gives us a cost of either \$13,572 or \$17,052.\footnote{These were the cheapest we found that would meet our needs.}    

{\bf Importantly, in having solved most of the variance in different seeded trials, we have made it possible to either reduce the number of trials in future work, say 3 instead of 5, or only do a single point estimate.  This significantly reduces this compute cost.} Also note, we have already benefited from this in this work compared to the prior work of \citet{frydenlund-2024-mystery} and \citet{saparov2025transformers} who used 11 and 14 trials per experiment, respectively.

\begin{table}
    \centering
    \begin{tabular}{llrrrr}
         Experiment & Figure & Num Sub Exps.   & Num $k$   & Num Trials  & Total Runs \\ \hline
         ERG, Prefix LM, Baselines &  Fig. \ref{fig:er-baseline} &  10 & 10  & 5 & 500 \\
         ERG, Prefix LM & Fig. \ref{fig:er-new} &  8 & 10  & 5 & 400 \\
         ERG, Causal LM, Baseline & Fig.\ \ref{fig:er-causal-lm-baseline}  & 6 & 10  & 5 & 300 \\
         ERG, Causal LM & Fig. \ref{fig:er-new-decoder-only}  &  8 & 10   & 5 & 400 \\
         EG, Prefix LM & Fig.\ \ref{fig:eg_k} & 8 & 10  & 5 & 400 \\
         PSG, Prefix LM &  Fig.\ \ref{fig:psg_k} & 18  & 10   & 5 & 900 \\ \hline
         &  &  &  &  & 2900 \\
    \end{tabular}
    \caption{Calculation of the total number of runs needed for the main experiments broken down by figure.  Overlapping experiments were not double counted.}
    \label{tab:cost}
\end{table}

\section{Extended Experimental Results on Shortest Path Task}

\subsection{Full-Page Versions of the Main Results Figures}\label{appx:fullpage}

We reproduce the main results figures at full-page size for legibility.
The data are identical to the main text and only the scale differs.

\begin{sidewaysfigure}
    \centering
    \sbox{\figbox}{\includegraphics[trim={.35cm .35cm 3.cm 1.cm},clip, width=0.95\textheight]{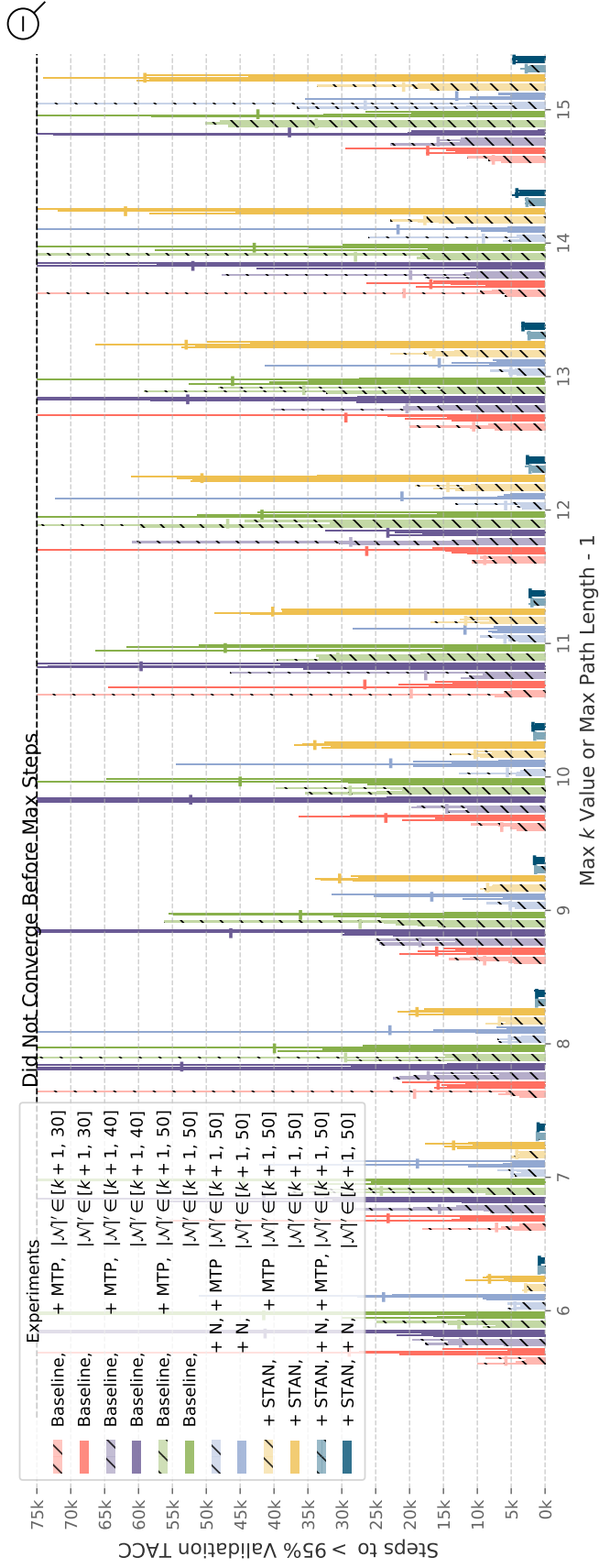}}%
    \usebox{\figbox}%
    \rlap{\hspace{0.6em}\raisebox{\dimexpr\ht\figbox-1.6ex\relax}[0pt][0pt]{\smallfig{fig:er-baseline}}}%
    \caption{Full-page version of Fig.\ \ref{fig:er-baseline}.}
    \label{fig:er-baseline-full}
\end{sidewaysfigure}

\begin{sidewaysfigure}
    \centering
    \sbox{\figbox}{\includegraphics[trim={.35cm .35cm 1.6cm .84cm},clip, width=0.95\textheight]{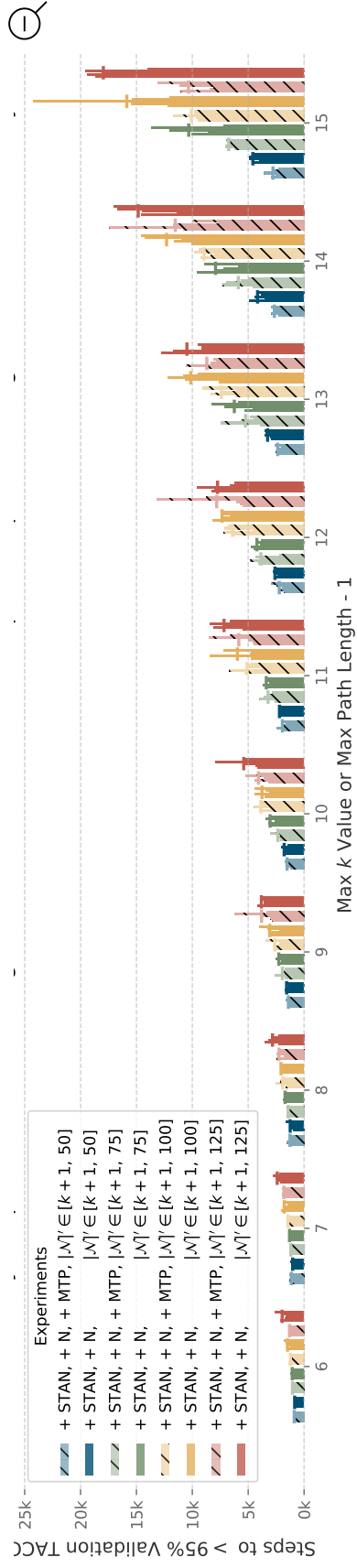}}%
    \usebox{\figbox}%
    \rlap{\hspace{0.6em}\raisebox{\dimexpr\ht\figbox-1.6ex\relax}[0pt][0pt]{\smallfig{fig:er-new}}}%
    \caption{Full-page version of Fig.\ \ref{fig:er-new}.}
    \label{fig:er-new-full}
\end{sidewaysfigure}

\begin{sidewaysfigure}
    \centering
    \sbox{\figbox}{\includegraphics[trim={.35cm .35cm 2.5cm .84cm},clip, width=0.95\textheight]{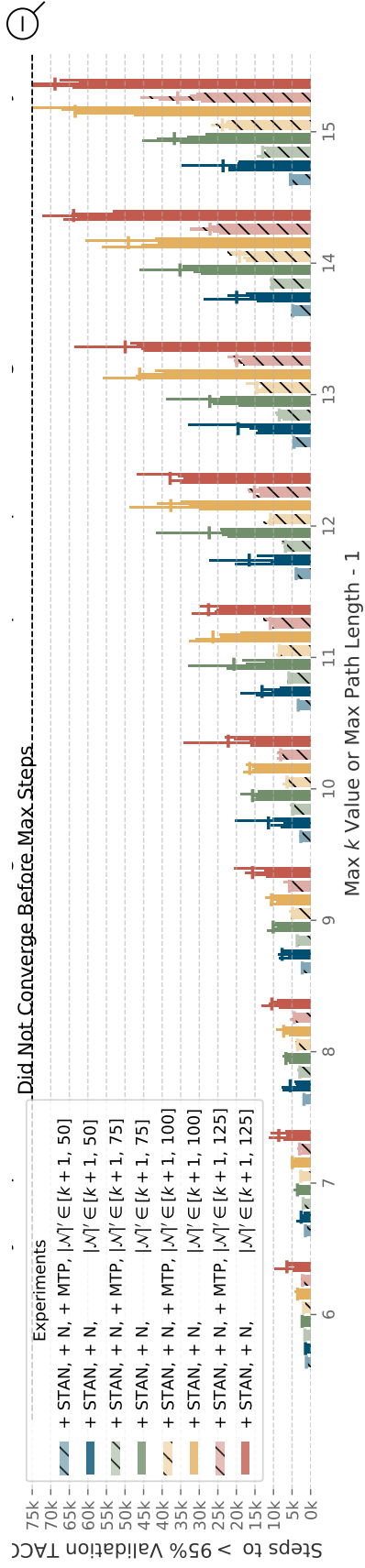}}%
    \usebox{\figbox}%
    \rlap{\hspace{0.6em}\raisebox{\dimexpr\ht\figbox-1.6ex\relax}[0pt][0pt]{\smallfig{fig:er-new-decoder-only}}}%
    \caption{Full-page version of Fig.\ \ref{fig:er-new-decoder-only}.}
    \label{fig:er-new-decoder-only-full}
\end{sidewaysfigure}

\begin{sidewaysfigure}
    \centering
    \sbox{\figbox}{\includegraphics[trim={.35cm .35cm 2.4cm .84cm},clip, width=0.95\textheight]{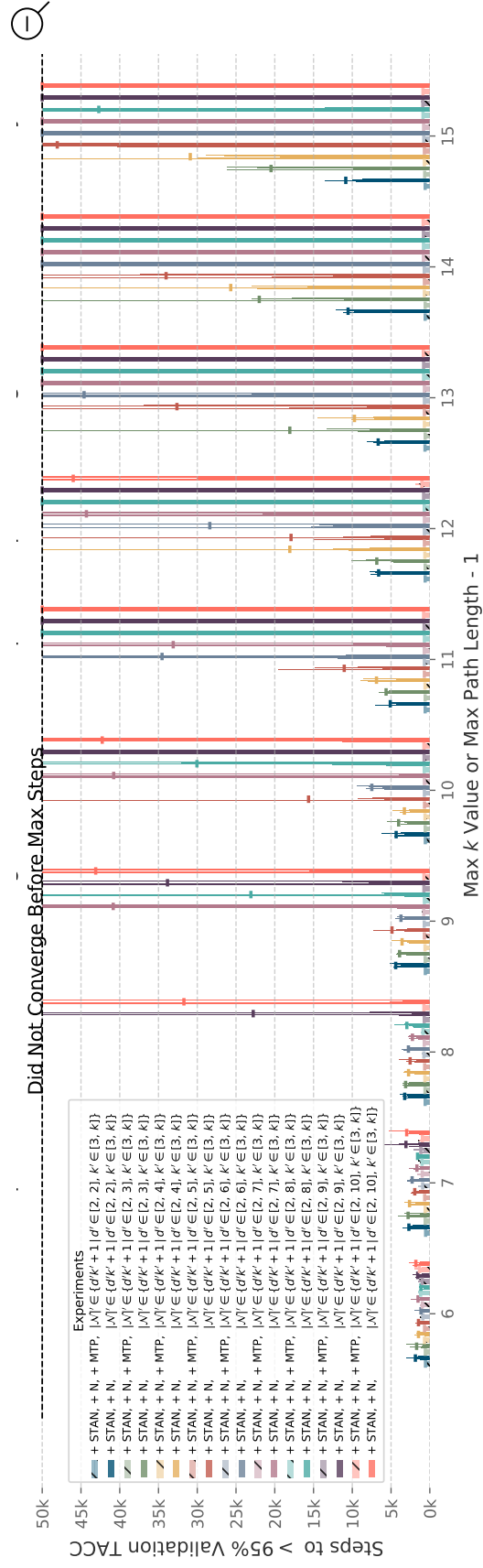}}%
    \usebox{\figbox}%
    \rlap{\hspace{0.6em}\raisebox{\dimexpr\ht\figbox-1.6ex\relax}[0pt][0pt]{\smallfig{fig:psg_k}}}%
    \caption{Full-page version of Fig.\ \ref{fig:psg_k}.}
    \label{fig:psg_k-full}
\end{sidewaysfigure}

\begin{sidewaysfigure}
    \centering
    \sbox{\figbox}{\includegraphics[trim={.35cm .35cm .35cm .86cm},clip, width=0.95\textheight]{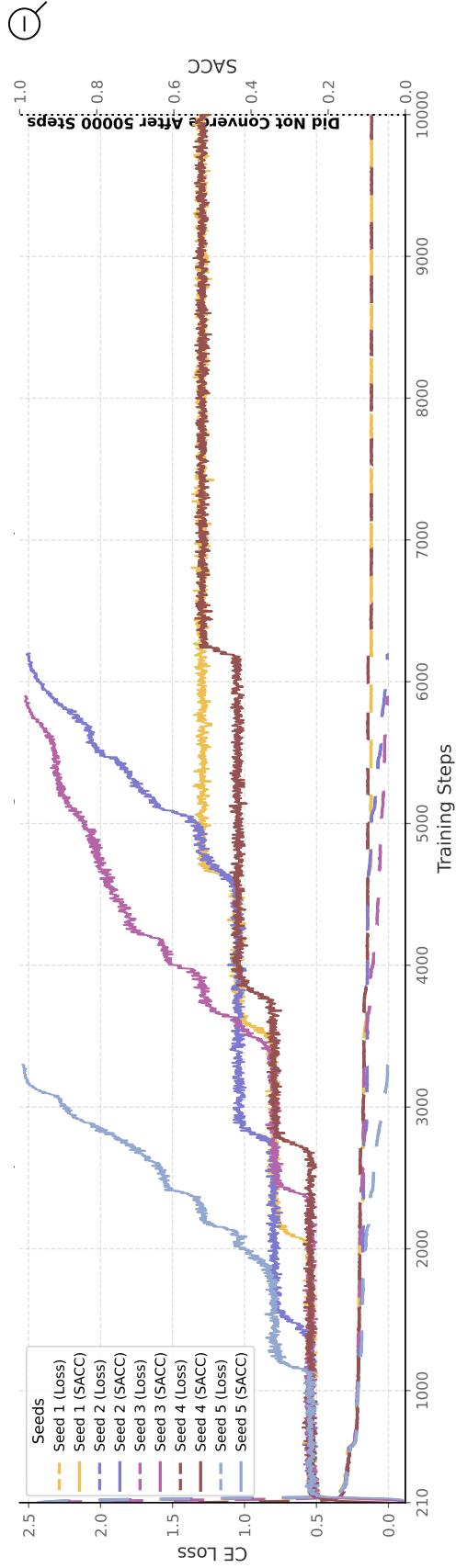}}%
    \usebox{\figbox}%
    \rlap{\hspace{0.6em}\raisebox{\dimexpr\ht\figbox-1.6ex\relax}[0pt][0pt]{\smallfig{fig:psg_step}}}%
    \caption{Full-page version of Fig.\ \ref{fig:psg_step}.  Note that, unlike in \citet{bachmann2024the} who did not use MMHT or a mixture of $d'$, the plateaus are above $1/d$ because SACC marginalizes over the $(k',\, d')$ mixture.  That is, failure to learn the task no longer results in an accuracy of $1/d$.}
    \label{fig:psg_step-full}
\end{sidewaysfigure}


\subsection{Causal LM baseline}\label{appx:er-causal-lm-baseline}
\begin{figure}
    \centering
    \includegraphics[scale=0.39]{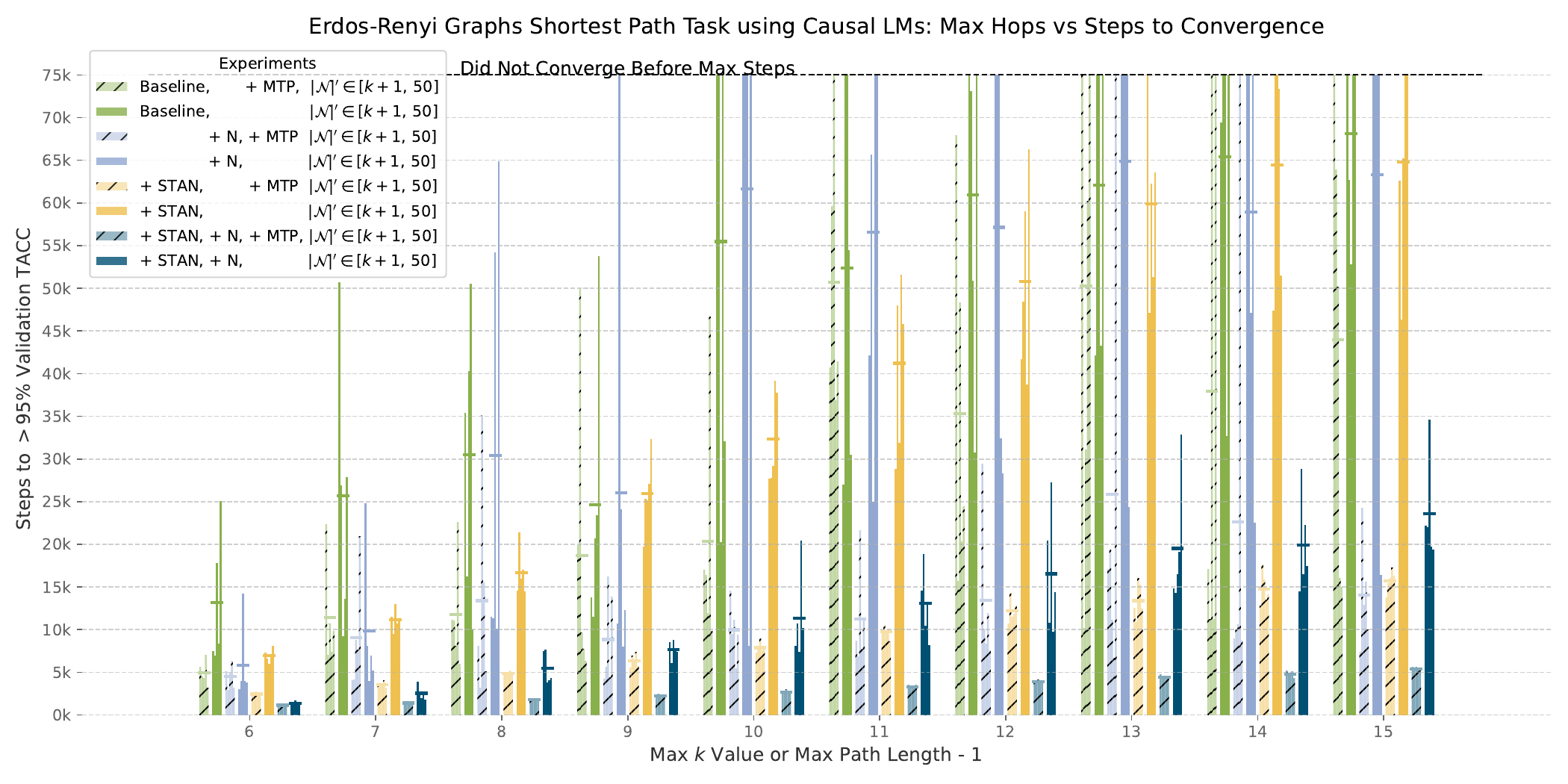}
    \caption{The baseline and ablation for the causal LM.}
    \label{fig:er-causal-lm-baseline}
\end{figure}

As in Fig.\ \ref{fig:er-baseline}, we consider the baseline and ablation of STAN and $N$ at $|\mathcal{N}| = 50$, in Fig.\ \ref{fig:er-causal-lm-baseline}.  Here we observe different patterns from the causal LM, both in the baseline and in the ablations.  First, unlike in the prefix LM, we see the expected pattern where task difficulty grows with $k$ and $|\mathcal{N}|$.  It seems like $N$ still leads to the greatest increase in performance at low $k$, but not at high $k$ compared to STAN.  We expect this is because, like the causal LMs, it is noisy, and the increased difficulty of using the causal LM adds to that noise.  Unlike the LM prefix, STAN consistently outperforms the baseline.  In many ways, these results are more expected than the results from the prefix LM baseline.  This exposes a clear difference in performance between the two models due to parameterization.

\subsection{Results on Euclidean Graphs}\label{appx:eg}

\begin{figure}
    \centering
    \includegraphics[scale=0.39]{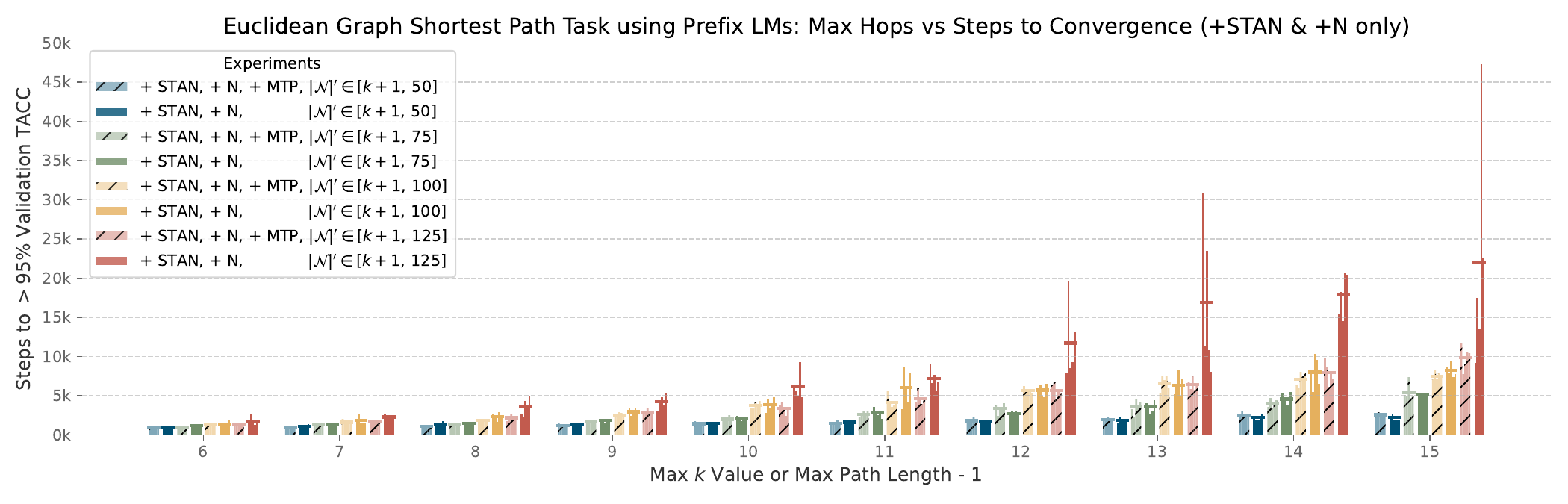}
    \caption{Results using Euclidean Graphs with Prefix LMs}
    \label{fig:eg_k}
\end{figure}

Fig.\ \ref{fig:eg_k} shows the results of the EGs with Prefix LMs.  As we can see, MTP only improves learning once we scale to 125 nodes.  We expect that this would be a trend had we experimented with larger graphs.  

\subsection{Restricted MMHT Path-Star Graph Experiments}\label{appx:restricte-psg}

\begin{figure}
    \centering
    \includegraphics[scale=0.39]{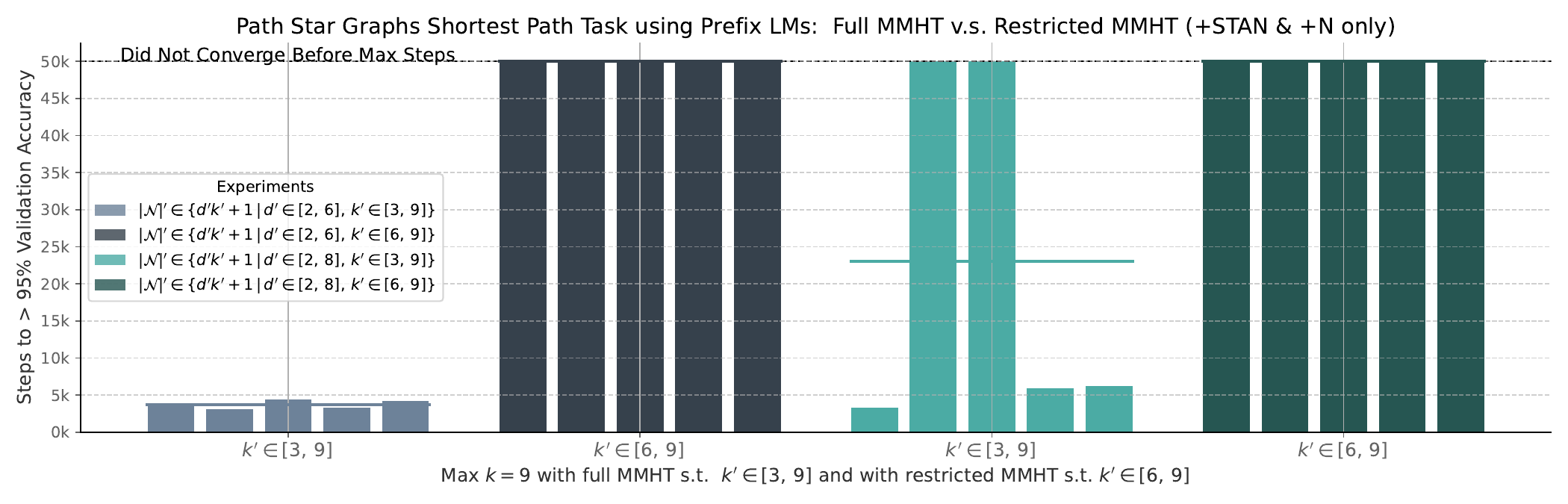}
    \caption{Restricting the mixture of MMHT to higher $k'$ fails to learn the task.}
    \label{fig:restricte-psg}
\end{figure}

Here we empirically validate our {\em lockout} hypothesis on PSGs.  We consider two experiments from Fig.\ \ref{fig:psg_k}: where $k=9,\, d=6$ and where $k=9,\, d=8$.  Both use full MMHT s.t.\ $k' \in [3,\, 9]$.  The learning dynamics of the $d=8$ experiments are plotted in Fig.\ \ref{fig:psg_step}.   We see all the $d=6$ trials solve the task, but only $3$ trials for $d=8$ solve the task and two fail.  Thus, we have one set in which all succeed and another in which some fail.
We then consider restricting the mixture of MMHT to higher $k' \in [6,\, 9]$ for both $d=6$ and $d=8$ in Fig.\ \ref{fig:restricte-psg}.  We see that all such models fail, which is behaviour consistent with PSGs when they are not provided with MMHT \citep{bachmann2024the, frydenlund-2025-language}.

\section{Auxiliary Multihop Tasks: Khops, Khops-Gen, and BFS Scratchpads}\label{appx:aux-tasks}

We consider three additional latent multihop reasoning tasks, Khops, Khops-Gen (a novel generative variant), and a BFS scratchpad (BFS-SP), for two reasons. First, they act as further testbeds to establish MTP's benefits in other symbolic multihop settings. Second, prior work has proposed scratchpads/chain-of-thought as a scalable way to sidestep the difficulty of latent multihop search \citep{saparov2025transformers}. We show this is not only impractical, since BFS-SP's sequence length grows too quickly to scale, but, more importantly, that it \textbf{changes the underlying computational problem}. A scratchpad that presents graph structure in traversal order collapses latent search into a much simpler linear-chain traversal task, one that is, in fact, computationally equivalent to Khops-Gen and Khops (\A\,\ref{appx:sp-collapse}). 

We define each task below; Khops (\A\,\ref{appx:khops}), Khops-Gen (\A\,\ref{appx:khops-gen}), and BFS-SP (\A\,\ref{appx:bfs-sp}), then return to this unifying argument in \A\,\ref{appx:sp-collapse}.

\subsection{Khops}\label{appx:khops}

Khops is an in-context latent multihop reasoning task introduced by \citet{sanford2024understanding}, extending {\em induction heads}, a well-studied attention mechanism that drives in-context learning by matching a symbol to its most recent prior occurrence and copying the symbol that followed it \citep{elhage2021mathematical}. Formally, let $x_1,\, \dots,\, x_N$ be a series of $N$ symbols in some vocabulary $\mathcal{V}$. An induction head implements $f(x_i) = x_{j+1}$, where $j$ is the most recent prior position with $x_j = x_i$. Khops extends this to $k$-fold composition.

{\em It is a sequential labelling task, not a sequential generation task}, given to causal LMs. For some $k$, the model must output the final induction symbol after $k$ hops at each of the $N$ positions where applicable (early positions may lack enough context to form $k$ hops, and hence output no symbol). The targets are $y = y_1,\, \dots,\, y_N$, where $y_i = f_k(\dots f_2(f_1(x_i)))$. $k$ is provided as part of the task description, similar to the query in shortest path search. Thus, each instance is really $N$ separate sub-tasks, each with a different, and often reduced, context.  The task is illustrated in Fig.\ \ref{fig:khops-panels}.

\citet{sanford2024understanding} show Khops can be expressed and learned by transformers with $\operatorname{O}(\log k)$ layers.\footnote{A similar result exists for shortest path search on path-star graphs \citep{frydenlund-2024-mystery}.  Both learn a solution which does pointer-doubling, hence only requiring a logarithmic number of layers, rather than the more intuitive linear number.} They also induce MMHT by sampling $k' \in [1,\, k]$, but do not evaluate its effect.  We perform such an experiment in \A \ref{appx:khops-results}, which shows that the training requires an exponential number of steps as $k$ grows without MMHT.  We also introduce a sequential-generation variant, Khops-Gen, needed to properly compare against MTP (\A \ref{appx:khops-gen}).

\newlength{\khopsw}
\setlength{\khopsw}{1.0\linewidth}   

\begin{figure}[]
    \centering
    \begin{subfigure}[b]{1.0\textwidth}
        \centering
        \includegraphics[width=\khopsw]{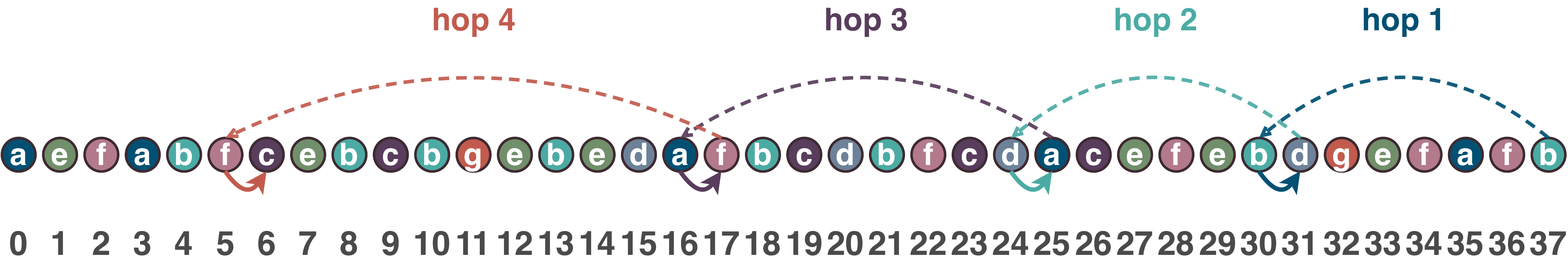}
        \caption[Khops: the task.]{The task at query position $37$ for $k'=4$ hops. Each hop matches the most recent occurrence of the current value (dashed), then reads one position to its right (solid). A {\em hop} is the composition of the two. Each hop searches strictly left of the {\em previous match} due to the causal constraint, so the window of potential matches, and thus the search space, shrinks monotonically.  {\em Note that the node IDs are actually their position i.e.\ we have nodes 0-37, and the letters are their values or semantic content.}}  
        \label{fig:khops-panelA}
    \end{subfigure}
    \begin{subfigure}[b]{1.0\textwidth}
        \centering
        \includegraphics[width=\khopsw]{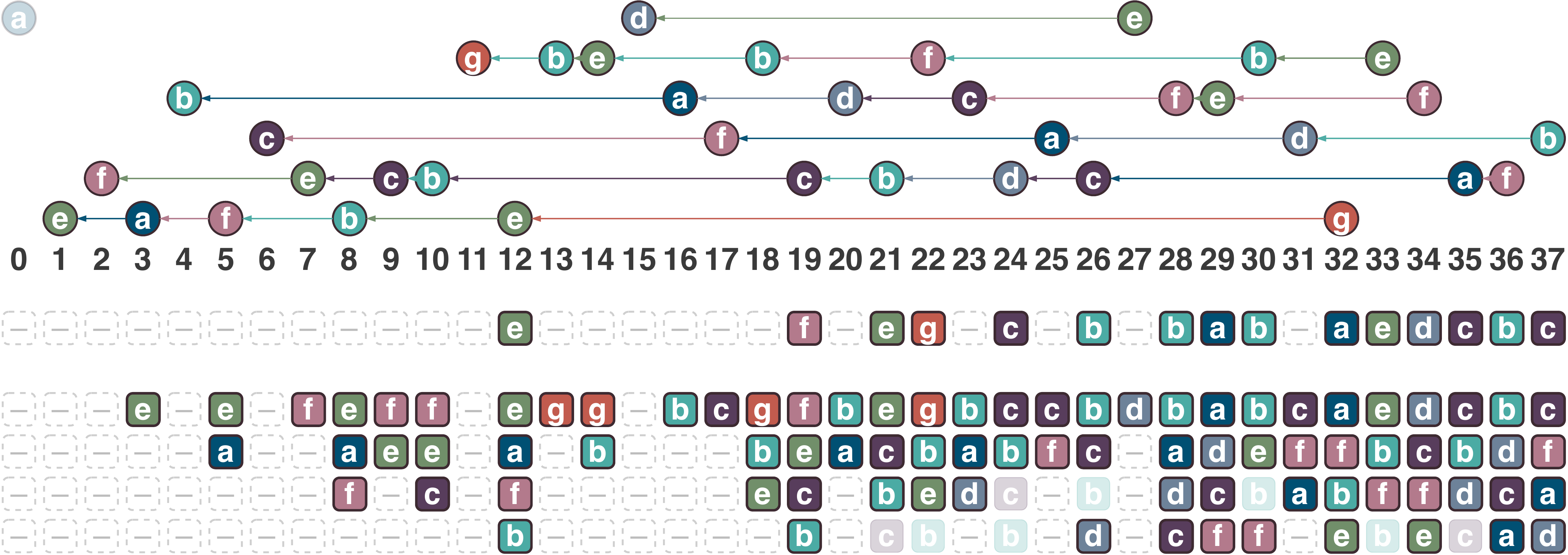}
        \caption[Khops: the reasoning graph.]{The latent reasoning graph induced by (a), which is a {\bf disjoint union of linear chains}.  The length of these chains defines the search space under the causal constraint.  
        Position $0$ can not connect and is ghosted. We show the single label that results after $k'=4$ hops.  These are masked out if there is no viable 4-hop solution.  Beneath this is the {\bf ranking labels} used in our MTP surrogate (\protect\A \ref{appx:ranking-sur}).  We ghost any repeated labels, which are masked out from the ranking ground-truth labels.}
        \label{fig:khops-panelB}
    \end{subfigure}
    \begin{subfigure}[b]{1.0\textwidth}
        \centering
        \includegraphics[width=\khopsw]{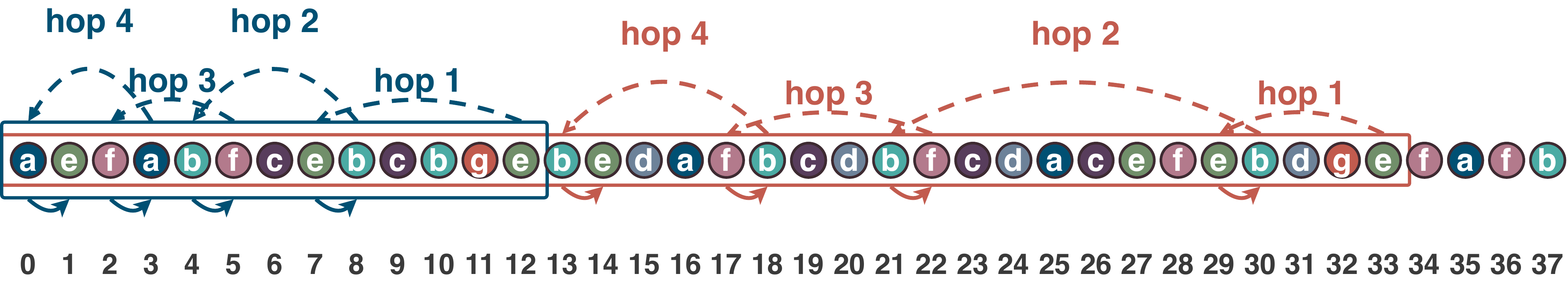}
        \caption[Khops: implicit MMLT.]{Two queries from the {\em same} instance, at positions $12$ (blue) and $33$ (red), both supporting a full $k=4$ chain.  This shows how predicting the task for multiple positions at once induces implicit MMLT, with some positions have a reduced search space relative to others (\protect\A \ref{appx:MMLT}).  Or in other words, a longer context contains far more {\em distractor node}, which have the same value as the sought after value that must not be matched.}
        \label{fig:khops-panelC}
    \end{subfigure}
    \caption[The Khops task.]{{\bf The Khops task}, $k=4$, one instance throughout. Colours and letters denote symbol {\em values}, the numbers denote {\em positions} i.e.\ node IDs.  Dashed arrows are the latent content-based match while solid is the deterministic $+1$ shift.}
    \label{fig:khops-panels}
\end{figure}

\subsubsection{Causal LMs and Mixed Multi-Length Training}\label{appx:MMLT}

Just as MMHT provides a mixture of hop-distances $k' \leq k$ within a single sample, we can define an analogous {\em mixed multi-length training} (MMLT), a mixture of task sizes $N' \leq N$. Where MMHT varies the difficulty of a fixed-size instance, MMLT varies the size of the instance itself. Both are hypothesized to aid learning by exposing the model to a range of task difficulties.

We find that Khops induces MMLT implicitly (in much the same way that multiple choice nodes in a path implicitly induce MMHT (\S\,\ref{sec::SSE})). Because each target $y_i$ only depends on symbols before position $i$, every position $i$ sees a different amount of preceding context. This creates an implicit mixture of task lengths, uniform across $1,\,\dots,\,N$ (ignoring invalid positions where $i$ is too small to support $k$ hops). This behaviour is both inherent to the task and a consequence of using a causal LM.  Note that the task is not defined for prefix LMs as it relies on causality to provide implicit left-right ordering.

Consider the state space of the task; position $i$ has $i$ tokens available before it, giving a state space of $i^k$, up to $N^k$ at the final position. Thus MMLT gives a mixture of examples spanning a wide range of difficulty for free, without any explicit sampling of $k'$ or $N'$.

Isolating the effect of this implicit MMLT is difficult on Khops itself, since every position is trained on simultaneously within a single instance. We instead add $\tilde{N}$ noise symbols to the front of the sequence, increasing the state or search space at position $i$ from $i^k$ to $(\tilde{N}+i)^k$. This has only a small effect on learnability. We find a much larger effect of MMLT on Khops-Gen (\A\,\ref{appx:khops-gen}), which, unlike Khops, is trained on only a single instance per example, so MMLT there must be induced explicitly.

\subsubsection{A Surrogate for MTP on Khops}\label{appx:ranking-sur}

MTP works by predicting several future tokens at once, exploiting the sequence structure of a generation task. Khops has no such structure to exploit directly because it is a labelling task with a single target per position. To meaningfully compare against MTP, we need a sequence-like structure standing in for ``future tokens.'' The intermediate compositional labels, $f_1,\, \dots,\, f_k$, at each position provide exactly this as they form an ordered sequence of increasingly-composed targets, even though only the final one, $f_k$, is the task's actual output.\footnote{This is also a nice example of how MTP can be framed as a multi-label task.} We therefore apply a ranking loss over these labels, preferring the final composition $f_k$ (the original single target) most, then $f_{k-1}$, and so on, mirroring how MTP ranks the immediate next token above later ones.

{\bf These labels are not new supervision.}  Since training already uses MMHT, each intermediate label $f_{k'}$ is already seen as a target whenever $k'$ is sampled. What changes is that we now supply all of them together, per instance, rather than one at a time across different samples. This mirrors how MTP increases supervision density per instance for NTP; however, this also converts batch-level inter-supervision to instance-level intra-supervision.
The gain therefore comes from the increase in per-instance supervision density, which is the mechanism we attribute to MTP throughout this work.  Separating the two is not possible, as the ranking loss is what makes the denser supervision expressible.

{\bf Importantly, we implement this with the same Plackett-Luce ranking loss used for MTP.}  The only difference is that our model does not need to output multiple labels; we only need to supervise them during training.  There is one issue, which is that we may have repeated symbols in the ground-truth composition.  We handle this by masking out lower-preference labels once a symbol repeats. Note that this is unnecessary for MTP on our other tasks, where all tokens are unique by task definition. Finally, unlike shortest path search, we do not add a new prediction head, and the loss is applied directly to the last layer's hidden states, as with softmax-CE.

\subsubsection{Khops Experiments and Results}\label{appx:khops-results}

We use models with $L=7$ layers and train for a maximum of 100k steps. We use $N=250$ and $\tilde{N} \in \{100,\, 200\}$. Unlike prior work, we use $|\mathcal{V}|=11$ rather than 4 \citep{pmlr-v235-sanford24a, saunshi2025reasoning}, chosen to better connect these results to the breadth-first search experiments (\A\,\ref{appx:bfs-sp}). To our knowledge, no prior work has considered the effect of $|\mathcal{V}|$ on this task. Note that a larger vocabulary makes it less likely to sample a sequence supporting a large $k$ without also increasing $N$.

\begin{figure}[h]
    \centering
    \includegraphics[scale=0.39]{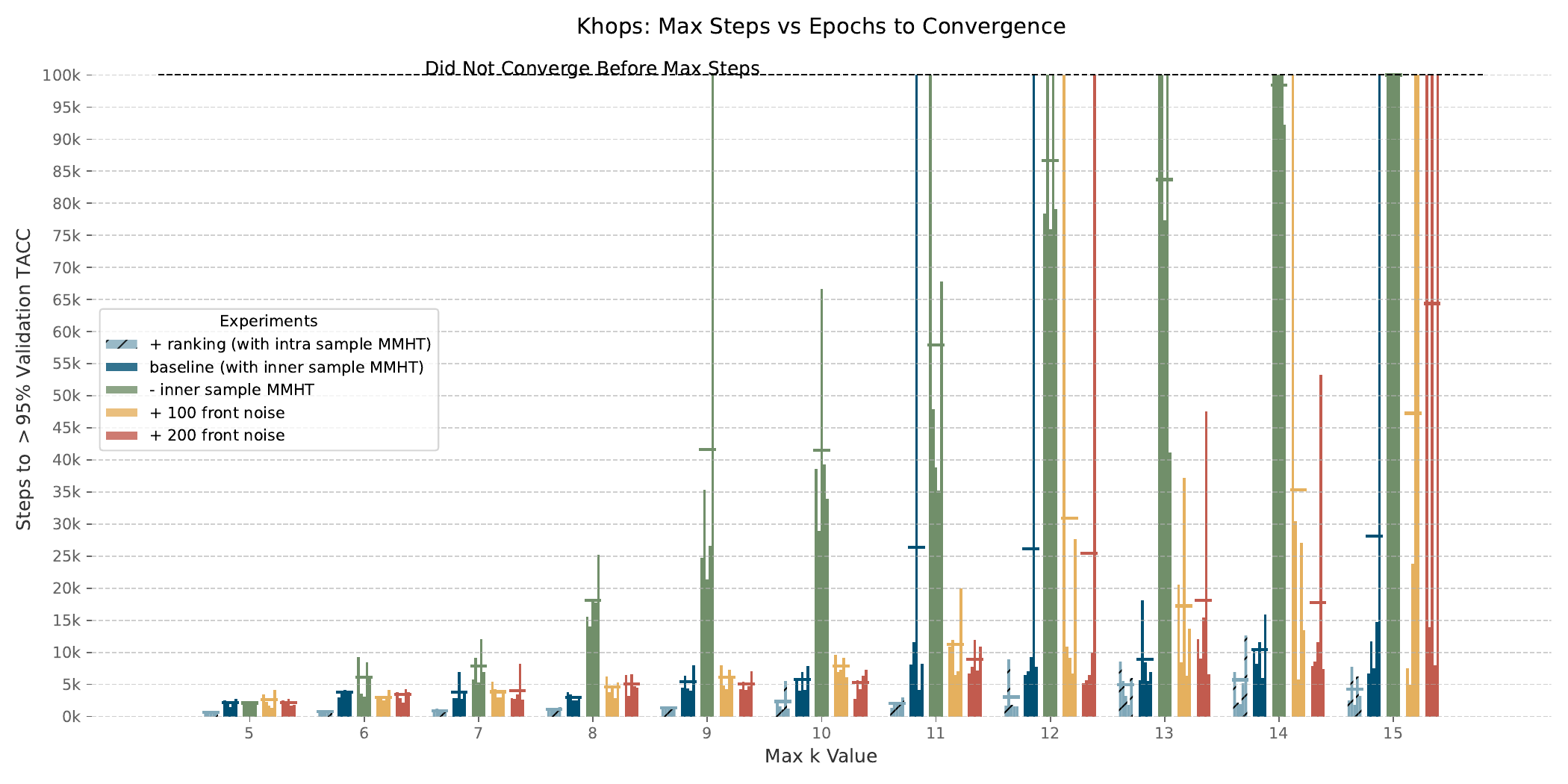}
    \caption{{\bf Khops experiments.} Prior work sampled $k' $, providing inner sample MMHT.  We show that this is necessary for learning.  Using multi-label ranking provides implicit intra sample MMHT.}
    \label{fig:khops}
\end{figure}

\begin{figure}[h]
    \centering
    \includegraphics[scale=0.39]{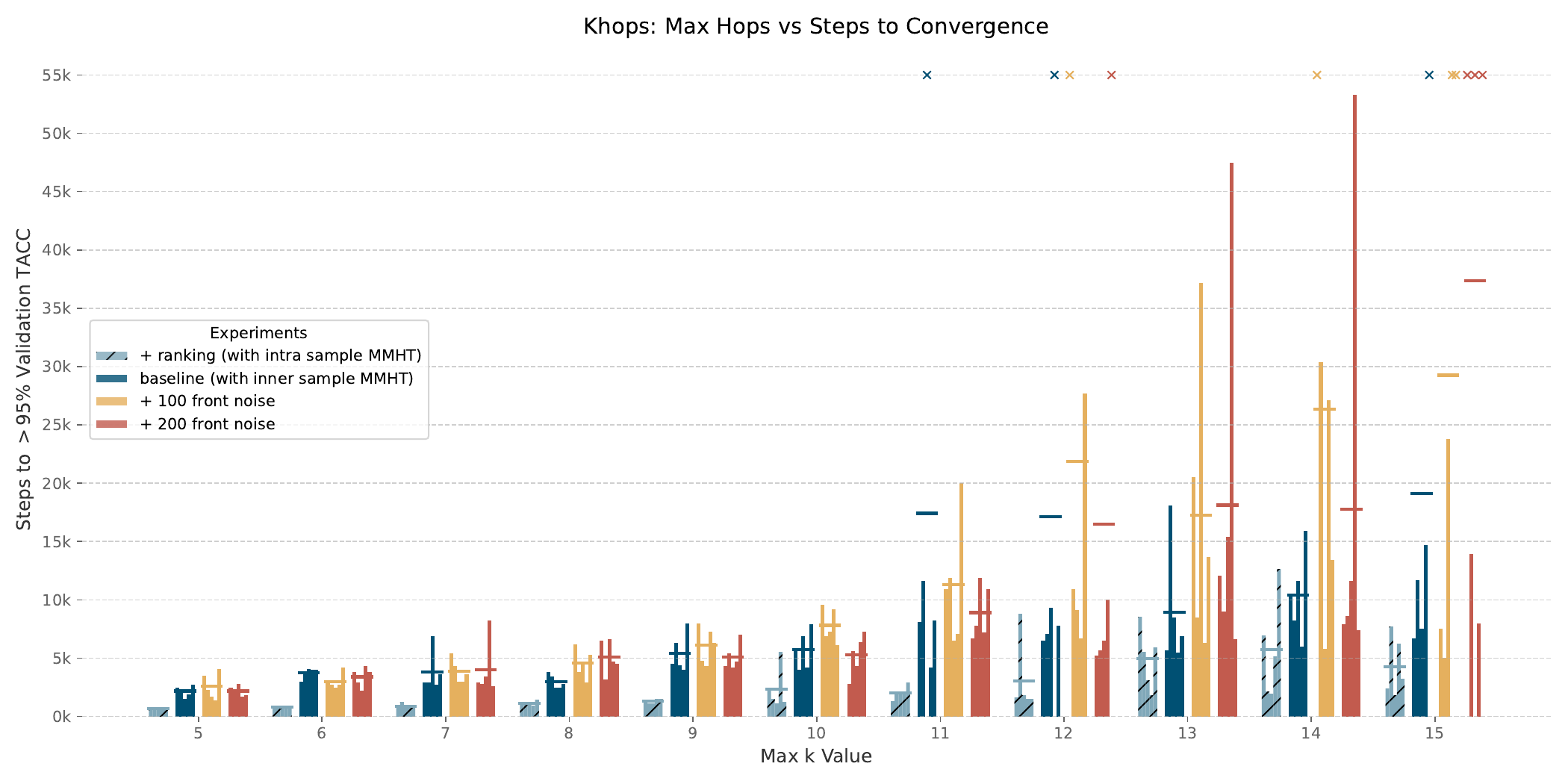}
    \caption{A simplified version of Fig.\ \ref{fig:khops} with the where we have removed the `- inner sample MMHT' experiment and any experiment which did not converge (`x's at top).  This better allows us to see the effect of the ranking loss vs. the baseline.}
    \label{fig:khops-simp}
\end{figure}

Our results are in Fig.\,\ref{fig:khops} (simplified in Fig.\,\ref{fig:khops-simp}). Without MMHT, the task scales very poorly, providing further empirical evidence that MMHT is critical for learning multihop tasks at scale \citep{sanford2024understanding, pmlr-v291-wang25a, frydenlund-2025-language, yao-etal-2025-language-models}. Using a ranking loss outperforms the baseline.  Although this is not strictly MTP, it captures two MTP-like benefits: increased supervision density, though not quadratic density, and implicit MMHT. Adding front noise makes learning more difficult, and this effect grows more pronounced as $k$ increases. This suggests that disrupting the implicit MMLT provided by causal Khops hurts learning, an effect we will see more clearly in Khops-Gen (\A\,\ref{appx:khops-gen}).

\subsection{Khops-Gen}\label{appx:khops-gen}

\begin{figure}[]
    \centering
    \includegraphics[width=\linewidth]{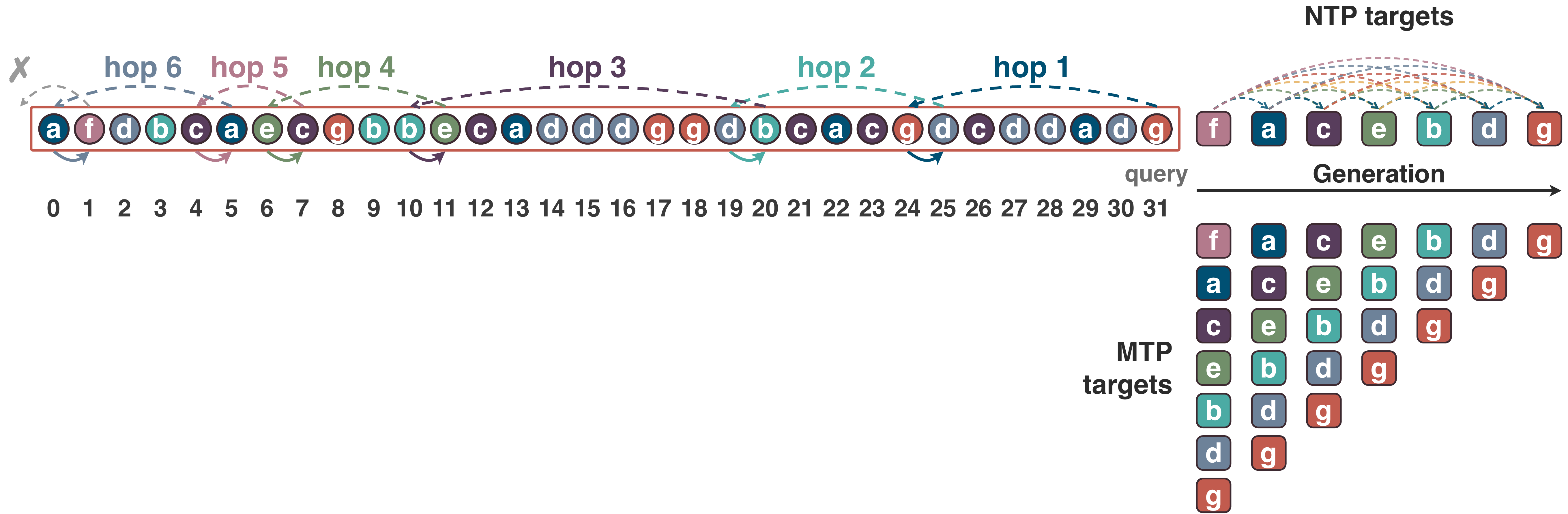}
    \caption[Khops-Gen.]{{\bf Khops-Gen.} Sequential generation of the full compositional chain from the single query $x_N$ (position 31). The source-side is on the left in red, with the target-side being generated.
    Unlike Khops, only one chain is considered in a single instance of the task.  Because of this, and as the source-side spans the entire $N$ nodes, there is no implicit MMLT training as in Khops.  Thus, MMLT must instead be induced explicitly by sampling $N' \leq N$ (\protect\A \ref{appx:MMLT}).  This also indicates that there is much less supervision in Khops-Gen as there is in Khops.  However, the supervision is different.  Whereas Khops has implicit MMLT but no implicit MMHT, Khops-Gen is the opposite. Unlike Khops, the chain runs until the context is exhausted and $k$ is not supplied to the model.  Note that we sample $k'$ uniformly, but that does not mean the context window between hops is the same size.  On the target side, we show both NTP and MTP.   Because this is a genuine generation task, MTP applies directly rather than through the ranking-loss stand-in required for Khops (\protect\A \ref{appx:ranking-sur}).  Importantly, MTP induces the same quadratic-to-$k$ supergraph (rainbow) as it does in path search (and will for any multihop reasoning task). The MTP target block is a triangle of $k(k+1)/2$ terms, matching Eq.\ \ref{eq:mtp}.}
    \label{fig:khops-gen-task}
\end{figure}

Khops is a sequential labelling task, whereas MTP is designed for sequential generation. Thus, to better compare against MTP, {\bf we introduce Khops-Gen, a generative variant of Khops.} Instead of predicting the final $k$-hop label at every position, Khops-Gen starts from a single symbol, $x_N$ (the last one), and generates the full compositional trace.  We illustrate the task in Fig.\ \ref{fig:khops-gen-task}.

The target sequence is ordered from the hardest prediction to the easiest:
$f_k,\, f_{k-1},\, \dots,\, f_1$. This is important. {\em If we reverse the generation order, the task becomes trivial under teacher forcing, since each previously generated target directly reveals the next one-hop predecessor.} Thus, the ordering is what preserves the need for multihop reasoning.  This is similar to the reverse `solution' for path-star graphs \citep{bachmann2024the}.

There are multiple changes from Khops.  Unlike Khops, $k$ is not provided explicitly. Instead, $k$ is determined by the available context where we construct $x$ so that exactly $k$ hops can be followed before the compositional chain runs out. This gives Khops-Gen an implicit MMHT structure, since generating $f_k,\, f_{k-1},\, \dots,\, f_1$ decomposes the task into progressively shorter-hop targets within the same sequence. We additionally sample $k'$ by uniformly sampling integer partitions of an $N$-length sequence \citep{Kenneth2013RandomIP}.

Unlike Khops, we construct sequences so that no symbol repeats within the generated chain. This (a) allows for a clean version of MTP, similar to the shortest path task, and (b) allows for a direct comparison to the BFS scratchpad task, which also has no repeats and where solving the shortest path from the scratchpad reduces to following a linear backtrack chain (\A\,\ref{appx:bfs-sp}). {\bf In this sense, Khops-Gen is the sequential-generation analogue of Khops and the core multihop subtask of BFS-SP once the graph has been linearized by the scratchpad} (\A\,\ref{appx:sp-collapse}).

Unlike Khops, Khops-Gen does not receive implicit MMLT via the task design. Since we generate from a single starting symbol, the entire $N$-length sequence is observed before generation begins, rather than producing labels at every position. Thus, we explicitly sample different $N' \leq N$ to induce MMLT for comparison with Khops.

\subsubsection{Khops-Gen Experiments and Results}
We use the same model and training settings as Khops.  We consider experiments at $N \in \{50,\, 100\}$.
The one exception is the vocabulary size.
In both Khops and Khops-Gen, positions act as the node identities, and the chain moves strictly leftward, so positions never repeat within a chain by construction (Fig.\,\ref{fig:khops-panels}).
The constraint we add for Khops-Gen applies to the emitted symbols rather than to the positions, and it requires at least as many symbols as there are links in the chain.
We therefore use $|\mathcal{V}| = 11$ for $k \leq 11$ and $|\mathcal{V}| = 16$ for $k > 11$, with special tokens in addition to these.
Note that $N$ is the number of positions and is independent of $|\mathcal{V}|$, as symbols repeat freely outside the chain.
Khops-Gen is much less supervision-efficient than Khops since Khops provides a label at every valid position in the sequence, while Khops-Gen generates only a single compositional chain from $x_N$. Thus, a single Khops instance provides many supervised subproblems, whereas a single Khops-Gen instance provides only one generated sequence. We therefore use smaller $N$ so the task can be learned efficiently.

\begin{figure}[h]
    \centering
    \includegraphics[scale=0.39]{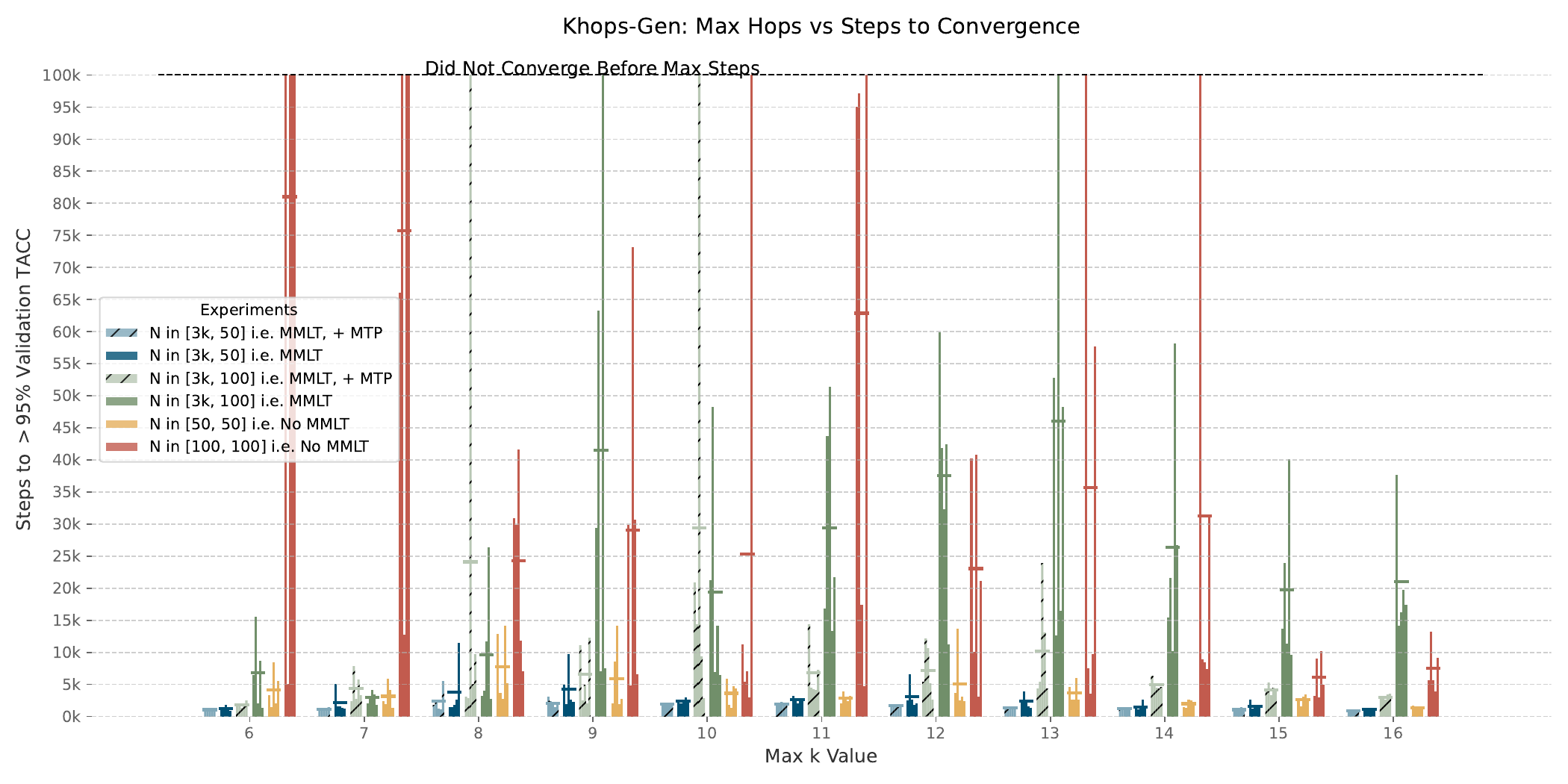}
    \caption{{\bf Khops-Gen experiments.} MTP improves learnability at scale.  Explicit MMLT is also necessary for efficiently learning the task.}
    \label{fig:khops-gen}
\end{figure}

\begin{figure}[h]
    \centering
    \includegraphics[scale=0.39]{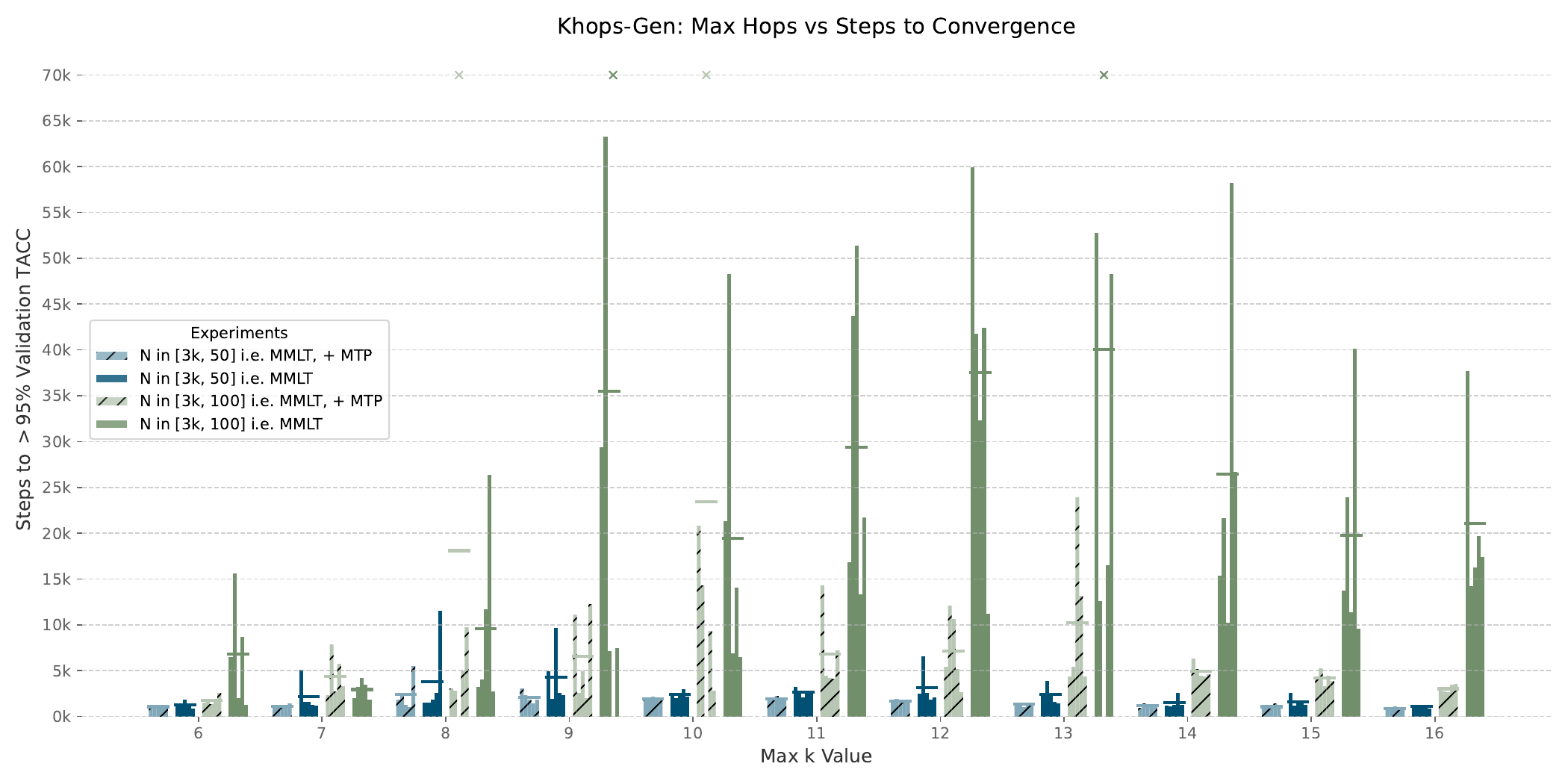}
    \caption{A simplified version of Fig.\ \ref{fig:khops-gen}}
    \label{fig:khops-gen-simp}
\end{figure}

Our results are in Fig.\,\ref{fig:khops-gen}, with a simplified version in Fig.\,\ref{fig:khops-gen-simp}. As expected, the task becomes harder as $N$ and $k$ increase. We also find that MTP significantly improves learning at scale.

MMLT is also important for Khops-Gen. Without explicit MMLT, the task scales poorly, but with a somewhat surprising pattern: lower-$k$ experiments fail while larger-$k$ experiments do not. We believe this occurs because the model can learn that each target lies to the left of the previously generated target, thereby reducing the effective state space. This resembles the implicit MMLT effect in Khops, but is much weaker since Khops has supervision at every position, so it directly exploits the causal LM's left-context inductive bias across many effective sequence lengths within each instance. Khops-Gen has the same causal inductive bias, but far less supervision per instance, so this structure must be learned from a much weaker signal.

Given this learned behaviour, larger $k$ divides the $N$-length context into fewer and shorter viable regions, reducing the state space for each prediction. This may explain why the baseline models are relatively uniform across $k$. It may also be an artifact of using $|\mathcal{N}|$ close to $k$, especially at $|\mathcal{N}| \leq 50$ and $k=15$, where teacher forcing reveals many prior path tokens and removes them from the candidate state space.

These results also highlight a broader point: difficulty, and therefore what counts as a `curriculum', can be multifaceted \citep{soviany2022curriculum}, with surface proxies not necessarily reflecting the relevant reasoning difficulty \citep{mordig2026rethinking}. In our setting, hop-depth and task length are distinct axes of difficulty. MMHT varies the number of hops, $k' \leq k$, while MMLT varies the task length, $N' \leq N$. To our knowledge, prior work has not considered the effect of length mixtures for Khops. This effect is difficult to isolate in the original task, since MMLT is implicitly induced by causal supervision at every position. Khops-Gen provides a cleaner setting for studying this effect, since it removes this implicit length mixture and requires MMLT to be induced explicitly. The poor scaling without MMLT in Fig.\,\ref{fig:khops-gen} suggests that length mixtures can be as important as hop mixtures for learning sequential multihop generation tasks.

We sample $N'$ uniformly, but this may not be optimal. From a state-space perspective, smaller $N'$ produces much easier instances, since the model must search over fewer possible positions at each compositional step. Thus, oversampling smaller $N'$ may help the model first learn the underlying leftward composition rule in a restricted state space before applying it to larger contexts. This is analogous to the power-law sampling effect of \citet{wang2026the}, where non-uniform sampling improves learning by restricting the effective sample space early in training. We do not test non-uniform length sampling here, but view it as a promising direction for future work.

For the shortest path task, we sample $\mathcal{N}$, which provides MMLT in this context.

\subsection{BFS Scratchpad}\label{appx:bfs-sp}

The shortest path task should, in principle, be solvable via scratchpads (SP) \citep{nye2022show}. \citet{saparov2025transformers}, who considered general path search (connectivity) rather than shortest path search, used a depth-first search and a `selection-inference' scratchpad.  As noted they found that these did not overcome the scaling issues in solving the task. Since we are concerned with shortest path search, we use a breadth-first search scratchpad (BFS-SP).

An SP breaks the task into two computational phases. First, the model generates the scratchpad. Second, given the scratchpad, the model solves the original task. During training, the scratchpad is provided as supervision, but it must be generated by the model during inference. Importantly, the scratchpad constitutes a task-specific inductive bias: it hand-supplies the model with an intermediate structure (here, the BFS ordering) rather than requiring the model to discover that structure on its own. As we will discuss in \A\,\ref{appx:sp-collapse}, supplying this structure can change the underlying computation required to solve the task.

BFS-SP therefore requires two types of generation: 1) generating the BFS topologically-ordered scratchpad, then 2) using it to generate the shortest path. The latter is done by tracing the path back from $t$ to $s$ through the scratchpad, which is (nearly) computationally equivalent to Khops-Gen.

We initially tokenize the SP as follows:

\begin{itemize}
    \item Include the special token $D1$.
    \item Tokenize $s$, followed by its adjacency list, $[u \in \mathcal{N} \mid A(s,\, u) = 1]$.
    \item For each subsequent depth, include a special token $D k$, followed by each unexplored node from the prior depth's adjacency lists, each followed in turn by its own adjacency list.
\end{itemize}

An example tokenization of this is for $\pi= 7,\, 20,\, 5,\, 1,\, 14,\, 12,\, 23,\, 10,\, 2,\, 11,\, 17,\,9$, where $s=7$ and $t=9$ is

$D1$ 7 [     24    20    ] $D2$    24    [     16    ]           20    [     5     ]
         $D3$   16    [     ]                 5     [     1     15    ]
         $D4$   1     [     14    ]           15    [     6     ]
         $D5$   14    [     12    ]           6     [     18    ]
         $D6$    12    [     23    ]           18    [     3     ]
        $D7$   23    [     10    ]           3     [     ]
         $D8$   10    [     2     ]
         $D9$   2     [     11    ]
         $D10$   11    [     17    8     ]
         $D11$  17    [     9     22    ]     8     [     ]

where we keep the adjacency list special tokens `[' and `]' even if it is empty.  One can see that backtracking the SP to generate the shortest path is similar to Khops-Gen. The difference is that the connecting symbol lies to the left of the inductive hop, at the head of the adjacency list, rather than adjacent to the symbol itself.

\subsubsection{Tokenization Matters for Scratchpads Too!}

We found that a number of seemingly minor tokenization decisions had a large effect on learnability.

First, we initially used a distinct depth marker for each depth, i.e.\ $D1,\, D2,\, \dots,\, Dk$. This caused the models to fail to learn the task. Using a single shared depth marker $D$ instead resolved this.

Second, when generating the SP, the model often failed to produce the correct adjacency lists. We found this can be addressed in two ways: repeating the adjacency list in reverse order, or including the BFS queue as part of the tokenization. Both prevented repetition mistakes, though at the cost of substantially increasing the length of the SP.

Third, the BFS itself is unordered with respect to the order of adjacency lists, and thus the order in which nodes are explored is not uniquely determined. We instead require the model to explore nodes in the order they appear in the edge list, matching the order used for the node list. This means the model must learn an in-context ordering of nodes based on their position in the edge list, and it makes the SP itself deterministic.  By forcing an order, we avoid the need of considering label smoothing over the permutations \citep[][\A \ref{appx:label-smoothing}]{frydenlund-2025-language}.

Thus, our final tokenization is 

 $D$     $\{$ 7 $\}$       7     [     24    20    ]
$D$     $\{$ 24 20 $\}$  24     [     16    ]        $\{$ 24 $\}$     20    [     5     ]
$D$     $\{$ 16 5 $\}$   16     [     ]              $\{$ 5 $\}$      5     [     1     15    ]
$D$    $\{$ 1 15 $\}$    1     [     14    ]        $\{$ 1 $\}$      15    [     6     ]
$D$    $\{$ 14 6 $\}$   14     [     12    ]        $\{$ 6 $\}$      6     [     18    ]
$D$     $\{$ 12 18 $\}$  12     [     23    ]        $\{$ 18 $\}$     18    [     3     ]
$D$ $\{$ 23 3 $\}$   23     [     10    ]        $\{$ 3 $\}$      3     [     ]
 $D$   $\{$ 10 $\}$     10    [     2     ]
 $D$    $\{$ 2 $\}$      2     [     11    ]
  $D$  $\{$ 11 $\}$     11    [     17    8     ]
  $D$  $\{$ 17 9 $\}$   17    [     9     22    ]   $\{$ 9 $\}$      9     [     ]

where special tokens `$\{$' and `$\}$' are the BFS queue special tokens.

\subsubsection{BFS-SP Experiments and Results}

\begin{figure}[h]
    \centering
    \includegraphics[scale=0.39]{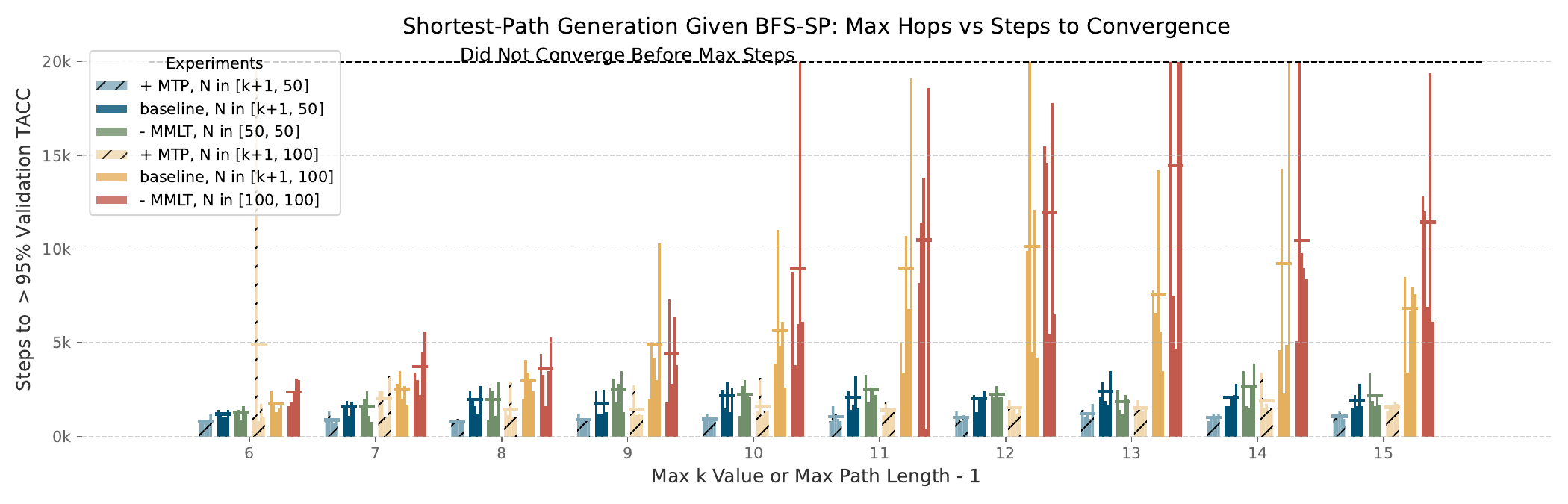}
    \caption{{\bf BFS-SP SP-only results.}  Given the SP, generate the shortest path, i.e.\ the partial task.  This task is (nearly) computationally  equivalent to Khops-Gen.}
    \label{fig:bfs-sp-sp-gen}
\end{figure}

For the partial task, we gave the model only the SP, omitting the graph description $[E, N]$ and the BFS queue, to reduce token count and let us scale to larger graphs. This lets us consider $|\mathcal{N}| \in \{50,\, 100\}$. The results are presented in Fig.\ \ref{fig:bfs-sp-sp-gen}. MTP makes the task much easier at scale. Unlike Khops-Gen, we do not see smaller $k$ being harder to learn, nor is omitting MMLT catastrophic to learnability. Thus, although the two tasks share the same underlying expression, i.e.\ backtracking over a linear chain, their learnability differs in practice.

\begin{figure}[h]
    \centering
    \includegraphics[scale=0.39]{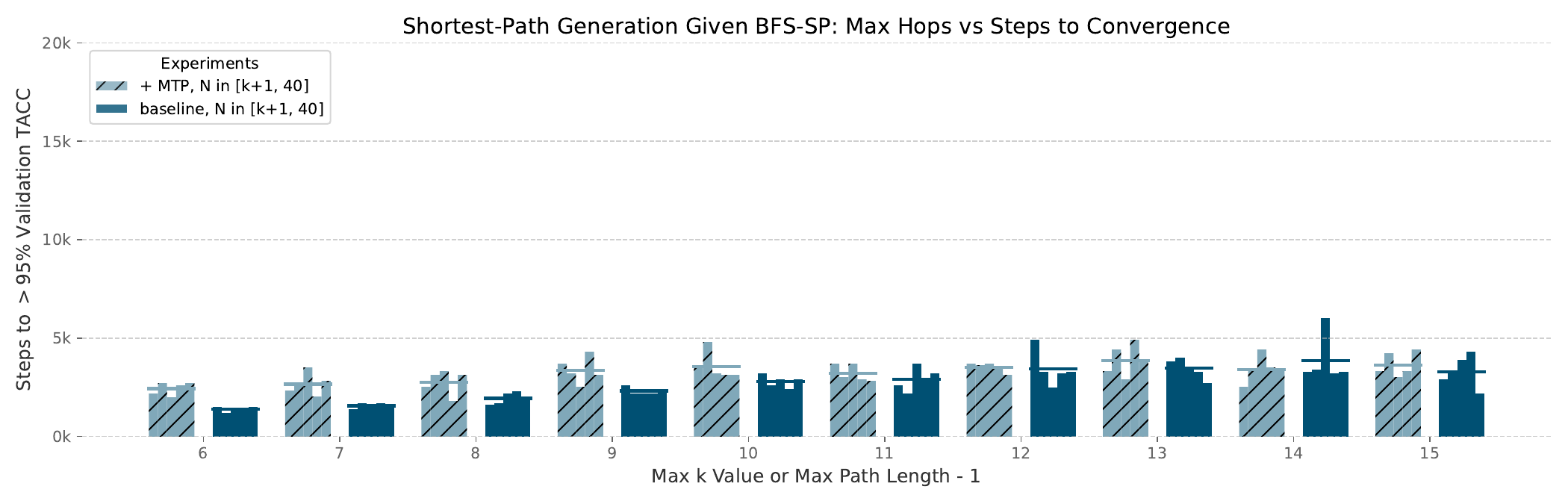}
    \caption{{\bf BFS-SP results.}  This is both the generation of the SP and the generation of the shortest path, i.e.\ the full SP task.}
    \label{fig:bfs-sp}
\end{figure}

For the full task, we only consider $|\mathcal{N}|=40$; we attempted $75$ and $50$, but training took too long to be practical. Given that directly learning the shortest path task is already scalable (\S\,\ref{sec:experiments}), we see little motivation to push further on scaling the SP experiments, since we already know they are impractical due to the increase in token length. The results are presented in Fig.\ \ref{fig:bfs-sp}, where the full task, generating the SP and then the shortest path, is learnable. This takes nearly twice as many steps as the partial task, indicating that learning to generate the SP is harder than solving the shortest path given the SP. Here, MTP does not outperform NTP, since SP generation does not itself require multihop reasoning.

Taken together, these results show that MTP still provides a clear benefit on the partial multihop task even though BFS-SP reduces the latent search problem to a simple chain traversal (\A\,\ref{appx:sp-collapse}). This is notable, since MTP's benefits are not restricted to cases involving non-trivial latent topology. 

\subsection{Scratchpads Substantially Change the Shortest Path Task and Break the Theoretical Multihop Reasoning Model}\label{appx:sp-collapse}

\begin{figure}[]
    \centering
    \begin{subfigure}[b]{0.40\linewidth}
        \centering
        \includegraphics[width=\linewidth]{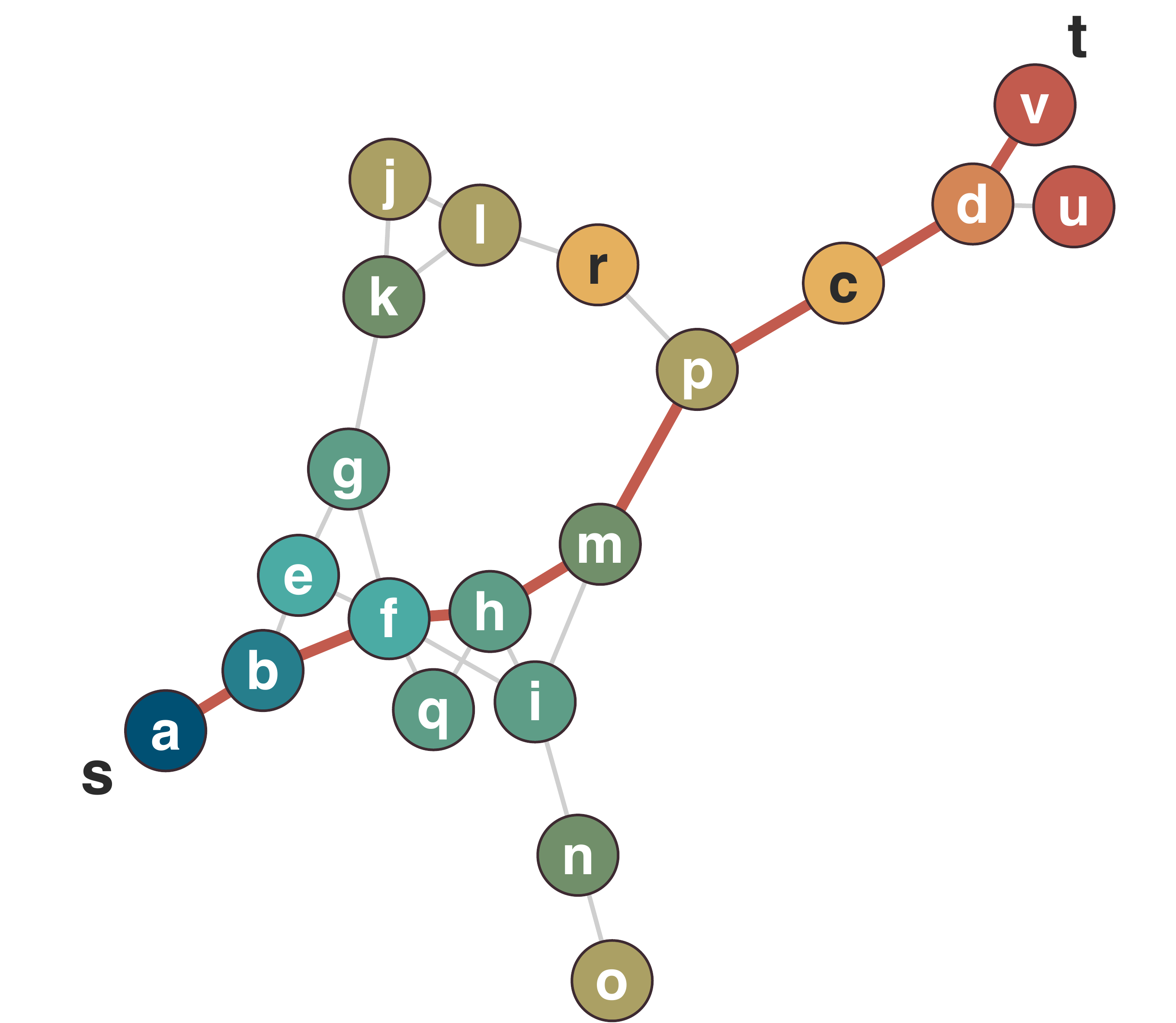}
        \caption[BFS-SP: the graph as given.]{A sample graph with a sampled path from $s$ to $t$ (red).  The graph has $20$ nodes and  $26$ edges, and is used in (b) and (c). Node colour is BFS depth from $s$.}
        \label{fig:bfs-sp-graph-a}
    \end{subfigure}
    \hfill
    \begin{subfigure}[b]{0.575\linewidth}
        \centering
        \includegraphics[width=\linewidth]{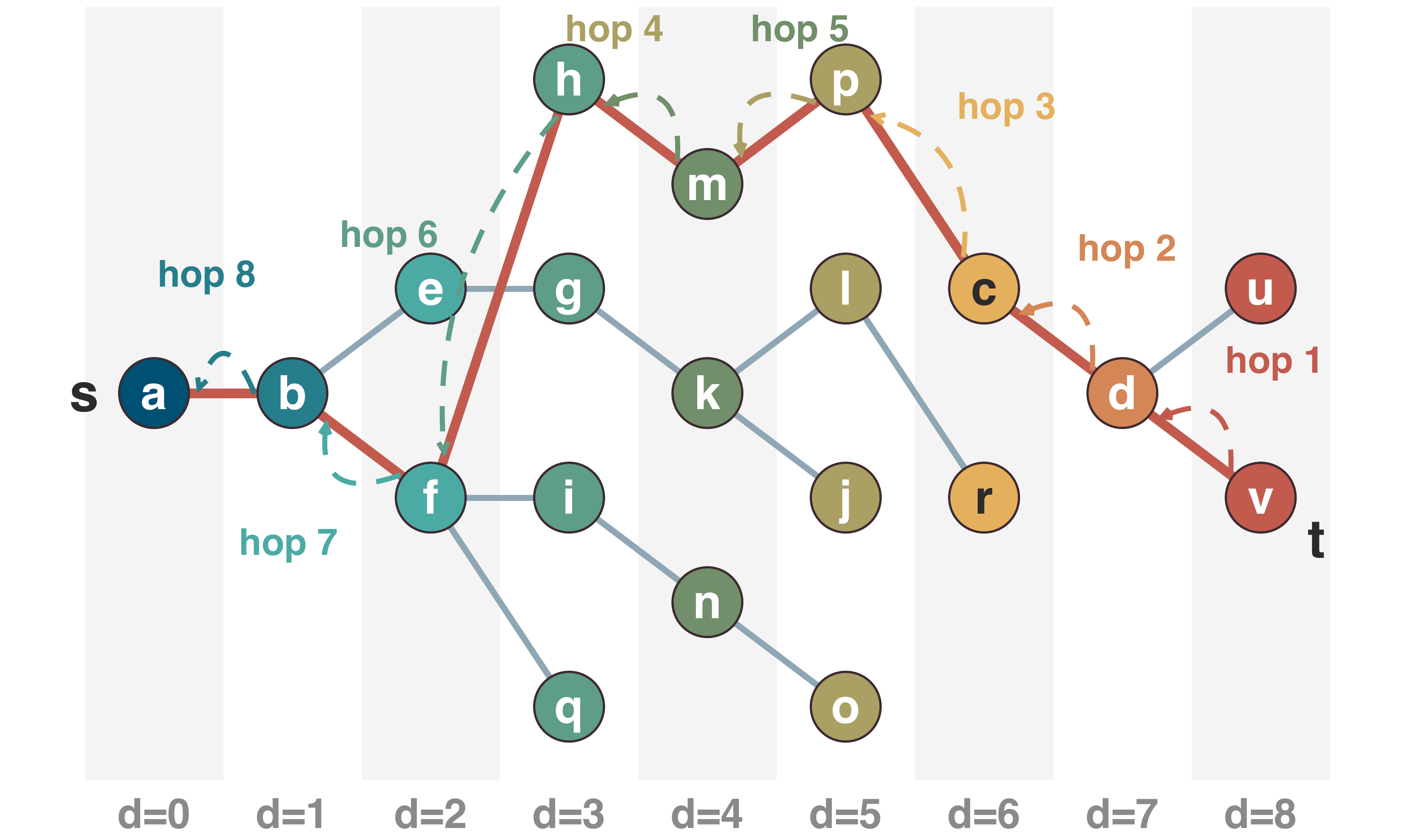}
        \caption[BFS-SP: after BFS layering.]{The same graph after the BFS ordering inductive bias is applied.  Note the dropped edges.  Ordering the graph only requires looking at neighbours of already-discovered nodes, and thus requires no multihop reasoning itself.  Tracing back from $t$ to $s$ does require multihop reasoning over a linear reasoning graph.  Thus, we break the theoretical model in two ways: we force a level topology on the graph turning it into a search tree, and then our actual latent reasoning graph reduces to a linear chain. }
        \label{fig:bfs-sp-graph-b}
    \end{subfigure}
    \\[6pt]
    \begin{subfigure}[b]{1.0\textwidth}
        \centering
        \includegraphics[width=1.\linewidth]{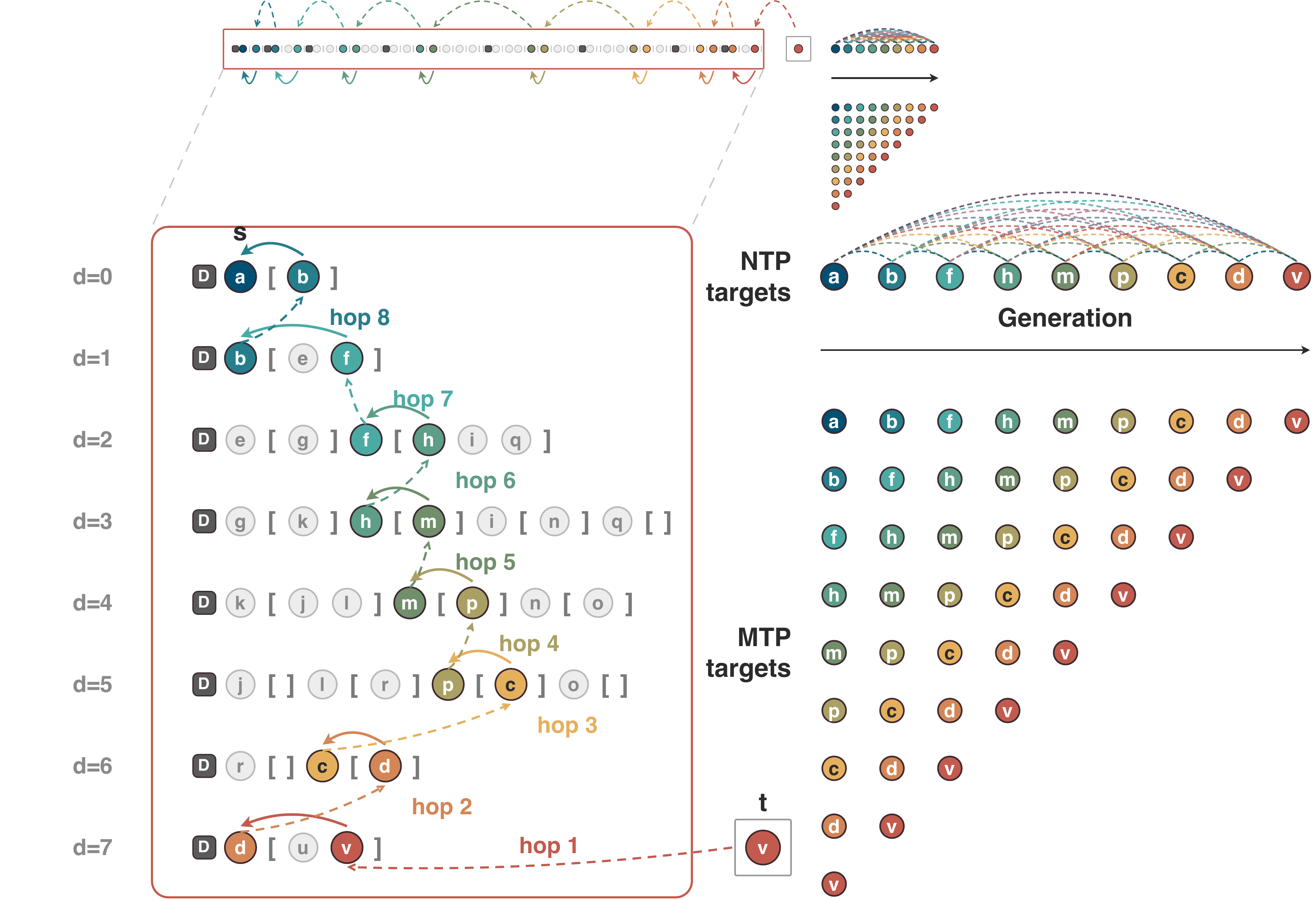}
        \caption[BFS-SP: the token sequence.]{The same sample as (a)--(b), now as the BFS-SP token sequence. {\bf Top:} the task mirrors Khops-Gen in Fig.\ \ref{fig:khops-gen-task}.  The only major difference is instead of adjacency-right connection, the connecting token is to the left of the start of the adjacency list, `['.   {\bf Bottom:} the same sequence magnified.  Here we see the backtrace reduces the latent reasoning graph to a linear chain.  We also show how MTP, once again, induces a quadratic supergraph (rainbow).}
        \label{fig:bfs-sp-tokens}
    \end{subfigure}
    \caption[BFS ordering changes the task topology.]{{\bf The BFS ordering inductive bias changes the topology of the task and reduces the latent reasoning graph to a linear chain.}}
    \label{fig:bfs-sp-fig}
\end{figure}

\textbf{BFS-SP reduces search to a linear traversal.} Recall that our theoretical model of multihop reasoning (\S\,\ref{sec:intro}) is deliberately agnostic to topology: it makes no assumption about the structure of the latent reasoning graph. The BFS-SP breaks this assumption in two ways.  First, it modifies the underlying topology to a search tree, starting at $s$.  Second, the latent reasoning graph reduces further to a linear chain, backtracking along this tree.  This backtracking is the structure of Khops-Gen (\A\,\ref{appx:khops-gen}). Thus, the general multihop search problem of \S\,\ref{sec:intro}, which requires resolving a choice node at each step among potentially many candidates in $\mathcal{E}$, is replaced by a traversal problem on a fixed topology where the resulting backtrack is still multihop, but it no longer involves any search. One may view this as the scratchpad not solving the original problem but rather substituting it with a different, strictly simpler problem in its place, one that is no longer a planning problem at all. {\bf Thus using a BFS-SP breaks the underlying theoretical model.}  This is illustrated in Fig.\ \ref{fig:bfs-sp-fig}.  

BFS-SP is still latent multihop reasoning/traversal.\footnote{\citet{frydenlund-2025-language} developed scratchpads that reduce path-star graphs to linear chains, and thus required multihop reasoning but not searching or planning.}\footnote{It is theoretically possible, though completely impractical, to create a BFS-SP which requires no multihop reasoning.  The key thing here is to repeat the full current branch/path at each depth as part of the tokenization and mark these with special tokens.  Then once $t$ is found, the model just needs to skip directly back to the start of this path and output it directly.  Alternatively, one can simply do two scratchpads, the BFS-SP, then a reverse SP.} Even though the backtrack removes any branching choice, the total computation performed across generating the SP and then the backtrack is still equivalent to $k$ hops of graph reasoning.  To output the correct node following $s$, the model must work back from $t$.  However, if we now reverse the output in generation, the task requires no multihop reasoning as each step can be produced with a single hop.  Thus, such a reverse `solution' works because it replaces latent computation with sequential computation.  Note that because backtracking is still a multihop task, this still allows the BFS-SP to act as a testbed for evaluating MTP.

{\bf Neither collapse in the original topology requires a shortest path scratchpad.} We could also do this by simply ordering the edge list $E$ by BFS distance from $s$ rather than shuffling it \citep{ge-etal-2025-graph}. A BFS-ordered edge list directly encodes, in its position, the same predecessor structure that the SP makes explicit. This is why we must shuffle $E$ before tokenization (\S\,\ref{sec:task}). These understandings of the task show that a correct output is not sufficient evidence of multihop reasoning; the task must also be solved for the right reason.

{\bf Not all tokenization changes break the model: STAN and $+N$ do not.} It would be natural to read \S\,\ref{sec:causal-constraint} and \S\,\ref{sec:STAN} as proposing the same kind of intervention as the BFS-SP scratchpad, since all three change the tokenization. They do not. The
theoretical model of \S\,\ref{sec:intro} forbids us from making any assumption about the topology of the latent reasoning graph.  We are therefore free in our choice of surface form insofar as that choice leaves the graph intact. STAN only modifies the representation of a single edge, and edges are still provided independently of one another. The node list is likewise trivially inferred from the edge list.  Neither is constructed as a function of the query; any query-conditioned structure must still be learnt by the model, whereas a BFS-SP is built from $s$ and $t$ and so supplies it directly in the input.  Note there are other tokenizations that break the theoretical model, such as those that use graph-global information, such as the graph Laplacian. \citep{bechler2026lost}.

\section{The $k$-Lookahead Issue}\label{appx:klookahead}

\citet{saparov2025transformers} note that, for pathfinding tasks, the characteristic difficulty measure should be the \emph{$k$-lookahead} -- the number of steps the model must search ahead from the start node to find the goal -- \textbf{not} the path length $|\pi| = k+1$. These are not equivalent: the model can shortcut prediction of $\pi$ once all competing arms are exhausted. For example, in a path-star graph with $\pi$ of length $8$ and all other arms of length $5$, once $4$ hops are taken, the model can determine $\pi$ is the only remaining viable path by elimination, without any further search.

A related issue arises when $s$ is not a choice node: the true number of required hops is then smaller than $k$. If $u_{i-1}$ is the first choice node in $\pi$, only $k' = k-i+1 < k$ hops are actually required to solve the task (\S \ref{sec::SSE}).

Both issues are easy to control for in \citet{saparov2025transformers}'s setup, since they consider only a restricted variant of path-star graphs: $s$ is a choice node by construction, and at least one non-$\pi$ arm can be constructed to be at least as long as $\pi$. This is much harder for general graph topologies, where we cannot freely construct $\mathcal{G}$ and then force the sampled path to have these properties. Consequently, a sampled path of length $k+1$ may not actually require $k$ hops to solve. This is a further illustration of why formally defining learnability for this task is difficult in the general setting.

\section{Multiple Shortest Paths and Semantic Label Smoothing}\label{appx:label-smoothing}

As mentioned above, multiple valid shortest paths can exist for a query $Q=(s,\,t)$ beyond the sampled ground-truth path $\pi$. During inference, this is not a problem: we validate a generated path against $\mathcal{G}$ directly rather than against $\pi$.\footnote{This requires checking against the distance matrix of $\mathcal{G}$.} During training, however, it is a problem. Under NTP, each token of $\pi$ is treated as the sole ground truth at that step, and CE tries to match this one-hot distribution. When multiple shortest paths exist, multiple targets are simultaneously valid at a given step, so the correct ground-truth distribution should be uniform over them. In expectation, this would not matter, since the model would still converge to the correct uniform distribution. In practice, though, we sample only a small number of graphs and paths.  As such, any alternative valid paths are rarely observed. CE then pushes the model to place all mass on the single sampled path even when others are equally valid, producing a loss spike whenever a well-trained model instead (correctly) spreads mass across them. This matters because, for much of training, we are in a low-loss regime where only a handful of tokens carry any signal, and these spikes are large enough to disrupt training dynamics.

Fortunately, the fix is simple.\footnote{This requires precomputing each graph's distance matrix to construct the correct ground truths, so we need the distance matrix at both training and inference time.} We treat NTP as a multi-label task (in the semantic dimension; see \A \ref{appx:mtp}): at each split point where multiple valid paths diverge, every node beginning a valid continuation is treated as a valid target, and we train with multi-label CE. See Fig.~\ref{fig:label-smoothing} for an example.

Then, because we use teacher-forcing, which conditions the sequence on the ground-truth path \citep{williams1989learning}, this will resolve multiple valid paths at each step without requiring a structured sequential distribution.  This means we do not need to model the joint distribution directly and can still factorize the joint distribution across time steps as in standard language modelling.

We do not apply this correction to the MTP loss.
Extending our rank-based MTP to multiple valid paths would require modelling ties correctly \citep{Frydenlund_Singh_Rudzicz_2022}.
While possible, it requires engineering overhead and, more importantly, we found this unnecessary in practice, given MTP's improved training dynamics compared to NTP.
The choice node analysis of \S \ref{sec::SSE} explains why the correction matters more for NTP.
A step with more than one valid continuation has more than one continuation, and is therefore a choice node.
The converse does not hold, as most choice nodes have a unique valid target, but it follows that conflicted targets can only land on the $c(\pi)$ positions that carry signal under NTP, and never on positions absorbed by the CHC.
MTP has $k(k+1)/2 - 1$ terms, and these are not restricted to choice nodes, since a skip of $j - i > 0$ requires multihop reasoning at non-choice nodes as well.
The same conflicted target is therefore a smaller fraction of the MTP loss and the spike is diluted.
Note that this correction improves NTP and is therefore applied in favour of the baseline, so it cannot account for the gains we attribute to MTP.
Evaluation is unaffected, as generated paths are validated against $\mathcal{G}$ for both losses.

\newlength{\lswidth}
\setlength{\lswidth}{0.8\linewidth}

\begin{figure}[h]
  \centering
  \begin{subfigure}[b]{1.0\textwidth}
  \centering
    \includegraphics[width=\lswidth]{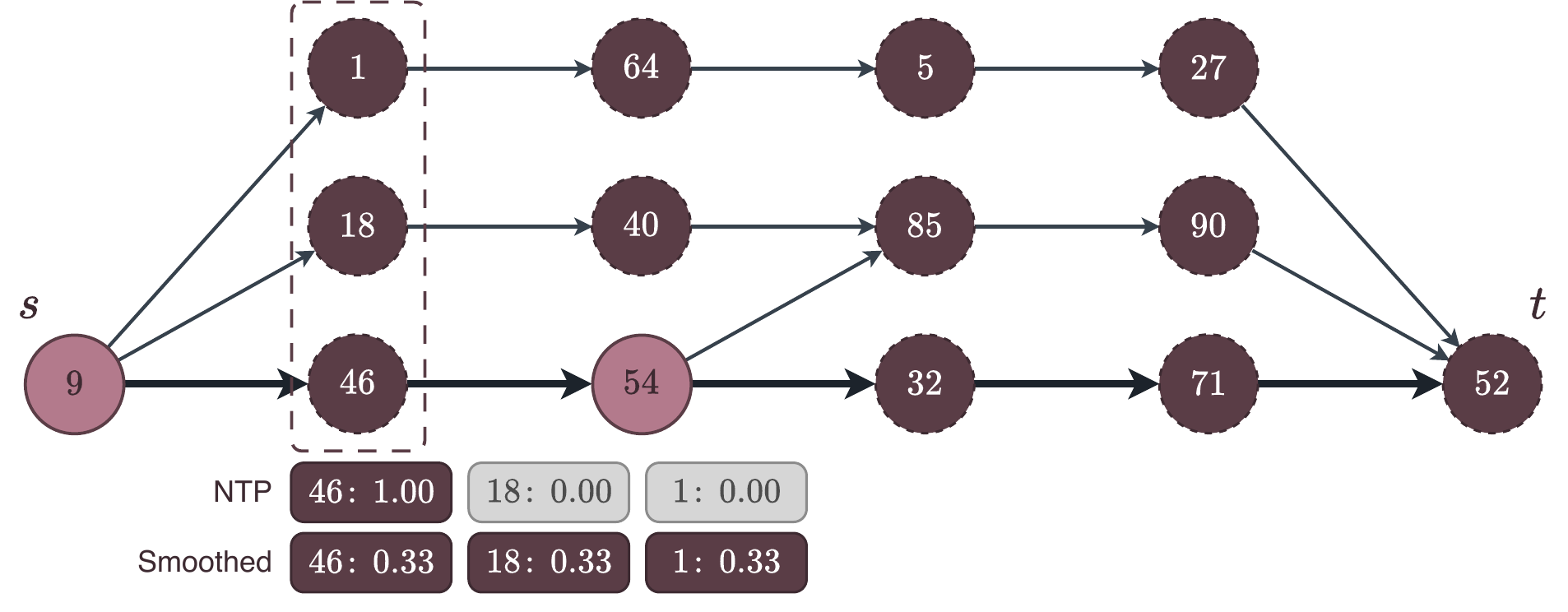}
    \caption{$\pi_{\leq}=9$. Each of $46$, $18$ and $1$ starts an equally valid
             shortest path, so $u_1$ has three correct targets.  The sampled ground-truth path is the bottom one with bold edges.}
    \label{fig:ls-a}
  \end{subfigure}%
  \\
  \begin{subfigure}[b]{1.0\textwidth}
  \centering
    \includegraphics[width=\lswidth]{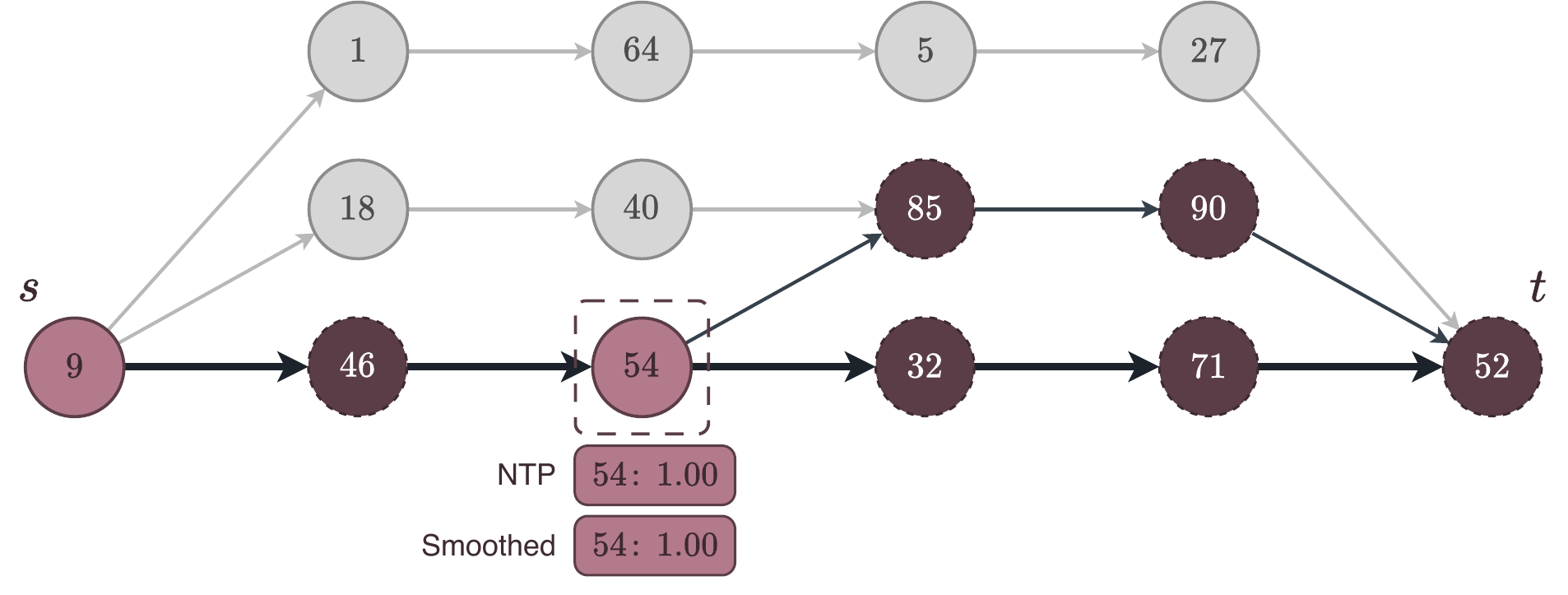}
    \caption{$\pi_{\leq}=9,46$. Conditioning on the ground-truth path resolves and removes the other two branches, and
             $46$ has a single continuation, so $u_2$ has one correct target.}
    \label{fig:ls-b}
  \end{subfigure}%
  \\
  \begin{subfigure}[b]{1.0\textwidth}
  \centering
    \includegraphics[width=\lswidth]{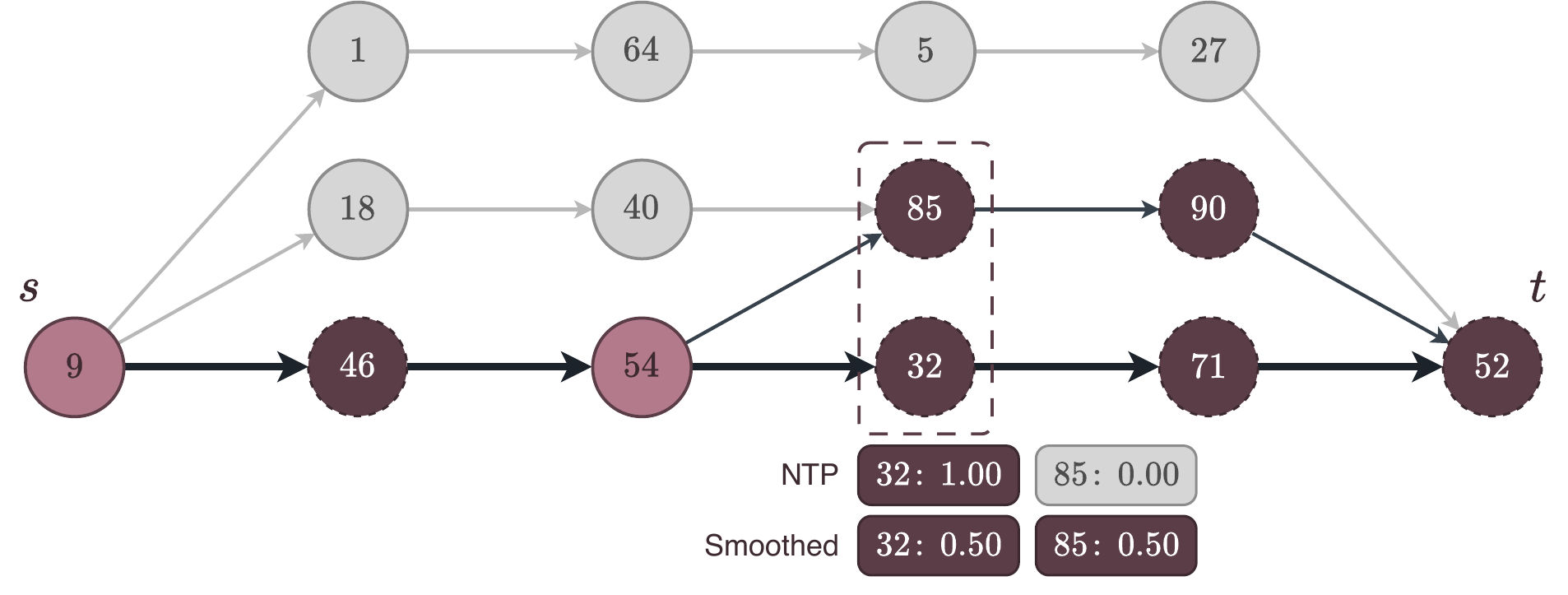}
    \caption{$\pi_{\leq}=9,46,54$. A second split occurs at the choice node $54$,
             so $u_3$ again has two correct targets, $32$ and $85$.}
    \label{fig:ls-c}
  \end{subfigure}
  \caption{Multiple valid shortest paths from $s=9$ to $t=52$ require label
    smoothing. Light marron solid-border nodes are {\em choice nodes},
    dark dashed-border nodes are not, and grey nodes have
    been removed from consideration by conditioning on $\pi_{\leq}$. The sampled ground-truth path $\pi$ is drawn with heavy edges. Below each
    choice node we give the NTP one-hot target and the correct uniform target
    over all valid next nodes, using the same encoding. The NTP row greys out
    the unchosen nodes, asserting they are off-path even though each begins an
    equally valid shortest path. Smoothing is needed only at the choice nodes,
    so the two rows coincide in (b).}
  \label{fig:label-smoothing}
\end{figure}

As an aside, {\bf semantic label smoothing} is related to MTP when considering MTP as a structured prediction task \citep{Frydenlund_Singh_Rudzicz_2022, iyer-etal-2026-beyond, zhang2026concept}.  The key difference here is that, in the former, the multiple target tokens are across the semantic dimension, while, in the latter, they are across the temporal or sequential dimensions.  However, in terms of implementation, the only change can be how the multiple targets are derived.  For example, the former created a self-supervised N-gram rule to derive multiple targets of semantically related words per time step. Whereas in this work, we simply use the generalization of the self-supervision rule used in NTP to derive the multiple targets for MTP.  However, both that work and this one use the same Plackett-Luce ranking loss to learn the underlying structured multi-label prediction problem.  \citet{zhang2026nitp} frames semantic prediction as a latent next-token prediction, where they hypothesize that future hidden-states can act as continuous supervision for the current state.  This can also be seen as a temporal form of self-distillation, in which the new loss encodes planning for future predictions, thereby relating it to MTP.

\section{Notes on Source-Side -- Full-Sequence -- Training with NTP}\label{appx:prefix}

Recently, both \citet{noroozizadeh2025deep} and \citet{trainin2026discrete} have shown positive results when allowing the model to predict the entire input, including the source-side prefix, instead of restricting NTP to the target side. This means the model must predict that $u$ follows $v$ to form edge $(u,\,v)$, for all edges in the randomly shuffled edge list.

\citet{noroozizadeh2025deep} considered the in-weights setting, where all training examples come from the same graph. Here, having the loss cover the entire input makes sense, since the model can and must learn the (single) graph's structure to complete the task.  However, \citet{trainin2026discrete}, like us, instead considered the in-context setting with path-star graphs, where every sample is a unique, unseen graph. This means the model cannot memorize information about any specific graph, as doing so would induce spurious correlations. They hypothesize that full-sequence training helps because the model learns the underlying generative process for creating graphs: ``the model is additionally rewarded for learning to generate legal graph descriptions,'' and that ``forcing the model to explicitly model the graph structure rather than merely attending to it is crucial for acquiring the necessary look-ahead mechanisms.''

We believe their hypothesis is correct, but argue the critical aspect is specifically that the graph must be \textbf{`legal.'} If the generative process were entirely random, as in Erdős–Rényi graphs, edge generation would be uninformative, and there should be no benefit from source-side training. However, the specific topology of path-star graphs imposes constraints on the generative process, and learning to respect them, we hypothesize, forces the model to engage in auxiliary forms of MMHT.

Concretely, consider the original path-star task, where the number of arms $d$ and arm length $k$ (excluding the start node $s$) are fixed across graphs, so $|\mathcal{N}| = dk+1$. This yields three legality constraints: $s$ must have degree $d$; every node has degree 2 if arm-interior or degree 1 if arm-final; and each arm must contain exactly $k$ unique nodes.

These degree and count constraints restrict which nodes can be selected from the vocabulary at each generation step, and satisfying them can require multihop reasoning. For instance, with $|\mathcal{N}|=51$ and $|\mathcal{V}|=100$, learning to respect the total node budget only requires counting prior nodes in the edge list, not traversing the graph. But learning to count how many nodes have been used \emph{within a given arm} does require multihop traversal. A further example which requires multihop reasoning is suppose an arm is missing exactly one edge $(u,\,v)$, with $(\cdot,\,u)$ and $(v,\,\cdot)$ already observed elsewhere in the list: the model can use counting to detect that an edge is missing, and multihop traversal to identify that it must be $(u,\,v)$ specifically. The number of hops required depends on where this edge falls in the arm.

Such constraints therefore create cases which provide a learning signal that may induce MMHT. If true, this offers further evidence that any source of MMHT signal can make the path-star task learnable, at least at limited scale \citep{frydenlund-2025-language}. We expect this particular signal to be minimal and not to scale \citep{trainin2026discrete}, but we see it as an interesting direction for future work.

\end{document}